\documentclass{article}

\usepackage{hyperref}

\usepackage{tcolorbox}
\usepackage{graphicx}
\usepackage{subfig}
\usepackage{float}
\usepackage{multicol}
\usepackage{lipsum}
\usepackage{enumitem}
\usepackage{textcomp}
\usepackage[utf8]{inputenc}
\usepackage{mdframed}
\usepackage{caption}
\usepackage{array}
\usepackage{multirow}

\usepackage[accepted]{icml2025}

\usepackage{amsmath}
\usepackage{amssymb}
\usepackage{mathtools}
\usepackage{amsthm}

\theoremstyle{plain}

\theoremstyle{definition}

\theoremstyle{remark}

\usepackage[textsize=tiny]{todonotes}

\icmltitlerunning{AutoChemSchematic AI: Agentic Framework for Chemical Process Scale-Up}

\begin{document}

\twocolumn[
\icmltitle{AutoChemSchematic AI: Agentic Physics-Aware Automation for Chemical Manufacturing Scale-Up}



\icmlsetsymbol{equal}{*}

\begin{icmlauthorlist}
\icmlauthor{Sakhinana Sagar Srinivas}{comp}
\icmlauthor{Shivam Gupta}{comp}
\icmlauthor{Venkataramana Runkana}{comp}
\end{icmlauthorlist}

\icmlaffiliation{comp}{Tata Research Development and Design Center, Bangalore}

\icmlcorrespondingauthor{Sakhinana Sagar Srinivas}{sagar.sakhinana@tcs.com}

\icmlkeywords{Machine Learning, ICML}

\vskip 0.3in
]



\printAffiliationsAndNotice{\icmlEqualContribution} % otherwise use the standard text.

\begin{abstract}
Recent advances in generative AI have accelerated the discovery of novel chemicals and materials. However, scaling these discoveries to industrial production remains a major bottleneck due to the synthesis gap---the need to develop entirely new manufacturing processes. This challenge requires detailed engineering blueprints: Process Flow Diagrams (PFDs) for equipment layouts and material/energy flows, and Piping and Instrumentation Diagrams (PIDs) for process plant operations. Current AI systems cannot yet reliably generate these critical engineering schematics, creating a fundamental obstacle to manufacturing scale-up of novel discoveries.
We present a closed-loop, physics-aware framework for automated generation of industrially viable PFDs and PIDs. The framework integrates three key components: (1) domain-specialized small language models (SLMs) trained for auto-generation of PFDs and PIDs, (2) a hierarchical knowledge graph containing process flow and instrumentation descriptions for 1,020+ chemicals for Graph Retrieval-Augmented Generation (GRAG), and (3) an open-source chemical process simulator for modeling, simulation, optimization, and analysis of novel chemical processes. The SLMs are trained through a multi-stage pipeline combining Supervised Fine-Tuning (SFT), Direct Preference Optimization (DPO), and Retrieval-Augmented Instruction Tuning (RAIT) on synthetic datasets, with process simulator-in-the-loop validation ensuring feasibility. To enhance computational efficiency, the framework implements structural pruning (width and depth) guided by importance heuristics to reduce language model size while preserving accuracy, followed by advanced inference optimizations including FlashAttention, Lookahead Decoding, PagedAttention with KV-cache quantization, and Test-Time Inference Scaling. Experimental results demonstrate that our framework generates simulator-validated process descriptions with high fidelity, outperforms baseline methods in correctness, and generalizes effectively to unseen chemicals. By bridging AI-driven molecular and material design with industrial-scale feasibility, this work significantly accelerates the path-to-production for AI-discovered chemicals.
\vspace{-2mm}
\end{abstract}

\section{Introduction}
\begin{figure}[ht!]
\vspace{-4mm}
\centering
\includegraphics[width=84.5mm,trim=0.0cm 4.0cm 0cm 4.0cm,clip]{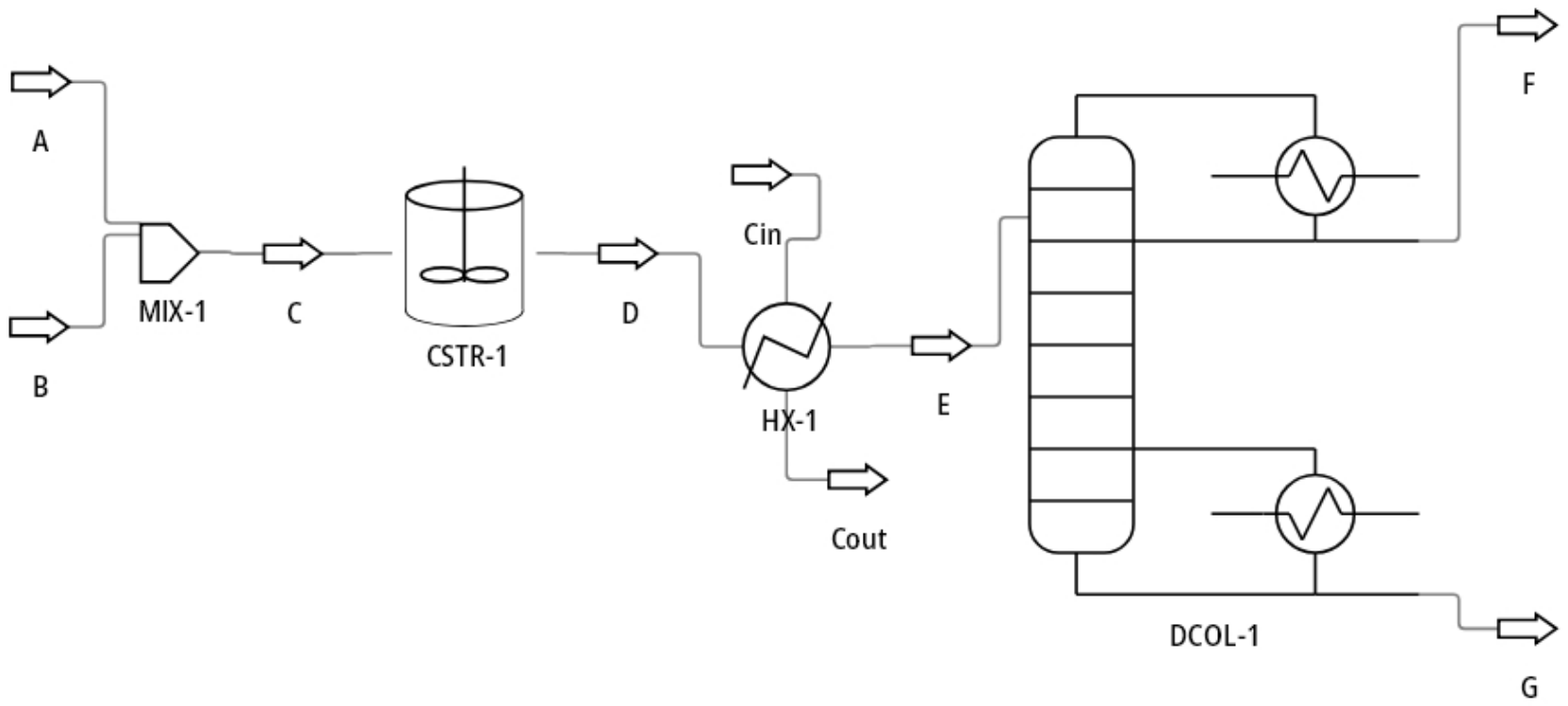}
\vspace{-4mm}
\caption{The figure shows a high-level schematic of a chemical process, depicting material flow from reactant inlets (A and B) through a mixer (MIX-1), a continuous stirred-tank reactor (CSTR-1), a heat exchanger (HX-1), and a distillation column (DCOL-1), yielding product streams F and G. Major equipment and stream connections are illustrated, excluding instrumentation and control logic. This schematic facilitates understanding of the core process operations and transformations.}
\label{fig:pfd_simple} 
\vspace{-2mm}
\end{figure}

\begin{figure}[ht!]
\vspace{-2mm}
\centering
\includegraphics[width=84.5mm,trim=0.0cm 3.75cm 0cm 3.85cm,clip]{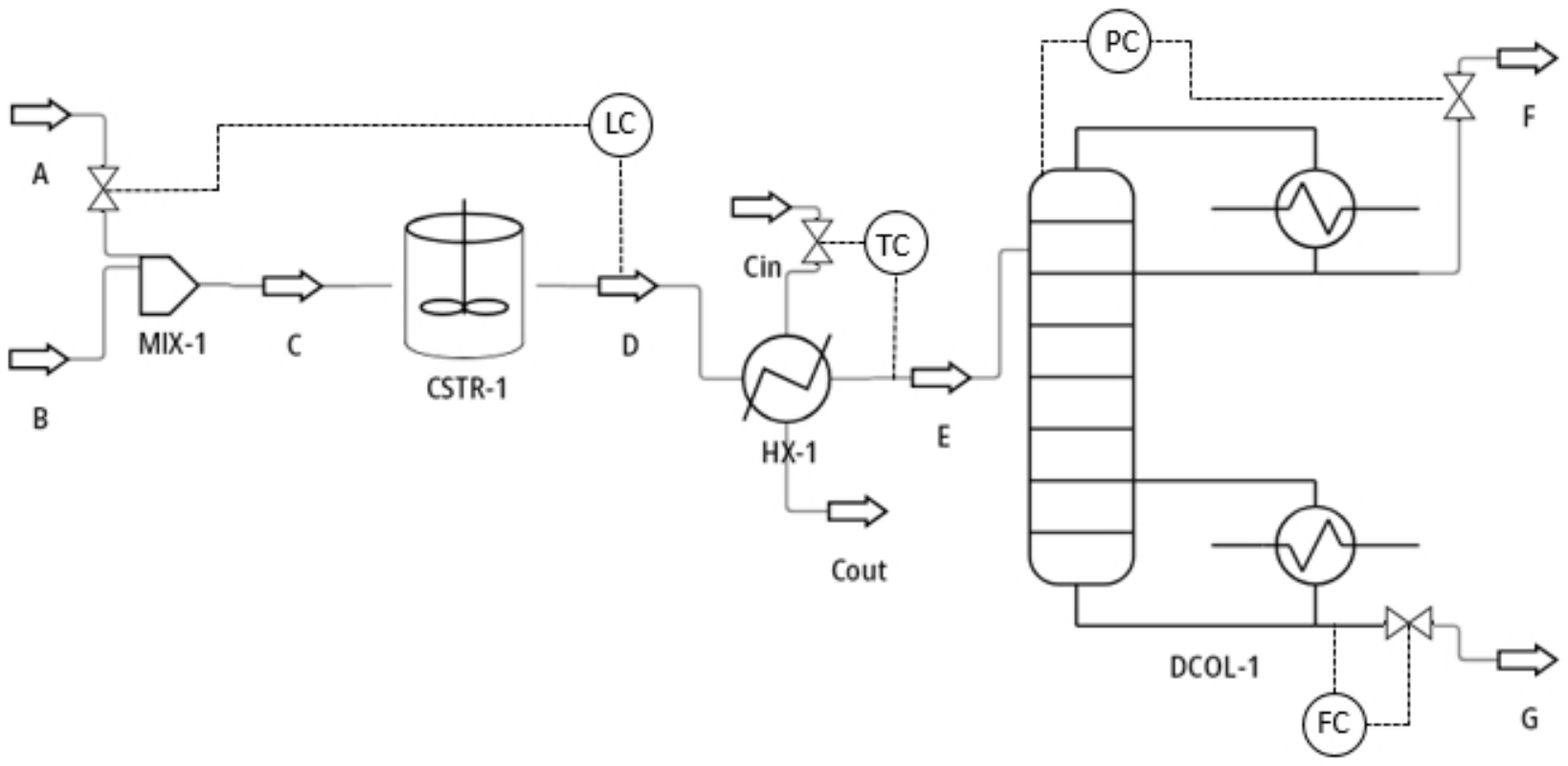}
\vspace{-4mm}
\caption{The figure shows the detailed PID of a chemical process showing instrumentation and control systems, including: level control (LC) on reactor CSTR-1 regulating feed A; temperature control (TC) on column feed E adjusting HX-1 utility flow; pressure control (PC) at DCOL-1 overhead controlling product F; and flow control (FC) on bottoms product G. The diagram specifies control strategies and safety-critical parameters.}
\label{fig:pid_detailed} % Label for referencing from the main text
\vspace{-2mm}
\end{figure}

Recent advancements in generative AI are transforming chemical and materials science \cite{chiang2024llamp, wang2024efficient, pan2024chemically, zhang2024honeycomb, guo2024saturn, kristiadi2024sober, sprueill2024chemreasoner, yang2024generative, kangretorinetext, yang2024generative}, accelerating the autonomous discovery of next-generation specialty chemicals and the development of high-performance, materials-based products. These advancements reduce dependence on manual, trial-and-error experimentation and computationally intensive first-principles simulation workflows, enabling faster and more sustainable innovation. However, many AI-discovered molecules and materials are not immediately manufacturable at scale. Transitioning them from computer simulations or wet-lab experiments to industrial production requires the development of new processes—a significant challenge in bringing better products to market rapidly. Bridging the gap from in silico design to industrial synthesis involves addressing the challenges of scalable process development. While generative AI has revolutionized molecular and materials discovery, its application to the design of scalable production processes remains largely underexplored. This gap is critical because Process Flow Diagrams (PFDs) and Piping and Instrumentation Diagrams (PIDs) serve as essential bridges between laboratory-scale innovations and industrial-scale manufacturing. These blueprints(or schematics) provide the foundational basis for the simulation, optimization, and control of chemical processes; thus, the ability to generate accurate PFDs and PIDs is fundamental to overcoming the scale-up bottleneck in AI-driven chemical innovation. PFDs and PIDs are standard engineering diagrams used in the chemical process industry. A PFD provides a high-level schematic of the flow of materials and energy through a chemical production process, depicting major equipment, process streams, and key operating conditions for specific units without detailing instrumentation or control systems. In contrast, PIDs build upon PFDs by offering a more detailed schematic of the instrumentation and control systems essential for monitoring, operational control, safety, and plant maintenance. The purpose of a PFD (see Figure~\ref{fig:pfd_simple}) is to illustrate what happens in the process—such as key physical or chemical transformations—and where it occurs (i.e., in which major equipment units), rather than how the process is controlled. Conversely, a PID (see Figure~\ref{fig:pid_detailed}) focuses on how the process operates and is controlled, including valves, sensors, and control loops, rather than just the transformations or equipment involved. 

Together, PFDs and PIDs serve as foundational documents for chemical process simulations, which drive the development of digital twins. These digital twins integrate first-principles or data-driven models with real-time sensor and actuator data, enabling dynamic monitoring, predictive control, and AI-driven automation for closed-loop process optimization. Current methods \cite{vogel2023learning, schulze2023data, hirretier2022towards, alimin2025talking, gowiakar2024agentic, srinivas2024accelerating} are not designed to auto-generate process flow schematics (e.g., PFDs) or instrumentation and control layouts (e.g., PIDs) for novel industrial-scale chemical production processes, significantly limiting their practical utility. These approaches also fail to incorporate essential process context: for PFDs, this includes high-level objectives—such as what the process achieves and in what sequence—while for PIDs, it requires operational details on how the process is monitored, controlled, and executed. Consequently, they cannot justify critical design choices or the control and instrumentation logic necessary for efficient plant operations. Another major limitation is the lack of integration with first-principles-based simulators to verify the physical and operational feasibility of generated PFDs and PIDs, further undermining their industrial reliability. Current AI-driven discovery pipelines frequently optimize molecular properties without production feasibility checks. Auto-generating and simulating PFDs (to verify unit operations, mass/energy balances, and phase behavior) and PIDs (to validate control logic, safety interlocks, and equipment specifications), chemical process simulators can flag scale-up issues like equipment sizing errors, utility mismatches, or unsafe designs before lab-scale synthesis. This moves manufacturability screening from retrospective correction to proactive design. Moreover, the reliance on manual, expertise-intensive creation of novel PFDs and PIDs introduces a bottleneck that adversely impacts simulation fidelity, digital twin accuracy, and scalable AI deployment in industrial manufacturing. To address these limitations, we present a closed-loop, self-driving lab framework for the auto-generation of high-fidelity process flow and instrumentation descriptions, accelerating the development of novel chemical processes. Implemented as an enterprise-grade, cloud-based SaaS solution, our framework significantly expedites the simulation-to-lab-to-pilot-to-plant scale-up pipeline, ensuring that only industrially viable, sustainable, and efficient processes advance to commercialization. Serving as an end-to-end process schematics modeling tool, the platform automates design, simulation, and optimization with minimal human intervention. By integrating first-principles, physics-aware modeling with iterative reflection and adaptive learning, the framework continuously self-improves, enhancing the reliability of AI-generated process schematics and control strategies.  Our approach combines three key innovations:

\begin{figure*}[ht!]
\vspace{-1mm}
\centering
\resizebox{0.80\textwidth}{!}{
\includegraphics[keepaspectratio,trim=0.0cm 0.6cm 0cm 1.5cm,clip]{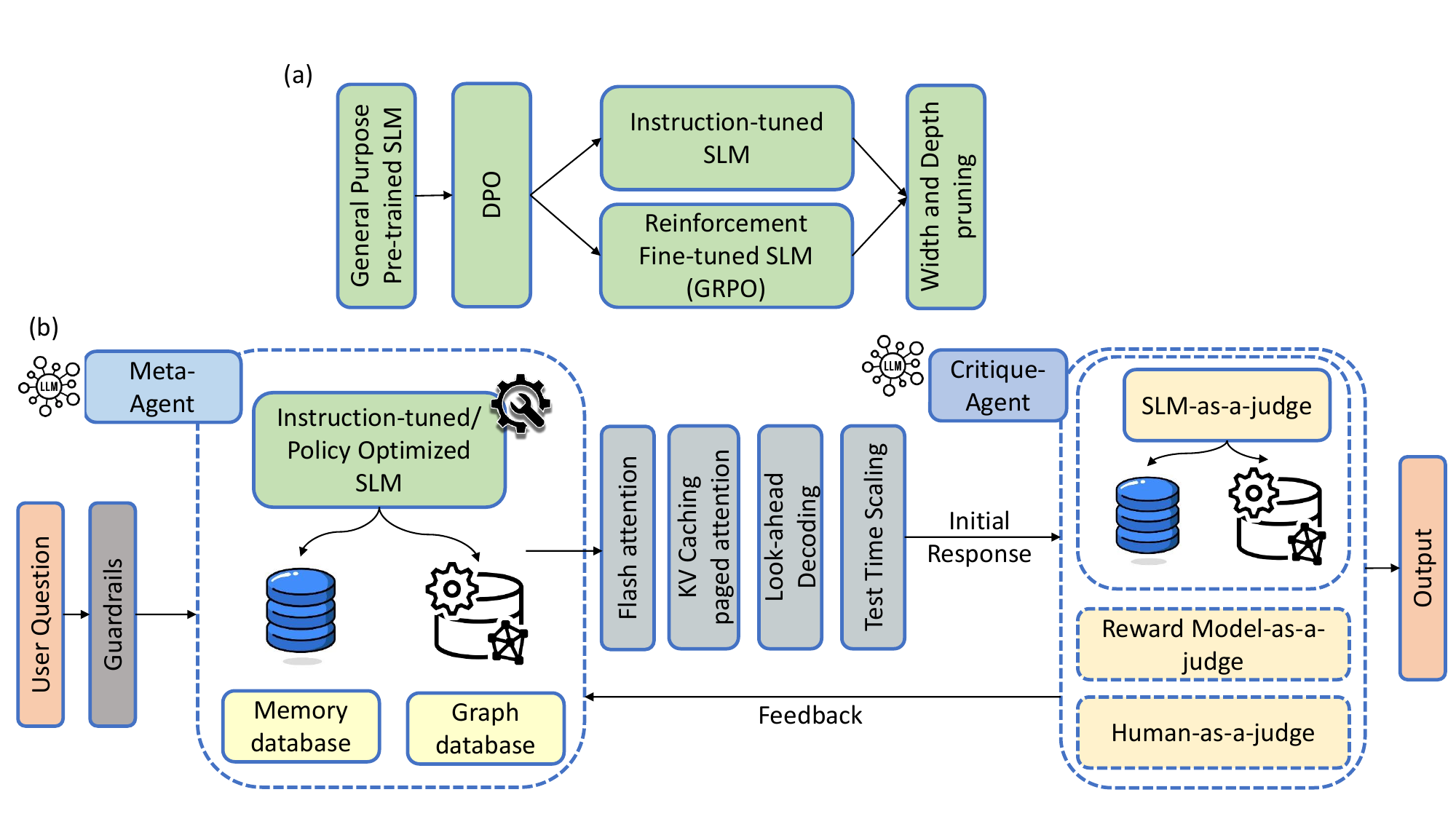} % trim = <left> <bottom> <right> <top>
}
\vspace{-2mm}
\caption{Overview of the integrated framework. (a) The SLM fine-tuning pipeline depicts initial DPO alignment followed by supervised instruction tuning or policy-gradient reinforcement learning, with optional width/depth pruning. (b) The operational RAG framework illustrates a Meta-Agent coordinating with the specialized SLM (from part a), which accesses memory and graph databases for context. The SLM's inference is accelerated via optimizations (FlashAttention, Paged KV Caching, Lookahead Decoding, Test-Time Scaling). Generated responses are refined iteratively through a feedback loop managed by a Critique-Agent employing diverse judges (e.g., Nemotron-4-340B reward model, LLM-as-a-judge like GPT-4o/Haiku, or human evaluation).}
\label{fig:framework}
\vspace{-2mm}
\end{figure*}

(1) custom chemical database curation and knowledge graph construction for Retrieval-Augmented Generation (RAG), (2) domain-specialized small-scale language models (SLMs) fine-tuned through multi-stage training, and (3) physics-aware simulator validation using \href{https://dwsim.org/}{DWSIM}. This closed-loop system enables robust generation and verification of the industrial-scale feasibility of scalable manufacturing processes and AI-driven discoveries. In the following sections, we present our methodology in detail, describe the experimental setup, and report results.

\section{Methodology}
Our methodology integrates data curation, advanced small language model (SLM) fine-tuning, knowledge graph construction for retrieval augmentation, inference optimization, and engineering validation to create specialized and efficient SLMs for chemical process engineering tasks---specifically, the interpretation, analysis, and generation of PFDs and PIDs (refer to Figure~\ref{fig:framework}). The pipeline begins with the curation of a custom database comprising over 1,120 chemicals drawn from sectors such as electronics, energy storage, pharmaceuticals, advanced manufacturing, and utilities, with a focus on chemicals essential to modern industrial applications. The data were programmatically extracted from product catalogs of leading manufacturers---including BASF, Dow Chemical, DuPont, Solvay, Mitsubishi Chemical, Bayer, Evonik, SABIC, and LyondellBasell---ensuring both reliability and broad industrial coverage. The dataset consists of two components: \textit{ChemAtlas} and \textit{ChemEval}. \textit{ChemAtlas} is a core collection of 1,020 chemicals. For each chemical in \textit{ChemAtlas}, we employ an AI-driven agentic web navigation framework that autonomously retrieves, interprets, and synthesizes multimodal data from diverse public sources to generate detailed descriptions of production processes (textual descriptions of both PFDs and PIDs). This structured data serves as the foundation for populating chemical knowledge graphs, where text chunks are processed by GPT-4o to extract semantic triples (subject--predicate--object). Entities are canonicalized based on high semantic similarity (via embeddings) and string similarity (via normalized Levenshtein distance), and the resulting graph is partitioned into hierarchical communities using the Leiden algorithm to optimize modularity for efficient retrieval. Our Graph RAG framework leverages this structured graph representation to enhance both contextual reasoning and retrieval efficiency, enabling SLMs to deliver accurate and context-sensitive answers. To ensure consistency and correctness, we further use advanced large language models (LLMs)---specifically, GPT-4o and Anthropic Claude Haiku---to generate and cross-validate chemical-specific production process descriptions derived from agentic web navigation, leveraging their pre-trained knowledge for automated validation. The second component, \textit{ChemEval}, comprises a held-out evaluation set of 100 chemicals curated to rigorously test the framework's zero-shot generalization performance in auto-generating process flow and instrumentation descriptions for chemicals not present in \textit{ChemAtlas}. Additionally, we adopt a teacher--student transfer learning approach by generating custom synthetic datasets from the \textit{ChemAtlas} database. This includes 20K instruction--response (QA) pairs created by teacher LLMs---specifically, GPT-4o and Anthropic Claude Haiku---to train SLMs such as LLaMA-3.2-1B and SmolLM-135M on complex, domain-specific process engineering tasks. These tasks include equipment and piping layout generation, sensor and instrumentation placement (i.e., the analysis, interpretation, and auto-generation of PFDs and PIDs). A small seed set of human-authored instruction--response pairs was used as demonstrations to initiate high-quality QA dataset generation through iterative synthesis, guided by predefined templates and a self-instruct bootstrapping strategy. The generated outputs are scored, validated, and filtered using NVIDIA's Nemotron-4-340B reward model. The resulting datasets span a diverse range of QA types, including factual knowledge, preference alignment, process flow and instrumentation interpretation, logical and multi-step chain-of-thought reasoning, sensor layout planning, comparative process analysis, and error detection and correction. The curated 20K synthetic QA dataset consists of six specialized subsets---\textit{Factual QA}, \textit{SynDIP}, \textit{LogiCore}, \textit{DPO}, \textit{Local RAIT}, and \textit{Global RAIT}---each systematically constructed to induce reasoning, alignment, and generation abilities in SLMs.
The \textit{Factual QA} dataset, constructed through hierarchical topic decomposition of chemical process engineering concepts, enhances foundational domain knowledge and factual recall. The \textit{SynDIP} dataset contains QA pairs describing process flow and instrumentation, equivalent in content to knowledge retrieved via the agentic web navigation framework, but instead generated from the pretrained knowledge of base LLMs. The \textit{LogiCore} dataset consists of multi-step reasoning QA pairs grounded in process flow and instrumentation descriptions. These pairs are crafted to justify process design choices, validate control logic, and explain flow sequencing within chemical process diagrams. The \textit{DPO} dataset comprises preference-labeled QA pairs, each including a preferred and a dispreferred response, distinguished using score differentials from a reward model to support alignment tuning via Direct Preference Optimization. The RAIT (Retrieval-Augmented Instruction Tuning) datasets are designed to enhance the SLMs' ability to incorporate retrieved context into generation for grounded and context-aware responses. \textit{Local RAIT} grounds QA pairs in individual \textit{SynDIP}-derived text chunks, enabling precise and context-specific information extraction. In contrast, \textit{Global RAIT} leverages semantically clustered groups of chunks---potentially spanning multiple \textit{SynDIP}-derived documents---to support cross-contextual reasoning and synthesis across related segments. The complete 20K synthetic QA dataset, encompassing all six categories, is randomly split into 80\% training, 10\% validation, and 10\% internal test sets for evaluating generalization performance.
In addition, we construct a 1.5K QA-pair out-of-distribution (OOD) benchmark dataset from \textit{ChemAtlas} using a self-instruct approach with teacher LLMs (OpenAI o3 and o1-mini) to generate synthetic QA pairs. These pairs are iteratively created from \textit{SynDIP}-retrieved information and filtered for quality using a reward model to evaluate whether fine-tuned SLMs can generalize across core capabilities, including factual knowledge, reasoning, instruction following, and the interpretation and analysis of process flow and instrumentation tasks. Finally, we evaluate the framework's ability to generate accurate PFD and PID descriptions for unseen chemicals using \textit{ChemEval}. Specifically, for each chemical in \textit{ChemEval}, GPT-4o and Claude Haiku produced process flow and instrumentation descriptions in the form of QA pairs using the same self-instruct bootstrapping method. These QA pairs served as reference targets (ground truth) for quantitative evaluation. Base SLMs---specifically, LLaMA-3.2-1B and SmolLM-135M---are customized using Quantized Low-Rank Adaptation (QLoRA) \cite{dettmers2023qlora, xu2023qa} with frozen base weights. We employ two distinct fine-tuning strategies on synthetic datasets. The first follows a sequential, modular pipeline: Supervised Fine-Tuning (SFT) on the combined \textit{Factual QA}, \textit{SynDIP}, and \textit{LogiCore} datasets; Direct Preference Optimization (DPO) on curated preference-labeled \textit{DPO} datasets; and Retrieval-Augmented Instruction Tuning (RAIT) on \textit{Local} and \textit{Global RAIT} datasets. The second strategy adopts a reinforcement learning approach using Group Relative Policy Optimization (GRPO) \cite{shao2024deepseekmath, guo2025deepseek, liu2024deepseek}, applied first to the SFT datasets and then refined on RAIT datasets. This approach optimizes a composite reward function combining ROUGE-L F1 scores, length ratio penalties, and LLM-as-a-judge quality scores, stabilized by KL divergence regularization. We compare these strategies to assess whether modular fine-tuning or end-to-end reinforcement learning better aligns SLMs with complex, multi-objective benchmarks. The fine-tuned SLMs are integrated with the structured knowledge graph through a Graph RAG framework. During inference, the framework retrieves relevant graph communities by comparing query embeddings to pre-computed community summaries, dynamically selects the most relevant communities, and constructs a subgraph containing interconnected entities, semantic relationships, and source text chunks. This contextual subgraph is then used for grounded, multi-hop reasoning. To enhance performance, a suite of inference optimization and reliability techniques is implemented: structural pruning (width and depth) guided by importance heuristics reduces model size; PagedAttention combined with KV cache quantization mitigates memory fragmentation and reduces cache footprint; Lookahead Decoding accelerates generation latency through parallel token speculation; FlashAttention optimizes the core attention computation to reduce memory bandwidth bottlenecks; and Test-Time Inference Scaling improves output reliability using self-consistency sampling, confidence-weighted entropy scoring, iterative self-reflection/revision, and consensus aggregation. Finally, the practical engineering feasibility of generated process flow and instrumentation descriptions is validated using the DWSIM open-source chemical process simulator, where PFDs are translated into flowsheets to verify material/energy balances and thermodynamic consistency, while PIDs are operationally validated  by implementing control loops in DWSIM's dynamic environment to evaluate stability and control performance (e.g., setpoint tracking, disturbance rejection). DWSIM validates AI-generated PFDs/PIDs by converting textual descriptions into executable simulations, verifying adherence to chemical engineering principles (mass/energy balances, thermodynamic consistency, and equipment feasibility) while flagging errors, inconsistencies, and optimization opportunities through first-principles analysis. Figure~\ref{fig:framework} visually outlines the overall architecture. Part (a) depicts the SLM fine-tuning pipeline, showing the progression from a general pre-trained model to initial preference alignment (DPO), followed by task-specific fine-tuning via either instruction tuning or reinforcement learning (GRPO), and concluding with model compression (pruning). Part (b) illustrates the operational RAG framework, where a user query passes through guardrails before being processed by a Meta-Agent. This agent employs the specialized SLM developed in part (a) as its core reasoning engine. Guided by the Meta-Agent, the SLM retrieves necessary context by accessing both a Memory database (e.g., for conversational history) and a Graph database (containing structured process knowledge), which informs its response generation. The SLM's inference process is enhanced by integrated optimizations (FlashAttention, PagedAttention KV caching, Lookahead Decoding, and Test-Time Scaling). An initial SLM-generated response is evaluated by a Critique-Agent using feedback mechanisms (SLM-as-a-judge, Reward Model-as-a-judge, or Human-as-a-judge) to potentially trigger refinement before final output delivery. In summary, our integrated framework combines knowledge graph-based retrieval augmentation, domain-specific SLM fine-tuning pipelines, comprehensive inference optimizations, and feedback-driven refinement. This approach achieves robust performance on complex reasoning tasks and demonstrates effective generalization through the generation of plausible, simulator-validated process descriptions for previously unseen chemicals.

\section{Experiments}

\subsection{Experimental Setup}
\begin{figure*}[ht!]
\vspace{-2mm}
\centering
\resizebox{0.85\textwidth}{!}{
\includegraphics[keepaspectratio,trim=0.0cm 0.5cm 0cm 3.75cm,clip]{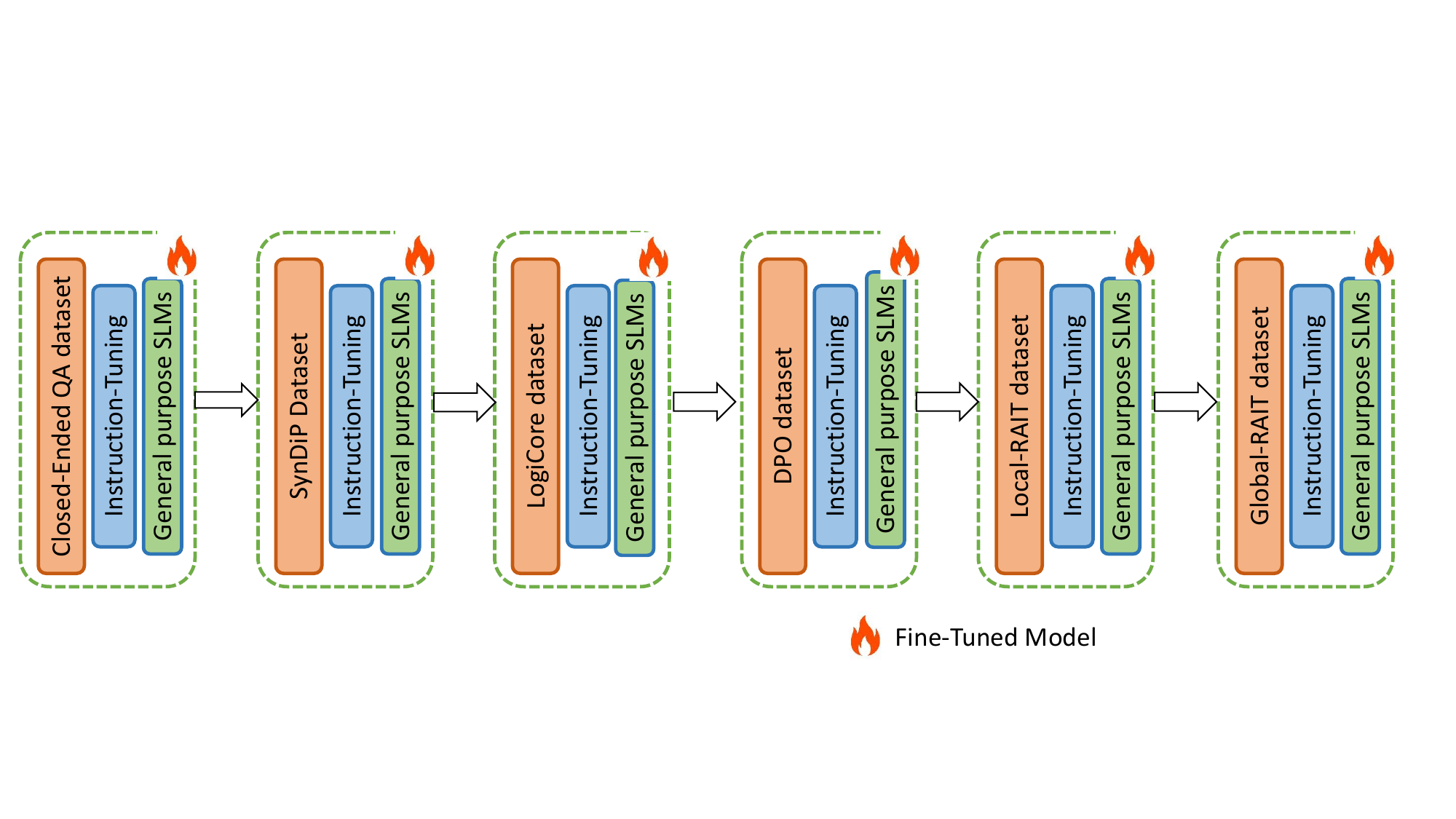} % trim = <left> <bottom> <right> <top>
}
\vspace{-14mm}
\caption{The figure illustrates the multi-stage instruction-tuning pipeline used to train specialized student models—such as Llama-3.2-1B and SmolLM2-135M—for PFD/PID interpretation tasks. The pipeline integrates synthetic datasets including \textit{Factual QA}, \textit{SynDIP}, \textit{LogiCore}, \textit{DPO}, \textit{Local-RAIT}, and \textit{Global-RAIT}—each generated using teacher LLMs (e.g., GPT-4o, Claude-3-Haiku) and validated with reward models such as NVIDIA's \textit{Nemotron-4-340B}. These datasets target diverse capabilities including factual question answering, process flow and instrumentation descriptions generation, logical reasoning, preference optimization, and retrieval-augmented comprehension. The combined instruction-tuning process refines general-purpose SLMs into domain-optimized models capable of performing chemical process engineering tasks with high fidelity.}
\label{fig:TP}
\vspace{-2mm}
\end{figure*}

Graph Retrieval-Augmented Generation (Graph RAG) integrates structured knowledge graphs with
large language models to enhance retrieval and reasoning. Our implementation begins with domain-specific documents—focused on chemical production processes—retrieved through autonomous agentic web navigation from the \textit{ChemAtlas} database. The raw text is segmented into overlapping chunks using a sliding window approach, preserving local context while ensuring cross-chunk continuity. Each text chunk is processed by GPT-4o to extract subject-predicate-object triples, forming semantic edges between entity nodes in the knowledge graph. Reference edges connect each entity to its source chunk, preserving document-graph alignment and enabling traceability. To resolve redundancy, we apply a canonicalization step: entities are merged only if they exhibit both high semantic similarity (measured via \texttt{text-embedding-3-small} embeddings) and high string similarity (evaluated using normalized Levenshtein distance), with both metrics exceeding predefined thresholds. For efficient retrieval, we partition the graph into hierarchical communities using the Leiden algorithm \cite{traag2019louvain}, optimizing for modularity to ensure semantically coherent clustering. Prior to inference, each community is summarized by GPT-4o, and these summaries are embedded for fast similarity comparison. Given a query, the framework retrieves the top-$K$ most relevant communities, dynamically constructing a subgraph that includes their interconnected entities, semantic relationships, and originating chunks. This structured context is then passed to the reasoning model, ensuring grounded, multi-hop generation. We fine-tuned the Llama-3-1B and SmolLM-135M models using Quantized Low-Rank Adaptation (QLoRA)~\cite{dettmers2023qlora}, which adapts low-rank matrices to key transformer projection layers while keeping the base model weights frozen in 4-bit NormalFloat (NF4) precision. All experiments used identical training parameters: an 8-bit AdamW optimizer ($\beta_1=0.9$, $\beta_2=0.999$), a learning rate of $2 \times 10^{-4}$ with linear decay, weight decay of 0.01, and an effective batch size of 8 (achieved via a per-device batch size of 2 with 4 gradient accumulation steps). We set a maximum sequence length of 4096 tokens, enabled by gradient checkpointing. Training was conducted on NVIDIA V100 GPUs using mixed precision (BF16 for matrix operations, FP16 for gradients). We explored two distinct fine-tuning strategies(refer Figure \ref{fig:TP}). The first strategy employed a sequential, multi-phase pipeline consisting of three stages: (1) Supervised Fine-Tuning (SFT) on the combined training splits of the \textit{Factual QA}, \textit{SynDIP}, and \textit{LogiCore} datasets for 15 epochs to integrate instruction-following capabilities and foundational domain knowledge into the SLMs; (2) Direct Preference Optimization (DPO) on the curated \textit{DPO} dataset's training split for 5 epochs to align the SLM's outputs with human preferences; and (3) Retrieval-Augmented Instruction Tuning (RAIT) leveraging the training splits of the \textit{Local} and \textit{Global} datasets for 15 epochs to enhance the SLM's ability to generate contextually grounded responses. The second strategy utilized the Group Relative Policy Optimization (GRPO) reinforcement learning algorithm \cite{shao2024deepseekmath}, adapted for direct policy optimization. This approach proceeded through two sequential stages: initially fine-tuning the pretrained base model on the training splits of the combined \textit{Factual QA}, \textit{SynDIP}, and \textit{LogiCore} datasets, followed by refining the resulting SFT checkpoint using the training splits of the \textit{Local RAIT} and \textit{Global RAIT} datasets. Both stages employed the same QLoRA configuration described earlier. The optimization process maximized a clipped surrogate objective \cite{schulman2017proximal}, conceptually similar to Proximal Policy Optimization (PPO), using normalized advantages derived from a composite reward function with three components: ROUGE-L F1 score (weight=0.3), a length ratio penalty to encourage similarity to reference response lengths (weight=0.2), and an LLM-as-a-Judge quality score evaluating answer correctness and relevance (weight=0.5). Rewards and advantages were computed relative to groups ($G=4$) of responses sampled from the policy for each input. Training stability was maintained through $\beta$-weighted KL divergence regularization against the relevant reference policy (either the pretrained base model or SFT checkpoint), with GRPO training running for 15 epochs per stage until convergence. To isolate the comparative effects of learning paradigms, we implemented parallel adaptation strategies under identical conditions: (1) supervised fine-tuning versus (2) GRPO-based reinforcement learning. Using fixed architectures and shared datasets (\textit{FactualQA}, \textit{SynDIP}, \textit{LogiCore}, and \textit{Local/Global RAIT}), this controlled experiment quantifies how each paradigm influences SLM performance metrics across knowledge acquisition, reasoning, and generation tasks. All implementations were developed in PyTorch using Hugging Face libraries, including \texttt{transformers}, \texttt{datasets}, \texttt{peft}, and \texttt{trl}. We evaluated fine-tuned SLMs through four key dimensions: (1) quantitative textual analysis comparing model outputs against ground-truth references using BLEU, ROUGE (1, 2, L), METEOR, SacreBLEU, BERTScore, and Sentence-BERT embedding cosine similarity; (2) qualitative scoring via the Nvidia Nemotron-4-340B reward model (0-4 ratings for correctness and coherence); (3) system-level efficiency benchmarks measuring inference latency (ms/token), throughput (tokens/sec), and GPU memory utilization; and (4) process engineering simulations in DWSIM to validate auto-generated PFDs and PIDs for industrial-scale feasibility. We investigated the trade-off between model compression and predictive fidelity (e.g., accuracy, reasoning) in SLMs through structural pruning techniques. Both width-level (neuron-level) and depth-level (layer-level) pruning were guided by importance heuristics computed during fine-tuning, enabling systematic parameter reduction while monitoring per-task downstream performance impact. To improve inference-time reliability—particularly for factual accuracy and reasoning consistency—we implemented a test-time scaling mechanism combining multi-path exploration through stochastic sampling, confidence-based candidate ranking, iterative self-reflection and revision, and consensus aggregation. These techniques collectively enhanced output robustness compared to standard deterministic decoding, as measured by qualitative reward model metrics. We conducted a systematic evaluation of optimization techniques to improve performance (speed, memory usage, throughput) during autoregressive inference of fine-tuned SLMs, including the LLaMA-3.2-1B architecture on NVIDIA V100 GPUs, focusing on three key inference-time methods: PagedAttention~\cite{kwon2023efficient} with KV-cache quantization for memory efficiency, Lookahead Decoding~\cite{fu2024break} for throughput improvement, and FlashAttention~\cite{dao2022flashattention,dao2023flashattention} for latency reduction. Benchmarking across metrics including inference throughput (tokens/sec), average generation latency (sec), maximum batch size, and peak GPU memory usage (GB) demonstrated their complementary benefits for demanding engineering applications. PagedAttention addressed memory fragmentation and throughput limitations by organizing the Key-Value cache into non-contiguous blocks, improving memory efficiency and enabling larger batch sizes compared to traditional contiguous caching. Lookahead Decoding reduced end-to-end latency through parallel token generation and verification within each forward pass while maintaining output equivalence with greedy decoding, with its effectiveness quantified through comparative measurements of generation latency and throughput. FlashAttention optimized attention computation by alleviating memory bandwidth bottlenecks in Transformer architectures through its I/O-aware approach, with improvements evaluated on both training/inference throughput and peak memory consumption during training. Finally, we validated engineering feasibility using the   DWSIM~\cite{dwsim}, an open-source chemical process simulator to construct and simulate PFDs/PIDs from auto-generated textual descriptions of novel chemical processes. DWSIM functions as a virtual chemical plant, enabling users to design, simulate, and analyze chemical processes. It supports both steady-state and dynamic modeling, computes material and energy balances, and simulates various physical and chemical operations such as mixing, reactions, separations, and etc. Additionally, it predicts key properties like phase behavior, heat duties, and equipment sizing while offering optimization and sensitivity analysis for process improvement. DWSIM enables chemical process design and simulation through drag-and-drop PFD creation, supporting unit operations like pumps, reactors, distillation columns, and heat exchangers. Users define components and select thermodynamic models (e.g., Peng-Robinson, NRTL, PC-SAFT) to simulate systems, obtaining flow rates, temperatures, pressures, and compositions. The software performs advanced calculations: phase diagrams, enthalpy-entropy charts, and property tables. It includes optimization tools for cost/yield/efficiency, sensitivity analysis, and reaction modeling (CSTRs, PFRs, Gibbs reactors) with conversion/yield tracking. Equipment sizing (PSVs, vessels, exchangers) and dynamic simulations (startups, shutdowns, upsets) are also supported, allowing real-time process change analysis. In our work, DWSIM provides rigorous validation of auto-generated PFD and PID descriptions by converting textual process information into executable simulations. The software performs multi-level verification through material/energy balance calculations, thermodynamic consistency checks (using appropriate property packages like NRTL or Peng-Robinson), and equipment compatibility analysis. It identifies structural gaps in process descriptions by mapping unit operations to mathematical models and detecting missing connections or undefined parameters. Through steady-state and dynamic simulation, DWSIM evaluates operational feasibility, verifying control strategies, equipment specifications, and safety limits against simulated performance. The validation process flags inconsistencies in phase behavior, stream properties, and process conditions, while convergence analysis ensures numerical robustness. This systematic approach transforms textual process descriptions into validated, implementable designs by bridging the gap between conceptual documentation and physical realizability. In summary, DWSIM goes beyond checking if the outputs 'look right' textually—it proves they would operate as executable chemical processes by subjecting them to rigorous physical/chemical laws and engineering constraints. This bridges the gap between LLM-generated text and real-world implementability.

\vspace{-2mm} 
\subsection{Results}
\begin{figure*}[ht!]
\vspace{-5mm}
\centering
\captionsetup[subfloat]{font=tiny,labelfont=bf}
\resizebox{1.0\textwidth}{!}{
\subfloat[Model performance comparison on the ChemEval benchmark for PFD/PID generation, evaluated using the NVIDIA/Nemotron-4-340B reward model (0–4 scale). Performance ranking: GPT-4o (highest), fine-tuned Llama-3.2 1B (second, with lower verbosity but higher variance), and SmolLM2-135M (lowest, though matching Llama-3.2 1B in complexity and verbosity).]{\includegraphics[width=55mm]{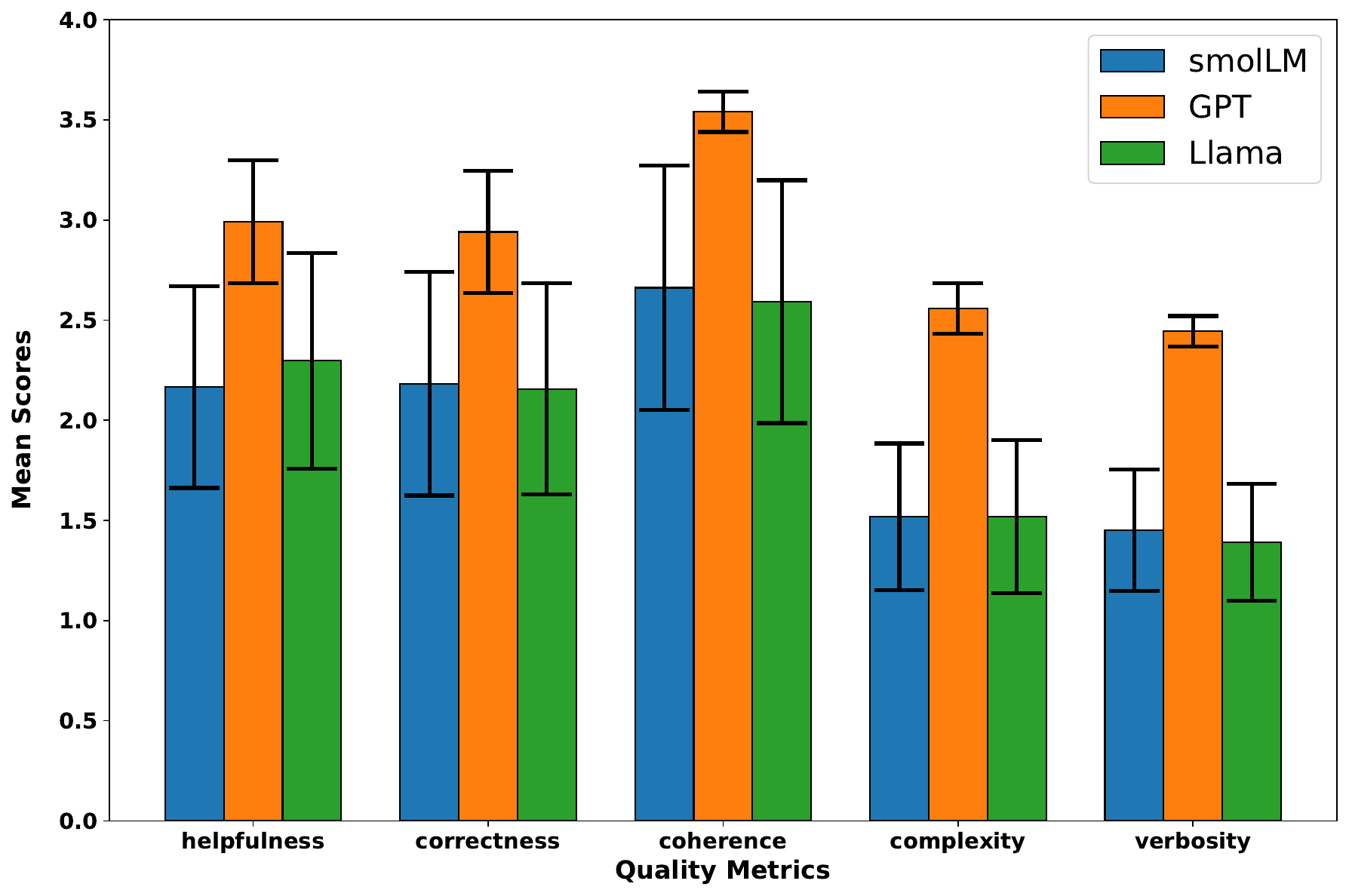}}
\hspace{0.05\textwidth}
\subfloat[Performance comparison of three Llama-3.2 1B configurations on the ChemEval benchmark. Performance ranking: fully enhanced variant with fine-tuning, GraphRAG, and feedback (green, highest) outperforms GraphRAG-only version (blue, middle) and base pretrained model (orange, lowest), demonstrating the cumulative benefits of retrieval and feedback mechanisms.]{\includegraphics[width=55mm]{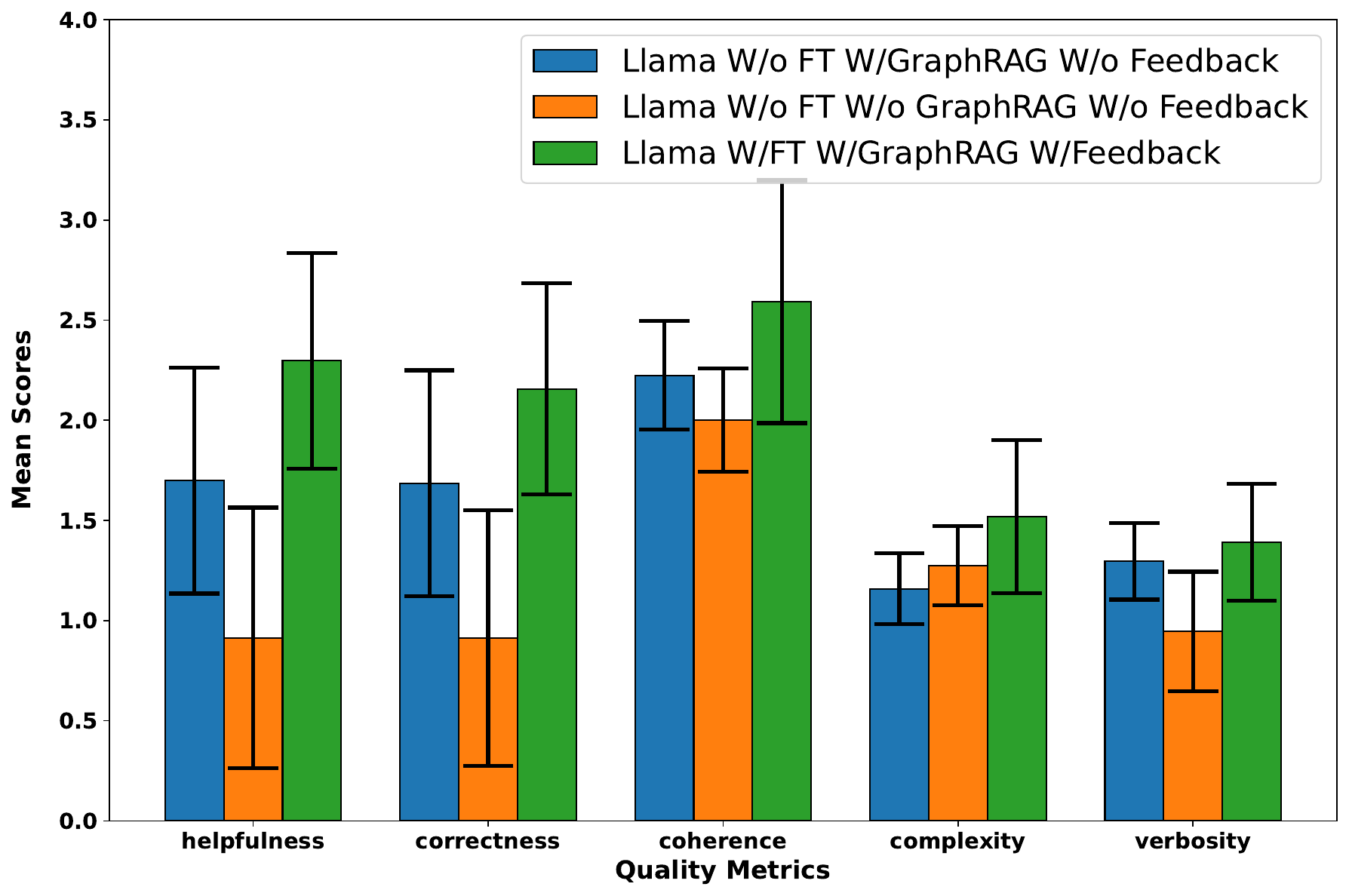}}
}
\resizebox{0.425\textwidth}{!}{
\subfloat[Quantitative comparison of fine-tuned Llama-3.2 1B and SmolLM2-135M using BLEU, METEOR, ROUGE, SacreBLEU, BERTScore, and cosine similarity metrics on the ChemEval benchmark. Llama-3.2 1B achieves superior overlap-based scores, while both models demonstrate comparable semantic similarity performance.]{\includegraphics[width=50mm]{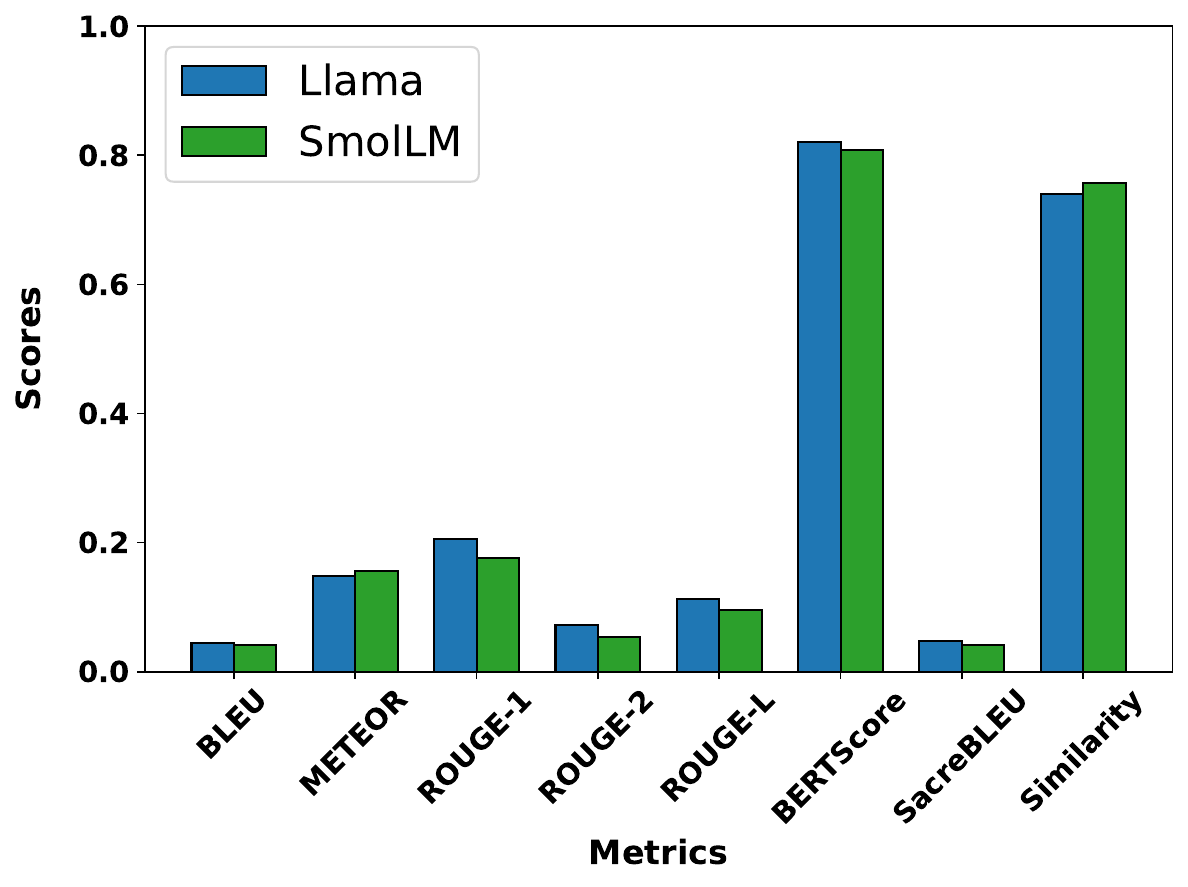}}
}
\vspace{-1mm}
\caption{Comprehensive evaluation of model performance on the ChemEval benchmark for automatic PFD/PID generation. (a) Compares GPT-4o, fine-tuned Llama-3.2 1B, and fine-tuned SmolLM2-135M using reward model evaluation (ranked by performance). (b) Analyzes the impact of fine-tuning, GraphRAG, and feedback components on Llama-3.2 1B performance. (c) Benchmarks Llama-3.2 1B against SmolLM2-135M using standard NLP metrics.}
\label{fig:uc}
\vspace{-3mm}
\end{figure*}

\begin{figure*}[ht!]
    \centering
    \begin{tcolorbox}[colframe=blue!70, colback=blue!10,arc=5mm,width=140mm]
        % Outer box
        \begin{center}
            \includegraphics[width=130mm,trim=0.5cm 6.0cm 0cm 6.0cm,clip]{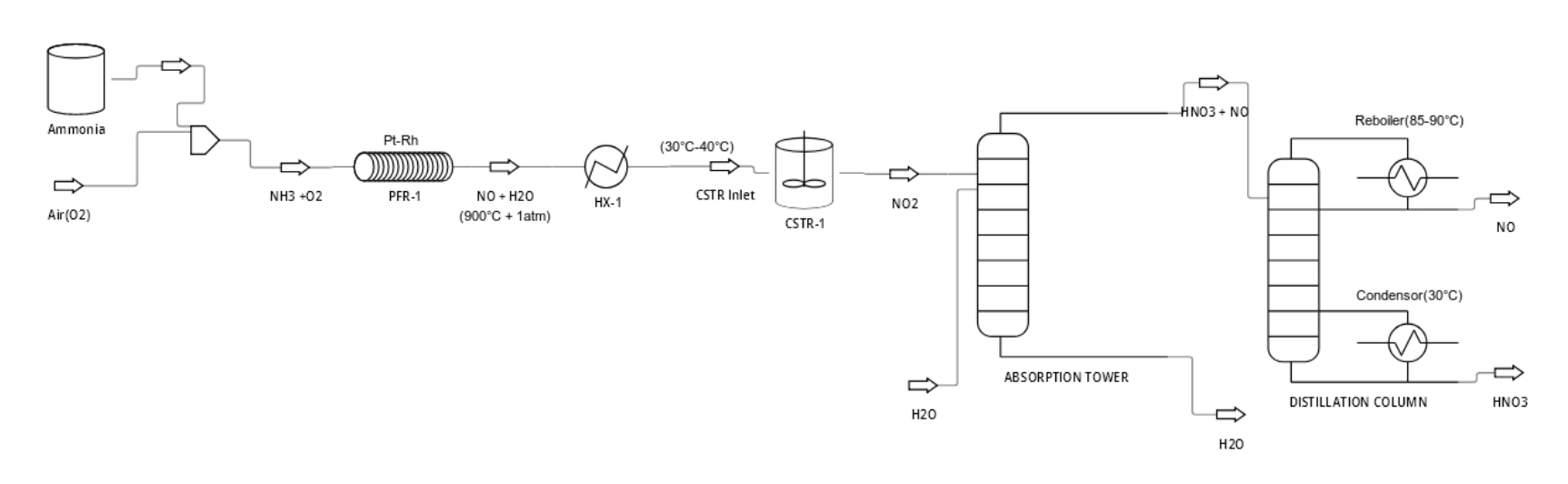} % Replace with your image file
        \end{center}
        \begin{tcolorbox}[colframe=blue!50, colback=white,arc=2mm,width=130mm]
        % Inner box
        The production of nitric acid (HNO\textsubscript{3}) follows a systematic sequence of steps. The process begins with feed preparation, where ammonia (NH\textsubscript{3}) from storage and compressed air (O\textsubscript{2}) are introduced. Ammonia is stored at ambient temperature and atmospheric pressure, while air is compressed to 1–2 atm. In the next step, ammonia undergoes oxidation in a plug flow reactor (PFR) using a platinum-rhodium (Pt-Rh) catalyst, converting NH\textsubscript{3} and O\textsubscript{2} into nitric oxide (NO) and water vapor at 900\textdegree C and 9 atm. The hot gas stream is then cooled to 30–40\textdegree C using a heat exchanger (HX1). Nitric oxide (NO) is subsequently oxidized to nitrogen dioxide (NO\textsubscript{2}) in a continuous stirred tank reactor (CSTR) at atmospheric pressure and a temperature of 30–40\textdegree C. The resulting NO\textsubscript{2} gas is absorbed in water inside an absorption tower, where it reacts to form nitric acid (HNO\textsubscript{3}) and nitric oxide (NO) at 60–70\textdegree C and 1–2 atm. The nitric acid solution is then purified in a multi-stage distillation column, concentrating it to 60–68\% while separating impurities, with the reboiler operating at 85–90\textdegree C and the condenser at 30\textdegree C. Key operational conditions include maintaining optimal temperatures and pressures in reactors and separation units to enhance efficiency. This optimized nitric acid production process ensures high efficiency, minimizes environmental impact, and is well-suited for large-scale industrial applications. 
        \end{tcolorbox}
    \end{tcolorbox}
\caption{The figure shows the nitric acid (HNO\textsubscript{3}) PFD showing key unit operations (NH\textsubscript{3} oxidation, NO/NO\textsubscript{2} conversion, absorption, distillation) with operating conditions. Generated in DWSIM from framework text.}
\label{fig:nitric}
\end{figure*}

\begin{figure*}[ht!]
    \centering
    \begin{tcolorbox}[colframe=blue!70, colback=blue!10,arc=5mm,width=140mm]
        % Outer box
        \begin{center}
            \includegraphics[width=130mm,trim=0.5cm 6.0cm 0cm 6.0cm,clip]{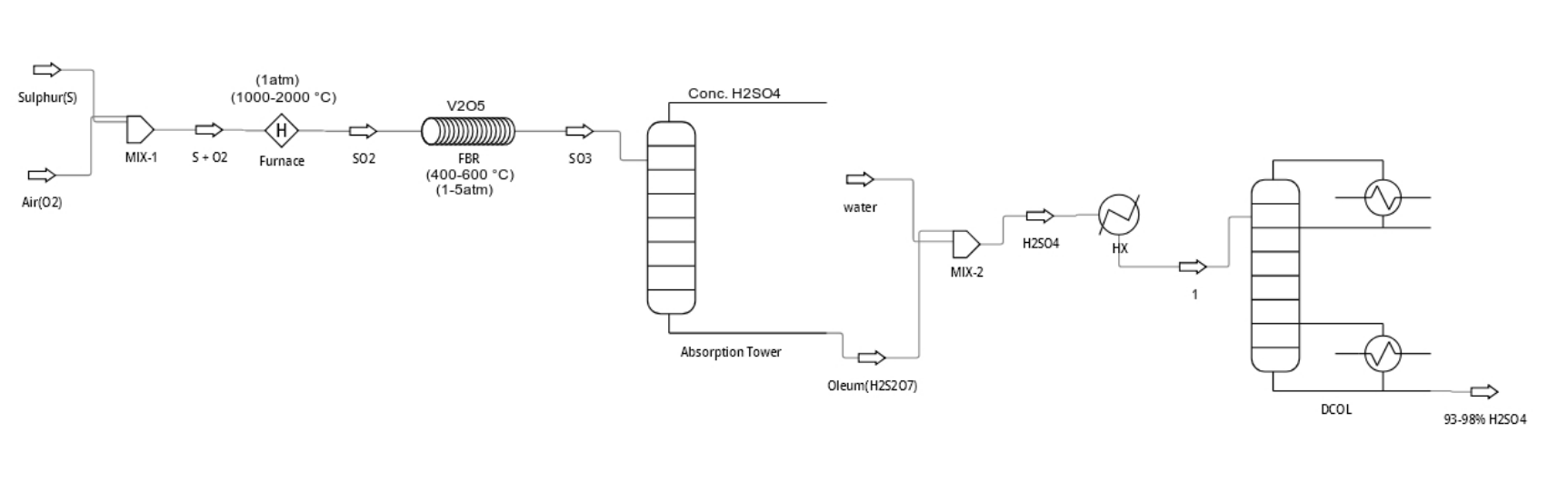} % Replace with your image file
        \end{center}
        \begin{tcolorbox}[colframe=blue!50, colback=white,arc=2mm,width=130mm]
            % Inner box
            The Contact Process for sulfuric acid (H\textsubscript{2}SO\textsubscript{4}) synthesis involves several key steps: sulfur combustion, sulfur dioxide oxidation, sulfur trioxide absorption, oleum dilution, and final purification. Initially, elemental sulfur (S) combusts with oxygen (O\textsubscript{2}) in a furnace at 1000–1200\textdegree C under atmospheric pressure, producing sulfur dioxide (SO\textsubscript{2}). The SO\textsubscript{2} then enters a series of fixed-bed reactors, where it undergoes catalytic oxidation with vanadium pentoxide (V\textsubscript{2}O\textsubscript{5}) at 400–600\textdegree C and 1–5 atm to form sulfur trioxide (SO\textsubscript{3}). Next, SO\textsubscript{3} is absorbed in concentrated sulfuric acid within a packed absorption tower at 30–60\textdegree C, forming oleum (H\textsubscript{2}S\textsubscript{2}O\textsubscript{7}). The oleum is then diluted with water in a mixing tank to produce concentrated sulfuric acid. A heat exchanger cools the reactor effluents, and a distillation column purifies the final product, yielding 93–98\% pure H\textsubscript{2}SO\textsubscript{4}. Safety measures include gas detection, automated controls, emergency protocols, and corrosion-resistant materials. Potential bottlenecks include catalyst deactivation in fixed-bed reactors, foaming in absorption towers, and inefficient heat recovery. This optimized process flow ensures efficient, large-scale sulfuric acid production with energy recovery and environmental sustainability.  
        \end{tcolorbox}
    \end{tcolorbox}
\caption{The figure illustrates the PFD of sulfuric acid (H\textsubscript{2}SO\textsubscript{4}) production, dynamically simulated in DWSIM. It details critical stages—including sulfur (S) combustion, catalytic SO\textsubscript{2} oxidation, SO\textsubscript{3} absorption, and oleum (H\textsubscript{2}S\textsubscript{2}O\textsubscript{7}) dilution—along with associated operating parameters (temperature, pressure, flow rates).}
\label{fig:sulfuric}
\end{figure*}

\begin{figure*}[ht!]
    \centering
    \begin{tcolorbox}[colframe=blue!70, colback=blue!10,arc=5mm,width=140mm]
        \begin{center}
            \includegraphics[width=130mm,trim=0.0cm 4.0cm 0cm 4.0cm,clip]{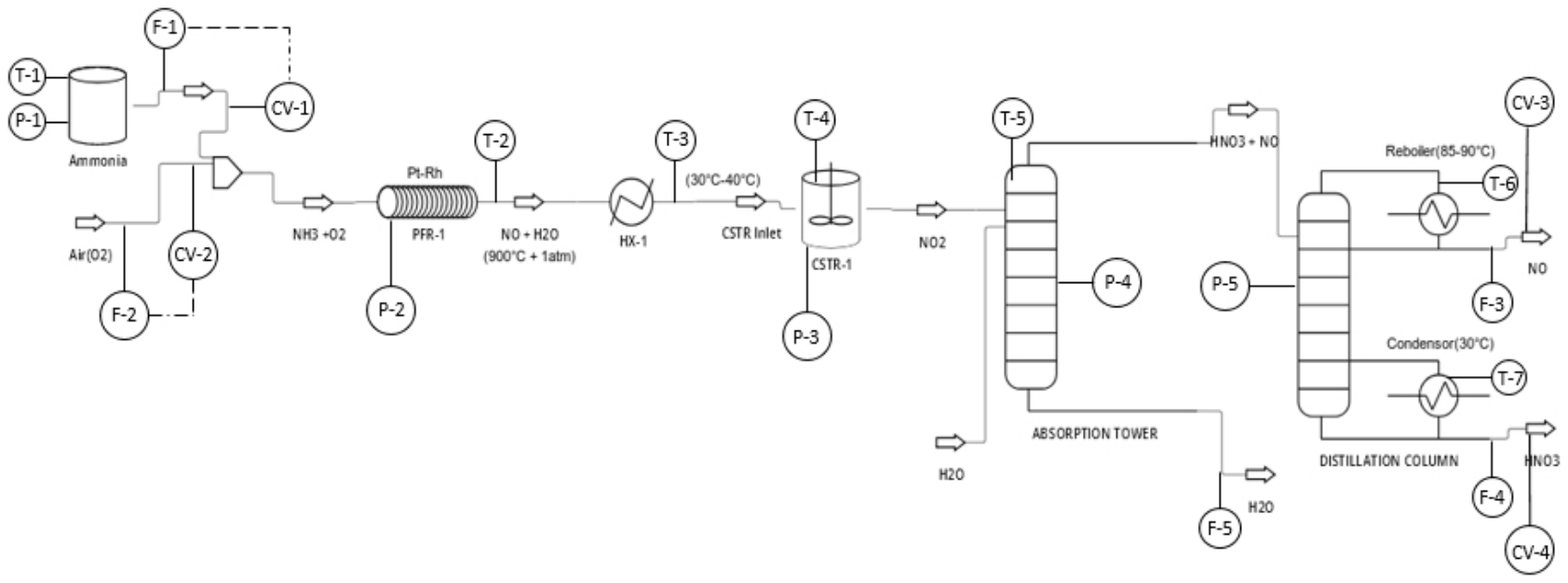}
        \end{center}
        \begin{tcolorbox}[colframe=blue!50, colback=white,arc=2mm,width=130mm]
        The optimized PID for nitric acid synthesis via the Ostwald process presents a comprehensive layout incorporating essential components, sensors, control mechanisms, and safety systems to facilitate efficient process monitoring and compliance with industry regulations. The system comprises key equipment such as the Ammonia Storage Tank, Air Compressor, Plug Flow Reactor, Heat Exchangers, Continuous Stirred Tank Reactor, Absorption Tower, Distillation Column, Gas Recycling System, and Wastewater Treatment Unit. Instrumentation includes temperature sensors (T-1 to T-7) placed at the ammonia tank, PFR outlet, HX1 outlet, CSTR, absorption tower, and distillation column reboiler and condenser; pressure sensors (P-1 to P-5) at critical points such as the ammonia tank, PFR, CSTR, absorption tower, and distillation column; flow meters (F-1 to F-5) for monitoring ammonia feed, air feed, NO and nitric acid product flow, and wastewater; and level sensors (L-1, L-2) for the ammonia and nitric acid storage tanks. The control infrastructure features valves (CV-1 to CV-4) to regulate ammonia and air feeds, NO, and nitric acid flow, with electric or pneumatic actuators deployed as required. Control strategies employ feedback control via PID controllers to stabilize PFR and CSTR temperatures and pressures, feedforward control to adjust downstream conditions based on upstream flow, and cascade control for distillation column temperature regulation. Recommended piping materials include carbon steel with coatings or stainless steel (e.g., 316L) for ammonia and NO, glass-lined or high-alloy stainless steel (e.g., Hastelloy) for nitric acid, and titanium or stainless steel for heat exchangers handling corrosive streams.          
        \end{tcolorbox}
    \end{tcolorbox}
\caption{The figure shows the PID for nitric acid production via the Ostwald process, generated using Visual Paradigm Online. The diagram highlights key process units—including the ammonia storage tank, plug flow reactor (PFR), absorption tower, and distillation column—along with instrumentation (temperature, pressure, flow, and level sensors) and control systems (valves, PID controllers, and cascade control). The design reflects process monitoring requirements and compliance with industry standards..}
\label{fig:nitricpid}
\end{figure*}

\begin{figure*}[ht!]
   \centering
    \begin{tcolorbox}[colframe=blue!70, colback=blue!10,arc=5mm,width=140mm]
        \begin{center}
            \includegraphics[width=130mm,trim=0.5cm 4.0cm 0cm 4.0cm,clip]{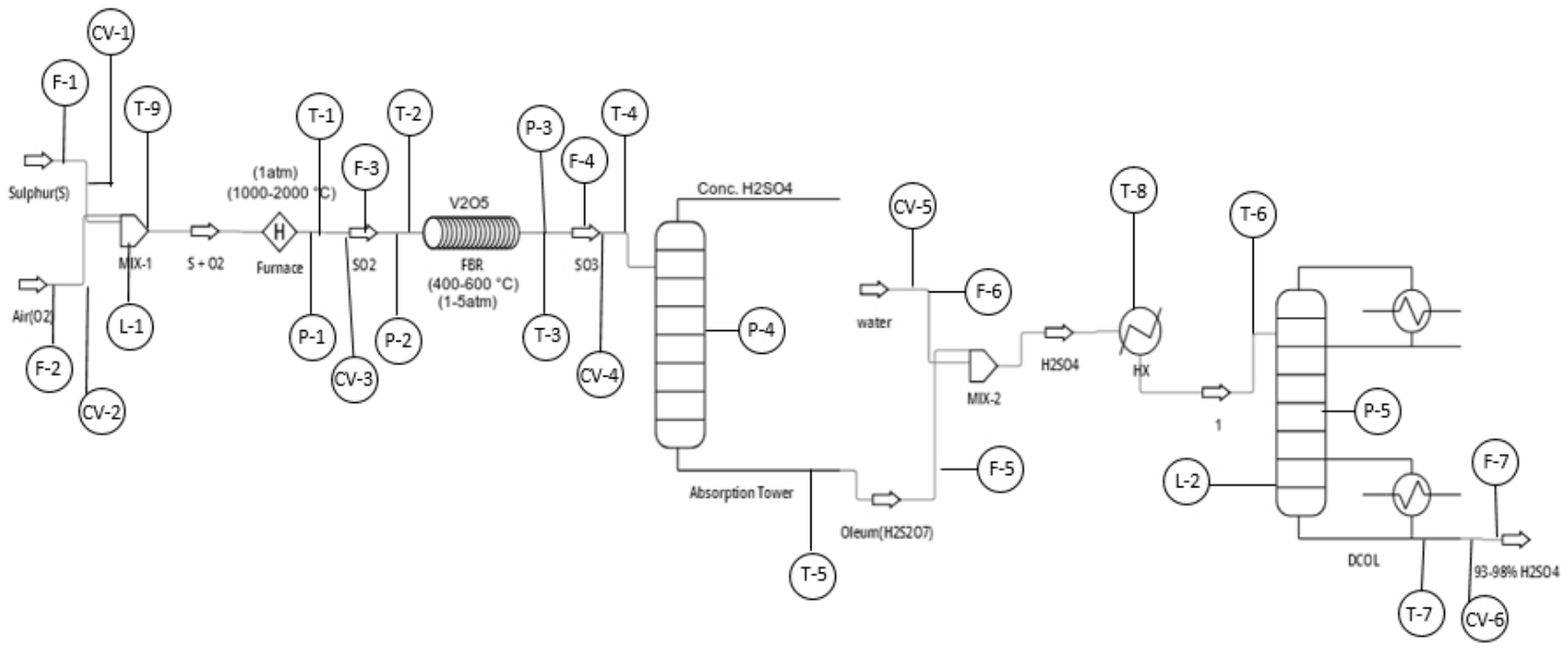}
        \end{center}
        \begin{tcolorbox}[colframe=blue!50, colback=white,arc=2mm,width=130mm]
        Creating an optimized PID for the synthesis of sulfuric acid via the Contact Process involves integrating best practices, emphasizing critical sensors, control elements, redundancy, reliability, piping materials, and control systems integration. The equipment and piping layout should include a multi-tube furnace for sulfur combustion, a series of fixed-bed reactors with heat exchangers for SO$_2$ oxidation, and a packed absorption tower with cooling jackets for absorbing SO$_3$ into concentrated H$_2$SO$_4$. A mixing tank for oleum dilution must be equipped with level sensors and flow control for water and oleum, while a heat exchanger is needed for cooling and heat recovery, monitored by temperature and flow sensors. The system should also feature a distillation column with reboiler and condenser controls for sulfuric acid purification, a scrubber system with gas detection for unreacted SO$_2$, and a filtration system for removing solid impurities. Instrumentation must include temperature sensors (T1 to T9) at critical points such as the furnace outlet, reactor inlets/outlets, absorption tower, distillation column, heat exchanger, and mixing tank. Pressure sensors (P1 to P5) should be installed at the furnace outlet, reactors, absorption tower, and distillation column, while flow sensors (F1 to F7) should monitor sulfur, air, SO$_2$, SO$_3$, oleum, water, and final H$_2$SO$_4$ flows. Level sensors (L1 and L2) should monitor the mixing tank and distillation column sump. Control valves (CV1 to CV6) must regulate feeds of sulfur, air, SO$_2$, SO$_3$, water, and oleum, operated by electric or pneumatic actuators for fast, reliable responses.  Control strategies should include feedback control through PID loops for temperature and pressure in critical areas, feedforward control to adjust sulfur and air feed rates based on production goals and data analytics, and cascade control for reactor pressure with temperature as the inner loop. Safety instrumentation is vital. In conclusion, this optimized PID framework for sulfuric acid synthesis via the Contact Process ensures efficient, safe, and reliable industrial-scale production. Incorporating redundancy, advanced control, and real-time monitoring significantly enhances both operational efficiency and safety. 
        \end{tcolorbox}
    \end{tcolorbox}
\caption{The figure presents the PID for sulfuric acid production via the Contact Process, created using Visual Paradigm Online from framework-generated descriptions. It highlights core equipment including the multi-tube furnace, fixed-bed reactors, absorption tower, and distillation column, along with critical instrumentation (temperature/pressure/flow sensors, control valves) and control strategies (PID loops, feedforward control) for efficient, safe operation.}
\label{fig:sulfuricpid}
\end{figure*}

Figure~\ref{fig:uc} presents a comprehensive evaluation of customized SLMs on the \textit{ChemEval} benchmark for automatic PFD/PID generation, using the NVIDIA/Nemotron-4-340B reward model and standard NLP metrics. Note: Ground truth for the \textit{ChemEval} benchmark is generated using OpenAI's advanced reasoning models o3/o3‑mini. We compare fine-tuned Llama-3.2 1B and SmolLM2-135M against GPT-4o to assess zero-shot generation quality. Figure~\ref{fig:uc}(a) reports mean reward scores (0–4 scale) across five dimensions: helpfulness, correctness, coherence, complexity, and verbosity. GPT-4o establishes the performance upper bound, while Llama-3.2 1B achieves the second-best results, outperforming SmolLM2-135M in helpfulness and coherence with more concise outputs but greater variance. SmolLM2-135M scores lowest overall yet performs comparably in complexity and verbosity. Figure~\ref{fig:uc}(b) examines architectural components within Llama-3.2 1B across three configurations: the base pretrained model, the model with GraphRAG, and the fully enhanced variant with fine-tuning, GraphRAG, and feedback. Both retrieval and feedback contribute independently to performance improvements, with their combination yielding the strongest gains. Figure~\ref{fig:uc}(c) presents quantitative evaluation using BLEU, METEOR, ROUGE, SacreBLEU, BERTScore, and cosine similarity. Llama-3.2 1B achieves higher overlap-based scores, while both models demonstrate strong semantic similarity alignment, confirming that appropriately fine-tuned smaller LLMs can preserve semantic fidelity.
Figures~\ref{fig:nitric} and~\ref{fig:sulfuric} present high-level PFDs for nitric acid and sulfuric acid production, respectively. These diagrams were constructed in DWSIM based on textual outputs generated by our framework and manually assembled using DWSIM’s unit operation blocks, thermodynamic models, and stream configuration tools. The nitric acid PFD (Figure~\ref{fig:nitric}) illustrates a structured sequence of operations, beginning with feed mixing and catalytic oxidation, followed by gas cooling, intermediate conversion, absorption, and final distillation—all represented through interconnected unit operations and material flow paths. Similarly, the sulfuric acid PFD (Figure~\ref{fig:sulfuric}) outlines key stages, including sulfur combustion, catalytic oxidation, $\text{SO}_3$ absorption, oleum dilution, and product purification, arranged in a logical progression of process units. Figures~\ref{fig:nitricpid} and~\ref{fig:sulfuricpid} illustrate PIDs for the industrial synthesis of nitric acid via the Ostwald process and sulfuric acid via the Contact Process, respectively. Each diagram details key equipment, instrumentation (temperature, pressure, flow, and level sensors), control elements (valves, PID controllers, cascade and feedforward strategies), and piping materials—all designed to ensure efficient, safe, and regulation-compliant chemical production. These flowsheets reflect realistic industrial workflows and were configured in DWSIM for simulation-based verification. The resulting simulations enable rigorous evaluation of material and energy balances, phase behavior, and equipment performance. By translating language-model-generated flowsheet descriptions into executable DWSIM simulations, we ensure engineering feasibility, identify configuration issues, and support process optimization in accordance with fundamental chemical engineering principles.  

\vspace{-3mm}
\section{Conclusion}
\vspace{0mm}
Automating the generation of industrially viable PFDs and PIDs is critical for accelerating chemical process scale-up. Current AI-assisted drug and materials discovery pipelines often prioritize molecular property optimization while neglecting production feasibility. Integrating early-stage auto-generation and validation of PFDs—which capture unit operations, material balances, and thermodynamic consistency—and PIDs—which define instrumentation, control logic, and safety systems—enables process simulation tools to detect scale-up conflicts in equipment sizing, utility demands, and hazardous material handling before experimental work begins. This proactive, concurrent design-for-manufacturing approach replaces post-hoc feasibility checks, mitigating late-stage reengineering risks. Our closed-loop framework addresses this gap by integrating domain-adapted small language models (SLMs) with physics-aware validation to enable end-to-end automation. The approach combines multi-stage SLM fine-tuning—leveraging synthetic datasets and retrieval augmentation from a hierarchical chemical knowledge graph—with rigorous simulation-based verification using DWSIM. Results demonstrate the framework’s robust performance in zero-shot synthesis of novel chemical production processes and its strong capabilities in core engineering QA tasks, including PFD/PID interpretation and analysis. By unifying generative AI with first-principles engineering constraints, the framework effectively bridges the gap between digital discovery and industrial deployment, addressing key R\&D bottlenecks.

%%%%%%%%%%%%%%%%%%%%%%%%%%%%%% Unknown Chemicals  %%%%%%%%%%%%%%%%%%%%%%%%%%%%%%%

% %% the bibliography file.
\bibliography{example_paper}

\begin{thebibliography}{65}
\providecommand{\natexlab}[1]{#1}
\providecommand{\url}[1]{\texttt{#1}}
\expandafter\ifx\csname urlstyle\endcsname\relax
  \providecommand{\doi}[1]{doi: #1}\else
  \providecommand{\doi}{doi: \begingroup \urlstyle{rm}\Url}\fi

\bibitem[Abbott \& Zardini(2024)Abbott and Zardini]{abbott2024flashattention}
Abbott, V. and Zardini, G.
\newblock Flashattention on a napkin: A diagrammatic approach to deep learning io-awareness.
\newblock \emph{arXiv preprint arXiv:2412.03317}, 2024.

\bibitem[Alimin et~al.(2025)Alimin, Goldstein, Balhorn, and Schweidtmann]{alimin2025talking}
Alimin, A.~A., Goldstein, D.~P., Balhorn, L.~S., and Schweidtmann, A.~M.
\newblock Talking like piping and instrumentation diagrams (p\&ids).
\newblock \emph{arXiv preprint arXiv:2502.18928}, 2025.

\bibitem[Balachandran et~al.(2025)Balachandran, Chen, Chen, Garg, Joshi, Lara, Langford, Nushi, Vineet, Wu, et~al.]{balachandran2025inference}
Balachandran, V., Chen, J., Chen, L., Garg, S., Joshi, N., Lara, Y., Langford, J., Nushi, B., Vineet, V., Wu, Y., et~al.
\newblock Inference-time scaling for complex tasks: Where we stand and what lies ahead.
\newblock \emph{arXiv preprint arXiv:2504.00294}, 2025.

\bibitem[Bi et~al.(2024)Bi, Han, Liu, Tang, and Wang]{bi2024forest}
Bi, Z., Han, K., Liu, C., Tang, Y., and Wang, Y.
\newblock Forest-of-thought: Scaling test-time compute for enhancing llm reasoning.
\newblock \emph{arXiv preprint arXiv:2412.09078}, 2024.

\bibitem[Chen et~al.(2025)Chen, Ren, Chen, Yang, Sun, and Ar{\i}k]{chen2025sets}
Chen, J., Ren, J., Chen, X., Yang, C., Sun, R., and Ar{\i}k, S.~{\"O}.
\newblock Sets: Leveraging self-verification and self-correction for improved test-time scaling.
\newblock \emph{arXiv preprint arXiv:2501.19306}, 2025.

\bibitem[Chen et~al.(2024)Chen, Liu, Wu, Zheng, Cong, Jiang, Wu, Su, and Yang]{chen2024int}
Chen, S., Liu, Z., Wu, Z., Zheng, C., Cong, P., Jiang, Z., Wu, Y., Su, L., and Yang, T.
\newblock Int-flashattention: Enabling flash attention for int8 quantization.
\newblock \emph{arXiv preprint arXiv:2409.16997}, 2024.

\bibitem[Chiang et~al.(2024)Chiang, Hsieh, Chou, and Riebesell]{chiang2024llamp}
Chiang, Y., Hsieh, E., Chou, C.-H., and Riebesell, J.
\newblock {LLAMP: Large language model made powerful for high-fidelity materials knowledge retrieval and distillation}.
\newblock \emph{arXiv preprint arXiv:2401.17244}, 2024.

\bibitem[Dao(2023)]{dao2023flashattention}
Dao, T.
\newblock Flashattention-2: Faster attention with better parallelism and work partitioning.
\newblock \emph{arXiv preprint arXiv:2307.08691}, 2023.

\bibitem[Dao et~al.(2022)Dao, Fu, Ermon, Rudra, and R{\'e}]{dao2022flashattention}
Dao, T., Fu, D., Ermon, S., Rudra, A., and R{\'e}, C.
\newblock Flashattention: Fast and memory-efficient exact attention with io-awareness.
\newblock \emph{Advances in neural information processing systems}, 35:\penalty0 16344--16359, 2022.

\bibitem[Dettmers et~al.(2023)Dettmers, Pagnoni, Holtzman, and Zettlemoyer]{dettmers2023qlora}
Dettmers, T., Pagnoni, A., Holtzman, A., and Zettlemoyer, L.
\newblock Qlora: Efficient finetuning of quantized llms.
\newblock \emph{Advances in neural information processing systems}, 36:\penalty0 10088--10115, 2023.

\bibitem[Edge et~al.(2024)Edge, Trinh, Cheng, Bradley, Chao, Mody, Truitt, Metropolitansky, Ness, and Larson]{edge2024local}
Edge, D., Trinh, H., Cheng, N., Bradley, J., Chao, A., Mody, A., Truitt, S., Metropolitansky, D., Ness, R.~O., and Larson, J.
\newblock From local to global: A graph rag approach to query-focused summarization.
\newblock \emph{arXiv preprint arXiv:2404.16130}, 2024.

\bibitem[Fu et~al.(2024)Fu, Bailis, Stoica, and Zhang]{fu2024break}
Fu, Y., Bailis, P., Stoica, I., and Zhang, H.
\newblock Break the sequential dependency of llm inference using lookahead decoding.
\newblock \emph{arXiv preprint arXiv:2402.02057}, 2024.

\bibitem[Gao et~al.(2024)Gao, Liu, Zhang, Du, and Xia]{gao2024bypass}
Gao, Y., Liu, Z., Zhang, W., Du, B., and Xia, G.-S.
\newblock Bypass back-propagation: Optimization-based structural pruning for large language models via policy gradient.
\newblock \emph{arXiv preprint arXiv:2406.10576}, 2024.

\bibitem[Gowiakar et~al.(2024)Gowiakar, Iyengar, Segal, and Kalyanaraman]{gowiakar2024agentic}
Gowiakar, S., Iyengar, S., Segal, S., and Kalyanaraman, S.
\newblock An agentic approach to automatic creation of p\&id diagrams from natural language descriptions.
\newblock \emph{arXiv preprint arXiv:2412.12898}, 2024.

\bibitem[Guo et~al.(2025)Guo, Yang, Zhang, Song, Zhang, Xu, Zhu, Ma, Wang, Bi, et~al.]{guo2025deepseek}
Guo, D., Yang, D., Zhang, H., Song, J., Zhang, R., Xu, R., Zhu, Q., Ma, S., Wang, P., Bi, X., et~al.
\newblock Deepseek-r1: Incentivizing reasoning capability in llms via reinforcement learning.
\newblock \emph{arXiv preprint arXiv:2501.12948}, 2025.

\bibitem[Guo \& Schwaller(2024)Guo and Schwaller]{guo2024saturn}
Guo, J. and Schwaller, P.
\newblock Saturn: Sample-efficient generative molecular design using memory manipulation.
\newblock \emph{arXiv preprint arXiv:2405.17066}, 2024.

\bibitem[Han et~al.(2024)Han, Wang, Shomer, Guo, Ding, Lei, Halappanavar, Rossi, Mukherjee, Tang, et~al.]{han2024retrieval}
Han, H., Wang, Y., Shomer, H., Guo, K., Ding, J., Lei, Y., Halappanavar, M., Rossi, R.~A., Mukherjee, S., Tang, X., et~al.
\newblock Retrieval-augmented generation with graphs (graphrag).
\newblock \emph{arXiv preprint arXiv:2501.00309}, 2024.

\bibitem[He et~al.(2024)He, Tian, Sun, Chawla, Laurent, LeCun, Bresson, and Hooi]{he2024g}
He, X., Tian, Y., Sun, Y., Chawla, N., Laurent, T., LeCun, Y., Bresson, X., and Hooi, B.
\newblock G-retriever: Retrieval-augmented generation for textual graph understanding and question answering.
\newblock \emph{Advances in Neural Information Processing Systems}, 37:\penalty0 132876--132907, 2024.

\bibitem[Hirretier et~al.(2022)Hirretier, Balhorn, and Schweidtmann]{hirretier2022towards}
Hirretier, E., Balhorn, L.~S., and Schweidtmann, A.~M.
\newblock Towards automatic generation of piping and instrumentation diagrams (p\&ids) with artificial intelligence.
\newblock \emph{arXiv preprint arXiv:2211.05583}, 2022.

\bibitem[Kang et~al.()Kang, Liu, and Guo]{kangretorinetext}
Kang, C., Liu, X., and Guo, F.
\newblock Retrointext: A multimodal large language model enhanced framework for retrosynthetic planning via in-context representation learning.
\newblock In \emph{The Thirteenth International Conference on Learning Representations}.

\bibitem[Kendapadi et~al.(2024)Kendapadi, Zaman, Menon, and Srivastava]{kendapadi2024interact}
Kendapadi, A., Zaman, K., Menon, R.~R., and Srivastava, S.
\newblock Interact: Enabling interactive, question-driven learning in large language models.
\newblock \emph{arXiv preprint arXiv:2412.11388}, 2024.

\bibitem[Kim et~al.(2024)Kim, Kim, Kim, Castells, Choi, Shin, and Song]{kim2024shortened}
Kim, B.-K., Kim, G., Kim, T.-H., Castells, T., Choi, S., Shin, J., and Song, H.-K.
\newblock Shortened llama: Depth pruning for large language models with comparison of retraining methods.
\newblock \emph{arXiv preprint arXiv:2402.02834}, 2024.

\bibitem[Kristiadi et~al.(2024)Kristiadi, Strieth-Kalthoff, Skreta, Poupart, Aspuru-Guzik, and Pleiss]{kristiadi2024sober}
Kristiadi, A., Strieth-Kalthoff, F., Skreta, M., Poupart, P., Aspuru-Guzik, A., and Pleiss, G.
\newblock A sober look at llms for material discovery: Are they actually good for bayesian optimization over molecules?
\newblock \emph{arXiv preprint arXiv:2402.05015}, 2024.

\bibitem[Kwon et~al.(2023)Kwon, Li, Zhuang, Sheng, Zheng, Yu, Gonzalez, Zhang, and Stoica]{kwon2023efficient}
Kwon, W., Li, Z., Zhuang, S., Sheng, Y., Zheng, L., Yu, C.~H., Gonzalez, J., Zhang, H., and Stoica, I.
\newblock Efficient memory management for large language model serving with pagedattention.
\newblock In \emph{Proceedings of the 29th Symposium on Operating Systems Principles}, pp.\  611--626, 2023.

\bibitem[Li(2025)]{li2025survey}
Li, X.
\newblock A survey on llm test-time compute via search: Tasks, llm profiling, search algorithms, and relevant frameworks.
\newblock \emph{arXiv preprint arXiv:2501.10069}, 2025.

\bibitem[Lin et~al.(2025)Lin, Lin, Xie, and Ji]{lin2025cppo}
Lin, Z., Lin, M., Xie, Y., and Ji, R.
\newblock Cppo: Accelerating the training of group relative policy optimization-based reasoning models.
\newblock \emph{arXiv preprint arXiv:2503.22342}, 2025.

\bibitem[Liu et~al.(2024)Liu, Feng, Xue, Wang, Wu, Lu, Zhao, Deng, Zhang, Ruan, et~al.]{liu2024deepseek}
Liu, A., Feng, B., Xue, B., Wang, B., Wu, B., Lu, C., Zhao, C., Deng, C., Zhang, C., Ruan, C., et~al.
\newblock Deepseek-v3 technical report.
\newblock \emph{arXiv preprint arXiv:2412.19437}, 2024.

\bibitem[Liu et~al.(2025{\natexlab{a}})Liu, Zhou, Xu, Huang, Wang, Zhang, Poon, and Chen]{liu2025metascale}
Liu, Q., Zhou, W., Xu, N., Huang, J.~Y., Wang, F., Zhang, S., Poon, H., and Chen, M.
\newblock Metascale: Test-time scaling with evolving meta-thoughts.
\newblock \emph{arXiv preprint arXiv:2503.13447}, 2025{\natexlab{a}}.

\bibitem[Liu et~al.(2025{\natexlab{b}})Liu, Wu, He, Gao, Chen, Bi, Zhang, Huang, and Hooi]{liu2025efficient}
Liu, Y., Wu, J., He, Y., Gao, H., Chen, H., Bi, B., Zhang, J., Huang, Z., and Hooi, B.
\newblock Efficient inference for large reasoning models: A survey.
\newblock \emph{arXiv preprint arXiv:2503.23077}, 2025{\natexlab{b}}.

\bibitem[Lu et~al.(2024)Lu, Zhou, Liu, Wang, Mahoney, and Yang]{lu2024alphapruning}
Lu, H., Zhou, Y., Liu, S., Wang, Z., Mahoney, M.~W., and Yang, Y.
\newblock Alphapruning: Using heavy-tailed self regularization theory for improved layer-wise pruning of large language models.
\newblock \emph{Advances in Neural Information Processing Systems}, 37:\penalty0 9117--9152, 2024.

\bibitem[Mamou et~al.(2024)Mamou, Pereg, Korat, Berchansky, Timor, Wasserblat, and Schwartz]{mamou2024dynamic}
Mamou, J., Pereg, O., Korat, D., Berchansky, M., Timor, N., Wasserblat, M., and Schwartz, R.
\newblock Dynamic speculation lookahead accelerates speculative decoding of large language models.
\newblock \emph{arXiv preprint arXiv:2405.04304}, 2024.

\bibitem[Medeiros(2025)]{dwsim}
Medeiros, D.
\newblock Dwsim: Open source process simulator, 2025.
\newblock URL \url{https://dwsim.fossee.in}.
\newblock Accessed April 15, 2025.

\bibitem[OpenAI(2024)]{openai2024textembedding3}
OpenAI.
\newblock text-embedding-3-small model.
\newblock \url{https://platform.openai.com/docs/guides/embeddings}, 2024.
\newblock Accessed: August 2024.

\bibitem[Pan et~al.(2024)Pan, Kwon, Liu, Xie, Duan, Prein, Sheriff, Roman, Moliner, G{\'o}mez-Bombarelli, et~al.]{pan2024chemically}
Pan, E., Kwon, S., Liu, S., Xie, M., Duan, Y., Prein, T., Sheriff, K., Roman, Y., Moliner, M., G{\'o}mez-Bombarelli, R., et~al.
\newblock A chemically-guided generative diffusion model for materials synthesis planning.
\newblock In \emph{AI for Accelerated Materials Design--NeurIPS 2024}, 2024.

\bibitem[Prabhu et~al.(2024)Prabhu, Nayak, Mohan, Ramjee, and Panwar]{prabhu2024vattention}
Prabhu, R., Nayak, A., Mohan, J., Ramjee, R., and Panwar, A.
\newblock vattention: Dynamic memory management for serving llms without pagedattention.
\newblock \emph{arXiv preprint arXiv:2405.04437}, 2024.

\bibitem[Qu et~al.(2025)Qu, Yang, Setlur, Tunstall, Beeching, Salakhutdinov, and Kumar]{qu2025optimizing}
Qu, Y., Yang, M.~Y., Setlur, A., Tunstall, L., Beeching, E.~E., Salakhutdinov, R., and Kumar, A.
\newblock Optimizing test-time compute via meta reinforcement fine-tuning.
\newblock \emph{arXiv preprint arXiv:2503.07572}, 2025.

\bibitem[Rawat et~al.(2024)Rawat, Sadhanala, Rostamizadeh, Chakrabarti, Jitkrittum, Feinberg, Kim, Harutyunyan, Saunshi, Nado, et~al.]{rawat2024little}
Rawat, A.~S., Sadhanala, V., Rostamizadeh, A., Chakrabarti, A., Jitkrittum, W., Feinberg, V., Kim, S., Harutyunyan, H., Saunshi, N., Nado, Z., et~al.
\newblock A little help goes a long way: Efficient llm training by leveraging small lms.
\newblock \emph{arXiv preprint arXiv:2410.18779}, 2024.

\bibitem[Rehg(2024)]{rehg2024kv}
Rehg, I.
\newblock Kv-compress: Paged kv-cache compression with variable compression rates per attention head.
\newblock \emph{arXiv preprint arXiv:2410.00161}, 2024.

\bibitem[Sandri et~al.(2025)Sandri, Cunegatti, and Iacca]{sandri20252ssp}
Sandri, F., Cunegatti, E., and Iacca, G.
\newblock 2ssp: A two-stage framework for structured pruning of llms.
\newblock \emph{arXiv preprint arXiv:2501.17771}, 2025.

\bibitem[Schulman et~al.(2017)Schulman, Wolski, Dhariwal, Radford, and Klimov]{schulman2017proximal}
Schulman, J., Wolski, F., Dhariwal, P., Radford, A., and Klimov, O.
\newblock Proximal policy optimization algorithms.
\newblock \emph{arXiv preprint arXiv:1707.06347}, 2017.

\bibitem[Schulze~Balhorn et~al.(2023)Schulze~Balhorn, Hirretier, Luderer, and Schweidtmann]{schulze2023data}
Schulze~Balhorn, L., Hirretier, E., Luderer, L., and Schweidtmann, A.~M.
\newblock Data augmentation for machine learning of chemical process flowsheets.
\newblock \emph{arXiv e-prints}, pp.\  arXiv--2302, 2023.

\bibitem[Shah et~al.(2024)Shah, Bikshandi, Zhang, Thakkar, Ramani, and Dao]{shah2024flashattention}
Shah, J., Bikshandi, G., Zhang, Y., Thakkar, V., Ramani, P., and Dao, T.
\newblock Flashattention-3: Fast and accurate attention with asynchrony and low-precision.
\newblock \emph{Advances in Neural Information Processing Systems}, 37:\penalty0 68658--68685, 2024.

\bibitem[Shao et~al.(2024)Shao, Wang, Zhu, Xu, Song, Bi, Zhang, Zhang, Li, Wu, et~al.]{shao2024deepseekmath}
Shao, Z., Wang, P., Zhu, Q., Xu, R., Song, J., Bi, X., Zhang, H., Zhang, M., Li, Y., Wu, Y., et~al.
\newblock Deepseekmath: Pushing the limits of mathematical reasoning in open language models.
\newblock \emph{arXiv preprint arXiv:2402.03300}, 2024.

\bibitem[Singhi et~al.(2025)Singhi, Bansal, Hosseini, Grover, Chang, Rohrbach, and Rohrbach]{singhi2025solve}
Singhi, N., Bansal, H., Hosseini, A., Grover, A., Chang, K.-W., Rohrbach, M., and Rohrbach, A.
\newblock When to solve, when to verify: Compute-optimal problem solving and generative verification for llm reasoning.
\newblock \emph{arXiv preprint arXiv:2504.01005}, 2025.

\bibitem[Snell et~al.(2024)Snell, Lee, Xu, and Kumar]{snell2024scaling}
Snell, C., Lee, J., Xu, K., and Kumar, A.
\newblock Scaling llm test-time compute optimally can be more effective than scaling model parameters.
\newblock \emph{arXiv preprint arXiv:2408.03314}, 2024.

\bibitem[Sprueill et~al.(2024)Sprueill, Edwards, Agarwal, Olarte, Sanyal, Johnston, Liu, Ji, and Choudhury]{sprueill2024chemreasoner}
Sprueill, H.~W., Edwards, C., Agarwal, K., Olarte, M.~V., Sanyal, U., Johnston, C., Liu, H., Ji, H., and Choudhury, S.
\newblock Chemreasoner: Heuristic search over a large language model's knowledge space using quantum-chemical feedback.
\newblock \emph{arXiv preprint arXiv:2402.10980}, 2024.

\bibitem[Srinivas et~al.(2024)Srinivas, Das, Gupta, and Runkana]{srinivas2024accelerating}
Srinivas, S.~S., Das, A., Gupta, S., and Runkana, V.
\newblock Accelerating manufacturing scale-up from material discovery using agentic web navigation and retrieval-augmented ai for process engineering schematics design.
\newblock \emph{arXiv preprint arXiv:2412.05937}, 2024.

\bibitem[Sun et~al.(2023)Sun, Liu, Bair, and Kolter]{sun2023simple}
Sun, M., Liu, Z., Bair, A., and Kolter, J.~Z.
\newblock A simple and effective pruning approach for large language models.
\newblock \emph{arXiv preprint arXiv:2306.11695}, 2023.

\bibitem[Sun et~al.(2025)Sun, Song, Li, Yin, Zheng, and Liu]{sun2025curse}
Sun, W., Song, X., Li, P., Yin, L., Zheng, Y., and Liu, S.
\newblock The curse of depth in large language models.
\newblock \emph{arXiv preprint arXiv:2502.05795}, 2025.

\bibitem[Tang et~al.(2025)Tang, Sieberling, Kurtic, Shen, and Alistarh]{tang2025darwinlm}
Tang, S., Sieberling, O., Kurtic, E., Shen, Z., and Alistarh, D.
\newblock Darwinlm: Evolutionary structured pruning of large language models.
\newblock \emph{arXiv preprint arXiv:2502.07780}, 2025.

\bibitem[Tian et~al.(2025)Tian, Han, Chen, Wang, and Chawla]{tian2025beyond}
Tian, Y., Han, Y., Chen, X., Wang, W., and Chawla, N.~V.
\newblock Beyond answers: Transferring reasoning capabilities to smaller llms using multi-teacher knowledge distillation.
\newblock In \emph{Proceedings of the Eighteenth ACM International Conference on Web Search and Data Mining}, pp.\  251--260, 2025.

\bibitem[Traag et~al.(2019)Traag, Waltman, and Van~Eck]{traag2019louvain}
Traag, V.~A., Waltman, L., and Van~Eck, N.~J.
\newblock From louvain to leiden: guaranteeing well-connected communities.
\newblock \emph{Scientific reports}, 9\penalty0 (1):\penalty0 1--12, 2019.

\bibitem[Vogel et~al.(2023)Vogel, Balhorn, and Schweidtmann]{vogel2023learning}
Vogel, G., Balhorn, L.~S., and Schweidtmann, A.~M.
\newblock Learning from flowsheets: A generative transformer model for autocompletion of flowsheets.
\newblock \emph{Computers \& Chemical Engineering}, 171:\penalty0 108162, 2023.

\bibitem[Wang et~al.(2024)Wang, Skreta, Ser, Gao, Kong, Strieth-Kalthoff, Duan, Zhuang, Yu, Zhu, et~al.]{wang2024efficient}
Wang, H., Skreta, M., Ser, C.-T., Gao, W., Kong, L., Strieth-Kalthoff, F., Duan, C., Zhuang, Y., Yu, Y., Zhu, Y., et~al.
\newblock Efficient evolutionary search over chemical space with large language models.
\newblock \emph{arXiv preprint arXiv:2406.16976}, 2024.

\bibitem[Wu(2024)]{wu2024llm}
Wu, H.
\newblock Llm-bip: Structured pruning for large language models with block-wise forward importance propagation.
\newblock \emph{arXiv preprint arXiv:2412.06419}, 2024.

\bibitem[Xu et~al.(2023)Xu, Xie, Gu, Chen, Chang, Zhang, Chen, Zhang, and Tian]{xu2023qa}
Xu, Y., Xie, L., Gu, X., Chen, X., Chang, H., Zhang, H., Chen, Z., Zhang, X., and Tian, Q.
\newblock Qa-lora: Quantization-aware low-rank adaptation of large language models.
\newblock \emph{arXiv preprint arXiv:2309.14717}, 2023.

\bibitem[Yang et~al.(2025{\natexlab{a}})Yang, Song, Han, Bi, Wang, Liang, Song, Zhang, Niu, Peng, et~al.]{yang2025feature}
Yang, J., Song, J., Han, X., Bi, Z., Wang, T., Liang, C.~X., Song, X., Zhang, Y., Niu, Q., Peng, B., et~al.
\newblock Feature alignment and representation transfer in knowledge distillation for large language models.
\newblock \emph{arXiv preprint arXiv:2504.13825}, 2025{\natexlab{a}}.

\bibitem[Yang et~al.(2024)Yang, Batzner, Gao, Aykol, Gaunt, McMorrow, Jimenez~Rezende, Schuurmans, Mordatch, and Cubuk]{yang2024generative}
Yang, S., Batzner, S., Gao, R., Aykol, M., Gaunt, A., McMorrow, B.~C., Jimenez~Rezende, D., Schuurmans, D., Mordatch, I., and Cubuk, E.~D.
\newblock Generative hierarchical materials search.
\newblock \emph{Advances in Neural Information Processing Systems}, 37:\penalty0 38799--38819, 2024.

\bibitem[Yang et~al.(2025{\natexlab{b}})Yang, Ma, Lin, and Wei]{yang2025towards}
Yang, W., Ma, S., Lin, Y., and Wei, F.
\newblock Towards thinking-optimal scaling of test-time compute for llm reasoning.
\newblock \emph{arXiv preprint arXiv:2502.18080}, 2025{\natexlab{b}}.

\bibitem[Yu et~al.(2025)Yu, Wu, Zhao, Cohan, and Zhang]{yu2025z1}
Yu, Z., Wu, Y., Zhao, Y., Cohan, A., and Zhang, X.-P.
\newblock Z1: Efficient test-time scaling with code.
\newblock \emph{arXiv preprint arXiv:2504.00810}, 2025.

\bibitem[Zhang et~al.(2024)Zhang, Song, Hou, Miret, and Liu]{zhang2024honeycomb}
Zhang, H., Song, Y., Hou, Z., Miret, S., and Liu, B.
\newblock Honeycomb: A flexible llm-based agent system for materials science.
\newblock \emph{arXiv preprint arXiv:2409.00155}, 2024.

\bibitem[Zhang et~al.(2025)Zhang, Lyu, Sun, Wang, Zhang, Guo, Wang, King, Liu, and Ma]{zhang2025and}
Zhang, Q., Lyu, F., Sun, Z., Wang, L., Zhang, W., Guo, Z., Wang, Y., King, I., Liu, X., and Ma, C.
\newblock What, how, where, and how well? a survey on test-time scaling in large language models.
\newblock \emph{arXiv preprint arXiv:2503.24235}, 2025.

\bibitem[Zhao et~al.(2024)Zhao, Xie, Liang, Zhuang, and Gu]{zhao2024lookahead}
Zhao, Y., Xie, Z., Liang, C., Zhuang, C., and Gu, J.
\newblock Lookahead: An inference acceleration framework for large language model with lossless generation accuracy.
\newblock In \emph{Proceedings of the 30th ACM SIGKDD Conference on Knowledge Discovery and Data Mining}, pp.\  6344--6355, 2024.

\bibitem[Zhong et~al.(2023)Zhong, An, Chen, Han, and He]{zhong2023seeking}
Zhong, M., An, C., Chen, W., Han, J., and He, P.
\newblock Seeking neural nuggets: Knowledge transfer in large language models from a parametric perspective.
\newblock \emph{arXiv preprint arXiv:2310.11451}, 2023.

\bibitem[Zhu et~al.(2024)Zhu, Li, Liu, Ma, and Wang]{zhu2024survey}
Zhu, X., Li, J., Liu, Y., Ma, C., and Wang, W.
\newblock A survey on model compression for large language models.
\newblock \emph{Transactions of the Association for Computational Linguistics}, 12:\penalty0 1556--1577, 2024.

\end{thebibliography}
\bibliographystyle{icml2025}

\clearpage
\newpage

% If your work has an appendix, this is the place to put it.
\section{Technical Appendix}

%%%%%%%%%%%%%%%%%%%%%%%%%%%%%%%%%%%%%%%%%%%%%%%%%%%%%%%%%%%%%%%%%%%%%%%%%%

%%%%%%%%%%%%%%%%%%%%%%%%%%%%%%%%%%%%%%%%%%%%%%%%%%%%%%%%%%%%%%%%%%%%%%%%%%

\vspace{-3mm}
\subsection{Agentic Web Search for Automated Extraction and Synthesis of PFD/PID Descriptions for Chemical Processes}  
PFDs and PIDs are fundamental engineering schematics in the chemical process industry, serving as the primary graphical representations of chemical plants. A Process Flow Diagram (PFD) provides a high-level overview of a plant’s major process units, piping, and material/energy flows, illustrating the transformation of raw materials into final products. In contrast, a Piping and Instrumentation Diagram (PID) offers a detailed schematic of mechanical components, including valves, instrumentation, and control systems, which are essential for safe and efficient operation. To generate textual descriptions of PFDs and PIDs for chemical processes in the ChemAtlas database, we employ agentic web navigation—an advanced autonomous framework for web-based information retrieval. This system scrapes, parses, and synthesizes process engineering information from open-access web sources to build foundational knowledge about established manufacturing processes. The framework generates structured textual descriptions of process designs, including: PFDs (equipment layouts, stream connections, mass/energy balances) and PIDs (instrumentation tags, control logic, safety interlocks). At the core of the agentic web search framework is a meta-agent responsible for query decomposition, task delegation, and response integration. Given a complex input query \( Q \), the meta-agent decomposes it into a set of subtasks \( \{q_1, q_2, \ldots, q_n\} \), where each subtask represents a semantically coherent information need. For each subtask \( q_i \), the meta-agent selects the optimal expert agent---such as the Visual Miner Agent, Research Agent, Patent Agent, or Wiki Agent---based on the highest semantic similarity between the vector representation of the subtask and that of the agent's capability. This approach goes beyond naïve task-to-tool mapping by embedding both task intent and agent capabilities into a shared semantic space, enabling principled and adaptive agent selection.

\vspace{-1mm} 
\resizebox{0.985\linewidth}{!}{ 
\begin{minipage}{\linewidth} 
\begin{equation}
 t^*_j = \arg\max_{j} \text{sim}_{\text{cos}}(v(q_i), v(d_j)) \nonumber
\end{equation}
\vspace{-1mm}
where,
\vspace{-1mm}
\begin{equation}
 \text{sim}_{\text{cos}}(v(q_i), v(d_j)) = \frac{v(q_i) \cdot v(d_j)}{\|v(q_i)\| \|v(d_j)\|} \nonumber
\end{equation}  
\end{minipage}  
}

\vspace{1mm} 
Here, \( v(q_i) \) and \( v(d_j) \) denote the dense vector embeddings of the subtask and the expert agent's capabilities, respectively. The agent embedding \( v(d_j) \) encodes domain expertise (i.e., specialized knowledge and skills relevant to retrieving and interpreting information within a specific content domain), tool access (e.g., SerpAPI), and reasoning modality (e.g., extractive or abstractive). Each expert agent operates within a multimodal, domain-specific retrieval regime. The Visual Miner Agent uses SerpAPI to retrieve high-quality industrial schematics and parses them to generate semantic summaries using an LLM. The Research, Patent, and Wiki Agents also leverage SerpAPI to retrieve content from domain-specific corpora, including peer-reviewed scientific papers, technical reports, patents, and Wikipedia articles, respectively, and synthesize structured, contextual summaries using LLMs. Subtasks are then structured as nodes \( V = \{v_1, \ldots, v_n\} \) in a Directed Acyclic Graph (DAG) \( \mathcal{G} = (\mathcal{V}, \mathcal{E}) \), where edges \( e_{ij} \in \mathcal{E} \) represent precedence constraints. This introduces formalisms into agent planning, moving away from fixed chain-of-thought paths to dynamic computation graphs. The DAG allows for topological sorting, task parallelism, and dependency resolution, supporting robust and interpretable execution flows. In particular, when no edge exists between subtasks \( q_i \) and \( q_j \), their associated agents---such as the Visual Miner Agent, Research Agent, Patent Agent, or Wiki Agent---are executed in parallel to optimize latency and throughput. Each agent executes its assigned subtask \( q_i \), retrieving a set of \( k \) candidate results \( M = \{m_1, \ldots, m_k\} \), each scored using cosine similarity:

\resizebox{0.985\linewidth}{!}{
\begin{minipage}{\linewidth}
\begin{equation}
\text{sim}_{\text{cos}}(v(m_i), v(q_i)) = \frac{v(m_i) \cdot v(q_i)}{\|v(m_i)\| \|v(q_i)\|} \nonumber
\end{equation}
\end{minipage}
}

\vspace{1mm} 
The top-\( K \leq k \) candidates are selected by ranking the retrieved items \( m_i \in M \) in descending order of cosine similarity to \( v(q_i) \), retaining the most relevant results for language model-based synthesis. Each expert agent then leverages a language model to perform information synthesis, semantic abstraction, and contextual reasoning over the selected top-ranked results, producing a coherent sub-answer \( R_{q_i} \). The global answer \( A \) is constructed by integrating sub-answers: $ A = \mathcal{F}_{\text{Meta}}(\{R_{q_i}\}_{i=1}^{n})$. To enhance quality and alignment, the framework introduces an iterative refinement loop for a predefined number of iterations:

\vspace{-1mm} 
\resizebox{0.985\linewidth}{!}{
\begin{minipage}{\linewidth}
\begin{equation}
A_{i+1} = \mathcal{F}_{\text{Meta}}(A_i, F_i) \nonumber
\end{equation}
\end{minipage}
}

\begin{figure*}[ht!] % left bottom right top
\centering
\includegraphics[width=0.85\linewidth, trim=0 10 0 5, clip]{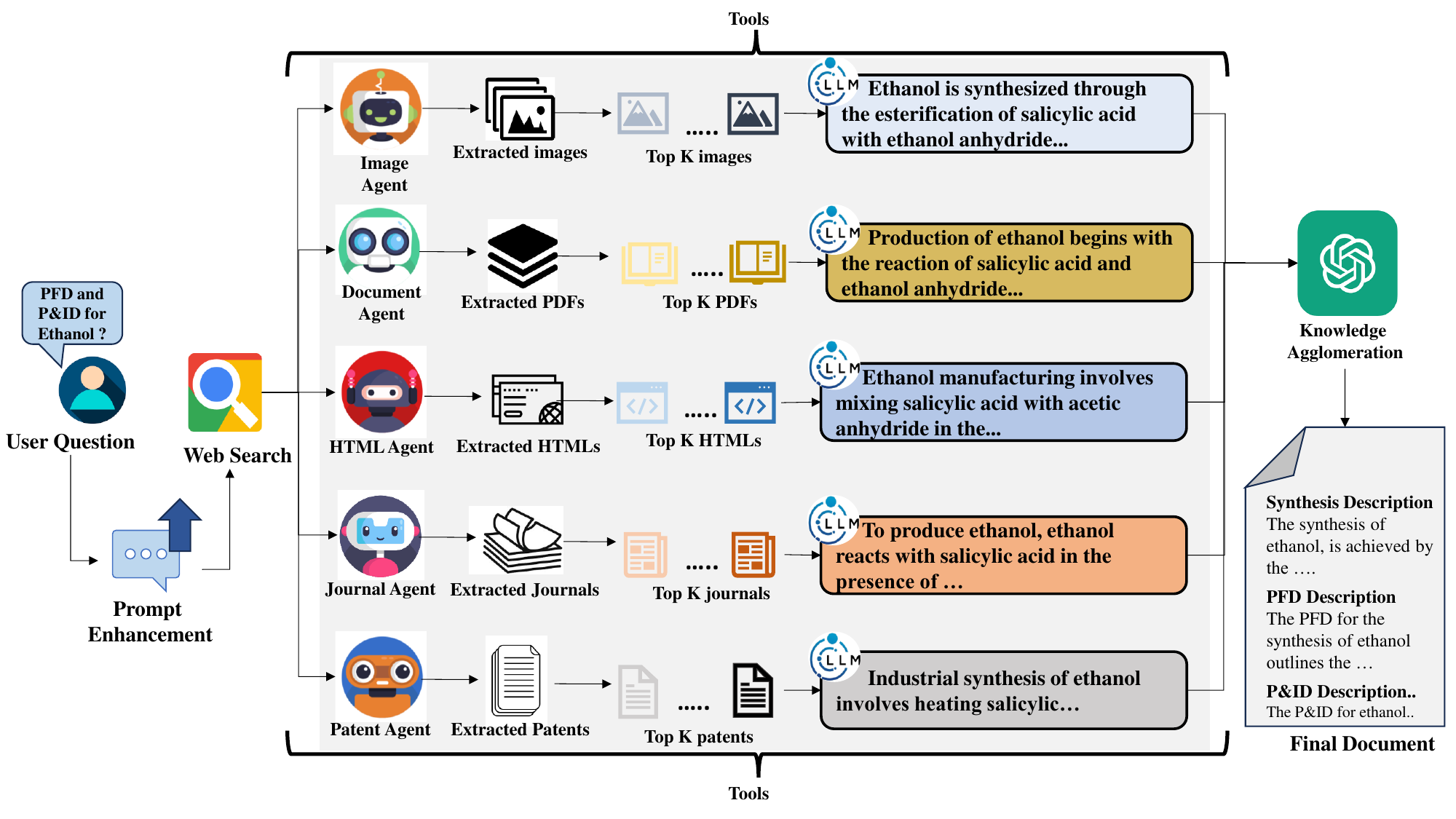}
\caption{The figure illustrates an autonomous framework for generating textual descriptions of PFDs and PIDs for user-specified chemical processes to construct property graphs. A meta-agent decomposes complex queries into subtasks, routes them to domain-specific expert agents (e.g., Visual Miner, Research), and structures execution using a DAG. The agents retrieve multimodal content (e.g., PDFs, patents, HTML documents), rank results by relevance, and synthesize summaries using LLMs. The outputs are iteratively refined through LLM-as-Judge feedback and reward models to ensure accuracy and coherence.}
\label{fig:agenticwebretrieval}
\vspace{-3mm}
\end{figure*}

\vspace{1mm}
Here, \( F_i \) includes feedback from: (a) LLM-as-Judge (e.g., GPT-4o, Anthropic Sonnet), applying ReAct-based reasoning and qualitative critique (e.g., correctness, coherence, and factuality); and (b) Reward Models (e.g., the Nemotron-4-340B multidimensional reward model), which score candidate outputs based on five key attributes: helpfulness, correctness, coherence, complexity, and verbosity. These mechanisms form a self-correcting feedback loop, enabling reward-aligned output generation and enhancing factuality and task relevance. This modular, explainable framework extends RAG from static retrieval to agentic, feedback-driven generation of high-quality textual descriptions---enabling automated generation of regulation-compliant PFD and PID descriptions for complex chemical synthesis pipelines. Figure~\ref{fig:agenticwebretrieval} outlines our agentic framework for automated PFD/PID synthesis via query decomposition, expert routing, and iterative refinement.

%%%%%%%%%%%%%%%%%%%%%%%%%%%%%%%%%%%%%%%%%%%%%%%%%%%%%%%%%%%%%%%%%%%%%

\subsection{Synthetic Datasets Generation for PFD/PID Analysis}
\begin{figure*}[ht!]
\centering
\begin{tcolorbox}[
  width=\textwidth,
  colback=white,            % Set background color to white (or match document background)
  colframe=black,           % Border color
  boxrule=1.5pt,            % Thicker border line
  arc=8pt,                  % Curved corners
  left=15pt,                % Left margin inside box
  right=15pt,               % Right margin inside box
  top=1pt,                  % Top margin inside box
  bottom=2pt,
]
  % First figure
  \centering
  \resizebox{0.80\textwidth}{!}{
    \includegraphics[keepaspectratio,trim=0.0cm 1.5cm 0cm 1.75cm,clip]{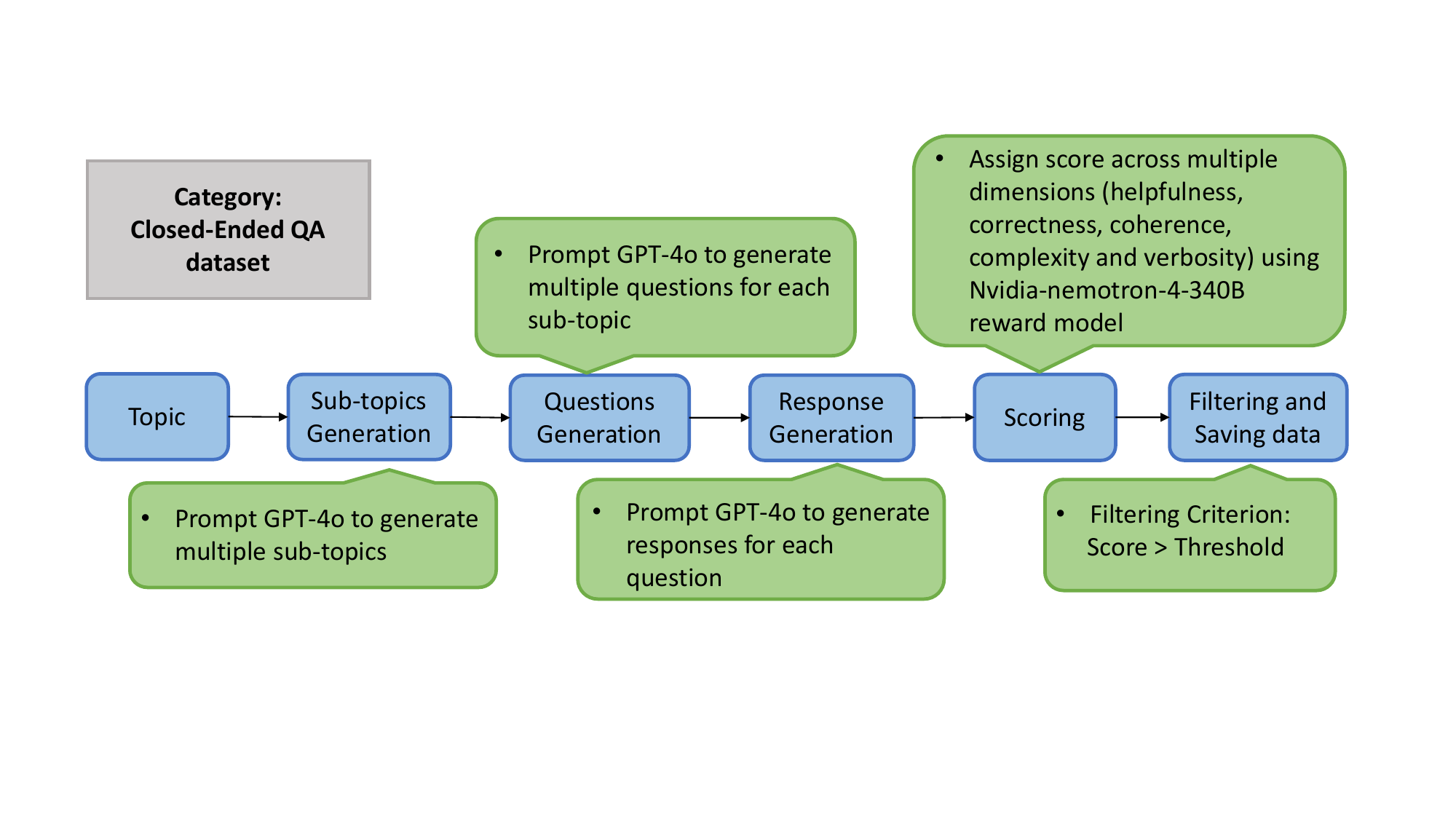}
  }
  \vspace{-15mm}
  \captionof{figure}{The figure shows the pipeline for generating synthetic \textit{Factual QA} dataset. GPT-4o or claude-3-Haiku decomposes domain topics into subtopics and creates question-answer pairs, which are filtered by the Nvidia Nemotron-4-340B reward model based on metrics like correctness, coherence and etc. Only high-scoring pairs are retained for the final dataset.}
  \label{fig:SynData1}
  \end{tcolorbox}
  \vspace{-5mm}
\end{figure*}

We adopt a teacher--student transfer learning framework \cite{zhong2023seeking, kendapadi2024interact, tian2025beyond, rawat2024little, yang2025feature} that leverages large language models (LLMs), such as OpenAI's GPT-4o and Anthropic's Claude Haiku, as teacher models to generate high-quality synthetic training data. This synthetic dataset is then used to fine-tune smaller, open-source student models such as Llama-3.2-1B and SmolLM2-135M, enhancing their ability to follow complex instructions, provide helpful and context-aware responses, and perform specialized domain tasks---particularly the interpretation, analysis, and generation of PFDs and PIDs for chemical processes. 
From a Bayesian learning perspective, the teacher model approximates a posterior distribution over possible outputs, while the student model learns a compressed yet effective representation of this distribution. Through this knowledge distillation process, the student model achieves performance comparable to that of the teacher model on out-of-distribution (OOD) tasks while being significantly more efficient to deploy.
Our data generation pipeline employs self-instruct prompting, where the teacher LLM is first conditioned on a small seed set of human-written instruction--response pairs, denoted as \(\mathcal{D}_{\text{seed}} = \{(x_i, y_i)\}_{i=1}^N\), and then recursively generates synthetic pairs \(\mathcal{D}_{\text{gen}} = \{(\tilde{x}_j, \tilde{y}_j)\}_{j=1}^{M}\), with \((\tilde{x}_j, \tilde{y}_j) \sim p_{\text{LLM}}(\cdot \mid \mathcal{D}_{\text{seed}})\). Here, \(\tilde{x}_j\) represents a synthetic instruction, \(\tilde{y}_j\) its corresponding generated response, and \(p_{\text{LLM}}\) denotes the teacher LLM's probability distribution. This bootstrapped approach generates structured instruction--response pairs without extensive human annotation, forming the core of our training corpus. In this section, we discuss the generation of multiple synthetic instruction--response datasets, all formatted as QA pairs, to support the development of expert language models for interpreting and generating PFD and PID descriptions in chemical process engineering. These datasets include: \textit{Factual QA}, which targets domain-specific factual knowledge; \textit{SynDIP}, designed to capture schematic-level descriptions of industrial processes; \textit{LogiCore}, which elicits multi-step reasoning and logical understanding; \textit{DPO}, comprising chosen--rejected response pairs for preference optimization; and \textit{Local} and \textit{Global RAIT}, which incorporate retrieval-augmented prompts with intra- and inter-cluster contextual grounding. All datasets are generated using a self-instruct bootstrapping pipeline with LLM-based prompting and validated through reward models to ensure alignment, informativeness, and response quality. (a) We generate a \textbf{factual QA dataset} (refer to Figure~\ref{fig:SynData1}) by first selecting a domain-level topic \( T \in \mathcal{T} \) (e.g., PFDs or PIDs), where \(\mathcal{T}\) denotes the set of all possible topics. The teacher model \( M \) (e.g., GPT-4o) decomposes \( T \) into subtopics \(\mathcal{S}_T = \{s_1, \dots, s_n\}\) and then synthesizes question--answer pairs \((\tilde{q}_{jk}, \tilde{a}_{jk})\) for each subtopic \( s_j \), where \( j = 1, \dots, n \) indexes the subtopics and \( k = 1, \dots, m_j \) indexes the QA pairs within subtopic \( s_j \). Each pair is generated as:

\vspace{-1mm}
\resizebox{0.985\linewidth}{!}{ 
\begin{minipage}{\linewidth} 
\begin{equation}
 (\tilde{q}_{jk}, \tilde{a}_{jk}) \sim M(\cdot \mid s_j, \mathcal{D}_{\text{seed}}^{\text{FQA}}) \nonumber
\end{equation}  
\end{minipage}  
}

\begin{figure*}[ht!]
\centering
\begin{tcolorbox}[
  width=\textwidth,
  colback=white,            % Set background color to white (or match document background)
  colframe=black,           % Border color
  boxrule=1.5pt,            % Thicker border line
  arc=8pt,                  % Curved corners
  left=15pt,                % Left margin inside box
  right=15pt,               % Right margin inside box
  top=1pt,                  % Top margin inside box
  bottom=2pt,
]
  % Second figure
  \vspace{-1mm}
  \centering
  \resizebox{0.80\textwidth}{!}{
    \includegraphics[keepaspectratio,trim=0.0cm 1.5cm 0cm 1.8cm,clip]{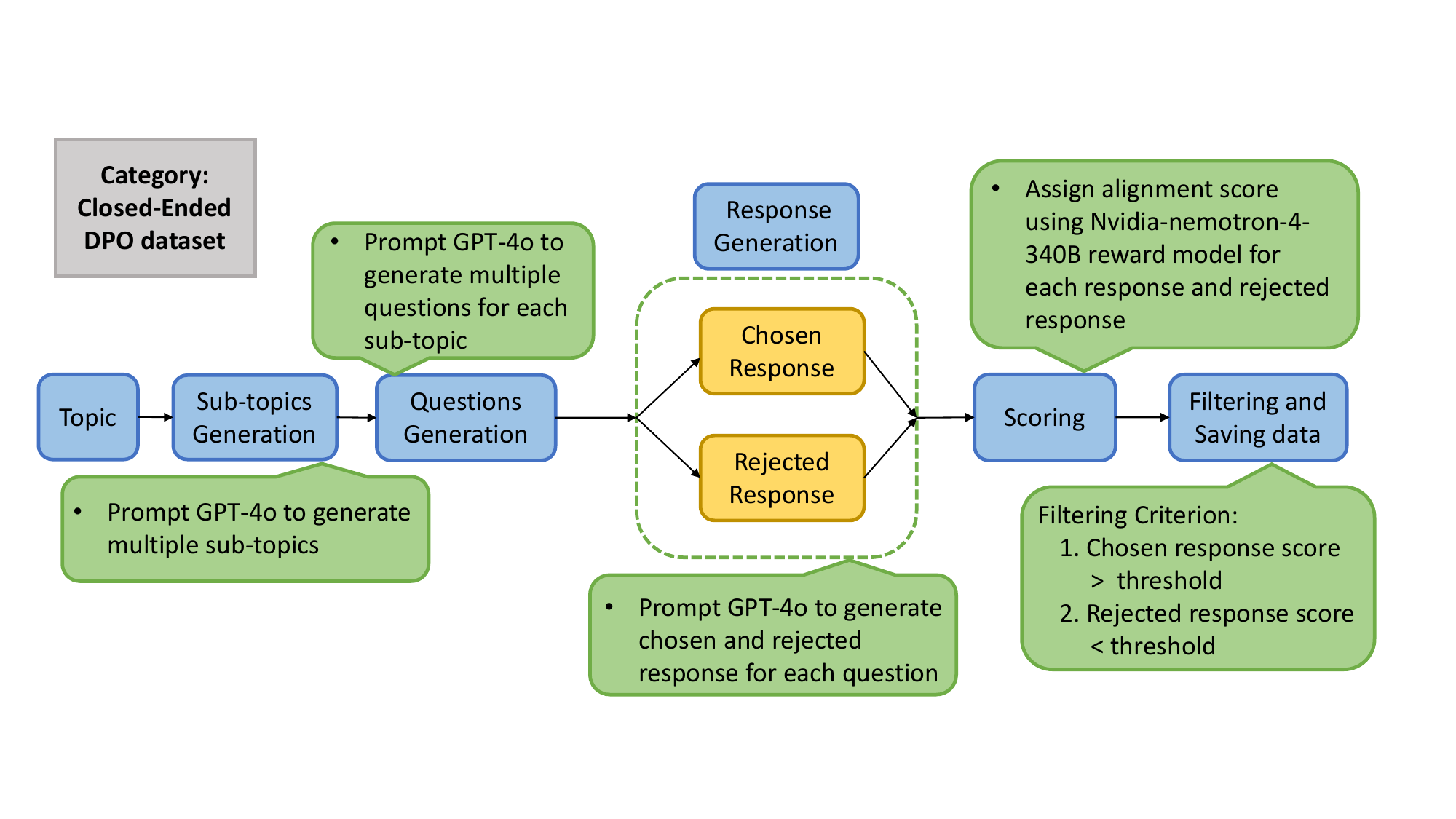}
  }
  \vspace{-5mm}
  \captionof{figure}{The figure illustrates the workflow for synthetic \textit{DPO} dataset generation. GPT-4o or claude-3-Haiku generates questions with paired preferred and dispreferred responses. The Nvidia Nemotron-4-340B reward model scores responses, and pairs are filtered to ensure the preferred response ranks significantly higher in quality.}
  \label{fig:SynData2}
  \end{tcolorbox}
  \vspace{-5mm}
\end{figure*}

Here, \(\mathcal{D}_{\text{seed}}^{\text{FQA}} = \{(x_i, y_i)\}_{i=1}^N\) denotes a seed set of human-written QA examples. The synthetic pairs form the dataset \(\mathcal{D}_{\text{gen}}^{\text{FQA}} = \{(\tilde{q}_{jk}, \tilde{a}_{jk})\}_{j,k}\), which is filtered via a reward model (e.g., Nemotron-4-340B-Reward), defined as: $ R(\tilde{q}, \tilde{a}) = \sum_{l=1}^5 \alpha_l \cdot \text{Metric}_l(\tilde{q}, \tilde{a})$, where \(\{\text{Metric}_l\}_{l=1}^5 = \{H, C, Co, Cx, V\}\) represent helpfulness, correctness, coherence, complexity, and verbosity, respectively, and \(\alpha_l \geq 0\) are predefined scalar weights. Only QA pairs satisfying the quality threshold \( R(\tilde{q}, \tilde{a}) \geq \tau \) are retained, ensuring the dataset meets the quality standards required for downstream student model fine-tuning. The resulting dataset \(\mathcal{D}_{\text{gen}}^{\text{FQA}}\) contains factual QA pairs related to chemical process engineering. (b) The \textbf{Direct Preference Optimization (DPO) dataset} (refer to Figure~\ref{fig:SynData2}) is generated using the teacher model \( M \) (e.g., GPT-4o or Claude Haiku) and the reward model \( R \). For each subtopic \( s_j \in \mathcal{S}_T \) (derived from a domain-level topic \( T \)) and each synthetic question \(\tilde{q}_{jk} \in \mathcal{D}_{\text{gen}}^{\text{FQA}} \), we sample two candidate responses from \( M \):

\vspace{-1mm}
\resizebox{0.985\linewidth}{!}{ 
\begin{minipage}{\linewidth} 
\begin{equation}
(\tilde{a}^+_{jk}, \tilde{a}^-_{jk}) \sim M(\cdot \mid \tilde{q}_{jk}, \mathcal{D}_{\text{seed}}^{\text{DPO}}) \nonumber
\end{equation}  
\end{minipage}  
}

Here, \(\tilde{a}^+_{jk}\) is the preferred response, \(\tilde{a}^-_{jk}\) is the dispreferred response, and \(\mathcal{D}_{\text{seed}}^{\text{DPO}}\) is a seed set of human-annotated preference pairs. The reward model \( R \) then computes the preference gap: $ \Delta R_{jk} = R(\tilde{q}_{jk}, \tilde{a}^+_{jk}) - R(\tilde{q}_{jk}, \tilde{a}^-_{jk}) $, where \( R \) is defined as a weighted sum over five quality metrics: \(\{H, C, Co, Cx, V\}\), representing helpfulness, correctness, coherence, complexity, and verbosity, respectively. Only preference triplets satisfying the quality threshold \(\Delta R_{jk} \geq \tau_{\text{DPO}} \) are retained, forming the final dataset:

\vspace{-1mm}
\resizebox{0.985\linewidth}{!}{ 
\begin{minipage}{\linewidth} 
\begin{equation}
\mathcal{D}_{\text{gen}}^{\text{DPO}} = \{(\tilde{q}_{jk}, \tilde{a}^+_{jk}, \tilde{a}^-_{jk}) \mid \Delta R_{jk} \geq \tau_{\text{DPO}}\} \nonumber
\end{equation}  
\end{minipage}  
}

\begin{figure*}[ht!]
\centering
\begin{tcolorbox}[
  width=\textwidth,
  colback=white,            % Set background color to white (or match document background)
  colframe=black,           % Border color
  boxrule=1.5pt,            % Thicker border line
  arc=8pt,                  % Curved corners
  left=15pt,                % Left margin inside box
  right=15pt,               % Right margin inside box
  top=1pt,                  % Top margin inside box
  bottom=2pt,
]
  % Third figure
  \centering
  \resizebox{0.80\textwidth}{!}{
    \includegraphics[keepaspectratio,trim=0.0cm 1.5cm 0cm 0.10cm,clip]{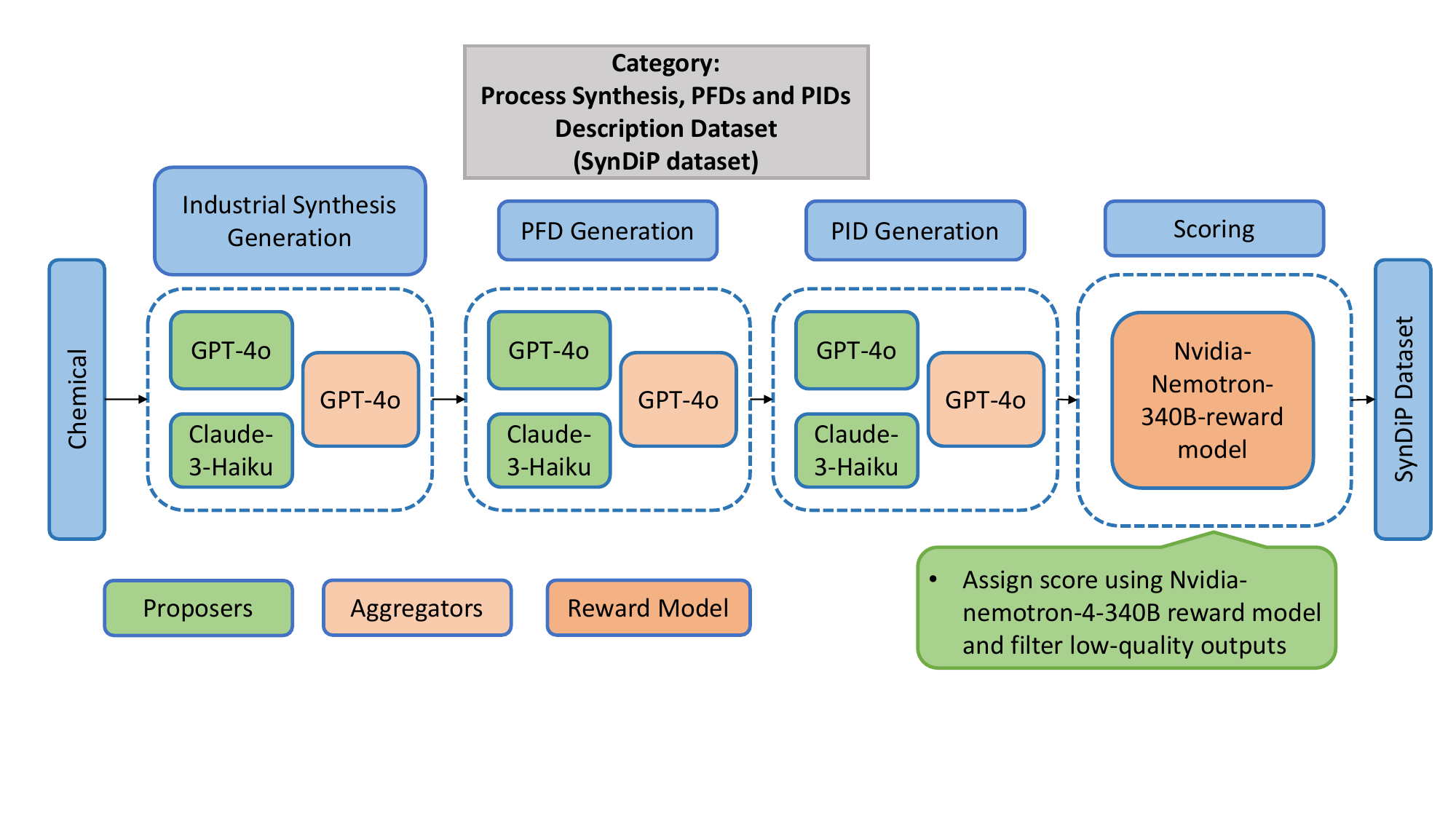}
  }
  \vspace{-10mm}
  \captionof{figure}{The figure outlines the SynDIP dataset generation process. Teacher models (GPT-4o, Claude-3-Haiku) generate PFD and PID descriptions. The Nvidia Nemotron-4-340B reward model validates, ensuring consistent quality across all outputs.}
  \label{fig:SynData3}
\end{tcolorbox}
\end{figure*}

\begin{figure*}[ht!]
\centering
\begin{tcolorbox}[
  width=\textwidth,
  colback=white,            % Set background color to white (or match document background)
  colframe=black,           % Border color
  boxrule=1.5pt,            % Thicker border line
  arc=8pt,                  % Curved corners
  left=15pt,                % Left margin inside box
  right=15pt,               % Right margin inside box
  top=1pt,                  % Top margin inside box
  bottom=2pt,
]
  % First figure
  \centering
\resizebox{0.80\textwidth}{!}{
\includegraphics[keepaspectratio,trim=0.0cm 1.5cm 0cm 0.75cm,clip]{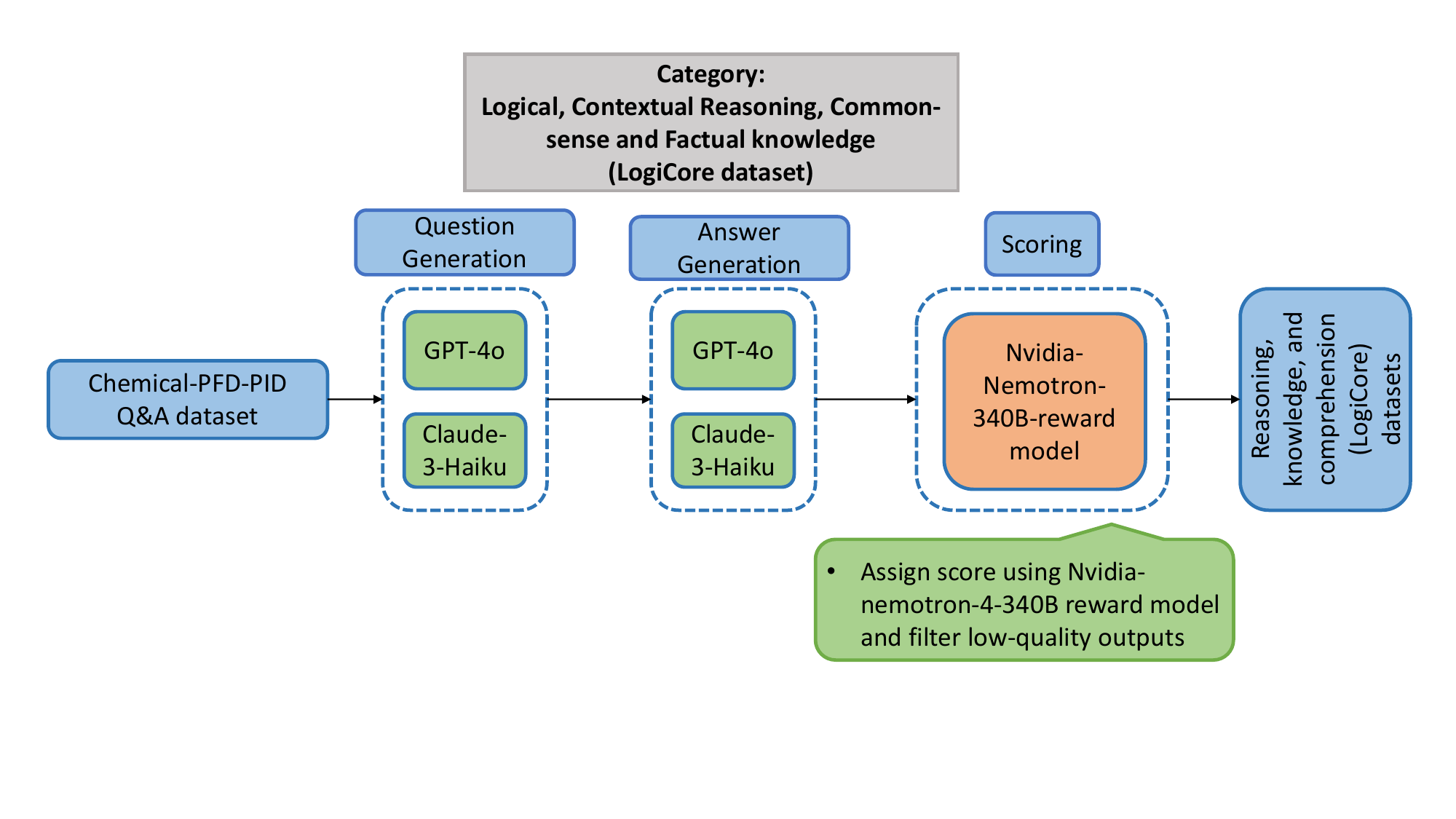} % trim = <left> <bottom> <right> <top>
}
\vspace{-10mm}
\caption{The figure outlines the generation pipeline for the \textit{LogiCore} dataset. Starting with the SynDIP dataset, GPT-4o and Claude-3-Haiku generate reasoning-augmented, Chain of thought (CoT) question-answer pairs. The Nvidia Nemotron-4-340B reward model scores outputs based on helpfulness, correctness, and coherence. High-quality responses are filtered to create a final dataset for advanced reasoning and contextual comprehension in process engineering.}
\label{fig:SynData4}
\end{tcolorbox}
\vspace{-5mm}
\end{figure*}

In summary, this pipeline automates the generation of high-quality preference-labeled datasets for PFD/PID analysis tasks by combining teacher-model synthesis \((\tilde{a}^+_{jk}, \tilde{a}^-_{jk}) \sim M\) with multi-metric reward-based filtering \( R \), resulting in a DPO-optimized dataset tailored for domain-specific student model training. (c) The \textbf{SynDIP dataset} (refer to Figure~\ref{fig:SynData3}) extends the teacher--student framework to generate chemical process context, PFDs, and PIDs textual descriptions, organized as sequential instruction--response pairs. The process context overview explains the why and how of a process design, covering its background, operation, engineering decisions, and control. It outlines unit operations, flow, reactions, and the rationale behind equipment and controls. For each target chemical, the teacher model \( M \) (e.g., GPT-4o or Claude-3-Haiku) generates a process blueprint \( \tilde{b}_k \) in response to a fixed instruction template \( x_k^{\text{SYN}} \) (e.g., ``Describe a chemical process for producing chemical X, including raw materials, reactions, and equipment''), with: $\tilde{b}_k \sim M(\cdot \mid x_k^{\text{SYN}}, \mathcal{D}_{\text{seed}}^{\text{SYN}})$ where \( \mathcal{D}_{\text{seed}}^{\text{SYN}} \) is a seed set of human-authored process blueprints. Each blueprint \( \tilde{b}_k \) is then processed in two stages:  (1) \textbf{PFD generation} via prompt \( x_k^{\text{PFD}} \) (e.g., ``Convert this blueprint to a PFD: [\( \tilde{b}_k \)]''), yielding: $ \tilde{f}_k \sim M(\cdot \mid x_k^{\text{PFD}}, \tilde{b}_k, \mathcal{D}_{\text{seed}}^{\text{PFD}})$ where \( \mathcal{D}_{\text{seed}}^{\text{PFD}} \) contains human-annotated PFD exemplars; and  (2) \textbf{PID generation} using prompt \( x_k^{\text{PID}} \) (e.g., ``Generate a PID for this PFD: [\( \tilde{f}_k \)]''), resulting in:

\vspace{1mm}
\resizebox{0.985\linewidth}{!}{ 
\begin{minipage}{\linewidth} 
\begin{equation}
\tilde{p}_k \sim M(\cdot \mid x_k^{\text{PID}}, \tilde{f}_k, \mathcal{D}_{\text{seed}}^{\text{PID}}) \nonumber
\end{equation}  
\end{minipage}  
} 

where \( \mathcal{D}_{\text{seed}}^{\text{PID}} \) contains human-annotated PID exemplars. The reward model \( R \) evaluates each instruction--response pair \( (x_k, \tilde{y}_k) \)---where \( \tilde{y}_k \in \{\tilde{b}_k, \tilde{f}_k, \tilde{p}_k\} \)---using the composite metric set \(\{H, C, Co, Cx, V\}\) (helpfulness, correctness, coherence, complexity, verbosity). The final SynDIP dataset is defined as:

\vspace{-1mm}
\resizebox{0.985\linewidth}{!}{ 
    \begin{minipage}{\linewidth} 
        \begin{align}
            \mathcal{D}_{\text{gen}}^{\text{SynDIP}} = \{&(x_k^{\text{SYN}}, \tilde{b}_k, x_k^{\text{PFD}}, \tilde{f}_k, x_k^{\text{PID}}, \tilde{p}_k) \mid 
            R(x_k^{\text{SYN}}, \tilde{b}_k) + R(x_k^{\text{PFD}}, \tilde{f}_k) + R(x_k^{\text{PID}}, \tilde{p}_k) \geq \tau_{\text{SYN}} \} \nonumber
        \end{align}  
    \end{minipage}  
}

ensuring that each entry includes validated process context, PFD, and PID descriptions for a complete chemical process representation. (d) The \textbf{LogiCore Dataset} (refer to Figure~\ref{fig:SynData4}) extends our teacher--student framework to generate multi-step reasoning question--answer pairs for PFD/PID analysis by building upon the $\mathcal{D}_{\text{gen}}^{\text{SynDIP}}$ dataset and extracting logical reasoning chains from its process descriptions. For each seed instruction \( x_i \in \mathcal{D}_{\text{seed}}^{\text{chem}} \) (human-annotated exemplars), the teacher model \( M \) (e.g., GPT-4o) generates multiple logical QA pairs \( (\tilde{q}_{ij}, \tilde{a}_{ij}) \sim M(\cdot \mid x_i, \mathcal{D}_{\text{seed}}^{\text{chem}}) \), where \( j \) indexes the generated pairs per seed and each \( \tilde{a}_{ij} \) contains explicit chain-of-thought reasoning. These pairs are filtered via the established reward model \( R \) (Nemotron-4-340B-Reward) using the same metrics: $ R(\tilde{q}_{ij}, \tilde{a}_{ij}) = \sum_{l=1}^4 \alpha_l \cdot \text{Metric}_l(\tilde{q}_{ij}, \tilde{a}_{ij})$,
where $\{\text{Metric}_l\}_{l=1}^4 = \{H, C, Co, Cx\}$ (helpfulness, correctness, coherence, complexity). The final dataset

\vspace{-1mm}  
\resizebox{0.985\linewidth}{!}{  
\begin{minipage}{\linewidth}  
\begin{equation}  
\mathcal{D}_{\text{gen}}^{\text{LogiCore}} = \{(\tilde{q}_{ij}, \tilde{a}_{ij}) \mid R(\tilde{q}_{ij}, \tilde{a}_{ij}) \geq \tau_{\text{logic}}\} \nonumber  
\end{equation}  
\end{minipage}  
}  

\begin{figure*}[ht!]
\centering
\begin{tcolorbox}[
  width=\textwidth,
  colback=white,            % Set background color to white (or match document background)
  colframe=black,           % Border color
  boxrule=1.5pt,            % Thicker border line
  arc=8pt,                  % Curved corners
  left=15pt,                % Left margin inside box
  right=15pt,               % Right margin inside box
  top=1pt,                  % Top margin inside box
  bottom=2pt,
]
  % Second figure
  \vspace{0mm}
  \centering
\resizebox{0.80\textwidth}{!}{
\includegraphics[keepaspectratio,trim=0.0cm 1.5cm 0cm 3.0cm,clip]{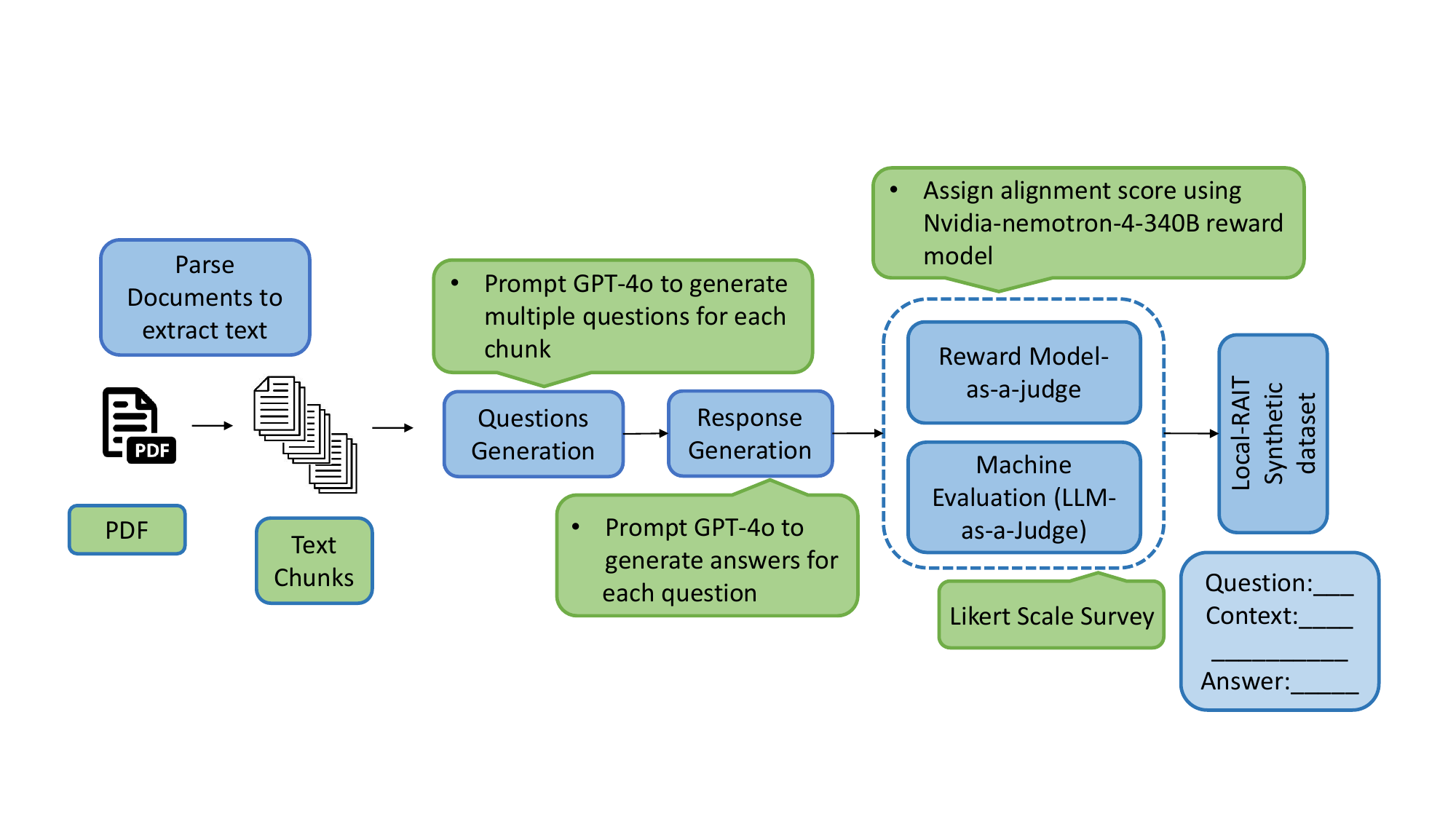} % trim = <left> <bottom> <right> <top>
}
\vspace{-4mm}
\caption{The figure depicts the workflow for the \textit{Local RAIT} dataset generation. Text chunks are extracted from the seed SynDIP dataset, and GPT-4o generates retrieval-grounded question-answer pairs. Outputs are evaluated using the Nemotron-4-340B reward model and additional LLM-based validation. High-quality pairs are retained to build a dataset for retrieval augmented instruction-tuning in process engineering tasks.}
\label{fig:SynData5}
\end{tcolorbox}
\vspace{-5mm}
\end{figure*}

\vspace{1mm}  
retains only high-quality reasoning chains, with logical validity implicitly ensured through \( C \) (factual alignment with PFD/PID schematics) and \( Co \) (stepwise flow coherence), maintaining full consistency with our synthetic data generation framework.
(e) The \textbf{Local RAIT Dataset} (refer to Figure~\ref{fig:SynData5}) extends our teacher--student framework to retrieval-augmented generation. Unlike $\mathcal{D}_{\text{gen}}^{\text{FQA}}$ and $\mathcal{D}_{\text{gen}}^{\text{SynDIP}}$, Local RAIT integrates retrieval mechanisms to

%%%%%%%%%%%%%%%%%%%%%%%%%%%%%%%%%%%%%%%%%%%%%%%%%%%%%%%%%%%%
%%%%%%%%%%%%%%%%%%%%%%%%%%%%%%%%%%%%%%%%%%%%%%%%%%%%%%%%%%%%

%%%%%%%%%%%%%%%%%%%%%%%%%%%%%%%%%%%%%%%%%%%%%%%%%%%%%%%%%%%%
%%%%%%%%%%%%%%%%%%%%%%%%%%%%%%%%%%%%%%%%%%%%%%%%%%%%%%%%%%%%

\clearpage
\newpage

\begin{figure*}[ht!]
\centering
\begin{tcolorbox}[
  width=\textwidth,
  colback=white,            % Set background color to white (or match document background)
  colframe=black,           % Border color
  boxrule=1.5pt,            % Thicker border line
  arc=8pt,                  % Curved corners
  left=15pt,                % Left margin inside box
  right=15pt,               % Right margin inside box
  top=1pt,                  % Top margin inside box
  bottom=2pt,
]  
  % Third figure
  \vspace{0mm}
  \centering
\resizebox{0.80\textwidth}{!}{
\includegraphics[keepaspectratio,trim=0.0cm 0.0cm 0cm 0.0cm,clip]{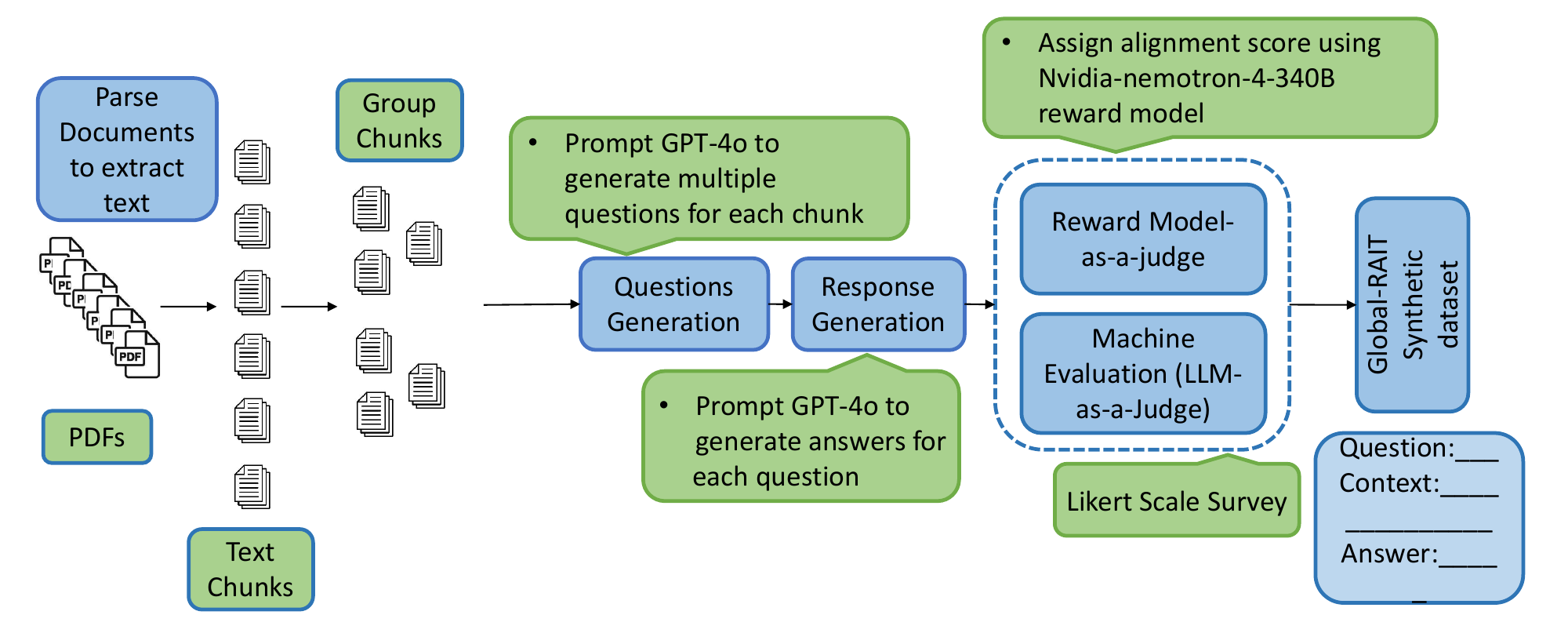} % trim = <left> <bottom> <right> <top>
}
\vspace{-3mm}
\caption{The figure illustrates the \textit{Global RAIT} dataset generation workflow. PDFs are parsed, chunked, and grouped via semantic clustering to preserve context. A retriever selects top-k relevant chunks using vector similarity of embeddings obtained from a sentence  embedding model. GPT-4o generates questions and multi-turn refined answers grounded in the cross-document chunks. Outputs are validated through the Nvidia Nemotron-4-340B reward model, LLM-based checks, and Likert-scale feedback, yielding a high-quality dataset for instruction-tuning.}
\label{fig:SynData6}
\vspace{-1mm}
\end{tcolorbox}
\end{figure*}

%%%%%%%%%%%%%%%%%%%%%%%%%%%%%%%%%%%%%%%%%%%%%%%%%%%%%%%%%%%%
%%%%%%%%%%%%%%%%%%%%%%%%%%%%%%%%%%%%%%%%%%%%%%%%%%%%%%%%%%%%

\clearpage
\newpage

ground $M$'s outputs in source documents, mitigating hallucination risks. For each chemical process description from the SynDIP datasets in the ChemAtlas database (stored as PDF documents containing process flow and instrumentation descriptions), the raw text $T$ is extracted and parsed into semantically coherent chunks $\mathcal{C}_T = \{c_1,...,c_K\}$, where $c_k \sim \text{Chunk}(T)$ and each $c_k$ retains contextual continuity with neighboring chunks. The teacher model $M$ (GPT-4o) then synthesizes QA pairs $(\tilde{q}_k, \tilde{a}_k) \sim M(\cdot \mid c_k, \mathcal{D}_{\text{seed}}^{\text{RAIT}})$, conditioned on seed human examples $\mathcal{D}_{\text{seed}}^{\text{RAIT}} = \{(x_i, y_i)\}$ that include both questions and gold-standard retrieval-augmented answers. This approach ensures $(\tilde{q}_k, \tilde{a}_k)$ are document-grounded, with $c_k$ providing explicit source references for generated answers—critical for technical domains where factual alignment with PFD/PID schematics is required.

\vspace{-1mm} 
\resizebox{0.985\linewidth}{!}{  
\begin{minipage}{\linewidth}  
\begin{equation}  
\mathcal{D}_{\text{gen}}^{\text{LocalRAIT}} = \{(\tilde{q}_k, c_k, \tilde{a}_k) \mid R(\tilde{q}_k, \tilde{a}_k) \geq \tau \land \mathcal{L}(\tilde{q}_k, \tilde{a}_k) \geq 4\} \nonumber  
\end{equation}  
\end{minipage}  
}  

\vspace{2mm} 
The QA pairs are filtered using the same reward model \(R\) and Likert scoring \(\mathcal{L}\) as \(\mathcal{D}_{\text{gen}}^{\text{FQA}}\), where the reward score \(R(\tilde{q}_k, \tilde{a}_k) = \sum_{l=1}^5 \alpha_l \cdot \text{Metric}_l(\tilde{q}_k, \tilde{a}_k)\) incorporates five metrics: \(H\)=Helpfulness, \(C\)=Correctness, \(Co\)=Coherence, \(Cx\)=Complexity, and \(V\)=Verifiability against \(c_k\). The Likert scale \(\mathcal{L}(\tilde{q}_k, \tilde{a}_k) \in \{1,\ldots,5\}\) (1=Poor, 3=Average, 5=Excellent) independently assesses answer quality across three dimensions: helpfulness, correctness, and coherence. Only instances meeting both criteria—\(\tau\) for \(R\) and 4+ for \(\mathcal{L}\)—are included in \(\mathcal{D}_{\text{gen}}^{\text{LocalRAIT}}\). (f) The \textbf{Global RAIT Dataset} (refer to Figure~\ref{fig:SynData6}) scales retrieval-augmented generation to both intra- and inter-document comprehension. Chunks $\mathcal{C}_T$ are clustered into semantically related groups $\mathcal{G}_j$ via cosine similarity $\text{sim}(\phi(c_i),\phi(c_j)) \geq \gamma$, where $\phi$ is a domain-tuned embedding model (fine-tuned on $\mathcal{T}$ using contrastive learning) optimized for cross-document semantic relationships. For cross-document groups, $\mathcal{G}_j$ aggregates chunks from multiple source PDFs. The teacher model $M$ generates answers $\tilde{a}_j \sim M(\cdot|\mathcal{G}_j,\mathcal{D}_{\text{seed}}^{\text{Global}})$,  
conditioned on seed examples $\mathcal{D}_{\text{seed}}^{\text{Global}}$ that include inter-document QA pairs.

\vspace{-1mm} 
\resizebox{0.985\linewidth}{!}{
\begin{minipage}{\linewidth}
\begin{align}
\hspace{-5mm}\mathcal{D}_{\text{gen}}^{\text{GlobalRAIT}} = \{&(\tilde{q}_j, \mathcal{G}_j, \tilde{a}_j) \, | \nonumber \\
&R(\tilde{a}_j) \geq \tau \land \mathcal{L} \geq 4\} \nonumber
\end{align}
\end{minipage}
}

\vspace{2mm}
where $\tilde{a}_j$ provides structured reasoning with evidence from multiple document chunks. Filtering follows the same criteria as $\mathcal{D}_{\text{gen}}^{\text{SynDIP}}$, applying both the reward threshold $\tau$ and a Likert score of $\mathcal{L} \geq 4$. By leveraging grouped chunk clusters, Global RAIT enables the student model to generate contextually grounded responses that synthesize information across intra- and inter-document contexts.

% %%%%%%%%%%%%%%%%%%%%%%%%%%%%%dataset generation Time and emissions%%%%%%%%%%%%%%%%%%%%

\subsubsection{Computational Time Analysis for Synthetic Dataset Generation}
\begin{figure}[ht!]
\vspace{-0mm}
\centering
\includegraphics[width=0.45\textwidth]{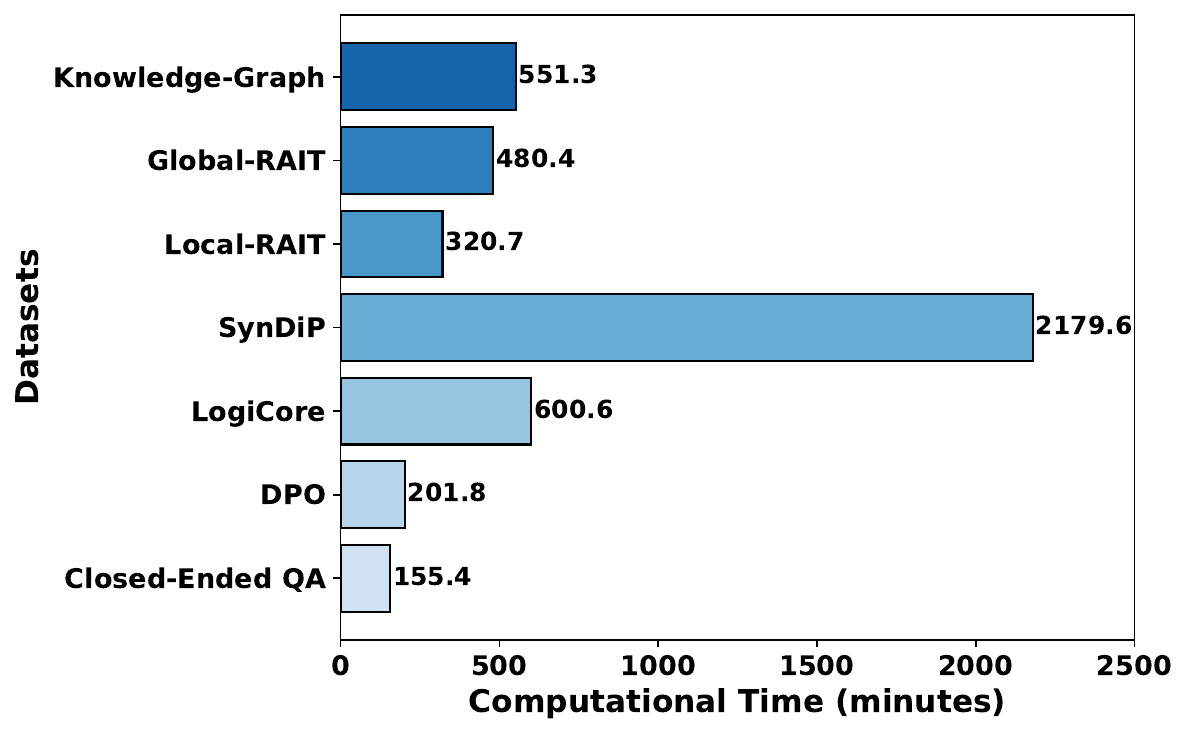}
\vspace{-5mm}
\caption{Computational time for generating self-instruct synthetic datasets, including QA pair creation, verification (using either the Nvidia Nemotron-4-340B reward model or an LLM-as-a-judge approach), and quality filtering. SynDiP’s multi-stage generation (process context $\rightarrow$  PFD $\rightarrow$ PID) requires significantly more time than simpler factual QA generation due to its iterative refinement process}
\label{fig:dataset_gen_time}
\vspace{0mm}
\end{figure}

Synthetic dataset generation follows a unified three-step pipeline: QA pair synthesis via teacher LLMs (e.g., GPT-4o, Claude Haiku), reward model validation (Nemotron-4-340B), and multi-metric filtering. SynDIP is the most time-intensive (2179.6 min) due to its sequential generation of PFDs and PIDs. LogiCore (600.6 min) emphasizes multi-step reasoning; Global RAIT (480.4 min) involves cross-document clustering; and Local RAIT (320.7 min) targets chunk-level QA generation. DPO (201.8 min) and Factual QA (155.4 min) are faster, reflecting their simpler generation logic. This reflects a clear trade-off between dataset complexity and computational cost (Figure~\ref{fig:dataset_gen_time}).

\subsubsection{Carbon Emissions for Synthetic Dataset Generation}
\begin{figure}[ht!]
\vspace{0mm}
\centering
\includegraphics[width=0.45\textwidth]{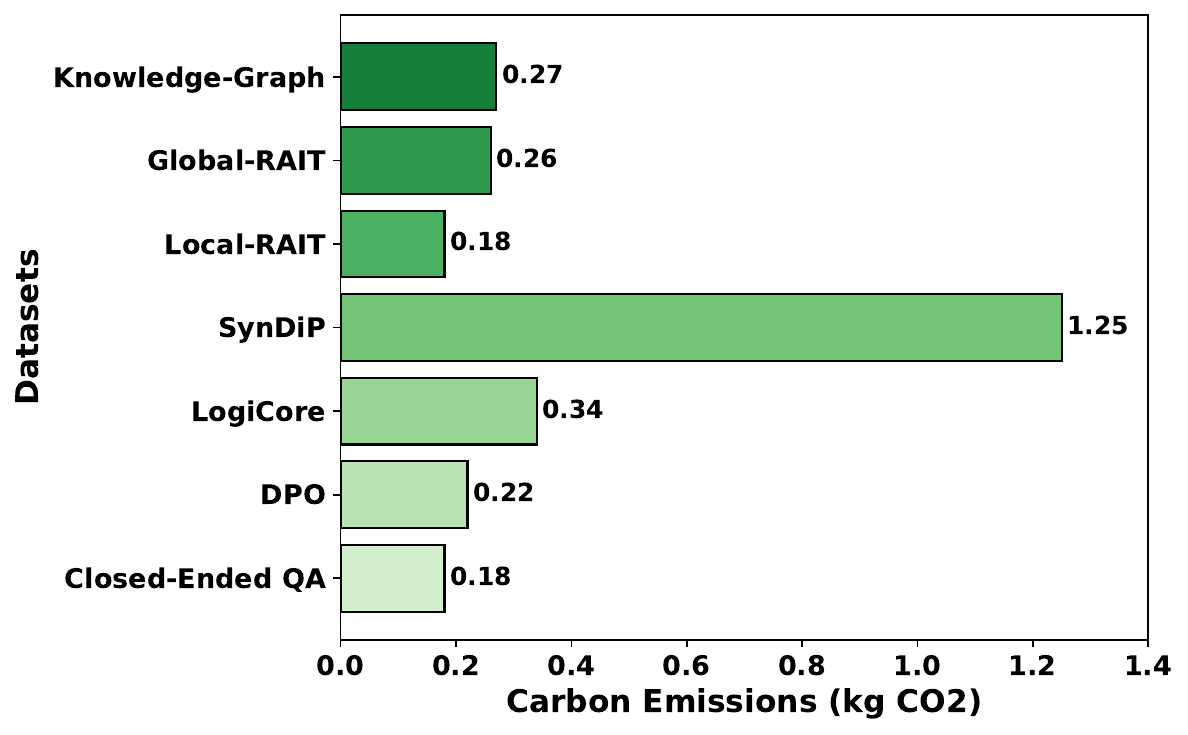}
\vspace{-5mm}
\caption{Carbon emissions (kg CO\textsubscript{2}) for synthetic dataset generation. SynDIP incurs the highest emissions, while Factual QA, DPO, and Local RAIT exhibit the lowest.}
\label{fig:dataset_carbon_emission}
\vspace{-0mm}
\end{figure}

Carbon emissions from synthetic dataset generation, tracked via CodeCarbon\footnote{\url{https://codecarbon.io/}}, vary by dataset type. SynDIP has the highest footprint ($\sim$1.25\,kg CO\textsubscript{2}) from its sequential PFD/PID generation. LogiCore ($\sim$0.34\,kg) and Global RAIT ($\sim$0.26\,kg) show moderate emissions, while DPO, Local RAIT, and Factual QA achieve $\sim$0.18--0.22\,kg through optimized workflows. Figure~\ref{fig:dataset_carbon_emission} illustrates these efficiency tradeoffs between data quality and sustainability.

\subsubsection{\textbf{Evaluation of Synthetic Datasets}}
\begin{figure}[htbp!]
\centering
\resizebox{0.375\textwidth}{!}{
\includegraphics{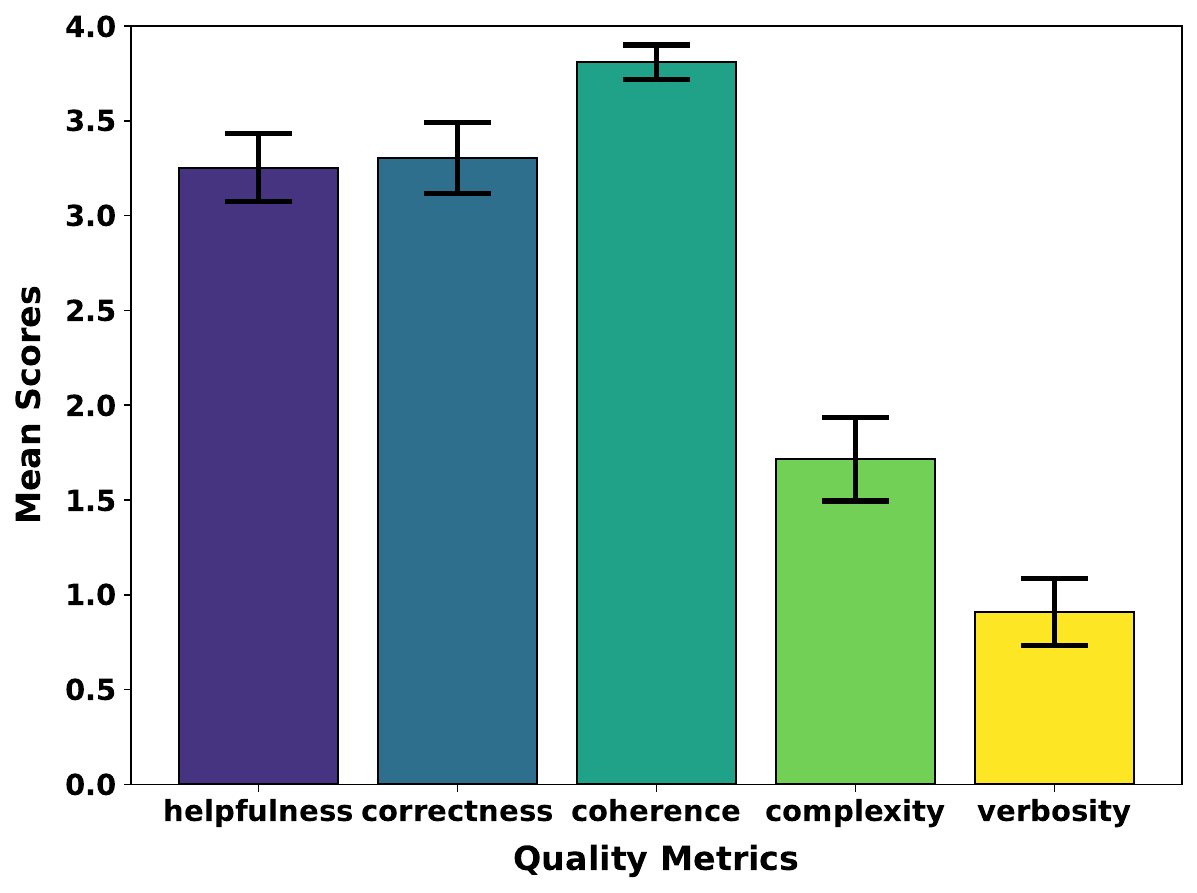}
}
\vspace{-3mm}
\caption{Evaluation results for the generated \textit{Factual QA} dataset using the NVIDIA-Nemotron-4-340B reward model. Each QA pair is scored on a 0--4 scale across five quality dimensions: helpfulness, correctness, coherence, complexity, and verbosity, ensuring high-quality data for instruction tuning.}
\label{fig:closed_QA}
\vspace{-3mm}
\end{figure}

\begin{figure}[htbp!]
\centering
\resizebox{0.375\textwidth}{!}{
\includegraphics{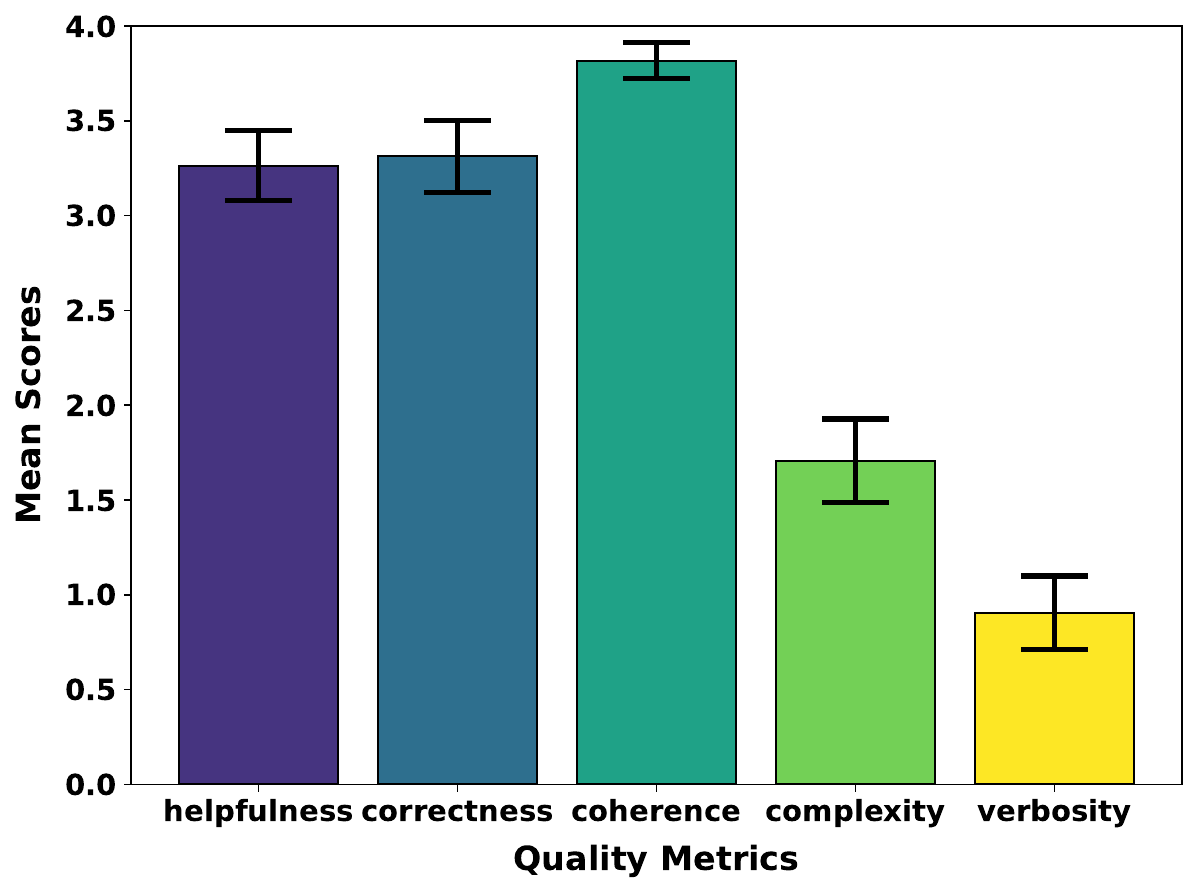}
}
\vspace{-3mm}
\caption{Evaluation of chosen responses from the \textit{DPO} dataset using the NVIDIA-Nemotron-4-340B reward model. High-scoring responses across quality dimensions (helpfulness, correctness, coherence) guide model fine-tuning toward human-preferred outputs.}
\label{fig:DPO_chosen}
\vspace{-4mm}
\end{figure}

\begin{figure}[ht!]
\centering
\resizebox{0.375\textwidth}{!}{
\includegraphics{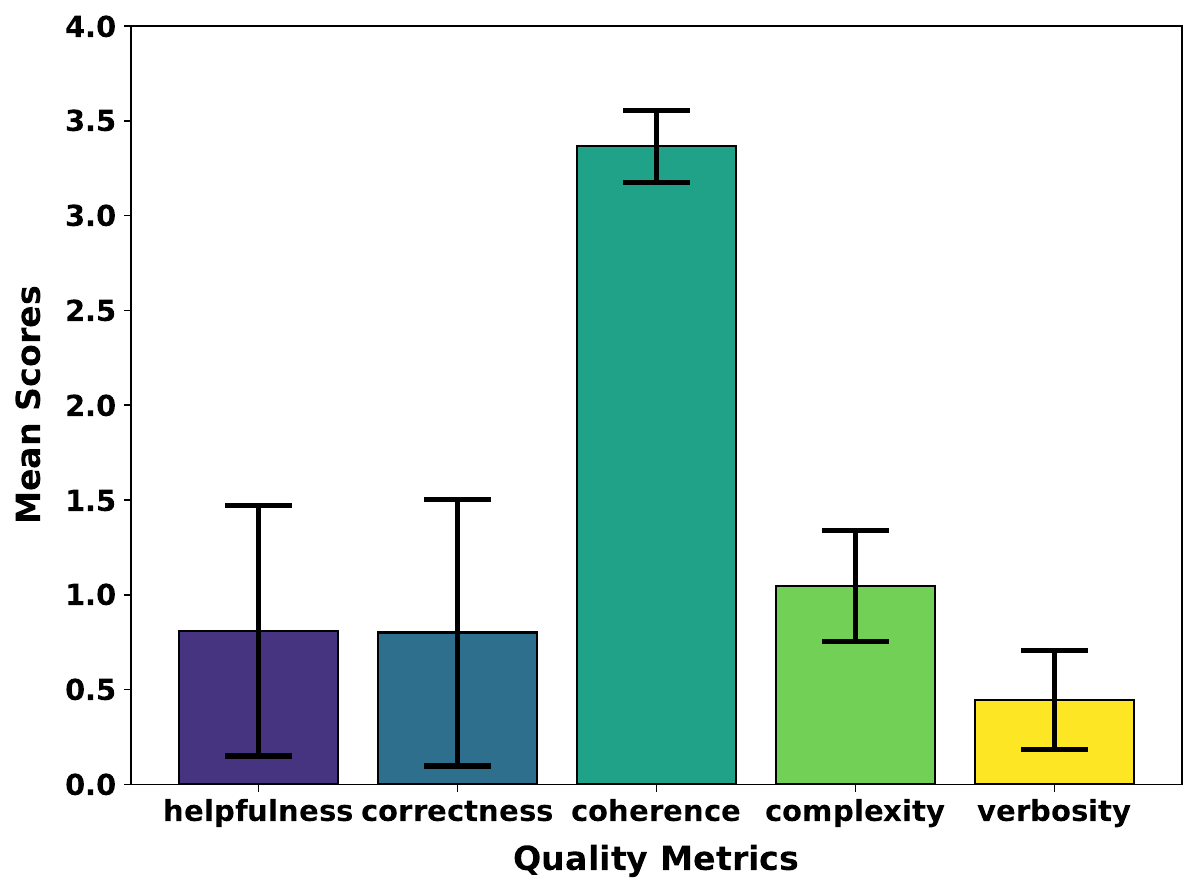}
}
\vspace{-3mm}
\caption{Evaluation of rejected responses from the \textit{DPO} dataset using the NVIDIA-Nemotron-4-340B reward model. Low-scoring responses across evaluation metrics demonstrate undesirable output characteristics for preference optimization.}
\label{fig:DPO_rejected}
\vspace{-4mm}
\end{figure}

\begin{figure}[ht!]
\centering
\resizebox{0.375\textwidth}{!}{
\includegraphics{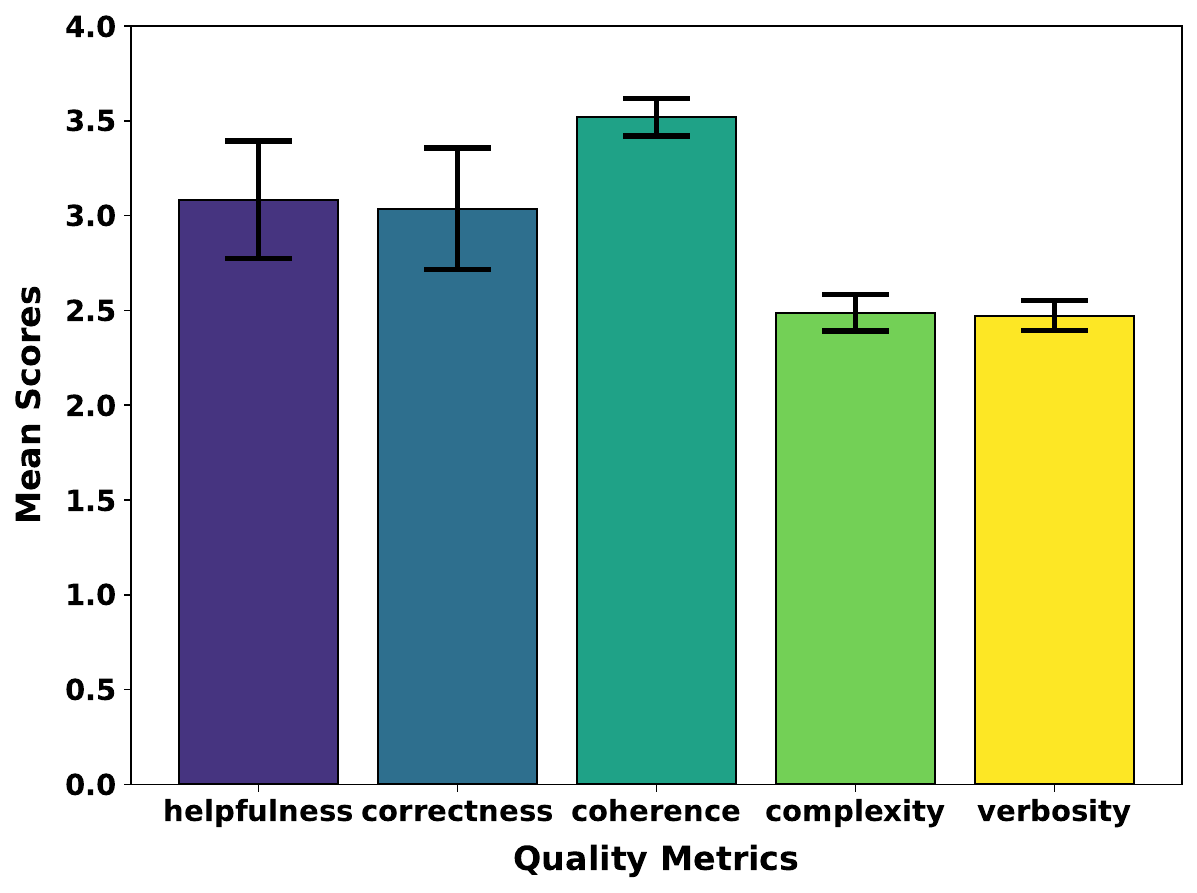}
}
\vspace{-3mm}
\caption{Quality evaluation of the synthetic \textit{SynDIP} dataset using the NVIDIA-Nemotron-4-340B reward model. Each chemical process description (PFD $\rightarrow$ PID) is scored across five key dimensions: helpfulness, correctness, coherence, complexity, and verbosity, validating alignment with actual process schematics.}
\label{fig:chemicals_QA}
\vspace{-3mm}
\end{figure}

\begin{figure}[ht!]
\vspace{-1mm}
\centering
\resizebox{0.375\textwidth}{!}{
\includegraphics{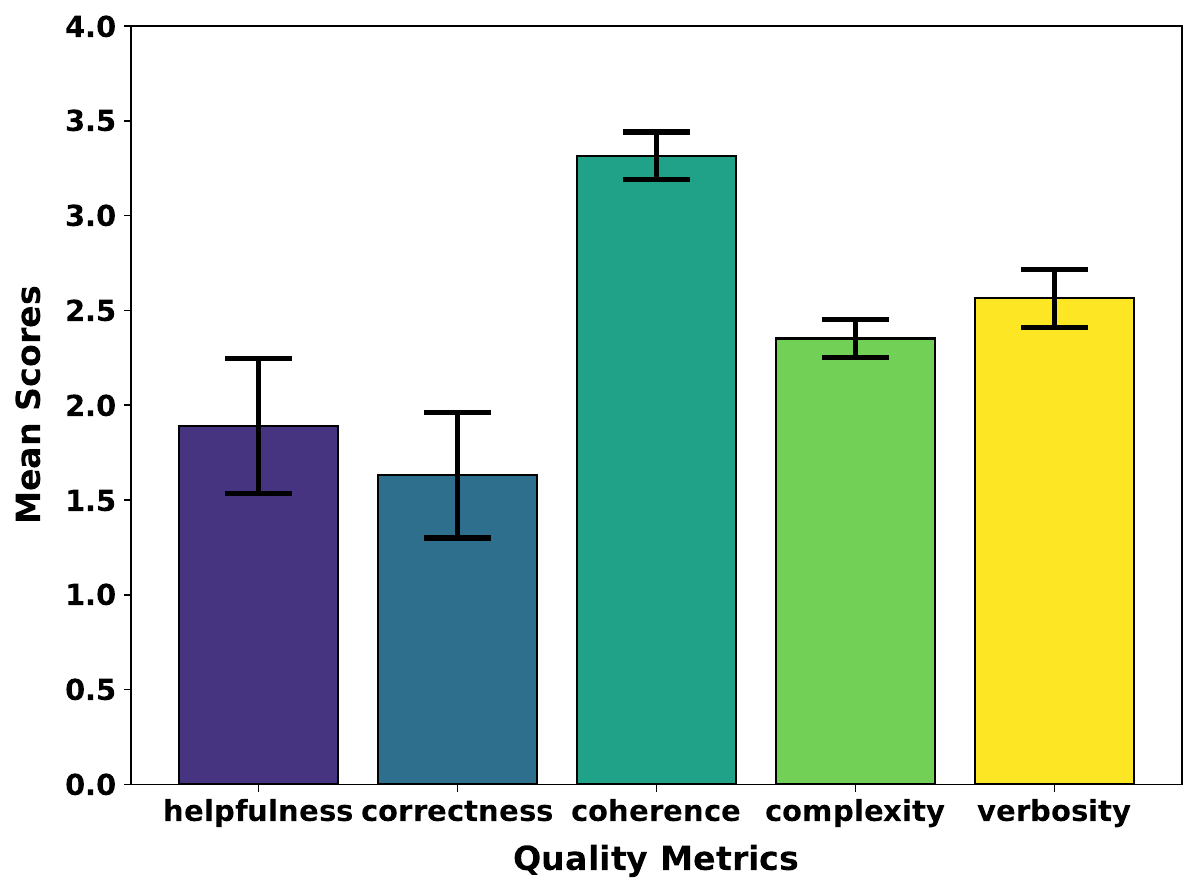}
}
\vspace{-2mm}
\caption{Quality evaluation of the reasoning-augmented \textit{LogiCore} dataset using the NVIDIA-Nemotron-4-340B reward model. Each multi-step response is scored across five dimensions (helpfulness, correctness, coherence, complexity, and verbosity) to ensure logical validity and faithful representation of PFD/PID schematics.}
\label{fig:logical_QA}
\vspace{-3mm}
\end{figure}

\begin{figure}[htbp!]
\centering
\resizebox{0.375\textwidth}{!}{
\includegraphics{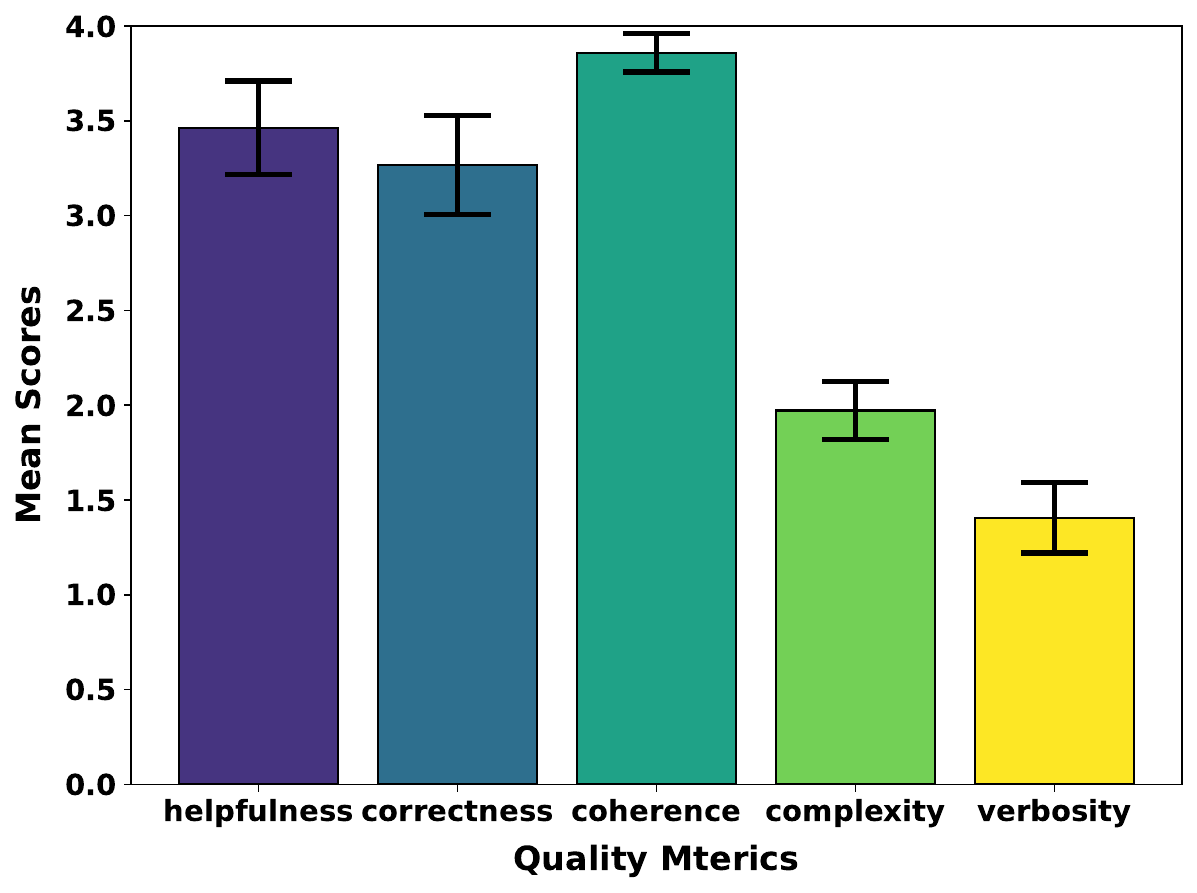}
}
\vspace{-3mm}
\caption{Quality evaluation of the \textit{Local RAIT} synthetic dataset using the NVIDIA-Nemotron-4-340B reward model. Performance across five metrics (helpfulness, correctness, coherence, complexity, and verbosity) demonstrates the quality of retrieval-augmented QA pairs grounded in individual document chunks.}
\label{fig:localRAG}
\vspace{-2mm}
\end{figure}

\begin{figure}[htbp!]
\vspace{-2mm}
\centering
\resizebox{0.375\textwidth}{!}{
\includegraphics{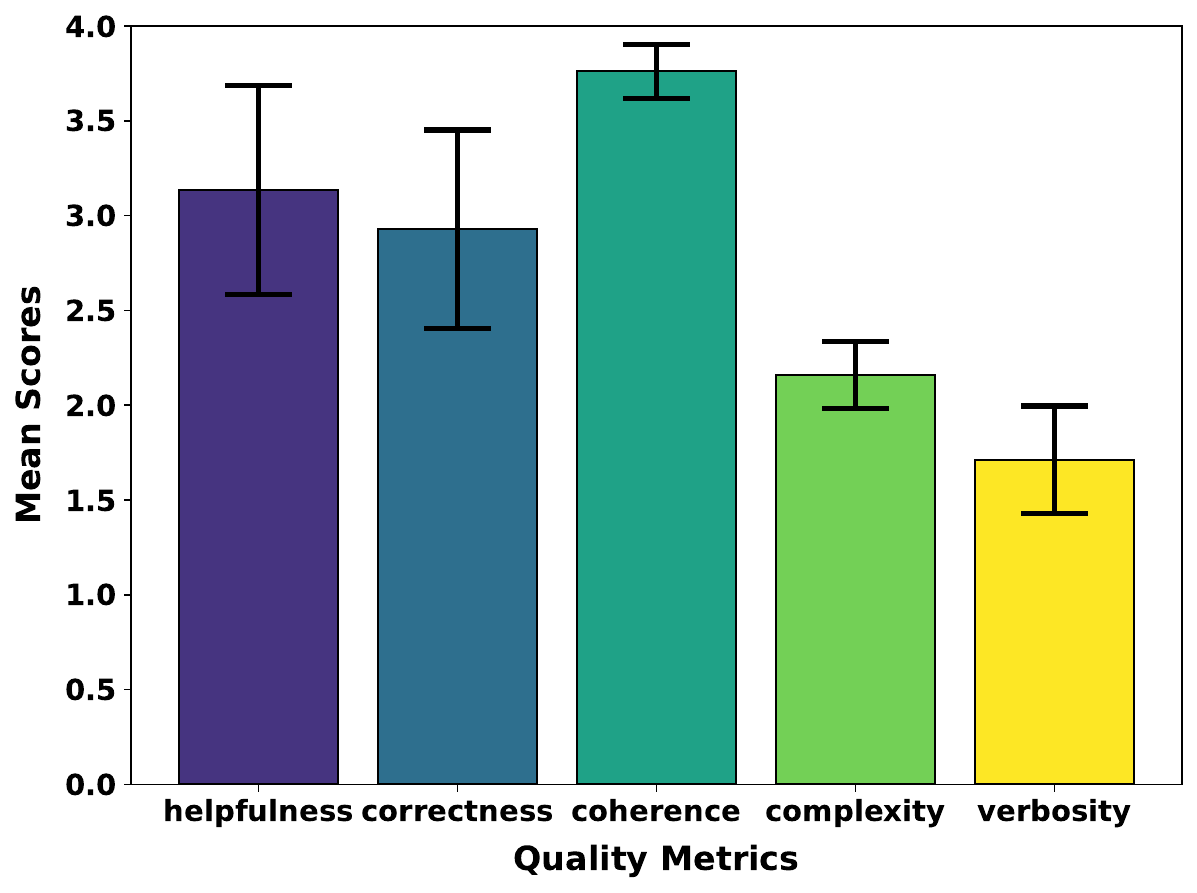}
}
\vspace{-3mm}
\caption{Quality evaluation of the synthetic \textit{Global RAIT} dataset using the NVIDIA-Nemotron-4-340B reward model. The scores reflect the effectiveness of answers generated from clustered document chunks, demonstrating robust intra-document and inter-document reasoning capabilities.}
\label{fig:GlobalRAG}
\vspace{-5mm}
\end{figure}

Our teacher-student transfer learning framework utilizes large language models (LLMs) - with GPT-4o and Claude-3-Haiku for generation and the NVIDIA-Nemotron-4-340B reward model for evaluation - to create high-quality synthetic datasets for fine-tuning SLMs including Llama 3.2-1B, Qwen 2.5-1.5B, and SmolLM2-135M. These models are specifically optimized for domain-specific tasks involving PFD and PID analysis, interpretation, and generation. The approach enables precise output ranking and filtering that aligns with human preference criteria throughout the synthetic dataset creation and evaluation process. We rigorously evaluated each synthetic dataset using the NVIDIA-Nemotron-4-340B reward model, which scores outputs on a 0-4 scale across five key metrics: helpfulness, correctness, coherence, complexity, and verbosity. Figure~\ref{fig:closed_QA} presents the evaluation results for the \textit{Factual QA} dataset, while Figures~\ref{fig:DPO_chosen} and~\ref{fig:DPO_rejected} show the performance comparison between chosen and rejected responses in the \textit{DPO} dataset. Figure~\ref{fig:chemicals_QA} demonstrates the quality of the multi-stage \textit{SynDIP} dataset generation process for producing PFDs and PIDs. Figure~\ref{fig:logical_QA} shows the evaluation of the multi-step reasoning in the \textit{LogiCore} dataset. Figures~\ref{fig:localRAG} (\textit{Local RAIT}) and~\ref{fig:GlobalRAG} (\textit{Global RAIT}) collectively demonstrate the quality and training-objective suitability of the retrieval-augmented datasets.

%%%%%%%%%%%%%%%%%%%%%%%%%%%%%%%%%%%%%%%%%%%%%%%%%%%%%%%%%%%%%%%
\vspace{-2mm}

\vspace{-2mm}

\vspace{-2mm}

\begin{figure*}[ht!]
\vspace{-2mm}
\centering
\resizebox{0.785\textwidth}{!}{
\includegraphics[keepaspectratio,trim=0.0cm 2.5cm 0cm 2.5cm,clip]{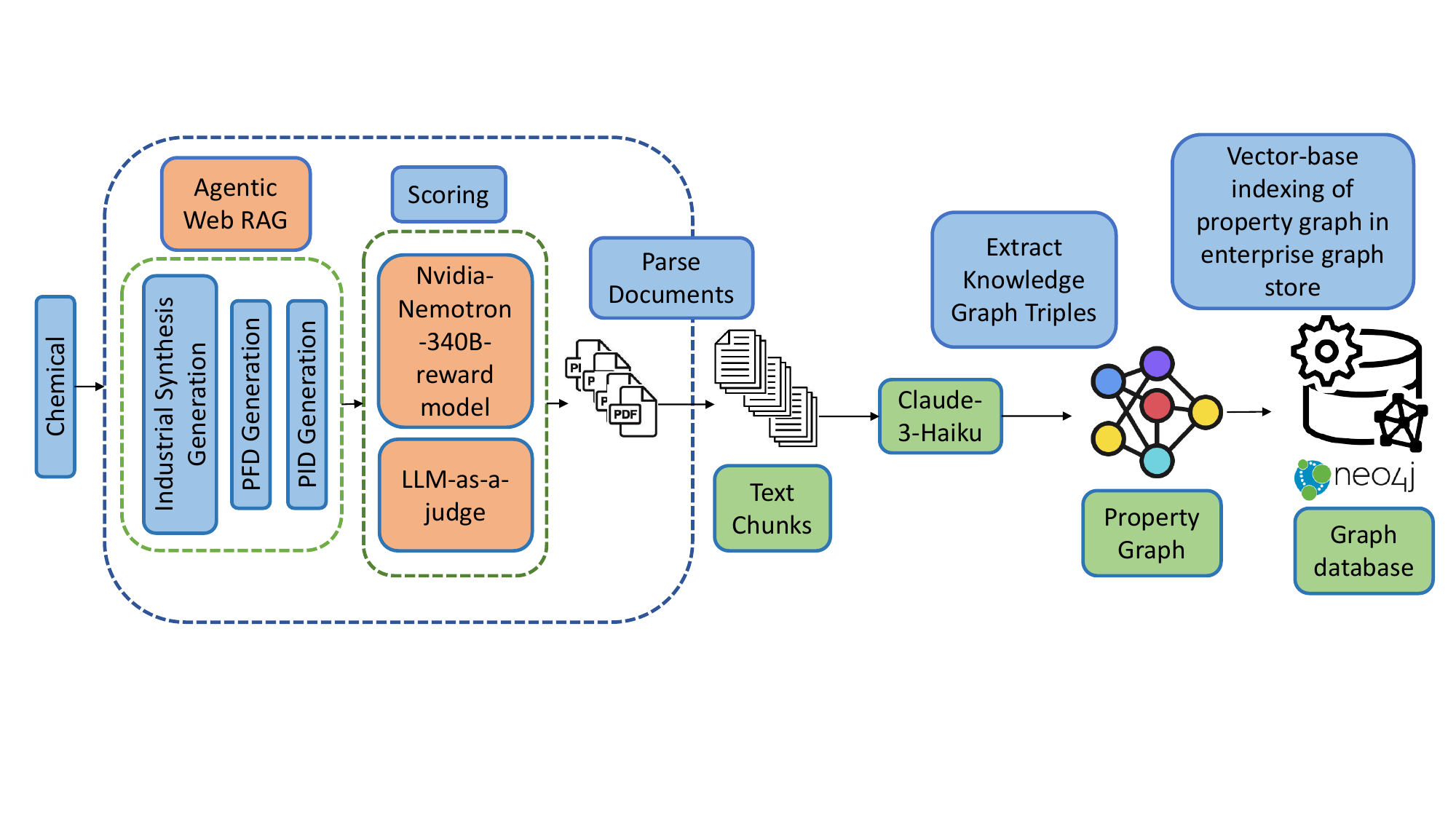} % trim = <left> <bottom> <right> <top>
}
\vspace{-8mm}
\caption{The figure illustrates the end-to-end Graph Retrieval-Augmented Generation (Graph RAG) pipeline for PFD/PID interpretation in chemical process engineering. A multimodal agentic framework—comprising expert agents coordinated by a meta-agent—retrieves and processes data for knowledge graph construction. Unstructured documents are parsed into text chunks, from which knowledge graph triples are extracted and structured into a property graph. The resulting graph is vector-indexed for similarity-based retrieval. Validation leverages LLM-as-a-Judge (GPT-4o) and reward models (NVIDIA Nemotron-4-340B) to optimize knowledge extraction, ensuring factual accuracy, coherence, and task relevance.}
\label{fig:kg}
\vspace{-2mm}
\end{figure*}

\vspace{3mm}
\subsection{Graph Retrieval-Augmented Generation (Graph RAG)}
Retrieval-Augmented Generation (RAG) enhances large language models (LLMs) by integrating external knowledge databases, enabling precise fact retrieval for domain-specific question answering. Graph RAG~\cite{han2024retrieval, edge2024local, he2024g}(refer Figure~\ref{fig:kg}) extends this paradigm by incorporating structured knowledge graphs, which offer three key advantages: (1) relational data organization for complex reasoning tasks; (2) explicit relationship traversal; and (3) multi-source information synthesis. This architecture supports multi-hop reasoning across interconnected knowledge nodes, significantly improving contextual understanding and response accuracy in open-domain question answering (ODQA). The structured representation leads to more precise and contextually grounded responses than conventional RAG approaches. As previously discussed, our framework employs specialized agents for autonomous web navigation to collect chemical-specific multimodal data from online sources, focusing on PFD and PID documentation. The aggregated web data is first stored as raw documents and then transformed into knowledge graphs. This transformation begins by processing unstructured documents into property graphs through the following steps. For each document \( t_i \), we first segment its text into smaller chunks using a sliding window approach. Let \( C_i = \{c_1, c_2, \ldots, c_M\} \) represent the set of text chunks from \( t_i \), where each chunk \( c_j \) has length \( |c_j| \). Using a window size \( w \) and stride \( s \), the sliding window technique generates chunks spanning positions \( p_j \) to \( p_j + w - 1 \), where \( p_j = 1 + (j - 1) \cdot s \). This overlapping segmentation preserves contextual continuity between chunks. To enhance semantic representation, we employ the language model \( \mathcal{M}_{\theta} \) to generate relational descriptions \( \mathcal{D}_j \) that capture inter-chunk relationships:

\vspace{-1mm}
\resizebox{0.985\linewidth}{!}{%
\begin{minipage}{\linewidth}
\begin{equation}
\mathcal{D}_j = \mathcal{M}_{\theta}(c_j, C_i \setminus \{c_j\}) \nonumber
\end{equation}
\end{minipage}
}

\vspace{1mm}
The enriched chunk \( c'_j = c_j \oplus \mathcal{D}_j \) combines original content with its relational context, where \( \oplus \) denotes concatenation, forming nodes in the knowledge graph. These augmented chunks support downstream graph operations via structured triple representations of the form (subject, predicate, object), where entities (subjects/objects) are connected through semantic predicates. For a given enriched chunk \( c'_j \), the extraction process involves the following steps: (1) Entities are represented as distinct nodes. Let \( E_j = \{e_{j1}, e_{j2}, \ldots, e_{jK_j}\} \) denote the set of entities extracted from \( c'_j \), where \( e_{jk} \) is the \( k \)-th entity and \( K_j = |E_j| \) is the entity count. (2) Inter-entity relations are represented as directed edges. Define the set of predicates as \( \mathcal{P}_j = \{r_{jkm} \mid 1 \leq k \neq m \leq K_j\} \), where \( r_{jkm} \) denotes the relation between entities \( e_{jk} \) and \( e_{jm} \). The extracted triples from \( c'_j \) are:

\vspace{-1mm}
\resizebox{0.985\linewidth}{!}{%
\begin{minipage}{\linewidth}
\begin{equation}
\mathcal{T}_j = \{(e_{jk}, r_{jkm}, e_{jm}) \mid 1 \leq k \neq m \leq K_j\} \nonumber
\end{equation}
\end{minipage}
}

\vspace{1mm}
Each triple \( (e_{jk}, r_{jkm}, e_{jm}) \) represents a directed relation from \( e_{jk} \) to \( e_{jm} \) via predicate \( r_{jkm} \). The union of triples from all enriched chunks \( C'_i = \{c'_1, c'_2, \ldots, c'_M\} \) forms the knowledge graph \( \mathcal{G}_i \), where entity nodes connect via predicate edges. Each entity \( e_{jk} \) is linked to its source chunk \( c'_j \) using an origin relation:

\vspace{-1mm}
\resizebox{0.985\linewidth}{!}{
\begin{minipage}{\linewidth}
\begin{equation}
(e_{jk}, \textsc{belongs\_to}, c'_j), \quad \forall e_{jk} \in E_j
\nonumber
\end{equation}
\end{minipage}
}

where \textsc{belongs\_to} denotes the entity-chunk association. The resulting knowledge graph captures both semantic relationships (via triples) and source attribution (via origin links). The knowledge graph \( \mathcal{G}_i \) is formally defined as a directed graph \( \mathcal{G}_i = (\mathcal{V}_i, \mathcal{E}_i) \), with:

\vspace{-1mm}
\resizebox{0.985\linewidth}{!}{
\begin{minipage}{\linewidth}
\begin{equation}
\mathcal{V}_i = \{c'_1, \ldots, c'_M\} \cup \{e_{jk} \mid j = 1, \ldots, M; k = 1, \ldots, K_j\} \nonumber
\end{equation}
\end{minipage}
}

\vspace{1mm}
comprising chunk nodes \( c'_j \) and entity nodes \( e_{jk} \). The edge set \( \mathcal{E}_i \) includes: (1) Semantic relation edges:

\vspace{-1mm}
\resizebox{0.985\linewidth}{!}{
\begin{minipage}{\linewidth}
\begin{equation}
\mathcal{E}_i^{\text{rel}} = \{(e_{jk}, r_{jkm}, e_{jm}) \mid j = 1, \ldots, M; 1 \leq k \neq m \leq K_j\} \nonumber
\end{equation}
\end{minipage}
}

(2) Structural containment edges:

\resizebox{0.985\linewidth}{!}{
\begin{minipage}{\linewidth}
\begin{equation}
\mathcal{E}_i^{\text{cont}} = \{(c'_j, e_{jk}) \mid j = 1, \ldots, M; k = 1, \ldots, K_j\} \nonumber
\end{equation}
\end{minipage}
}

\vspace{1mm}
The complete edge set is \( \mathcal{E}_i = \mathcal{E}_i^{\text{rel}} \cup \mathcal{E}_i^{\text{cont}} \). This heterogeneous graph structure—combining chunk and entity nodes with relational and containment edges—enables robust graph-based retrieval, reasoning, and generation. To improve knowledge retrieval accuracy, we implement a two-step entity resolution process to identify and merge duplicate entities referring to the same concept. Each entity \( e_{jk} \) is encoded as a vector embedding \( v_{jk} \) using a text embedding model to capture semantic representation. For any pair of entities \( e_{jk} \) (from chunk \( c'_j \)) and \( e_{j'k'} \) (from chunk \( c'_{j'} \)), we assess conceptual equivalence through sequential similarity evaluations. First, we compute cosine similarity between their embeddings:

\vspace{0mm}
\resizebox{0.985\linewidth}{!}{
\begin{minipage}{\linewidth}
\begin{equation}
\text{sim}(v_{jk}, v_{j'k'}) = \frac{v_{jk} \cdot v_{j'k'}}{\|v_{jk}\| \, \|v_{j'k'}\|} \nonumber
\end{equation}
\end{minipage}
}

\vspace{1mm}
If the semantic similarity exceeds a threshold \( \tau_{\text{sim}} \), we conduct a secondary evaluation using normalized Levenshtein distance:

\vspace{0mm}
\resizebox{0.985\linewidth}{!}{
\begin{minipage}{\linewidth}
\begin{equation}
\text{str\_sim}(e_{jk}, e_{j'k'}) = 1 - \frac{d_{\text{lev}}(e_{jk}, e_{j'k'})}{\max(|e_{jk}|, |e_{j'k'}|)} \nonumber
\end{equation}
\end{minipage}
}

\vspace{1mm}
Here, \( d_{\text{lev}}(e_{jk}, e_{j'k'}) \) is the Levenshtein distance between entity strings, and \( |e_{jk}| \), \( |e_{j'k'}| \) are their lengths. Entities are merged as duplicates only when both similarity metrics exceed their respective thresholds \( \tau_{\text{sim}} \) (semantic) and \( \tau_{\text{str}} \) (string-based), ensuring robust entity consolidation. We apply the hierarchical Leiden algorithm to detect communities \( \mathcal{C}_k \) at various granularities within the knowledge graph \( \mathcal{G}_i \), aiming to optimize modularity \( M_{\text{Mod}} \). Modularity measures the quality of a community structure by comparing the density of intra-community edges to the expected density if edges were placed randomly while preserving node degrees. It is defined as:

\vspace{0mm}
\resizebox{0.985\linewidth}{!}{
\begin{minipage}{\linewidth}
\begin{equation}
M_{\text{Mod}} = \frac{1}{2m} \sum_{i,j} \left[ \mathcal{A}_{ij} - \frac{d_i d_j}{2m} \right] \delta(c_i, c_j) \nonumber
\end{equation}
\end{minipage}
}

\vspace{1mm}
where \( \mathcal{A}_{ij} \) is the adjacency matrix (1 if an edge exists between nodes \( i \) and \( j \), 0 otherwise), \( d_i \) and \( d_j \) are the degrees of nodes \( i \) and \( j \), respectively, and \( m \) is the total number of edges in the graph. The term \( \frac{d_i d_j}{2m} \) represents the expected number of edges between \( i \) and \( j \) under the configuration model. The function \( \delta(c_i, c_j) \) is the Kronecker delta, equal to 1 if nodes \( i \) and \( j \) belong to the same community and 0 otherwise. A community \( \mathcal{C}_k = (\mathcal{V}_{\mathcal{C}_k}, \mathcal{E}_{\mathcal{C}_k}) \) is a subgraph where the nodes \( \mathcal{V}_{\mathcal{C}_k} \subseteq \mathcal{V}_i \) and edges \( \mathcal{E}_{\mathcal{C}_k} \subseteq \mathcal{E}_i \) are more densely connected internally than to nodes outside the community. These communities organize the graph into densely connected subgraphs, typically representing specific topics or contexts. This structure enhances retrieval by scoping searches within relevant communities and facilitates reasoning by grouping related facts necessary for multi-step inference.
The hierarchical Leiden algorithm decomposes \( \mathcal{G}_i = (\mathcal{V}_i, \mathcal{E}_i) \) into \( L \) disjoint communities \( \{\mathcal{C}_k\}_{k=1}^L \) through modularity maximization, where \( \mathcal{C}_k = (\mathcal{V}_{\mathcal{C}_k}, \mathcal{E}_{\mathcal{C}_k}) \) denotes the \( k \)-th community subgraph. This optimization proceeds iteratively through three phases: (1) local node reassignment to neighboring communities to improve modularity; (2) aggregation of communities into super-nodes to construct a reduced graph; and (3) repetition of this procedure on the coarse-grained graph until convergence, yielding a hierarchical community structure. For complex reasoning tasks, relevant information often spans multiple communities, necessitating efficient retrieval by identifying communities aligned with query-specific subgraphs. To achieve this, we rank the top-\( K \) communities \( \{\mathcal{C}_1, \mathcal{C}_2, \ldots, \mathcal{C}_K\} \) based on their cosine similarity to the user query \( \mathcal{Q} \) and the summaries of relationship paths within each community. Each community \( \mathcal{C}_k \) is summarized using the language model \( \mathcal{M}_{\theta} \), which encodes its relational edges \( \mathcal{E}_{\mathcal{C}_k}^{\text{rel}} \) (subject-predicate-object triples) into a summary \( s_k \). The summarization process is formalized as:

\resizebox{0.985\linewidth}{!}{
\begin{minipage}{\linewidth}
\begin{equation}
s_k = \mathcal{M}_{\theta}(\mathcal{E}_{\mathcal{C}_k}^{\text{rel}}) = \arg\max_{S} P(S \mid \mathcal{E}_{\mathcal{C}_k}^{\text{rel}}) \nonumber
\end{equation}
\end{minipage}
}

where \( P(S \mid \mathcal{E}_{\mathcal{C}_k}^{\text{rel}}) \) is the likelihood of generating a summary \( S \) conditioned on the set of relation edges. The summary \( s_k \) retains the semantic (predicate) relationships of the original subgraph. These summaries are then encoded into vector embeddings \( v(s_k) \) using a text-embedding model, enabling efficient similarity computation with the query embedding \( v(\mathcal{Q}) \):

\resizebox{0.985\linewidth}{!}{
\begin{minipage}{\linewidth}
\begin{equation}
\text{sim}(\mathcal{Q}, \mathcal{C}_k) = \frac{\langle v(\mathcal{Q}), v(s_k) \rangle}{\|v(\mathcal{Q})\| \, \|v(s_k)\|} \nonumber
\end{equation}
\end{minipage}
}

\begin{figure}[ht!]
\vspace{-4mm}
\includegraphics[width=\linewidth,trim=0.0cm 4.0cm 0cm 4.0cm,clip]{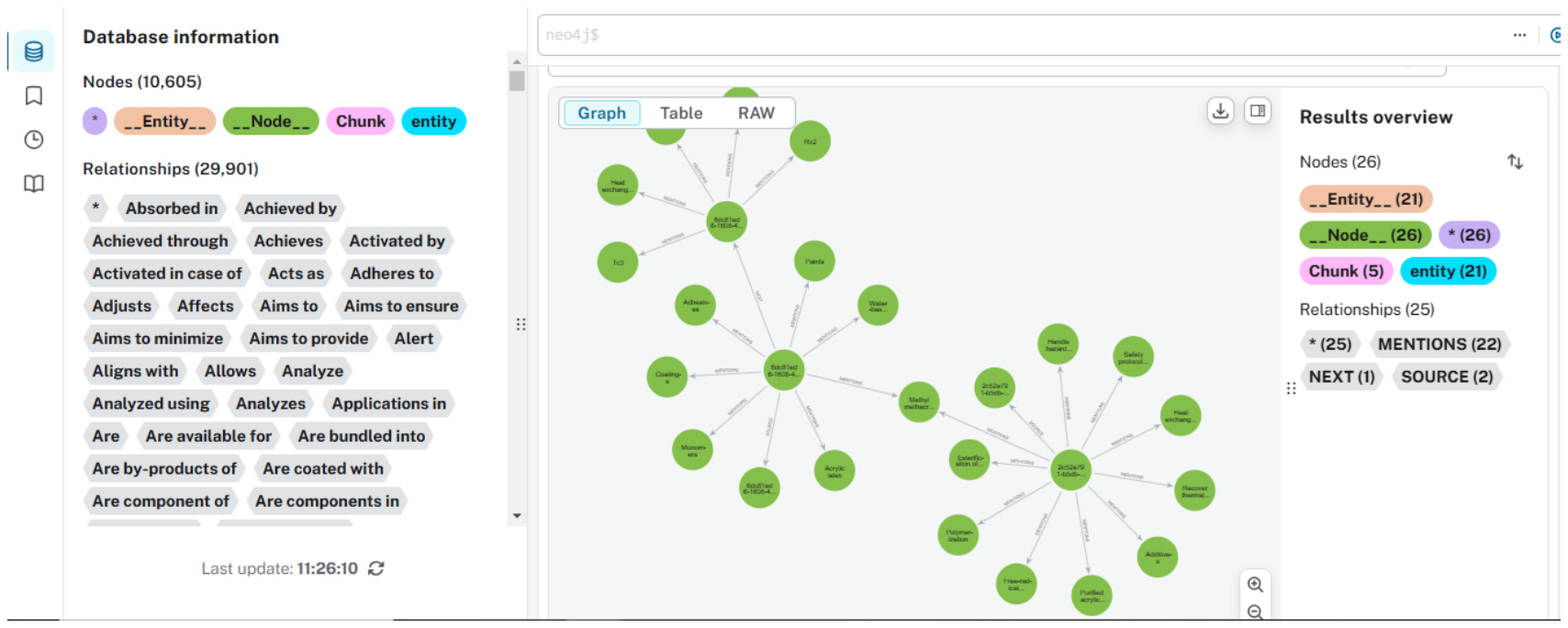}  % trim = <left> <bottom> <right> <top>
\vspace{-8mm}
\caption{Visualization of the Neo4j knowledge graph constructed for the Graph RAG framework, showing a subset of a larger graph containing 10,605 nodes and 29,901 edges. The graph includes two types of nodes: chunk nodes (text segments enriched with contextual relationships) and entity nodes (named concepts extracted from text). Edges represent \texttt{MENTIONS} (linking entities to their originating chunks) and semantic relationships between entities, modeled as subject–predicate–object triples. This structured organization supports multi-hop reasoning and community-based retrieval, enabling the generation of accurate, context-rich descriptions of chemical processes such as PFDs and PIDs.}
\label{fig:neo}
\vspace{-5mm}
\end{figure}

The top-\( K \) communities with the highest similarity scores are selected and combined into a query-specific subgraph \( \mathcal{G}_Q = (\mathcal{V}_Q, \mathcal{E}_Q) \), defined as:

\resizebox{0.985\linewidth}{!}{
\begin{minipage}{\linewidth}
\begin{equation}
\mathcal{V}_Q = \bigcup_{k=1}^K \mathcal{V}_{\mathcal{C}_k}, \quad \mathcal{E}_Q = \bigcup_{k=1}^K \mathcal{E}_{\mathcal{C}_k} \nonumber
\end{equation}
\end{minipage}
}

where \( \mathcal{V}_Q \) and \( \mathcal{E}_Q \) are the union of nodes and edges, respectively, from the selected top-\( K \) communities. This subgraph captures dependencies across disparate facts while retaining critical relationships necessary to answer the user query. Communities \( \mathcal{C}_k \) are precomputed via the Leiden algorithm, ensuring modularity-optimized clustering. The top-\( K \) selection scales sublinearly with graph size, as coarse-grained retrieval via community summaries reduces search space before fine-grained traversal. Finally, the language model \( \mathcal{M}_{\theta} \) generates the answer \( \hat{A} \) by conditioning on the query \( \mathcal{Q} \) and the subgraph \( \mathcal{G}_Q \):

\vspace{-1mm}
\resizebox{0.985\linewidth}{!}{
\begin{minipage}{\linewidth}
\begin{equation}
\hat{A} = \mathcal{M}_{\theta}(\mathcal{Q}, \mathcal{G}_Q) = \arg\max_{\hat{A}} P(\hat{A} \mid \mathcal{Q}, \mathcal{G}_Q) \nonumber
\end{equation}
\end{minipage}
}

where \( P(\hat{A} \mid \mathcal{Q}, \mathcal{G}_Q) \) denotes the likelihood of generating the answer grounded in the retrieved subgraph. Figure~\ref{fig:neo} visualizes the Neo4j knowledge graph's nodes (chunks and entities) and edges, supporting Graph RAG's reasoning and retrieval process.

\subsection{Additional Results}
\begin{figure*}[t!]
\centering

% Row 1: Llama QA and DPO
\subfloat[\small Llama-3.2-1B evaluated on test splits of \textit{Factual QA}, \textit{SynDIP}, and \textit{LogiCore} after Supervised Fine-Tuning (SFT).]{
    \includegraphics[width=0.45\textwidth]{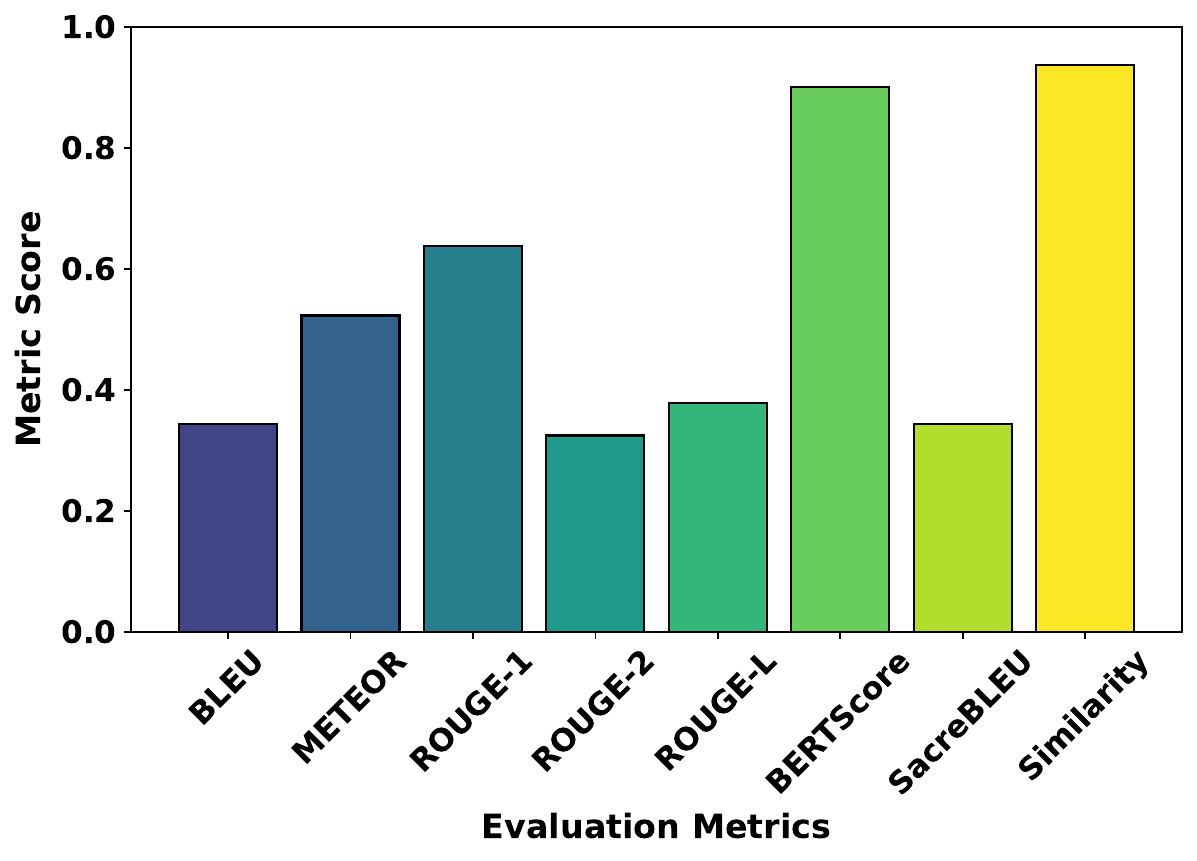}
    \label{fig:LlamaQA_metrics}
}
\hspace{8mm}
\subfloat[\small Llama-3.2-1B evaluated on the \textit{DPO} test split after Direct Preference Optimization (DPO).]{
    \includegraphics[width=0.45\textwidth]{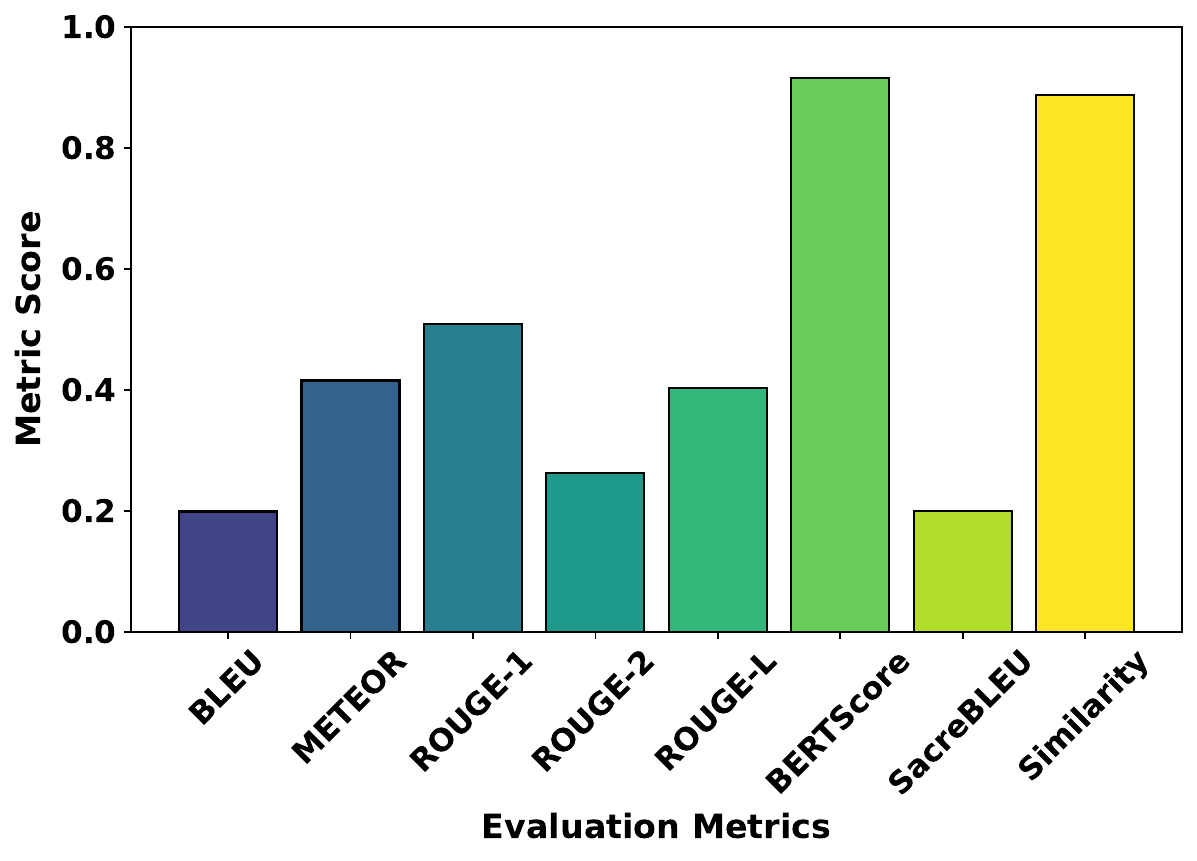}
    \label{fig:LlamaDPO_metrics}
}
\vspace{-2mm}

% Row 2: Llama RAIT and SmolLM QA
\subfloat[\small Llama-3.2-1B evaluated on test splits of \textit{Local} and \textit{Global RAIT} after Retrieval-Augmented Instruction Tuning (RAIT).]{
    \includegraphics[width=0.45\textwidth]{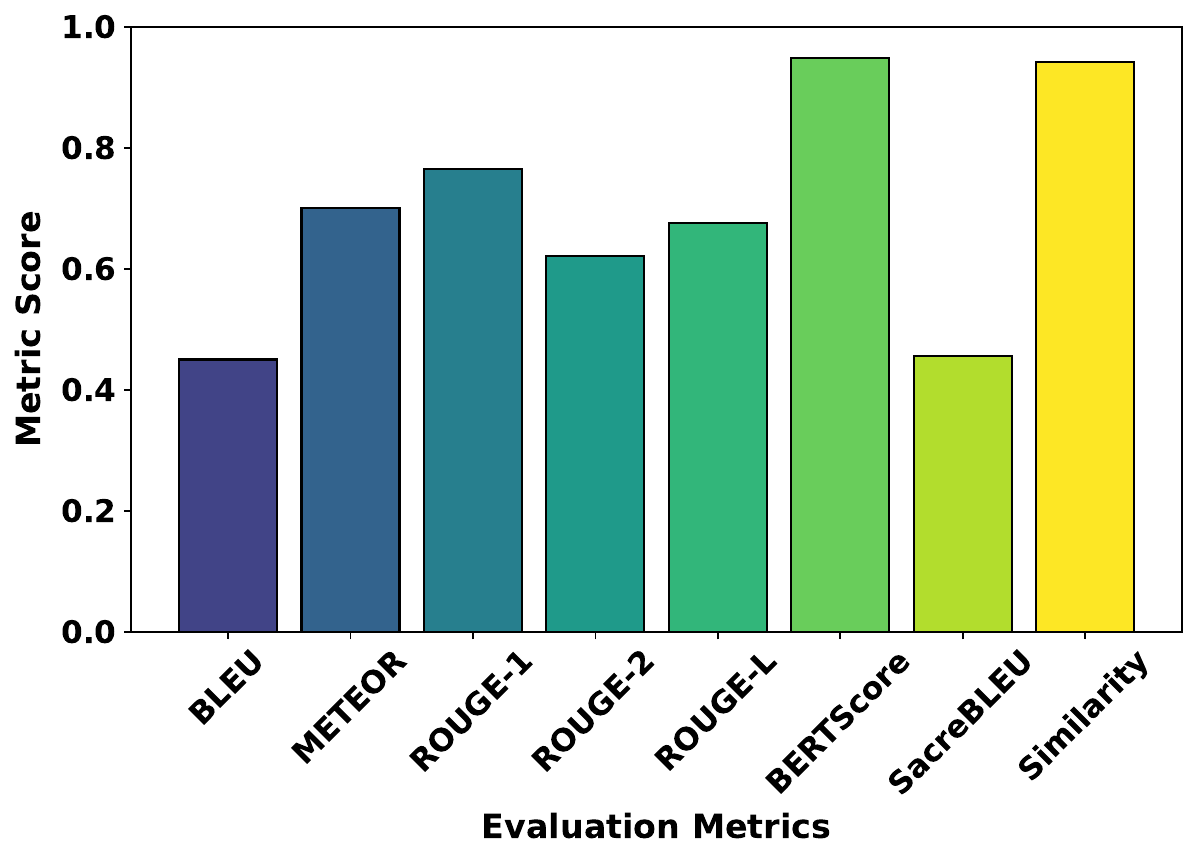}
    \label{fig:LlamaRAG_metrics}
}
\hspace{8mm}
\subfloat[\small SmolLM-135M evaluated on test splits of \textit{Factual QA}, \textit{SynDIP}, and \textit{LogiCore} after Supervised Fine-Tuning (SFT).]{
    \includegraphics[width=0.45\textwidth]{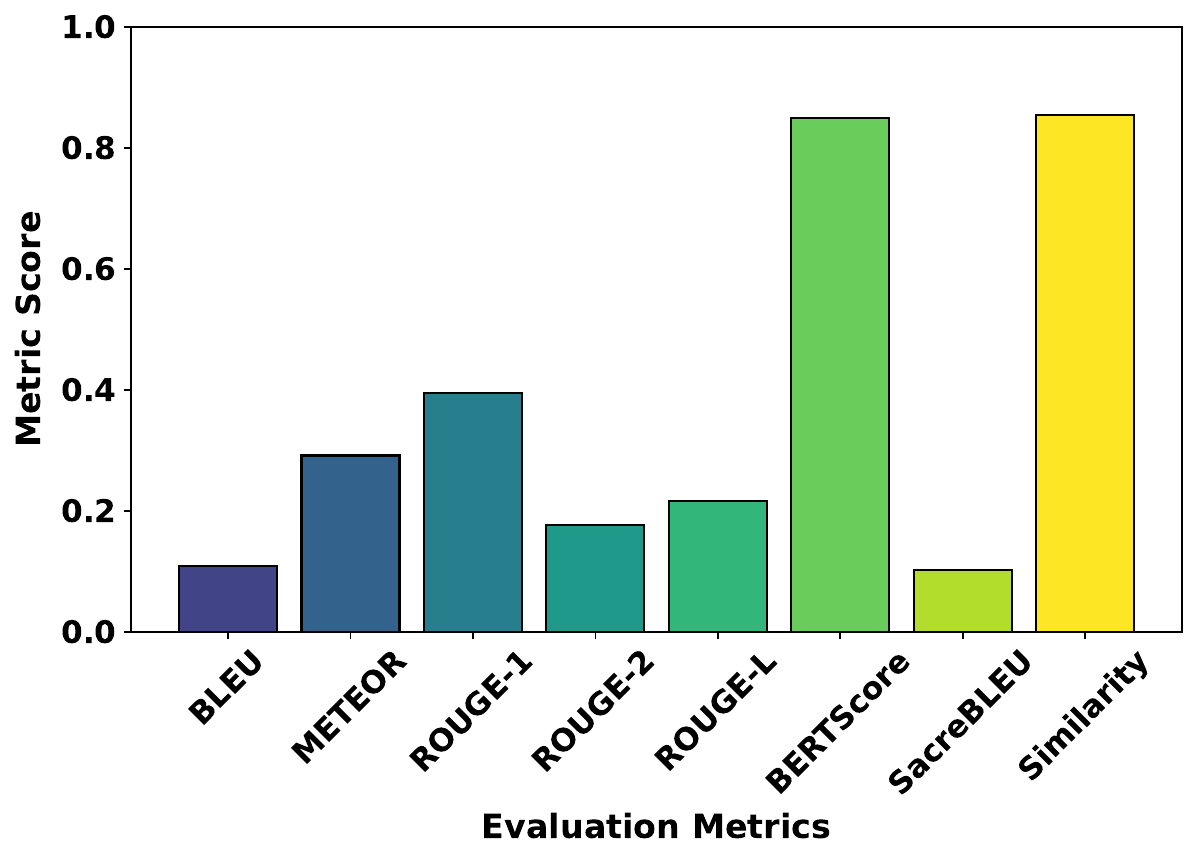}
    \label{fig:SmolLMQA_metrics}
}
\vspace{-2mm}

% Row 3: SmolLM DPO and RAIT
\subfloat[\small SmolLM-135M evaluated on the \textit{DPO} test split after Direct Preference Optimization (DPO).]{
    \includegraphics[width=0.45\textwidth]{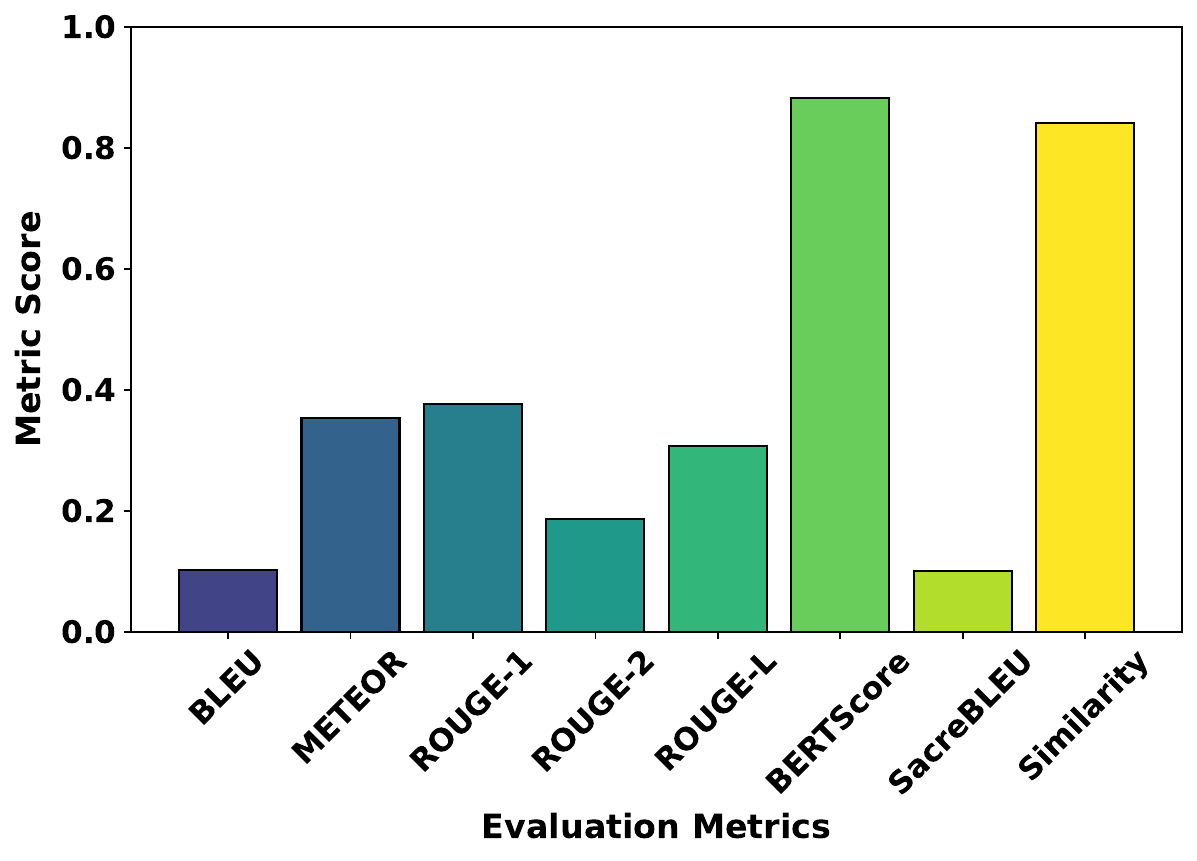}
    \label{fig:SmolLMDPO_metrics}
}
\hspace{8mm}
\subfloat[\small SmolLM-135M evaluated on test splits of \textit{Local} and \textit{Global RAIT} after Retrieval-Augmented Instruction Tuning (RAIT).]{
    \includegraphics[width=0.45\textwidth]{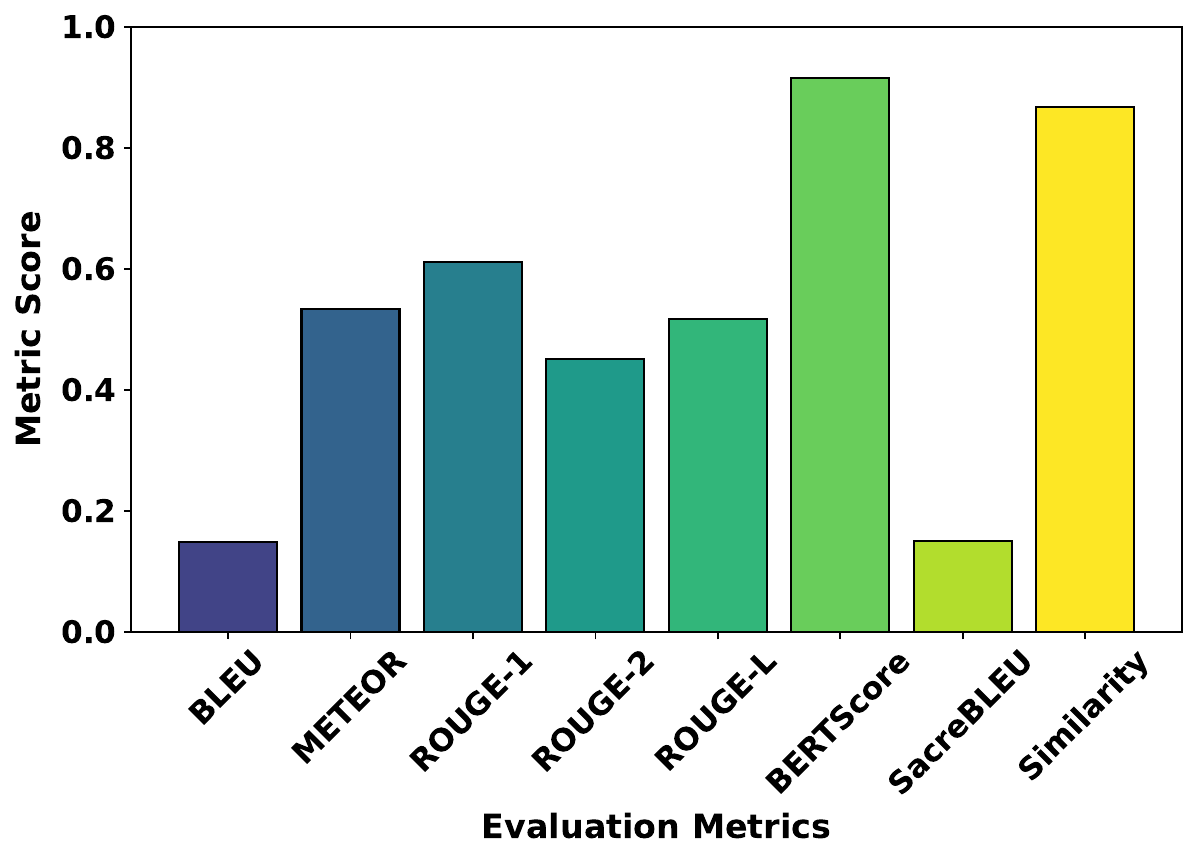}
    \label{fig:SmolLMRAG_metrics}
}

\vspace{0mm}
\caption{
Quantitative evaluation of Llama-3.2-1B and SmolLM-135M across three fine-tuning stages: (1) Supervised Fine-Tuning (SFT) on \textit{Factual QA}, \textit{SynDIP}, and \textit{LogiCore}; (2) Direct Preference Optimization (DPO) using the \textit{DPO} dataset; and (3) Retrieval-Augmented Instruction Tuning (RAIT) on \textit{Local} and \textit{Global RAIT}. Performance is evaluated on held-out test splits for each phase using both n-gram overlap metrics (BLEU, ROUGE, METEOR, SacreBLEU) and semantic similarity measures (BERTScore, sentence similarity).
}
\label{fig:emp}
\end{figure*}

\begin{figure*}[ht!]
\centering
\captionsetup[subfloat]{font=scriptsize,labelfont=bf}

\resizebox{0.85\textwidth}{!}{
\subfloat[Comparison of reward model helpfulness scores showing that both fine-tuning and retrieval augmentation improve practical utility, with \textit{Llama-3.2-1B} variants consistently outperforming \textit{SmolLM2-135M} across all configurations]{
    \includegraphics[width=50mm]{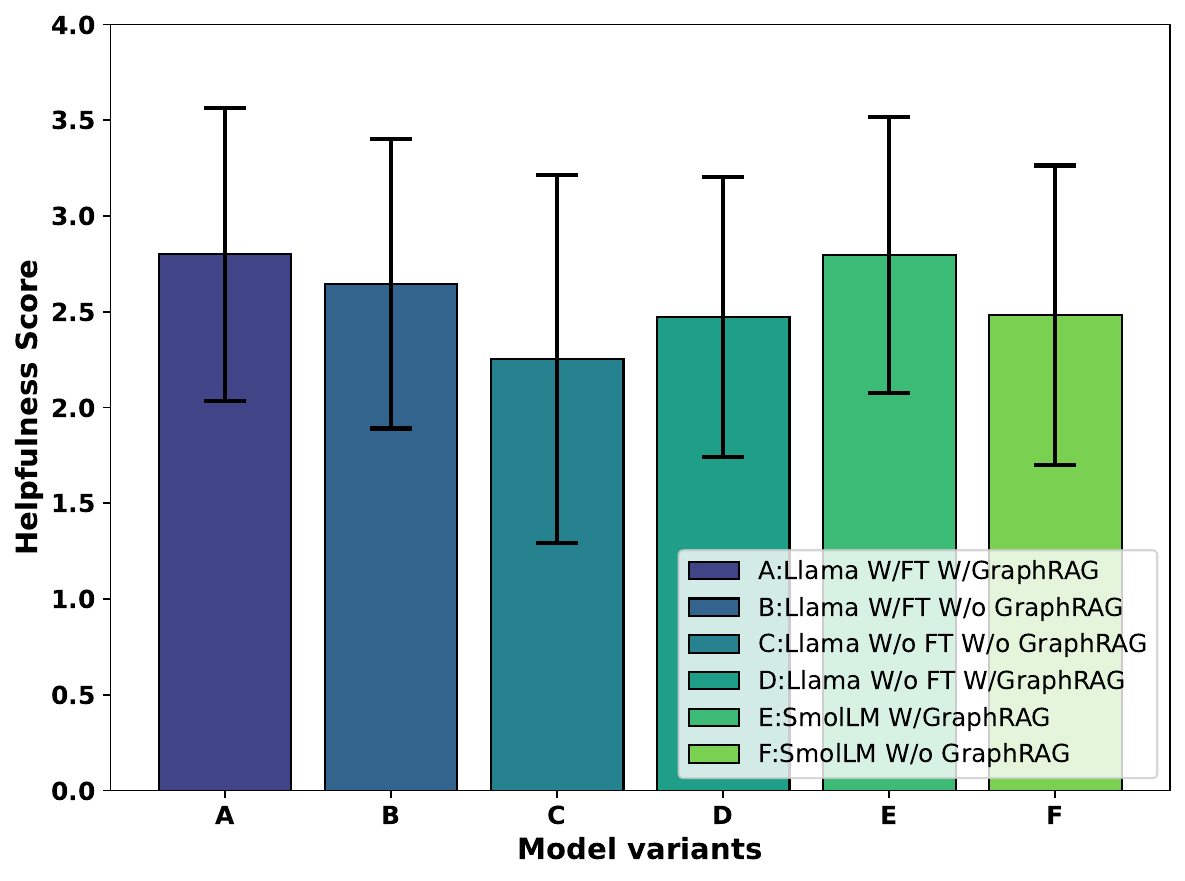}}
\hspace{0.05\textwidth}
\subfloat[Correctness evaluation demonstrating \textit{Graph RAG}'s substantial improvement in factual accuracy, particularly for \textit{Llama-3.2-1B}, confirming its effectiveness in reducing hallucinations for knowledge-intensive tasks]{
    \includegraphics[width=50mm]{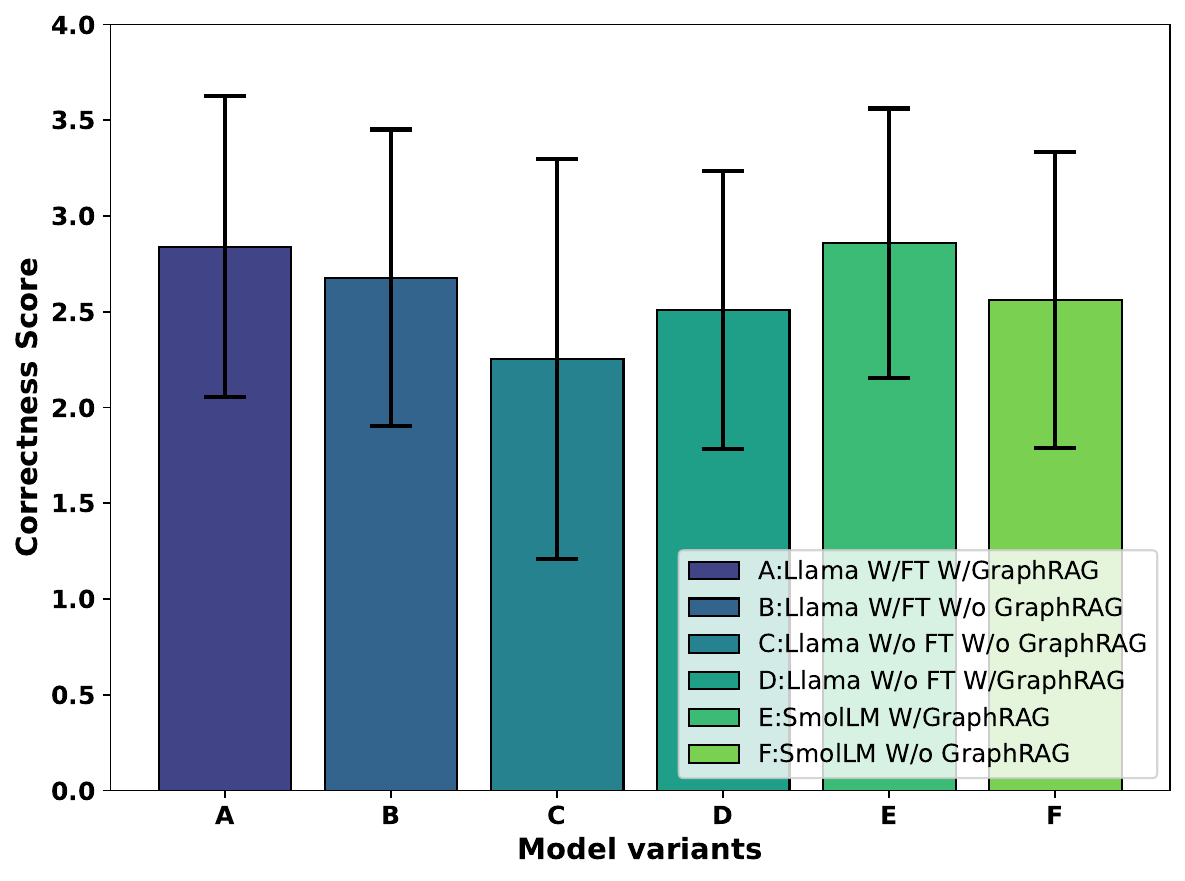}}
}

\vspace{-1mm}
\resizebox{0.85\textwidth}{!}{
\subfloat[Coherence analysis revealing that fine-tuned models produce more logically structured outputs, with \textit{Llama-3.2-1B} exhibiting superior contextual continuity and narrative fluency]{
    \includegraphics[width=50mm]{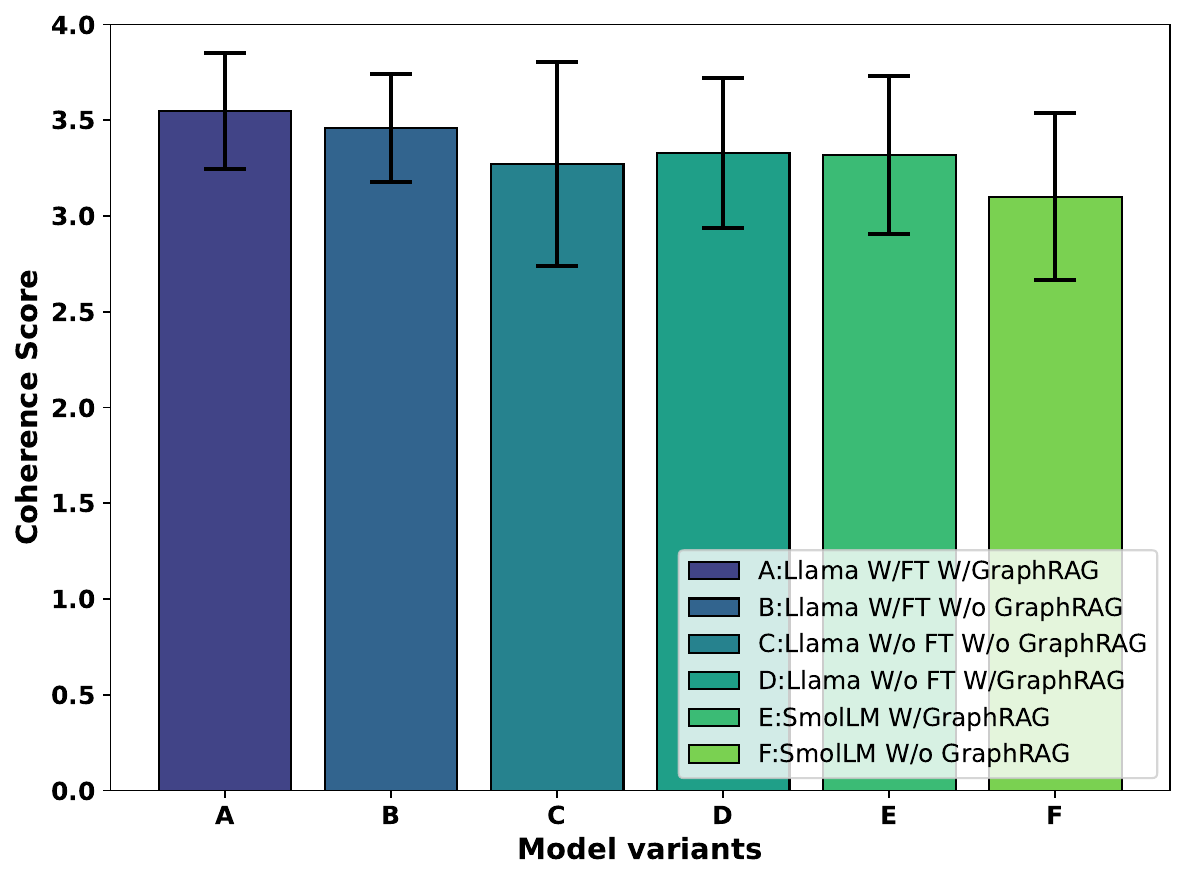}}
\hspace{0.05\textwidth}
\subfloat[Complexity scores showing fine-tuned models generate more detailed responses, while retrieval augmentation further enhances their capacity for multi-layered reasoning]{
    \includegraphics[width=50mm]{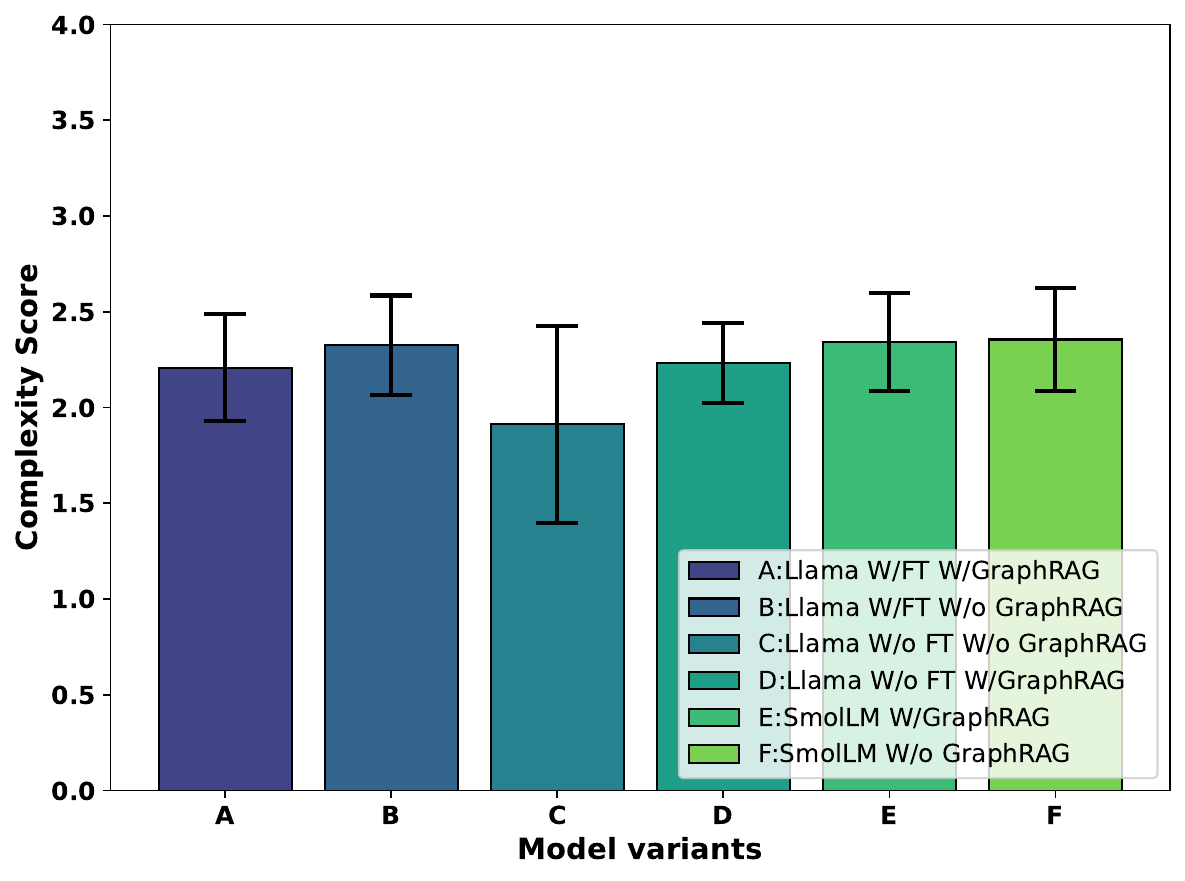}}
}

\vspace{-2mm}
\resizebox{0.425\textwidth}{!}{
\subfloat[Verbosity measurements indicating that fine-tuning increases response length, while \textit{Graph RAG} produces more concise yet informative completions by grounding generation in retrieved context]{
    \includegraphics[width=50mm]{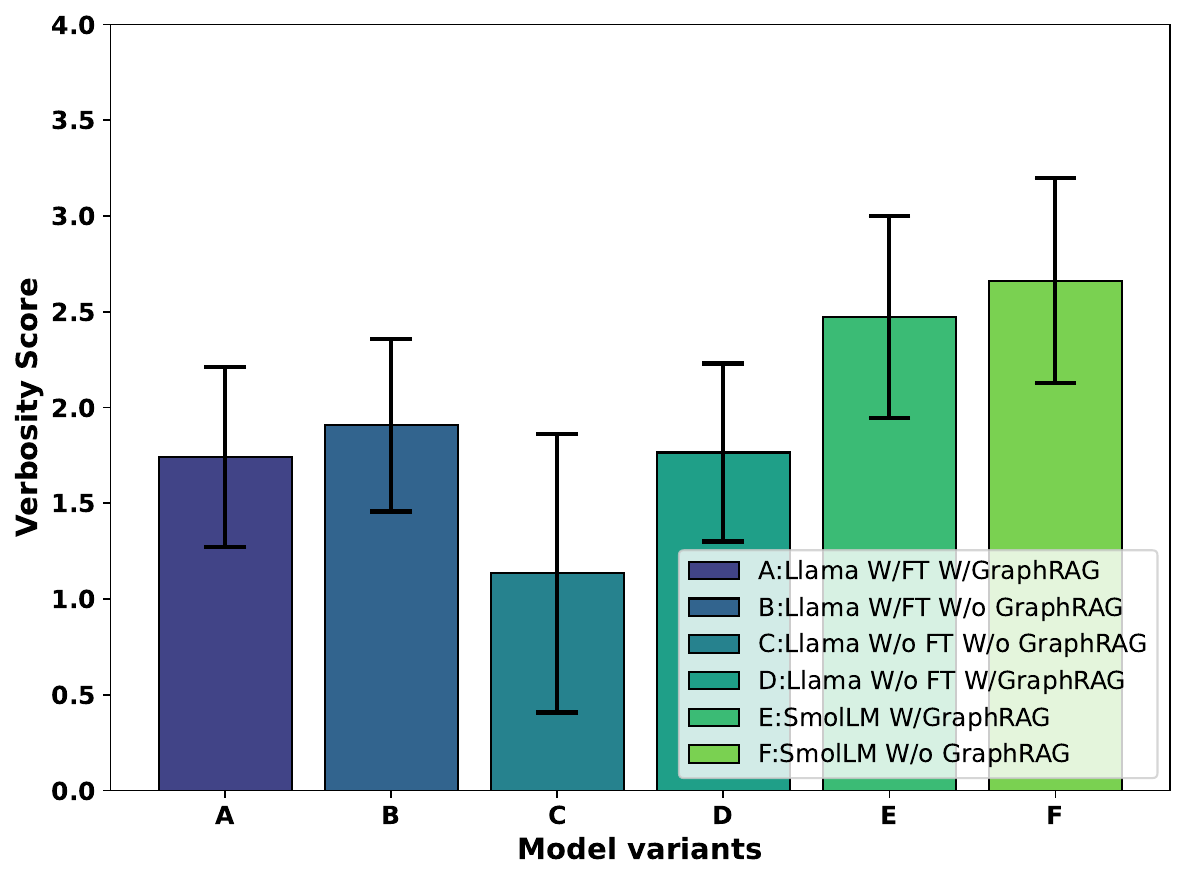}}
}

\vspace{0mm}
\caption{
Performance evaluation of six model configurations on a 1.5K QA-pair out-of-distribution benchmark, independent of all synthetic training datasets (\textit{Factual QA}, \textit{SynDIP}, \textit{LogiCore}, \textit{DPO}, and \textit{RAIT}). The \textit{Nvidia Nemotron-4-340B} reward model assessed five key dimensions: (1) helpfulness (practical utility), (2) correctness (factual accuracy), (3) coherence (logical flow), (4) complexity (content depth), and (5) verbosity (response length). Results demonstrate that fine-tuning enhances structural quality and content depth while \textit{Graph RAG} significantly improves factual precision. The \textit{Llama-3.2-1B} model combining both techniques achieves optimal performance across all dimensions, highlighting the complementary benefits of domain adaptation and structured knowledge retrieval for complex chemical process understanding.
}
\label{fig:sAF}
\end{figure*}

\begin{figure*}[ht!]
\centering
\subfloat[Supervised fine-tuning (SFT) loss for Llama 3.2 1B on \textit{Factual QA}, \textit{SynDIP}, and \textit{LogiCore} datasets. Training loss decreases from $\sim$1.4 to 0.35 in 5 epochs (blue) and reaches $\sim$0.1 after 15 epochs (red), showing consistent convergence.\label{fig:LlamaQA_loss}]{%
\includegraphics[width=0.40\textwidth]{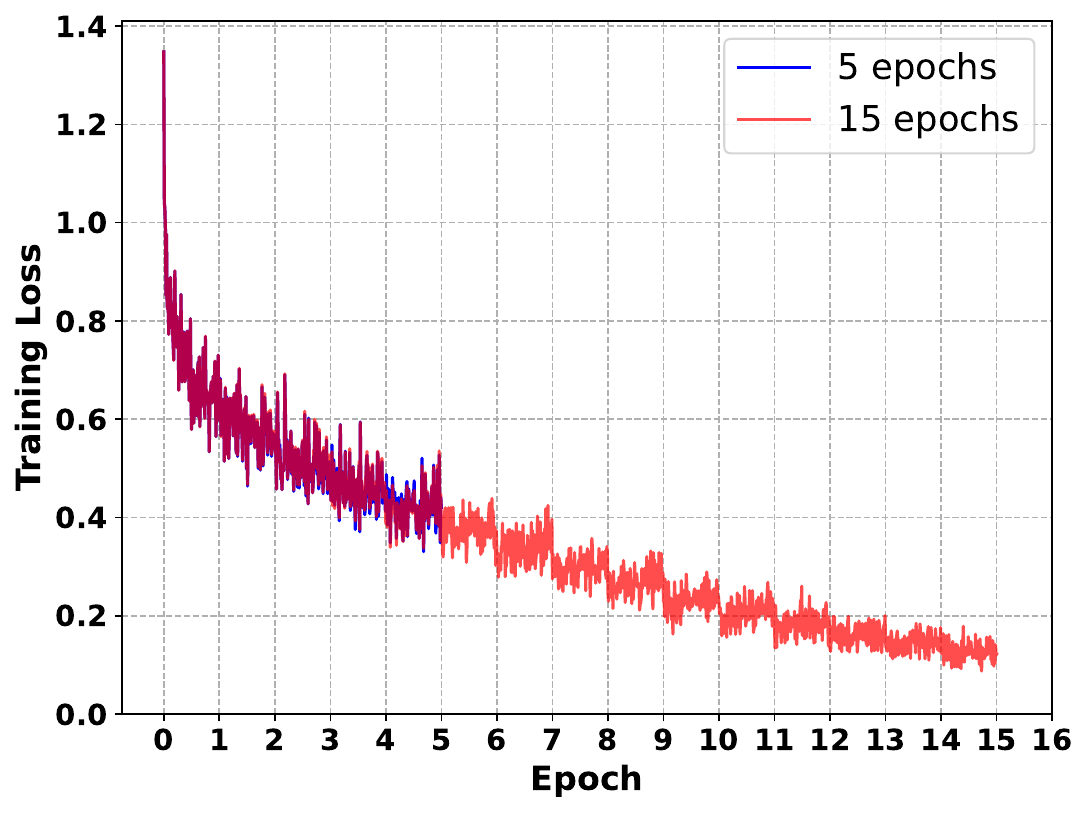}}
\hfill
\subfloat[Direct Preference Optimization (DPO) loss for Llama 3.2 1B. The loss converges to near-zero within one epoch and maintains stability through both 2-epoch (red) and 5-epoch (blue) training runs.\label{fig:LlamaDPO_loss}]{%
\includegraphics[width=0.40\textwidth]{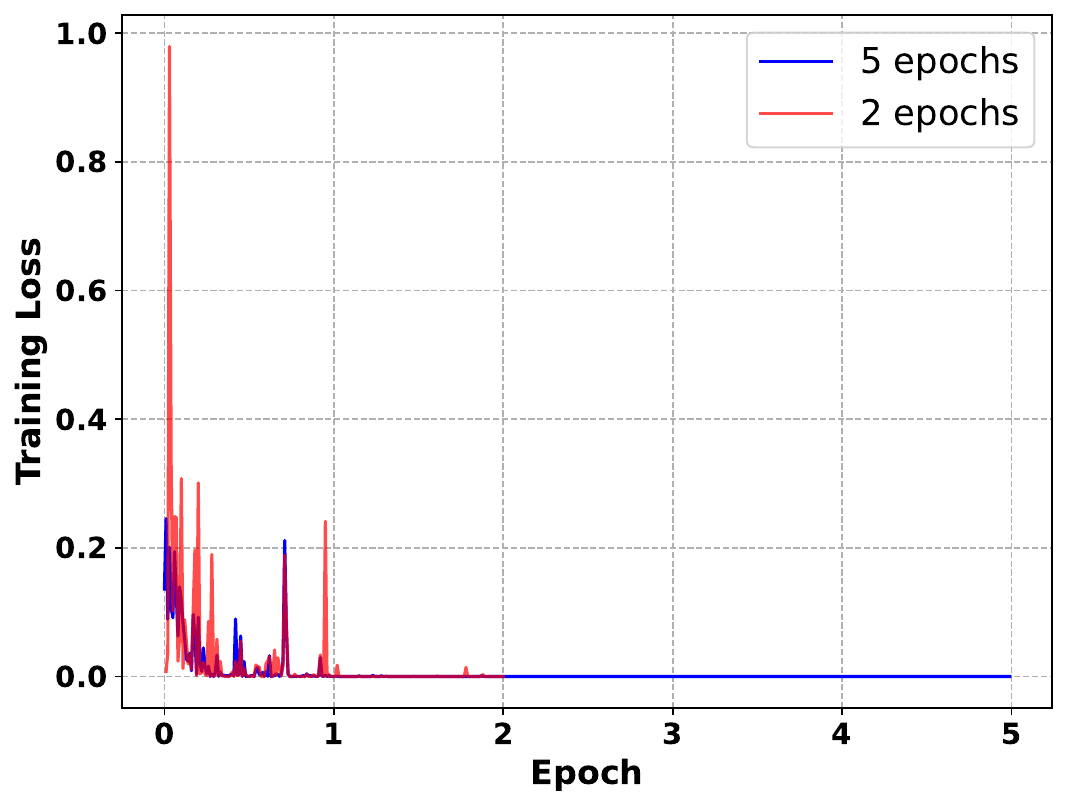}}

\vspace{0.25em}

\subfloat[Training loss for Llama 3.2 1B on multi-scale RAIT datasets (\textit{Local/Global RAIT}). The 5-epoch run (blue) achieves $\sim$0.15 loss, while 15 epochs (red) reduce loss below 0.05, indicating effective learning.\label{fig:LlamaRAG_loss}]{%
\includegraphics[width=0.40\textwidth]{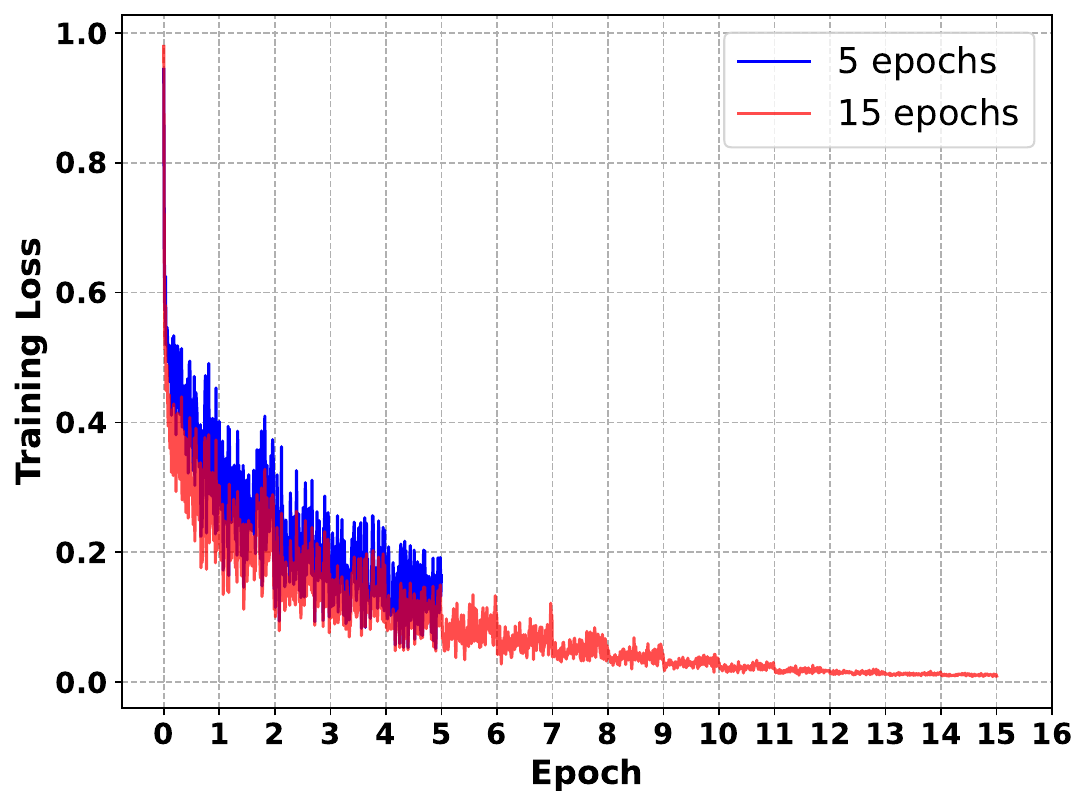}}
\hfill
\subfloat[SFT loss for SmolLM2-135M on \textit{Factual QA}, \textit{SynDIP}, and \textit{LogiCore} datasets. Loss decreases from $\sim$2.1 to $\sim$0.6 over 15 epochs, with higher variance than larger models.\label{fig:SmolLMQA_loss}]{%
\includegraphics[width=0.40\textwidth]{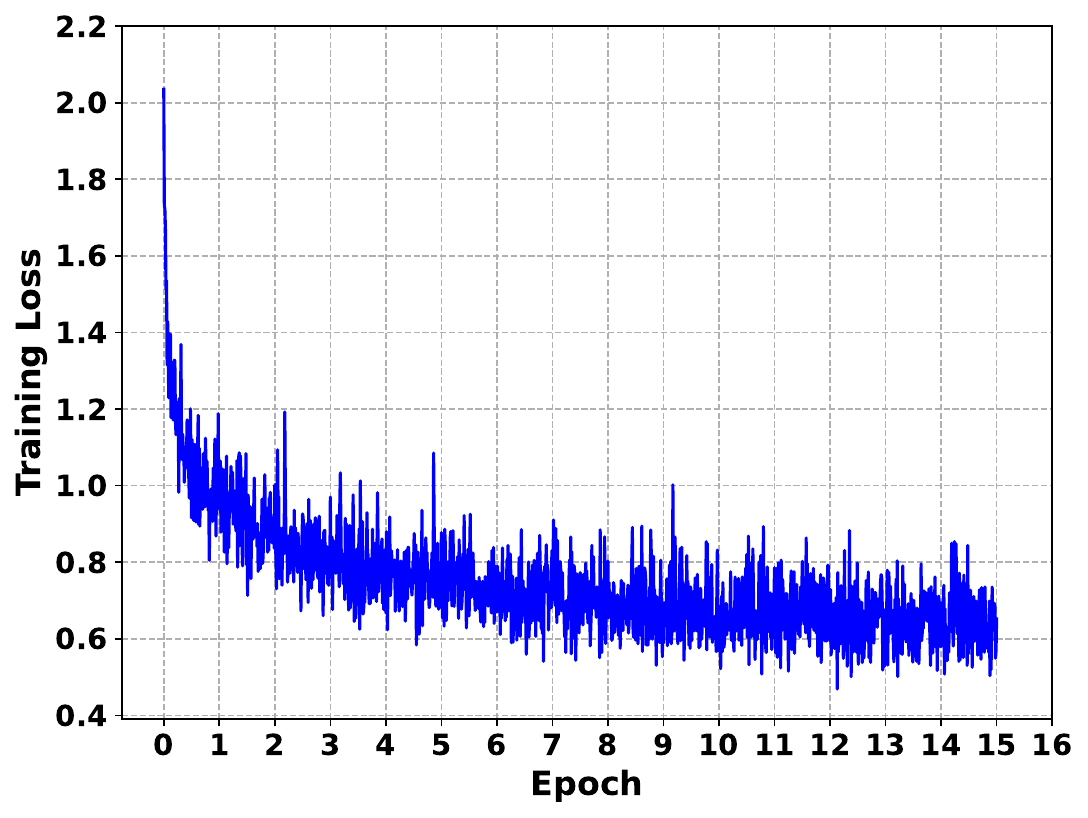}}

\vspace{0.25em}

\subfloat[DPO loss for SmolLM2-135M. Initial loss of 0.35 reaches near-zero within one epoch and remains stable through 5 epochs, demonstrating efficient preference learning.\label{fig:SmolLMDPO_loss}]{%
\includegraphics[width=0.40\textwidth]{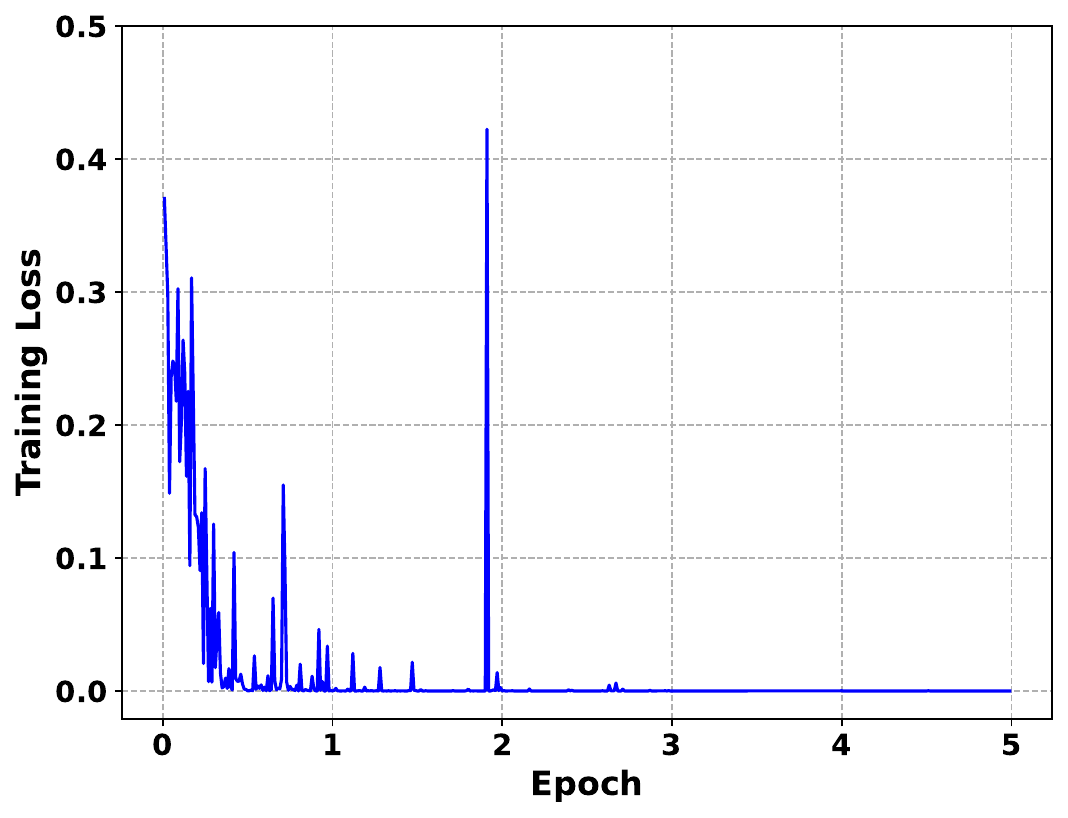}}
\hfill
\subfloat[Training loss for SmolLM2-135M on RAIT datasets (\textit{Local/Global RAIT}). Loss improves from $\sim$1.4 to 0.2 over 15 epochs despite higher noise, showing gradual learning.\label{fig:SmolLMRAG_loss}]{%
\includegraphics[width=0.40\textwidth]{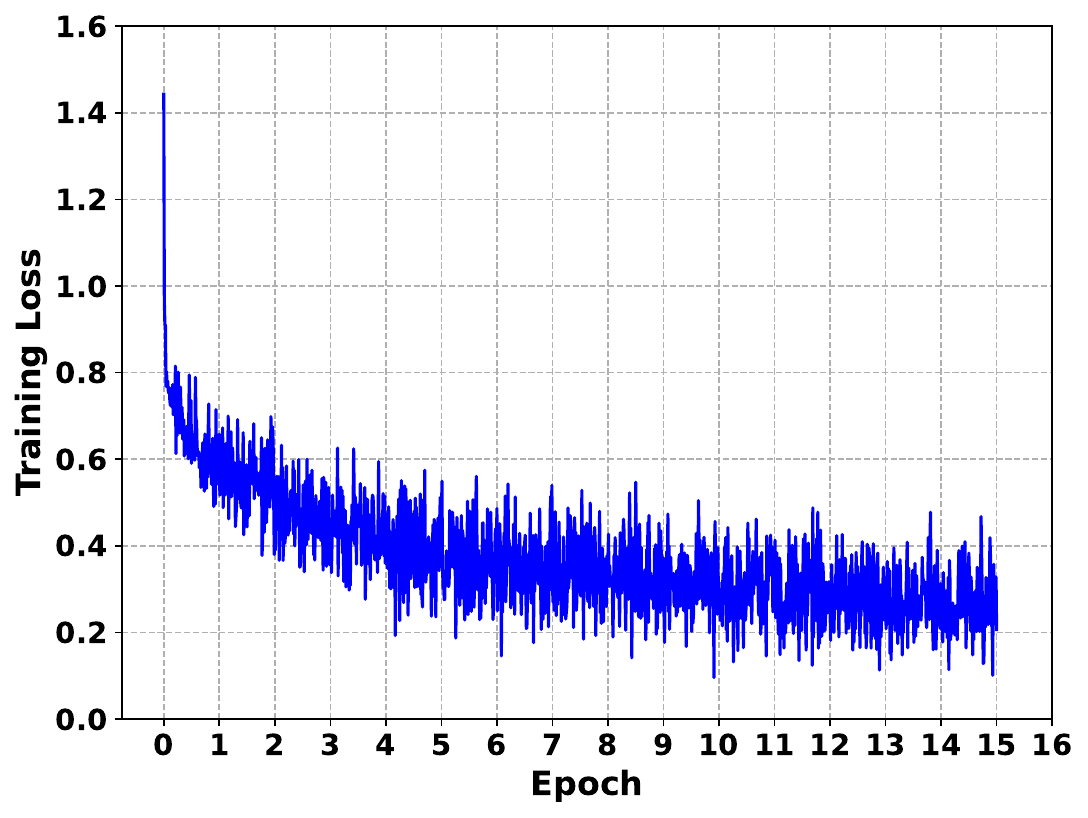}}

\caption{Training loss curves across different fine-tuning approaches and model sizes. Top row shows Llama 3.2 1B results for (a) supervised fine-tuning, (b) direct preference optimization, and (c) RAIT training. Bottom row presents corresponding results for SmolLM2-135M, demonstrating consistent learning patterns across model scales with expected variance differences.}
\label{fig:training_losses}
\vspace{-2mm}
\end{figure*}

\begin{figure*}[ht!]
\centering
\captionsetup[subfloat]{font=scriptsize,labelfont=bf}
% Row 1: Llama-3.2-1B Metrics
\resizebox{0.90\textwidth}{!}{%
    \subfloat[Computational requirements for \textit{Llama-3.2-1B}. \textit{RAIT} demanded the most wall-clock time (1463.4 min), followed by supervised \textit{QA} tuning (1315.2 min), with \textit{DPO} being the fastest (108.1 min).]{\includegraphics[width=50mm]{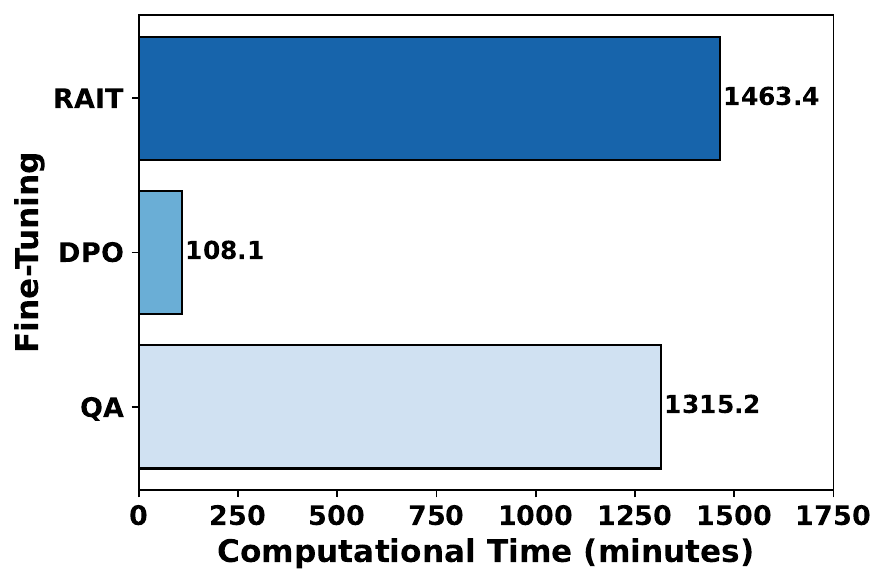}\label{fig:fttime_llama_time}}
    \hspace{0.05\textwidth}%
    \subfloat[Environmental impact for \textit{Llama-3.2-1B}. Supervised \textit{QA} tuning produced the highest CO\textsubscript{2} emissions (0.89 kg), followed by \textit{RAIT} (0.76 kg), with \textit{DPO} being the most efficient (0.06 kg).]{\includegraphics[width=50mm]{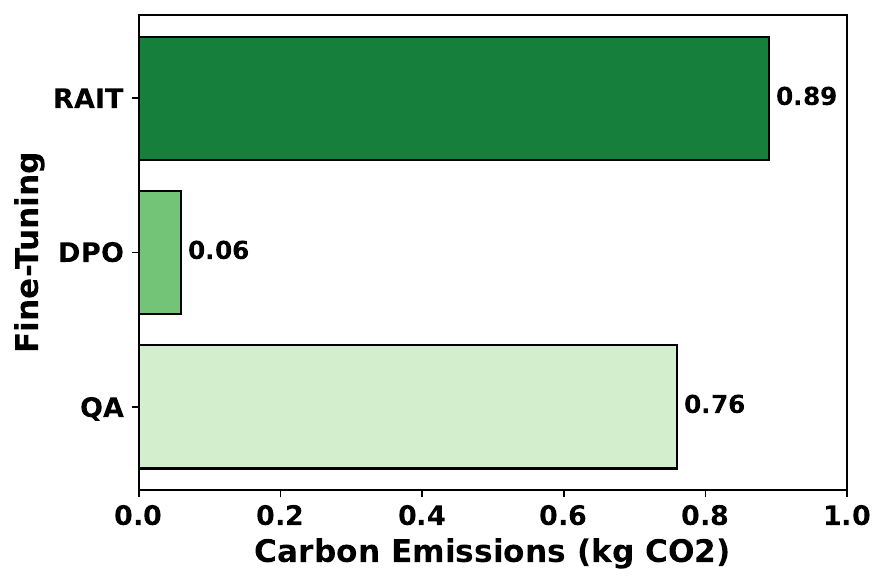}\label{fig:fttime_llama_ce}}
}
% Row 2: SmolLM2-135M Metrics
\resizebox{0.90\textwidth}{!}{%
    \subfloat[Computational requirements for \textit{SmolLM2-135M}. Supervised \textit{QA} tuning required the most time (640.3 min), followed by \textit{RAIT} (442.3 min), with \textit{DPO} being the fastest (41.5 min).]{\includegraphics[width=50mm]{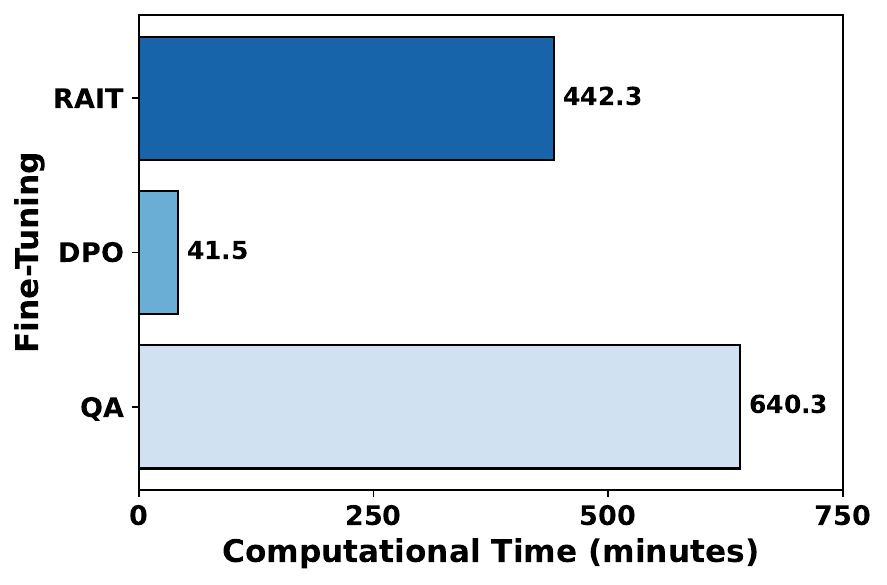}\label{fig:fttime_smol_time}}
    \hspace{0.05\textwidth}%
    \subfloat[Environmental impact for \textit{SmolLM2-135M}. \textit{RAIT} emitted the most CO\textsubscript{2} (0.41 kg), followed by supervised \textit{QA} tuning (0.26 kg), with \textit{DPO} being the most efficient (0.04 kg).]{\includegraphics[width=50mm]{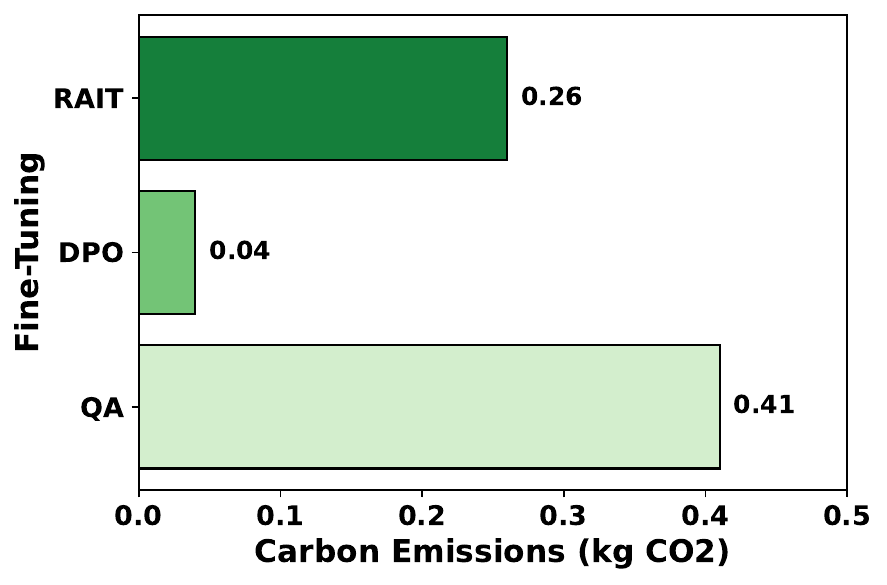}\label{fig:fttime_smol_ce}}
}
\caption{Comparison of computational efficiency and environmental impact for fine-tuning \textit{Llama-3.2-1B} (top) and \textit{SmolLM2-135M} (bottom) across three approaches: (1) supervised \textit{QA} tuning, (2) \textit{DPO}, and (3) \textit{RAIT}. Left panels (a,c) show wall-clock training time as a measure of computational requirements. Right panels (b,d) show the resulting CO\textsubscript{2} emissions as a measure of environmental impact. \textit{DPO} was consistently the most efficient method in both dimensions.}
\label{fig:fttime}
\end{figure*}

We present a comparative evaluation of Llama-3.2-1B and SmolLM-135M across successive fine-tuning stages on the respective test splits of SFT datasets (\textit{Factual QA}, \textit{SynDIP}, and \textit{LogiCore}), preference alignment fine-tuning datasets (\textit{DPO}), and retrieval-augmented fine-tuning (RAFT) datasets (\textit{Local/Global RAIT}). We report their performance against a comprehensive set of evaluation metrics. As shown in Figures~\ref{fig:LlamaQA_metrics}–\ref{fig:SmolLMRAG_metrics}, the evaluation was conducted using both token-level n-gram overlap metrics (BLEU, ROUGE-1/2/L, METEOR, SacreBLEU) and embedding-based semantic similarity metrics (BERTScore and Sentence-BERT cosine similarity), with all scores normalized to the [0,1] interval. We report the Llama-3.2-1B performance on the test splits of the \textit{Factual QA}, \textit{SynDIP}, and \textit{LogiCore} datasets (see Figure~\ref{fig:LlamaQA_metrics}). The language model demonstrates strong semantic alignment, evidenced by high BERTScore and sentence similarity, despite lower performance on n-gram metrics, indicating a preference for paraphrastic generation over lexical overlap. In contrast, SmolLM-135M performance on the same test splits (see Figure~\ref{fig:SmolLMQA_metrics}) exhibits relatively higher n-gram scores and sentence similarity while achieving moderate BERTScore, suggesting a tendency toward surface-level fidelity. When evaluated on the \textit{DPO} dataset test split, Llama-3.2-1B (refer to Figure~\ref{fig:LlamaDPO_metrics}) achieves high semantic similarity scores, whereas SmolLM-135M (Figure~\ref{fig:SmolLMDPO_metrics}) demonstrates balanced improvements across both lexical and semantic metrics, reflecting effective alignment via instruction tuning. For the retrieval-augmented tasks, Llama-3.2-1B performance on the test splits of the \textit{Local} and \textit{Global RAIT} datasets (refer to Figure~\ref{fig:LlamaRAG_metrics}) continues to show dominant semantic scores relative to n-gram metrics. SmolLM-135M (Figure~\ref{fig:SmolLMRAG_metrics}) exhibits comparatively lower scores across most metrics, with sentence similarity remaining the strongest, suggesting diminished generalization ability under retrieval-augmented long-context settings. These plots(see Figures \ref{fig:emp}a-e) provide phase-by-phase performance insights, highlighting how successive fine-tuning regimes induce distinct response behaviors across models in terms of semantic coherence, lexical fidelity, and alignment with training objectives. Additionally, we conduct a systematic evaluation of how fine-tuning (FT) and Graph RAG affect quantitative performance across six language model variants, comprising two architectures at different scales: the larger Llama-3.2-1B and the more compact SmolLM2-135M. Each variant represents a distinct configuration (where W/ = With and W/o = Without): (a) Llama-3.2-1B W/FT W/Graph RAG, (b) Llama-3.2-1B W/FT W/o Graph RAG, (c) Llama-3.2-1B W/o FT W/o Graph RAG, (d) Llama-3.2-1B W/o FT W/Graph RAG, (e) SmolLM2-135M W/FT W/Graph RAG, and (f) SmolLM2-135M W/FT W/o Graph RAG. We rigorously evaluate these variants using the NVIDIA Nemotron-4-340B reward model across five key quantitative dimensions: helpfulness (practical utility), correctness (factual accuracy), coherence (logical flow), complexity (depth of content), and verbosity (response length), with detailed results presented in Figures \ref{fig:sAF}a-e on the 1.5K QA-pair out-of-distribution benchmark. The evaluation reveals several key findings regarding model scale and methodological impact. Among Llama-3.2-1B variants, the FT+Graph RAG configuration (variant a) demonstrates superior performance, achieving peak scores in correctness and complexity by combining fine-tuned capabilities with retrieved knowledge, albeit with increased verbosity from incorporating supplementary knowledge graph content. The FT-only variant (b) maintains strong coherence and helpfulness but shows limitations in knowledge-intensive tasks without retrieval support. Notably, the Graph RAG-enabled Llama variant without FT (d) outperforms the baseline (c) in correctness, proving retrieval augmentation can partially compensate for missing task-specific tuning. The complete absence of both methods (variant c) yields the weakest performance, revealing the limitations of relying solely on pretrained knowledge. For SmolLM2-135M, Graph RAG improves correctness (variant e vs. f), but both configurations underperform relative to comparable Llama-3.2-1B variants across all metrics, particularly in coherence and complexity, highlighting scale's importance for effectively utilizing both techniques. Results demonstrate FT substantially enhances overall response quality by aligning models with domain requirements, while Graph RAG provides complementary factual accuracy benefits. This synergy proves especially valuable in specialized domains like chemical process synthesis, where both task adaptation and external knowledge integration are crucial. The optimal configuration—Llama-3.2-1B with both FT and Graph RAG—achieves balanced performance across all dimensions, successfully integrating structured retrieval with fine-tuned understanding while maintaining reasonable verbosity. These findings carry significant implications for deploying language models in technical domains requiring both factual precision and contextual understanding. Figures \ref{fig:LlamaQA_loss}--\ref{fig:SmolLMRAG_loss} present the training loss curves for \textit{Llama-3.2-1B} and \textit{SmolLM2-135M} models fine-tuned using QLoRA on synthetic datasets from the \textit{ChemAtlas} corpus. The \textit{Llama-3.2-1B} model shows strong convergence during supervised fine-tuning (SFT) on the \textit{Factual QA}, \textit{SynDIP}, and \textit{LogiCore} datasets (Figure \ref{fig:LlamaQA_loss}), with loss decreasing from $\sim$1.5 to 0.35 within 5 epochs (blue curve) and further improving to $\sim$0.1 after 15 epochs (red curve). Direct Preference Optimization (DPO) training (Figure \ref{fig:LlamaDPO_loss}) achieves near-zero loss within the first epoch and maintains stable performance throughout both 2-epoch and 5-epoch runs. For Retrieval-Augmented Instruction Tuning (RAIT) (Figure \ref{fig:LlamaRAG_loss}), the loss consistently decreases from $\sim$0.15 to below 0.05 over 15 epochs. In contrast, the smaller \textit{SmolLM2-135M} exhibits slower convergence with higher variance across all tasks. During SFT (Figure \ref{fig:SmolLMQA_loss}), its loss declines from $\sim$2.2 to 0.6 but shows significant training instability. While DPO fine-tuning (Figure \ref{fig:SmolLMDPO_loss}) also achieves near-zero loss rapidly, RAIT training (Figure \ref{fig:SmolLMRAG_loss}) demonstrates more gradual improvement ($\sim$1.5 to 0.2) with persistent fluctuations. These results highlight two key observations: (1) \textit{Llama-3.2-1B} benefits substantially from extended training durations, and (2) \textit{SmolLM2-135M} shows stronger dependence on fine-tuning methodology, with DPO yielding more stable convergence than SFT. The computational cost analysis (Figures \ref{fig:fttime_llama_time}--\ref{fig:fttime_smol_ce}) reveals DPO requires the fewest GPU hours, while SFT and RAIT costs vary with dataset complexity.

\subsubsection{Evaluation on a Generalization Benchmark}
\begin{figure}[ht!]
\centering
\includegraphics[width=82.5mm]{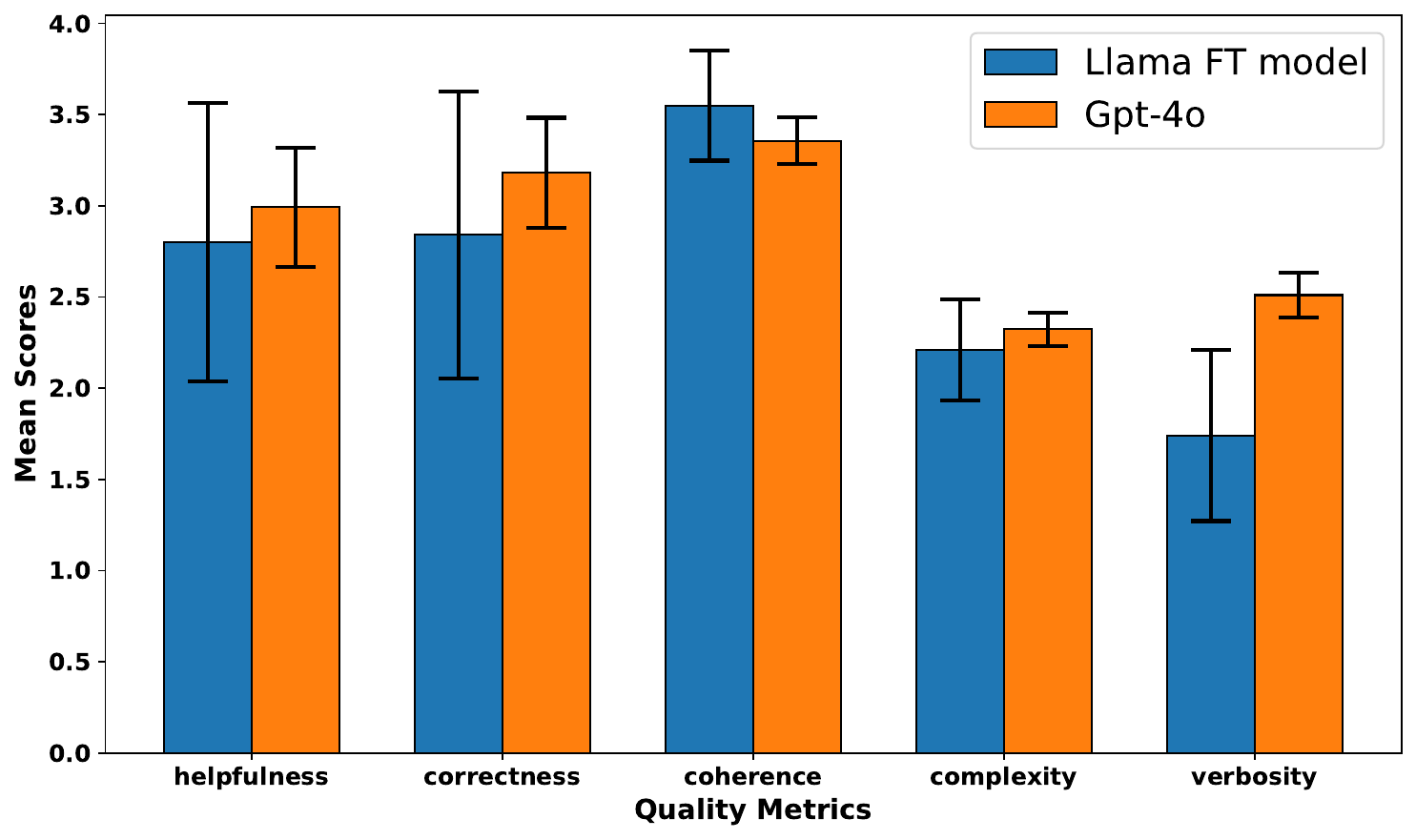}
\caption{Performance comparison between the fine-tuned Llama-3.2-1B model and GPT-4o on a held-out 1.5K QA-pair generalization benchmark, evaluated using the Nvidia/Nemotron-4-340B reward model. GPT-4o establishes a strong baseline, outperforming Llama-3.2-1B in most metrics (helpfulness, correctness, complexity). However, Llama-3.2-1B achieves comparable coherence and significantly lower verbosity. Larger error bars indicate higher variance in Llama-3.2-1B's responses.}
\label{fig:lvg}
\vspace{-3mm}
\end{figure}

We conduct a comparative evaluation of the fine-tuned Llama-3 1B model (\textit{Llama FT}) against GPT-4o using a held-out 1.5K QA-pair generalization benchmark dataset, as shown in Figure~\ref{fig:lvg}. Performance is assessed across five core metrics---helpfulness, correctness, coherence, complexity, and verbosity---each scored on a 0--4 scale using the \textit{Nvidia/Nemotron-4-340B} reward model. This out-of-distribution (OOD) benchmark is entirely disjoint from the synthetic datasets used during model development---including both training and evaluation phases---which comprise \textit{Factual QA}, \textit{SynDIP}, \textit{LogiCore}, and \textit{Local/Global RAIT} and \textit{DPO}.

%%%%%%%%%%%%%%%%%%%%%%%%%%%%%%%%% Training Data Metrics %%%%%%%%%%%%%%%%%%%%%%%

\clearpage
\newpage

%%%%%%%%%%%%%%%%%%%%%%%%%%%%%%%%%%%% Test Data Reward Model Scores %%%%%%%%%%%%%%%%%%%%%%%

\clearpage
\newpage

\begin{figure}[ht!]
\centering
\includegraphics[width=82.5mm]{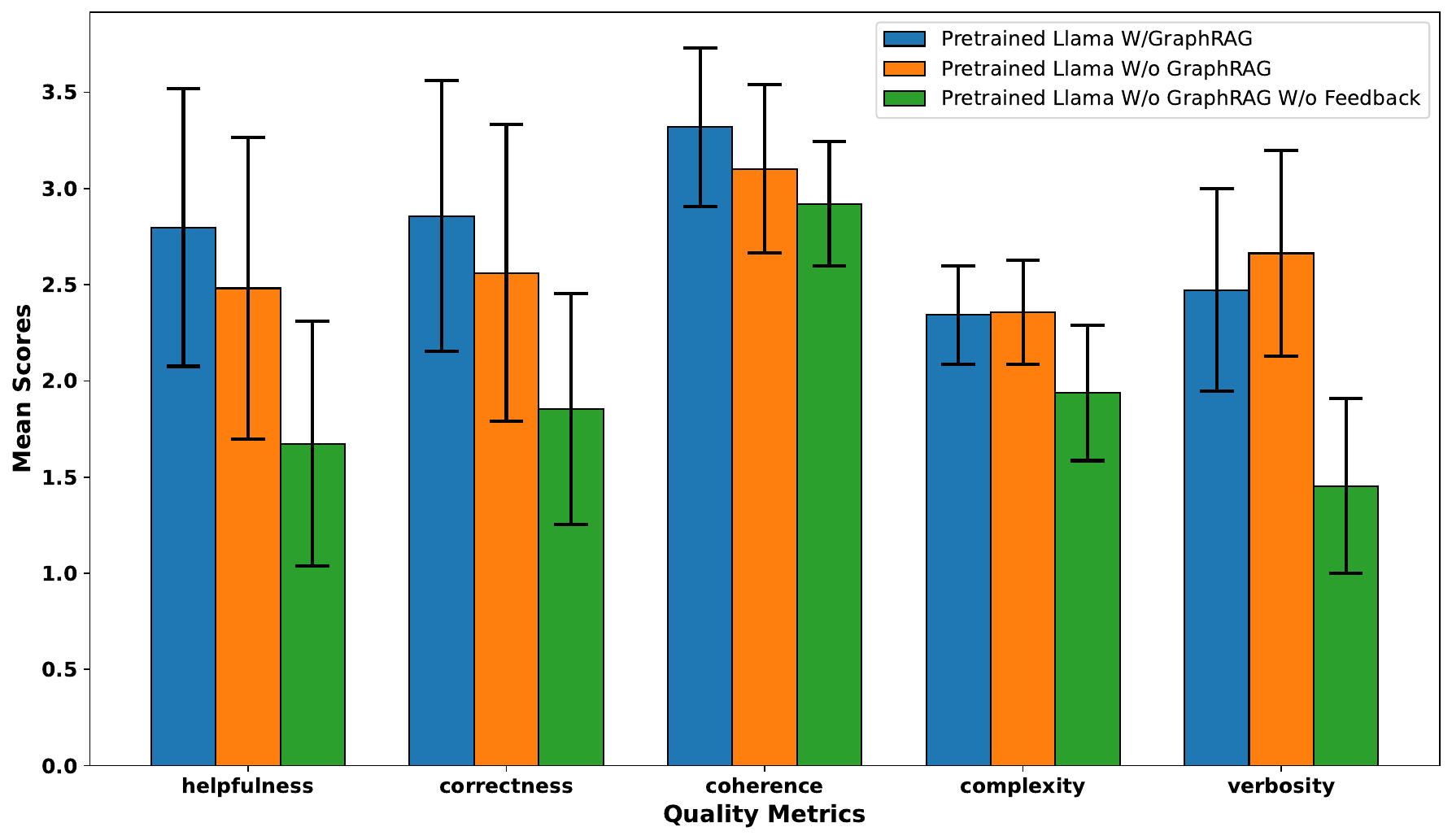}
\vspace{-2mm}
\caption{Zero-shot performance of the pretrained Llama-3.2-1B model across three configurations: (1) GraphRAG + feedback, (2) GraphRAG only, and (3) no enhancements. Evaluated on the same 1.5K QA benchmark with the \textit{Nvidia/Nemotron-4-340B} reward model (0--4 scale), the combined GraphRAG+feedback variant achieves the highest scores, particularly in helpfulness and correctness. Performance degrades progressively when either component is removed, demonstrating their synergistic role in zero-shot generalization.}
\label{fig:pmp}
\vspace{-2mm}
\end{figure}

Each dataset features predefined training, validation, and test splits. As illustrated in Figure~\ref{fig:lvg}, GPT-4o consistently achieves high scores across all metrics, establishing a strong performance baseline. The fine-tuned Llama-3 1B model demonstrates competitive results: it nearly matches GPT-4o in coherence, trails slightly in helpfulness and correctness, and produces significantly more concise responses, as reflected by lower verbosity scores. However, the larger error bars for Llama-3 1B suggest greater variability in performance across the generalization dataset. These results indicate that despite its smaller size, Llama-3 1B rivals GPT-4o in key quality dimensions while offering practical advantages in response brevity and computational efficiency. We further evaluate the zero-shot performance of the pretrained Llama-3 1B model, augmented with GraphRAG and feedback mechanisms, without any additional fine-tuning on synthetic datasets. As shown in Figure~\ref{fig:pmp}, we test three configurations on the same 1.5K QA-pair generalization benchmark dataset: (1) Llama-3 1B with both GraphRAG and feedback, (2) Llama-3 1B with GraphRAG but without feedback, and (3) Llama-3 1B without either GraphRAG or feedback. All configurations are evaluated using the \textit{Nvidia/Nemotron-4-340B} reward model across the same five metrics. The results demonstrate that the configuration incorporating both GraphRAG and feedback consistently outperforms the other two variants, with especially notable gains in helpfulness and correctness---approaching a reward score of 3.0. These findings underscore the synergistic benefit of retrieval and critique mechanisms, even in the absence of task-specific fine-tuning, for improving zero-shot generalization. While coherence remains largely similar across all configurations, the improvements in helpfulness and correctness are more pronounced. Overall, GraphRAG substantially enhances language model performance by enabling more accurate and useful responses, while feedback mechanisms independently contribute meaningful quality improvements.

\vspace{2mm}
\subsubsection{Ablation Study: Head-to-Head Multi-Metric Evaluation of Framework Variants.}
\begin{figure*}[ht!]
\centering
% Row 1
\begin{minipage}[t]{0.465\textwidth}
\centering
\includegraphics[width=55mm]{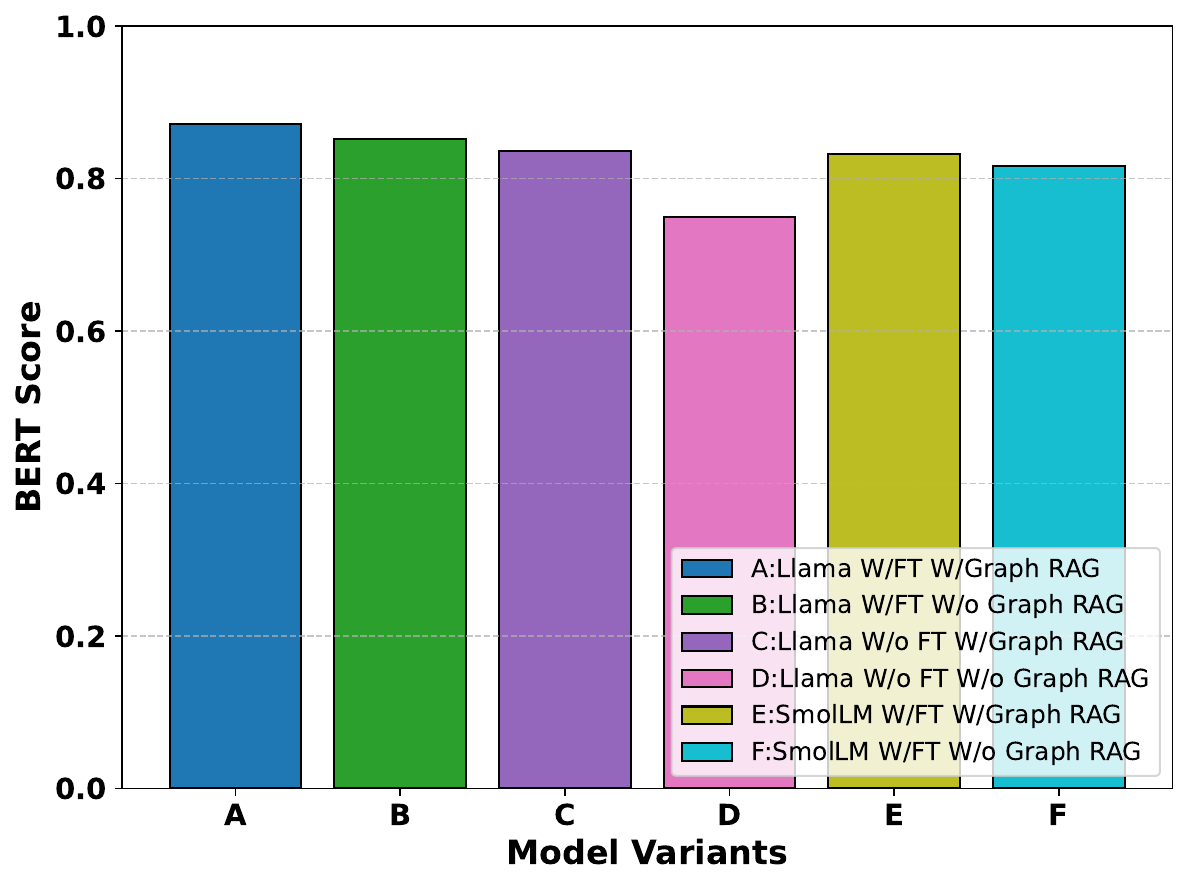}
\par \small \textbf{(a)} BERTScore evaluation across six framework variants on the 1.5K QA-pair generalization benchmark. Variant A (Llama-3.2 1B w/FT w/GraphRAG) achieves the highest semantic similarity ($\sim$0.9), demonstrating the combined benefit of fine-tuning and GraphRAG.
\end{minipage}
\hspace{0.05\textwidth}
\begin{minipage}[t]{0.465\textwidth}
\centering
\includegraphics[width=55mm]{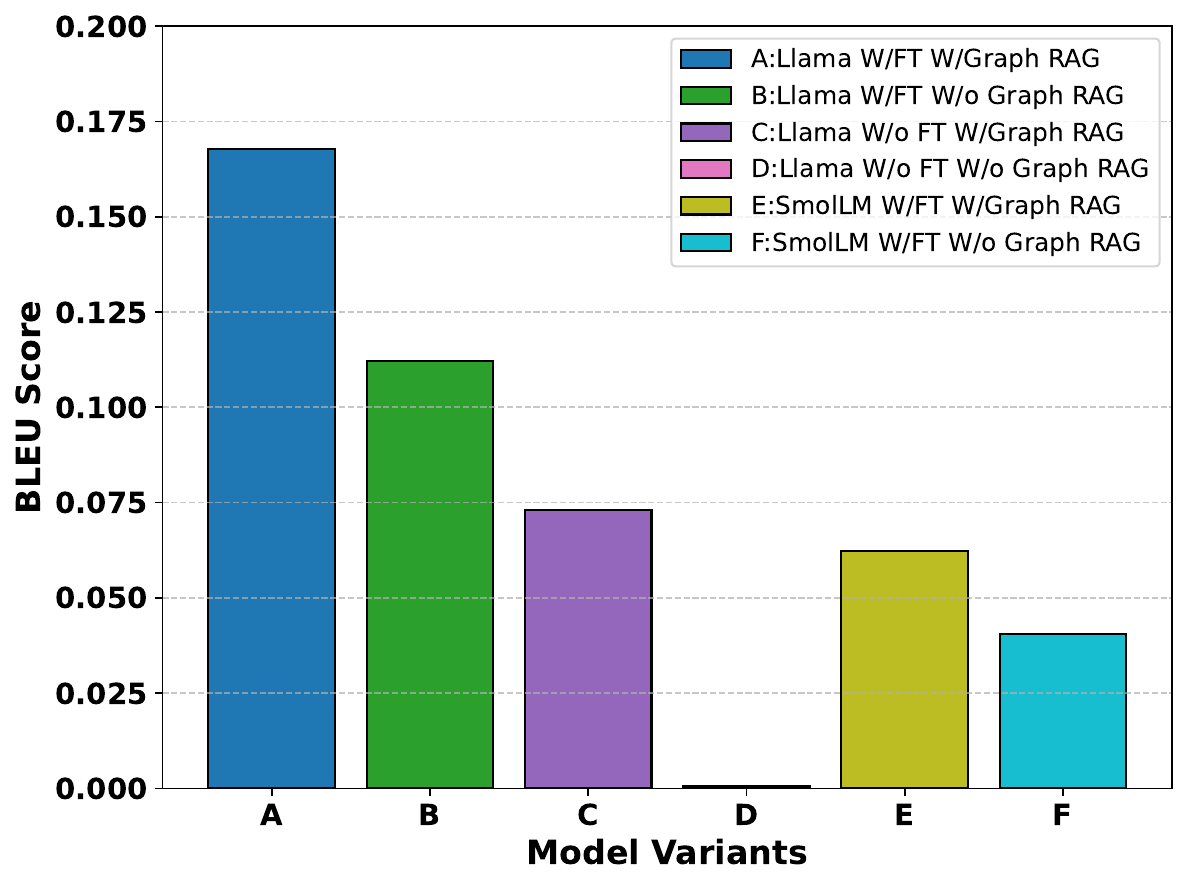}
\par \small \textbf{(b)} BLEU score analysis showing n-gram precision improvements. Variant A (Llama-3.2 1B w/FT w/GraphRAG) leads with a $\sim$0.17 score, outperforming other configurations by significant margins.
\end{minipage}

\vspace{2mm}

% Row 2
\begin{minipage}[t]{0.465\textwidth}
\centering
\includegraphics[width=55mm]{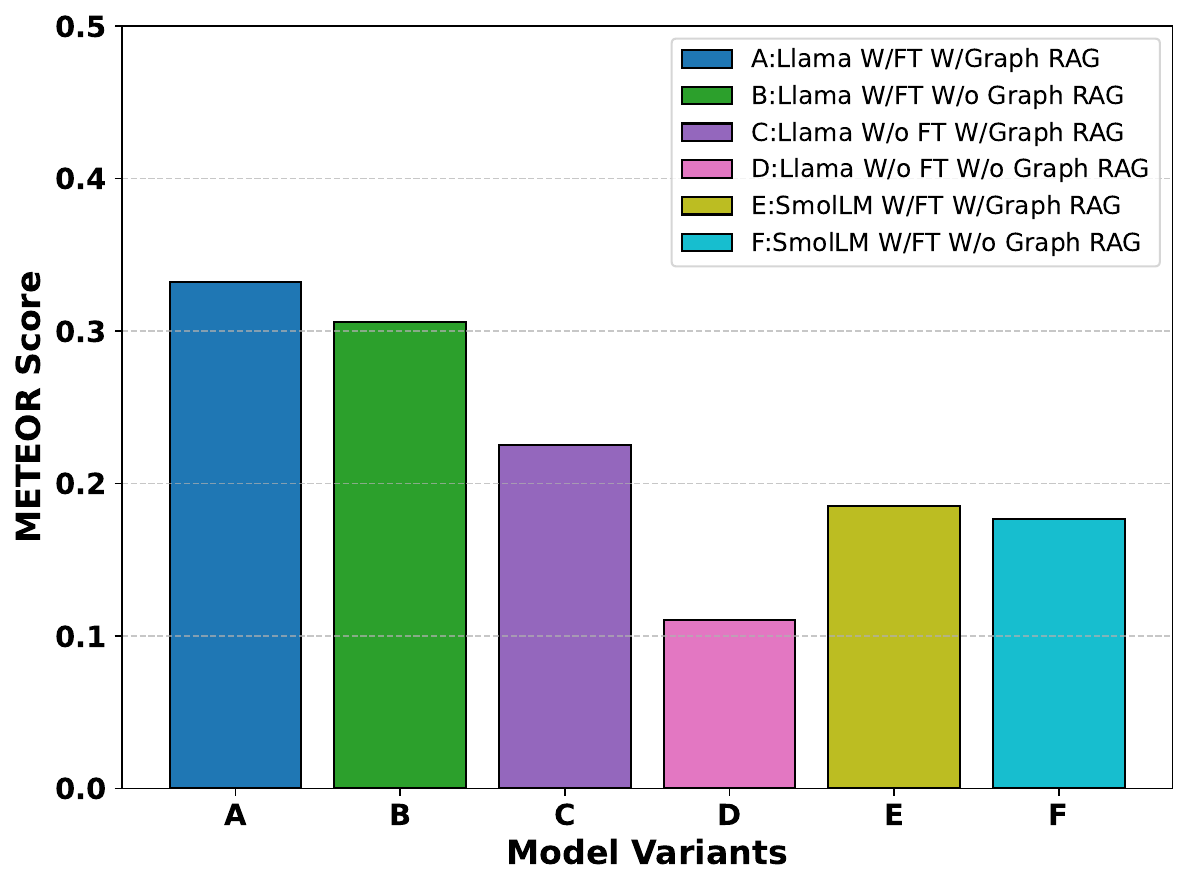}
\par \small \textbf{(c)} METEOR scores assessing lexical and semantic alignment. Fine-tuned Llama-3.2 1B variants (A and B) score $>$0.3, with GraphRAG providing additional gains (A vs. B).
\end{minipage}
\hspace{0.05\textwidth}
\begin{minipage}[t]{0.465\textwidth}
\centering
\includegraphics[width=55mm]{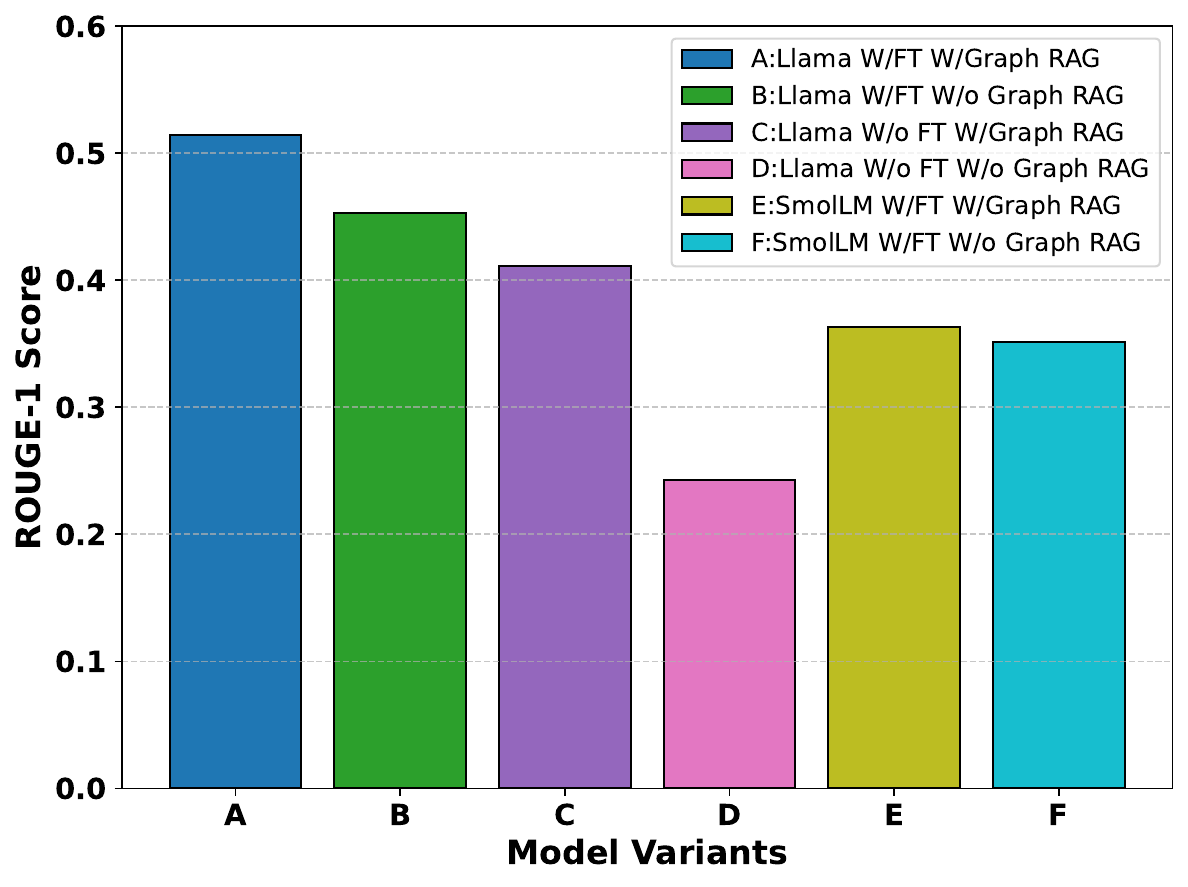}
\par  \small \textbf{(d)} ROUGE-1 evaluation of unigram overlap. Variant A (Llama-3.2 1B w/FT w/GraphRAG) achieves a $>$0.5 score, showing that both fine-tuning and GraphRAG independently improve performance.
\end{minipage}

\vspace{2mm}

% Row 3
\begin{minipage}[t]{0.465\textwidth}
\centering
\includegraphics[width=55mm]{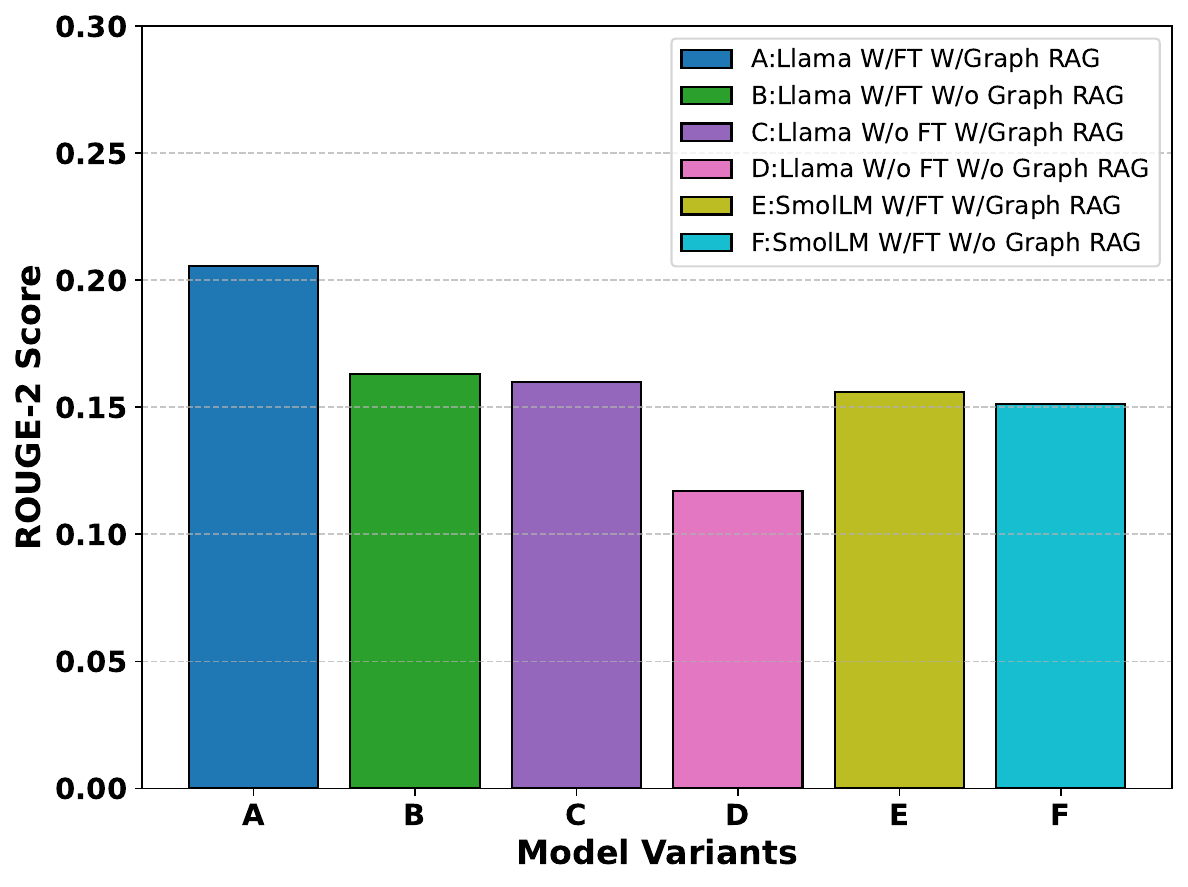}
\par \small \textbf{(e)} ROUGE-2 analysis of bigram overlap. Variant A maintains the lead ($>$0.20), with the fine-tuned SmolLM2-135M also benefiting from GraphRAG (E vs. F).
\end{minipage}
\hspace{0.05\textwidth}
\begin{minipage}[t]{0.465\textwidth}
\centering
\includegraphics[width=55mm]{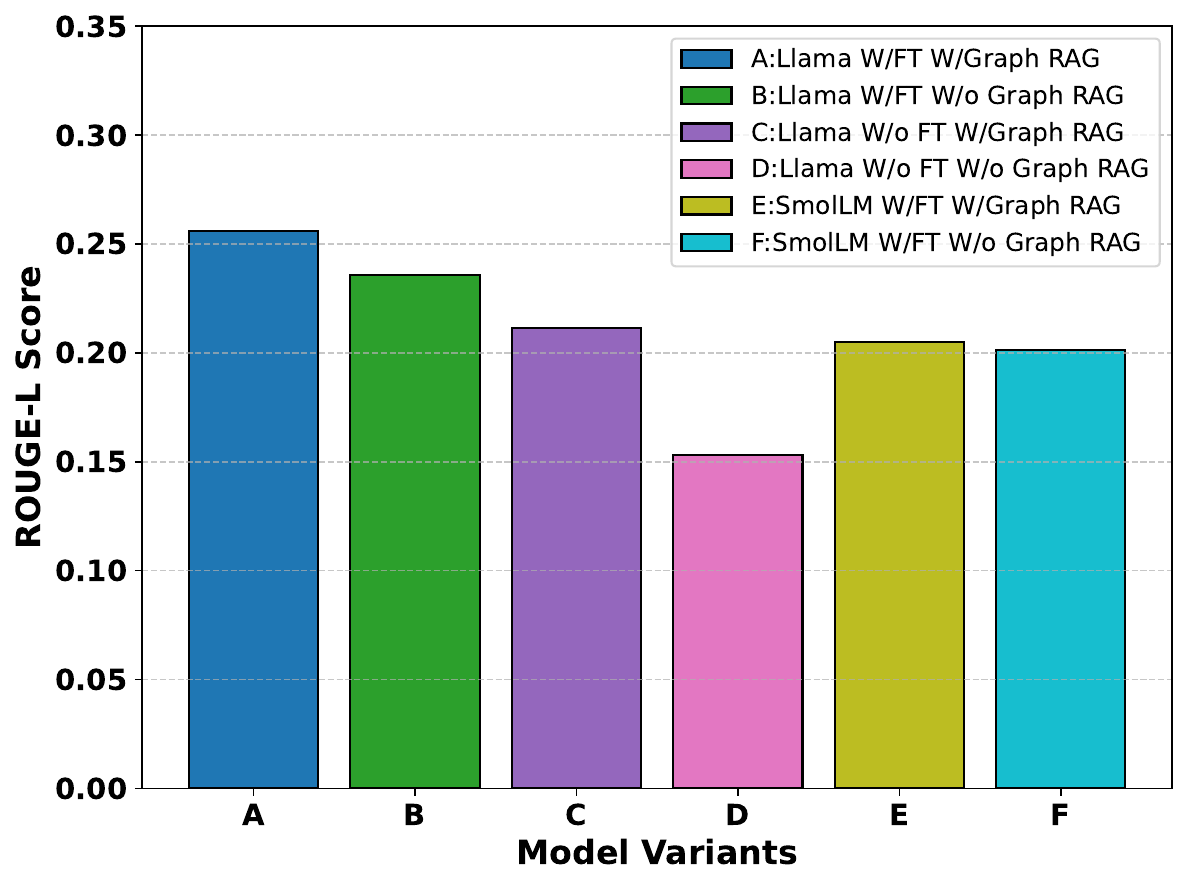}
\par \small \textbf{(f)} ROUGE-L assessment of longest common subsequence. Variant A shows the best performance ($\sim$0.26), with fine-tuning driving most improvement and GraphRAG providing complementary benefits.
\end{minipage}

\vspace{2mm}
\caption{Comprehensive evaluation of six framework variants (A--F) using standard NLP metrics on the 1.5K QA-pair generalization benchmark. Results demonstrate that Variant A consistently achieves the highest scores, with fine-tuning and GraphRAG offering complementary improvements. Configuration details: A=Llama-3.2 1B w/FT w/GraphRAG, B=Llama-3.2 1B w/FT w/o GraphRAG, C=Llama-3.2 1B w/o FT w/GraphRAG, D=baseline (Llama-3.2 1B w/o FT w/o GraphRAG), E=SmolLM2-135M w/FT w/GraphRAG, F=SmolLM2-135M w/FT w/o GraphRAG.}
\label{fig:em1}
\end{figure*}

\vspace{-2mm}
We evaluate six framework variants to analyze the individual and combined effects of fine-tuning (FT) and GraphRAG. Variant (A) represents the Llama-3.2 1B model w/ both fine-tuning and GraphRAG enabled. Variant (B) uses the fine-tuned Llama-3.2 1B model but excludes GraphRAG (w/o GraphRAG). Variant (C) employs the pre-trained Llama-3.2 1B model w/o fine-tuning but w/ GraphRAG, while variant (D) serves as the baseline, featuring the pre-trained Llama-3.2 1B model w/o fine-tuning and w/o GraphRAG. For the smaller model, variant (E) applies the fine-tuned SmolLM2-135M model w/ GraphRAG, and variant (F) represents the fine-tuned SmolLM2-135M model w/o GraphRAG. As shown in Figure~\ref{fig:em1}, across all metrics—BERT (semantic similarity), BLEU (n-gram precision), METEOR (lexical and semantic alignment), and ROUGE (unigram, bigram, and longest-sequence overlap)—the results demonstrate that Variant A (Llama-3.2 1B w/ both fine-tuning and GraphRAG) consistently achieves the highest performance. Both fine-tuning and GraphRAG independently improve performance beyond the baseline, while their combination achieves peak performance. Specifically, Figure~\ref{fig:em1}(a) presents BERT scores, which assess semantic similarity across the six framework variants. The results highlight the benefits of fine-tuning and GraphRAG: Variant A (Llama-3.2 1B w/FT w/GraphRAG) achieves the highest score (~0.9), indicating superior semantic alignment. Comparisons among the Llama-3.2 1B variants (A–D) show that fine-tuning and GraphRAG each independently improve performance over the baseline (D). A similar positive effect occurs for the fine-tuned SmolLM2-135M model, where GraphRAG enhances performance (E vs. F). These findings confirm that both methods improve semantic quality, with the optimized Llama-3.2 1B model (Variant A) delivering the best performance. Figure~\ref{fig:em1}(b) displays BLEU scores, measuring n-gram precision across the six framework variants. Variant A (Llama-3.2 1B w/FT w/GraphRAG) achieves the highest score (~0.17), outperforming other variants by a wide margin. Analysis of Llama-3.2 1B variants (A–D) shows that fine-tuning alone significantly improves precision over the pre-trained baseline (D), while the addition of GraphRAG further boosts performance (A vs. B). A comparable but smaller improvement occurs for the fine-tuned SmolLM2-135M model with GraphRAG (E vs. F). These results indicate that both fine-tuning and GraphRAG independently enhance precision, with their combined implementation in Variant A yielding optimal results. Figure~\ref{fig:em1}(c) presents METEOR scores evaluating lexical and semantic alignment across the six variants. Fine-tuned Llama-3.2 1B models (Variants A and B, both >0.3) significantly outperform non-fine-tuned counterparts (Variants C and D). GraphRAG provides additional gains for both Llama-3.2 1B (A vs. B, C vs. D) and fine-tuned SmolLM2-135M (E vs. F), confirming fine-tuning's primary role in score enhancement with GraphRAG offering secondary benefits. Notably, the top Llama configurations (A and B) consistently surpass all SmolLM2-135M variants. Figure~\ref{fig:em1}(d) shows ROUGE-1 unigram overlap results, with Variant A (Llama-3.2 1B w/FT w/GraphRAG) achieving the highest score (>0.5). Both fine-tuning and GraphRAG independently improve performance over the pre-trained baseline (Variant D), while GraphRAG also benefits the fine-tuned SmolLM2-135M (E vs. F), demonstrating their synergistic effect on unigram overlap optimization with Variant A delivering peak performance.

%%%%%%%%%%%%%%%%%%%%%%%%%%%%%%%%%%%%%% Training Loss%%%%%%%%%%%%%%%%%%%%%%%%

\clearpage
\newpage

%%%%%%%%%%%%%%%%%%%%%%Fine tuning time and emissions %%%%%%%%%%%%%%%%%%%%%%%%%%

%%%%%%%%%%%%%%%%%%%% Llama vs GPT-4o %%%%%%%%%%%%%%%%%%%%%%%%
\vspace{-3mm}

%%%%%%%%%%%%%%%%%%%% Pretrained Model Performance %%%%%%%%%%%%%%%%%%%%%%%
\vspace{-3mm}

\begin{figure}[ht!]
\vspace{-1mm}
\centering
\captionsetup[subfloat]{font=scriptsize,labelfont=bf}

\subfloat[SacreBLEU evaluation showing n-gram precision. Variant A ($\approx$0.168) demonstrates optimal performance, with fine-tuning providing major gains and GraphRAG further enhancing results.]{
    \includegraphics[width=0.45\textwidth]{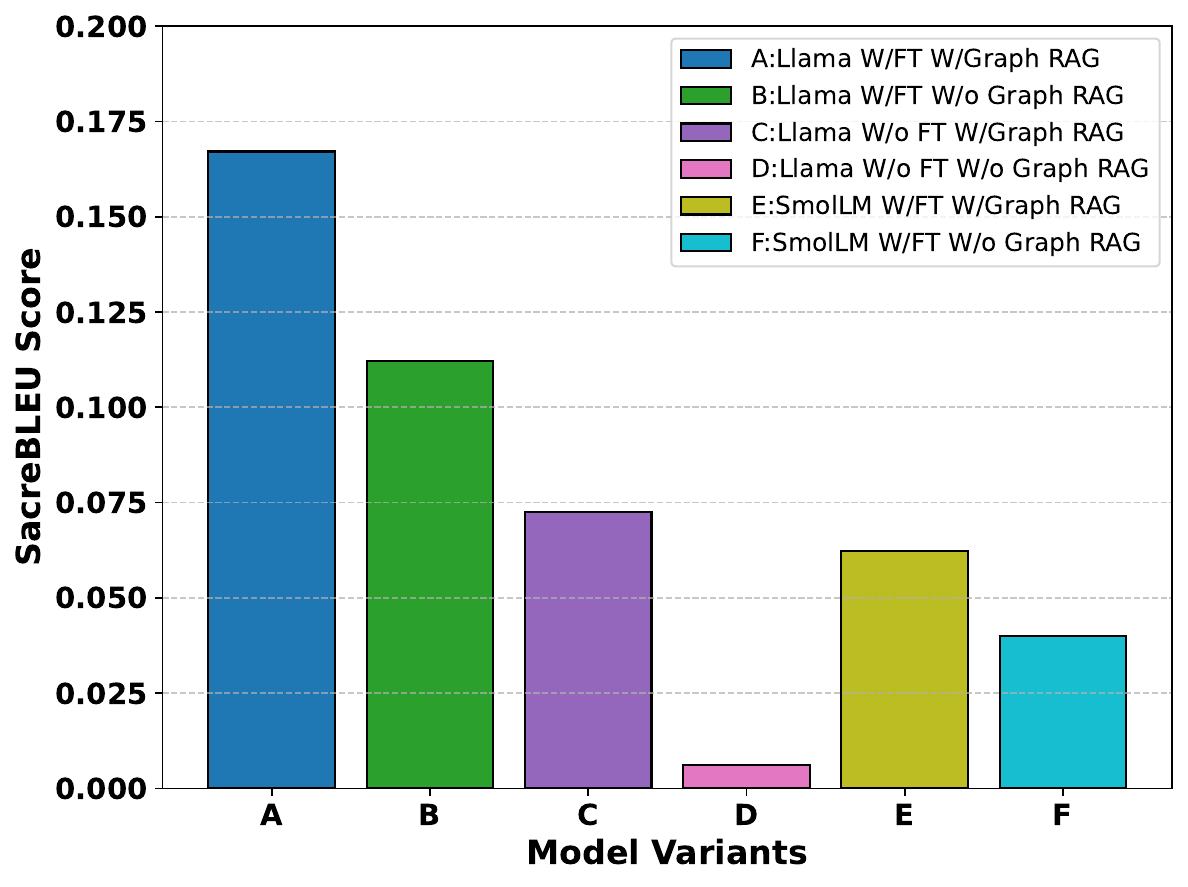}
    \label{fig:em2a}
}
\vspace{1mm} % adds spacing between the two figures
\subfloat[Similarity score analysis reveals high semantic alignment ($>$0.85) across variants. Variant A peaks at $\approx$0.93, slightly exceeding Variant B and significantly outperforming smaller model configurations.]{
    \includegraphics[width=0.45\textwidth]{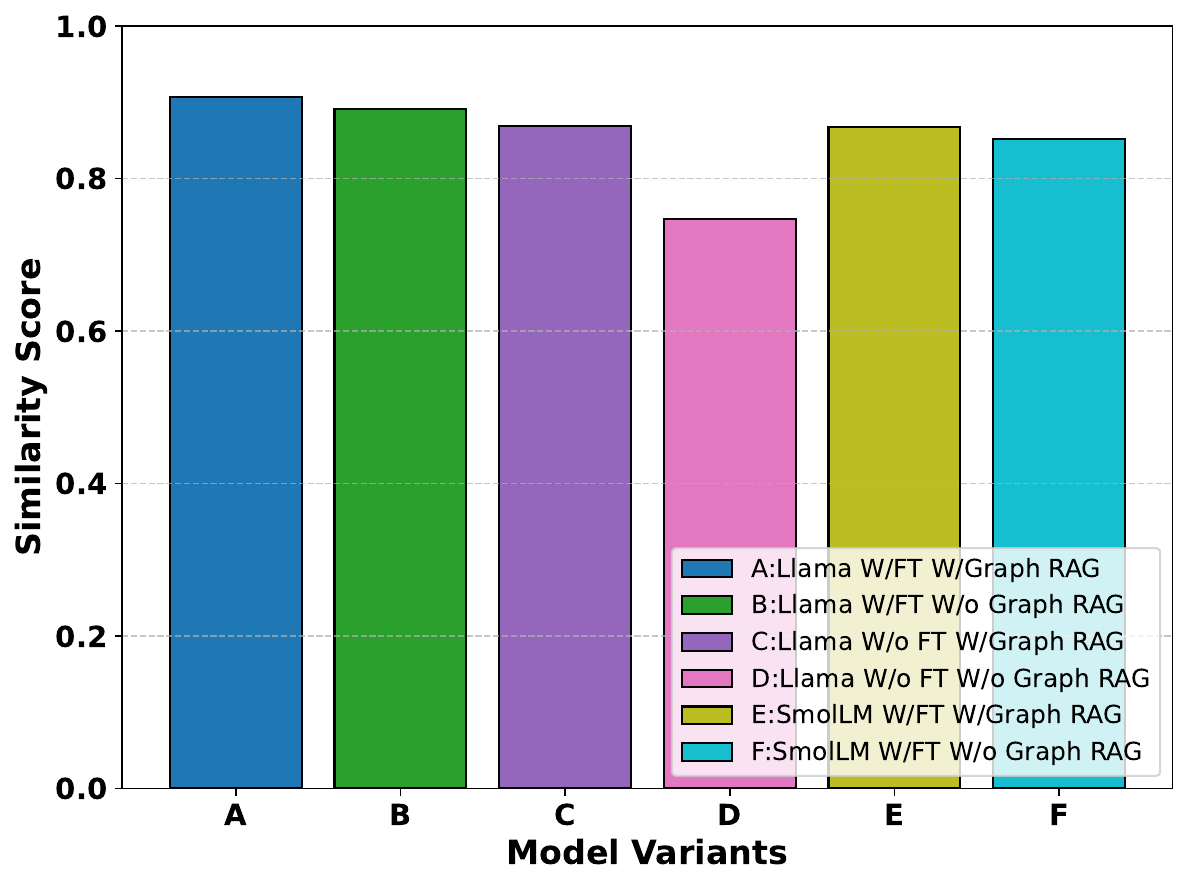}
    \label{fig:em2b}
}

\caption{Additional metric evaluation (SacreBLEU, Similarity Score) for the six framework variants on the 1.5K QA-pair generalization benchmark. Results confirm the pattern observed in Figure \ref{fig:em1}: (1) fine-tuned Llama-3.2 1B variants (A and B) consistently outperform SmolLM2-135M counterparts (E and F), (2) the baseline configuration (D) remains the weakest, and (3) Variant A (with both fine-tuning and GraphRAG) delivers optimal performance across all quality dimensions.}
\label{fig:em2}
\end{figure}

Figure~\ref{fig:em1}(e) displays ROUGE-2 scores measuring bigram overlap across the six framework variants. Variant A (Llama-3.2 1B with both fine-tuning and GraphRAG) achieves the highest score (>0.20). Both components independently enhance performance relative to the baseline (Variant D), with the fine-tuned SmolLM2-135M also showing GraphRAG benefits (E vs. F). Figure~\ref{fig:em1}(f) shows ROUGE-L scores evaluating sentence-level alignment, where Variant A again leads (~0.26). Fine-tuning drives most improvement for Llama-3.2 1B (A vs. B) while GraphRAG provides complementary gains, a pattern mirrored in SmolLM2-135M (E vs. F). These results demonstrate that Variant A's combined approach yields optimal performance, with fine-tuning contributing primary improvements and GraphRAG offering secondary enhancements across both metrics. The consistent pattern across ROUGE-2 and ROUGE-L confirms the synergistic effect of these components in improving both bigram matching and longer-sequence alignment. Figure~\ref{fig:em2}(a) reports SacreBLEU scores (n-gram precision) across six model variants. Variant A (Llama-3.2 1B w/FT w/GraphRAG) achieves superior performance ($\approx 0.168$). Fine-tuning alone substantially boosts Llama-3.2 1B's scores (B vs. D), with GraphRAG providing further enhancement (A vs. B). The pretrained baseline (D) performs weakest, while GraphRAG also benefits the fine-tuned SmolLM2-135M (E vs. F). Llama-3.2 1B consistently outperforms SmolLM2-135M across all variants. Figure~\ref{fig:em2}(b) shows semantic alignment scores ($>0.85$), with Variant A peaking at $\approx 0.93$. Fine-tuning drives most improvement (B vs. D), while GraphRAG provides smaller gains (A vs. B). The trend holds for both Llama-3.2 1B (A,B) and SmolLM2-135M (E,F), with baseline D performing weakest.

%%%%%%%%%%%%%%%%%%%%%%%%%%%%%% Test Data time and CE %%%%%%%%%%%%%%%%%%%%%%

\subsubsection{Computational Tradeoffs: Runtime and Carbon Costs Across Framework Variants}
\begin{figure}[ht!]
\centering
\captionsetup[subfloat]{font=scriptsize,labelfont=bf}
% Subfigure (a): Computational Time (minutes)
\subfloat[Model runtime (minutes) on the 1.5K QA-pair generalization benchmark. 
Llama-3.2 1B variants (A--D) require substantially longer inference time than SmolLM2-135M variants (E, F). 
Enabling GraphRAG increases model runtime across all variants (A vs B, D vs C, E vs F). 
Variant D is slowest; Variant F is fastest.]{
    \includegraphics[width=0.45\textwidth]{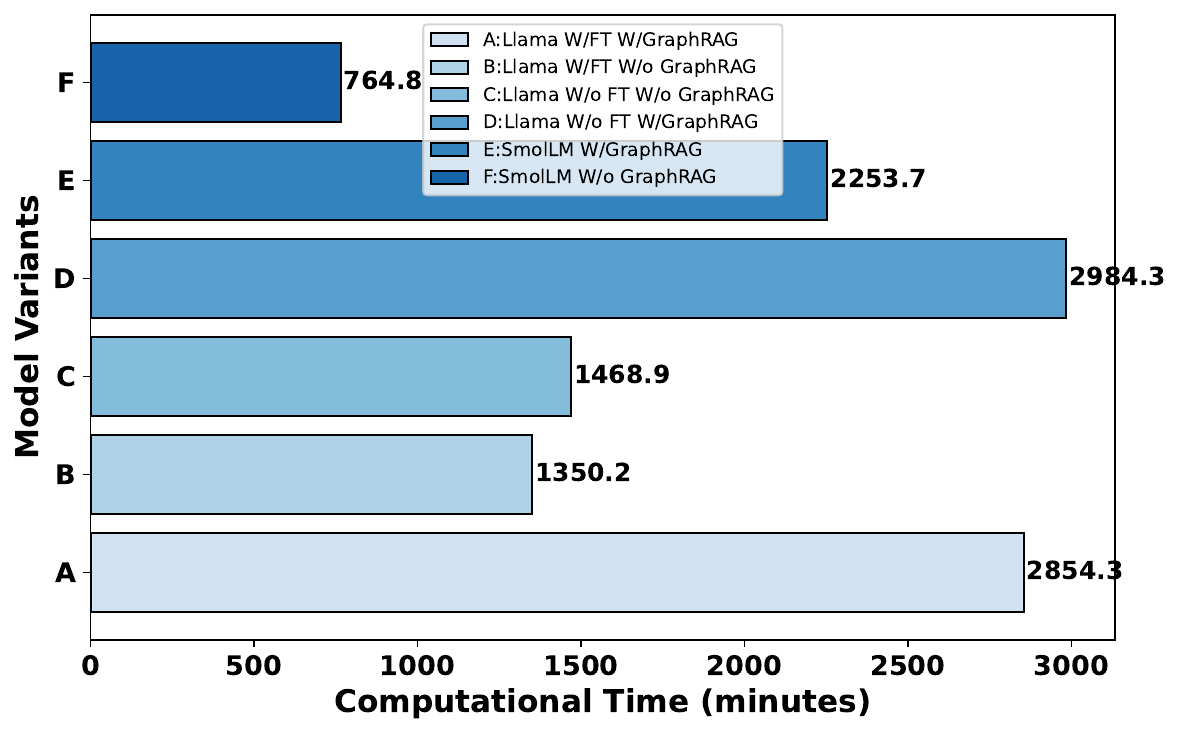}
    \label{fig:fet_runtime}
}
\vspace{1mm} % adds spacing between the two stacked figures
% Subfigure (b): Carbon Emissions (kg CO2)
\subfloat[Estimated model CO\textsubscript{2} emissions during evaluation. 
Llama-3.2 1B variants (A--D) show higher emissions than SmolLM2-135M variants (E, F). 
GraphRAG consistently increases environmental cost across all model variants (A vs B, D vs C, E vs F). 
Variant D yields the highest emissions; Variant F yields the lowest.]{
    \includegraphics[width=0.45\textwidth]{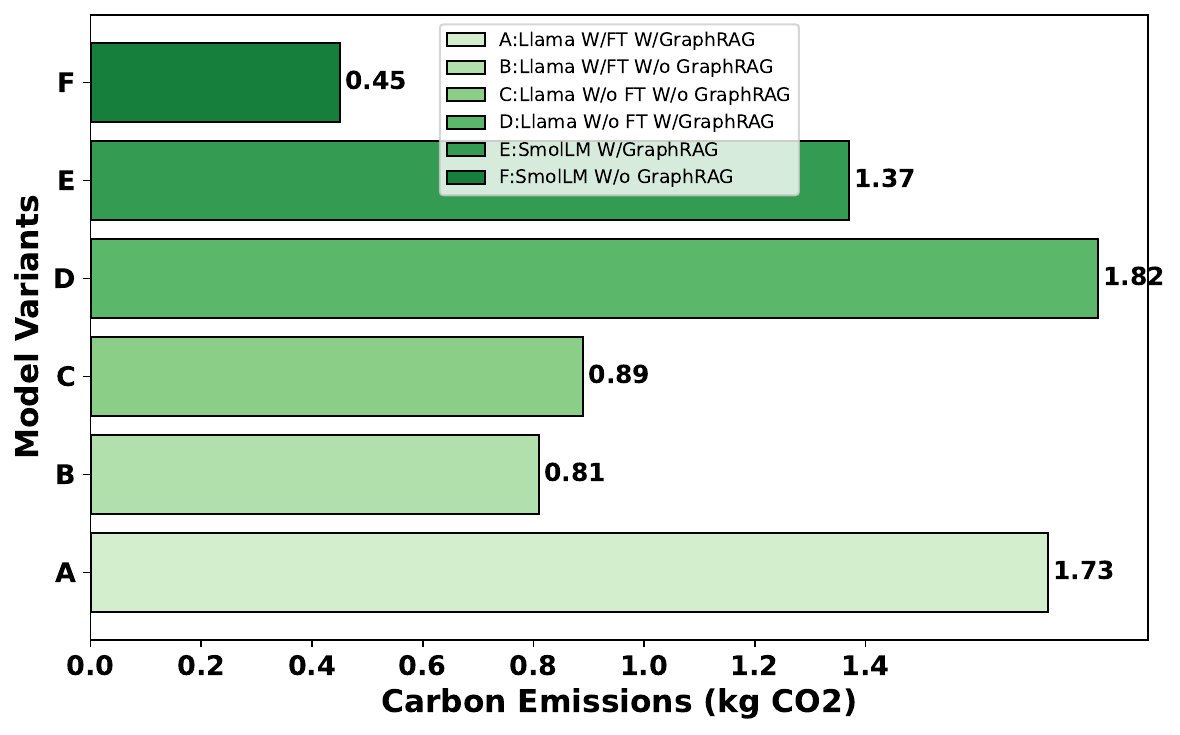}
    \label{fig:fet_emissions}
}

\caption{Evaluation-time computational cost and carbon impact across six framework variants (A--F) on the 1.5K QA-pair generalization benchmark. 
(a) Model runtime in minutes. 
(b) Estimated model CO\textsubscript{2} emissions in kg. 
SmolLM2-135M variants (E, F) are markedly more efficient than Llama variants. 
GraphRAG increases both runtime and emissions across all configurations.}
\label{fig:fet}
\vspace{-3mm}
\end{figure}

We analyze the computational efficiency and environmental footprint of the six framework variants during evaluation on the 1.5K QA-pair generalization benchmark. Figure~\ref{fig:fet} quantifies runtime and estimated CO\textsubscript{2} emissions across Llama-3.2 1B and SmolLM2-135M configurations. Larger Llama variants (A--D) consistently require more inference time (Figure~\ref{fig:fet}a) and produce higher carbon emissions (Figure~\ref{fig:fet}b) than compact SmolLM counterparts (E, F). Within each model family, GraphRAG increases computational overhead and emissions---evident from comparisons A vs B, D vs C, and E vs F. Variant D (Llama-3.2 1B w/o FT w/GraphRAG) incurs the highest computational cost and carbon output, while Variant F (SmolLM2-135M w/FT w/o GraphRAG) is the most resource-efficient. These results highlight a clear tradeoff between model size, retrieval augmentation, and evaluation efficiency.

%%%%%%%%%%%%%%%%%%%%%%%%%%%%%%%%%%%%%%%%%%%%%%%%%%%%%%%%%%%%%%%%%%%%%%%%%%%%%%%

\subsection{Inference Optimization Techniques}

\subsubsection{Width and Depth Pruning}
Transformer-based language models are computationally expensive, with inference cost proportional to model size multiplied by the sum of input and output tokens: $\text{Inference Cost} \propto \text{Model Size} \times (\text{Input Tokens} + \text{Output Tokens})$. Pruning\cite{kim2024shortened, sun2025curse, wu2024llm, zhu2024survey, sun2023simple, sandri20252ssp, tang2025darwinlm, gao2024bypass, lu2024alphapruning} is a model compression technique that removes less critical components to reduce model size and inference costs while preserving accuracy. This enables efficient deployment of smaller, faster, and more cost-effective models in resource-constrained environments. We consider a small-scale transformer-based language model represented as a parameterized function:

\vspace{-1mm}
\resizebox{0.985\linewidth}{!}{
\begin{minipage}{\linewidth}
\begin{equation}
\mathcal{F}_\theta: \mathbb{R}^{T \times d_{\text{model}}} \rightarrow \mathbb{R}^{T \times \mathcal{V}} \nonumber
\end{equation}
\end{minipage}
}

where $T$ is the sequence length, $d_{\text{model}}$ is the hidden dimensionality, and $\mathcal{V}$ is the vocabulary size. The model consists of $L$ stacked transformer blocks $\{\mathcal{T}_\ell\}_{\ell=1}^L$, each comprising a Grouped Query Attention (GQA) module, a feedforward network (FFN), residual connections, and pre-layer normalization.
Grouped Query Attention (GQA) separates the number of query heads from key-value heads to improve efficiency. Let $H_q$ and $H_{kv}$ denote the number of query and key-value heads, respectively, with $H_q \geq H_{kv}$. The grouping factor $g = H_q / H_{kv}$ represents the number of query heads that share each key-value head, and $d_h = d_{\text{model}} / H_q$ is the dimensionality per query head. Given input $\mathbf{X} \in \mathbb{R}^{T \times d_{\text{model}}}$, the linear projections are:

\vspace{-1mm}
\resizebox{0.985\linewidth}{!}{
\begin{minipage}{\linewidth}
\begin{equation}
\begin{split}
\mathbf{Q} = \mathbf{X} W^Q \in \mathbb{R}^{T \times H_q \times d_h}, \quad \mathbf{K} = \mathbf{X} W^K \in \mathbb{R}^{T \times H_{kv} \times d_h}, \\
\mathbf{V} = \mathbf{X} W^V \in \mathbb{R}^{T \times H_{kv} \times d_h} \nonumber
\end{split}
\end{equation}
\end{minipage}
}

For each query head $i \in \{1, \ldots, H_q\}$, its associated key-value head is determined by $k(i) = \lfloor (i-1) / g \rfloor$, which maps each query head to its corresponding key-value head by grouping $g$ query heads per key-value head. The attention output for head $i$ is:

\vspace{-1mm}
\resizebox{0.985\linewidth}{!}{
\begin{minipage}{\linewidth}
\begin{equation}
\mathbf{O}_i = \text{softmax} \left( \frac{\mathbf{Q}_i \mathbf{K}_{k(i)}^\top}{\sqrt{d_h}} \right) \mathbf{V}_{k(i)} \in \mathbb{R}^{T \times d_h} \nonumber
\end{equation}
\end{minipage}
}

The final GQA output is obtained by concatenating all attention heads and applying an output projection:

\vspace{-1mm}
\resizebox{0.985\linewidth}{!}{
\begin{minipage}{\linewidth}
\begin{equation}
\text{GQA}(\mathbf{X}) = \text{Concat}(\mathbf{O}_1, \ldots, \mathbf{O}_{H_q}) W^O \nonumber
\end{equation}
\end{minipage}
}

where $W^O \in \mathbb{R}^{d_{\text{model}} \times d_{\text{model}}}$ is the output projection matrix. The decoder language models implement FFNs using Gated Linear Units (GLUs), which apply an activation function to one projection and use it to gate another projection:

\vspace{-1mm}
\resizebox{0.985\linewidth}{!}{
\begin{minipage}{\linewidth}
\begin{equation}
\text{FFN}(\mathbf{h}) = W_2 \left( \phi(W_1^{(a)} \cdot \mathbf{h}) \odot (W_1^{(b)} \cdot \mathbf{h}) \right) \nonumber
\end{equation}
\end{minipage}
}

where $W_1^{(a)}, W_1^{(b)} \in \mathbb{R}^{d_{\text{ff}} \times d_{\text{model}}}$ are the gate and up-projection matrices, $W_2 \in \mathbb{R}^{d_{\text{model}} \times d_{\text{ff}}}$ is the down-projection matrix, $\phi$ is an activation function (typically SiLU or GELU), and $\odot$ denotes element-wise multiplication. The Width pruning reduces the intermediate FFN dimensionality $d_{\text{ff}}$ by eliminating unimportant neurons. For the $j$-th neuron output $z_j$ from the GLU, we estimate its importance using gradient-based scoring:

\vspace{-1mm}
\resizebox{0.985\linewidth}{!}{
\begin{minipage}{\linewidth}
\begin{equation}
I_j = \mathbb{E}_{\mathbf{x} \sim \mathcal{D}} \left[ \left| \frac{\partial \mathcal{L}}{\partial z_j} \cdot z_j \right| \right] \nonumber
\end{equation}
\end{minipage}
}

where $\mathcal{L}$ is the task loss and $\mathcal{D}$ is the data distribution. This importance score $I_j$ quantifies the average contribution of neuron $j$ to the task loss. Neurons with the lowest $I_j$ values are pruned, reducing the width to $\tilde{d}_{\text{ff}} < d_{\text{ff}}$ by removing corresponding rows in $W_1^{(a)}$ and $W_1^{(b)}$, and columns in $W_2$. The depth pruning removes entire transformer blocks based on their contribution to the task. For layer $\ell \in \{1, \ldots, L\}$, its importance is computed as:

\vspace{-1mm}
\resizebox{0.985\linewidth}{!}{
\begin{minipage}{\linewidth}
\begin{equation}
I^{(\ell)} = \mathbb{E}_{\mathbf{x} \sim \mathcal{D}} \left[ \left| \left\langle \frac{\partial \mathcal{L}}{\partial \mathbf{h}^{(\ell)}}, \mathbf{h}^{(\ell)} \right\rangle \right| \right] \nonumber
\end{equation}
\end{minipage}
}

where $\mathbf{h}^{(\ell)}$ is the residual output of block $\ell$, and $\langle \cdot, \cdot \rangle$ denotes the inner product. Layers with small $I^{(\ell)}$ values are removed, and the retained set is denoted $\mathcal{S} \subset \{1, \ldots, L\}$ with $|\mathcal{S}| = \tilde{L} \ll L$. For joint width and depth pruning, we introduce binary gates to control both layer and neuron retention: $\gamma^{(\ell)} \in \{0,1\}$ for layer $\ell$ retention and $g_j^{(\ell)} \in \{0,1\}$ for neuron $j$ retention in layer $\ell$. The forward computation becomes:

\vspace{-1mm}
\resizebox{0.985\linewidth}{!}{
\begin{minipage}{\linewidth}
\begin{equation}
\mathbf{h}^{(\ell)} = \mathbf{h}^{(\ell-1)} + \gamma^{(\ell)} \cdot \text{FFN}_\ell^{(g)}\left(\text{GQA}_\ell(\text{LN}(\mathbf{h}^{(\ell-1)}))\right) \nonumber
\end{equation}
\end{minipage}
}

where $\text{LN}$ denotes layer normalization, and the gated FFN is defined as:

\vspace{-1mm}
\resizebox{0.985\linewidth}{!}{
\begin{minipage}{\linewidth}
\begin{align}
\text{FFN}_\ell^{(g)}(\mathbf{h}) 
&= \sum_{j=1}^{d_{\text{ff}}} g_j^{(\ell)} \cdot \Big[ 
      W_2[:,j] \cdot \big( \phi(W_1^{(a)}[j,:] \cdot \mathbf{h}) \nonumber \\
& \hspace{27mm} \cdot (W_1^{(b)}[j,:] \cdot \mathbf{h}) \big) \Big] \nonumber
\end{align}
\end{minipage}
}

\begin{figure*}[ht!]
\centering
\resizebox{1.0\textwidth}{!}{
\subfloat[Impact of width pruning on fine-tuned model performance using reward model scores (0-4 scale) across five metrics on the 1.5K \textit{QA}-pair generalization benchmark. Higher pruning percentages degrade correctness and helpfulness most significantly.]{\includegraphics[width=75mm]{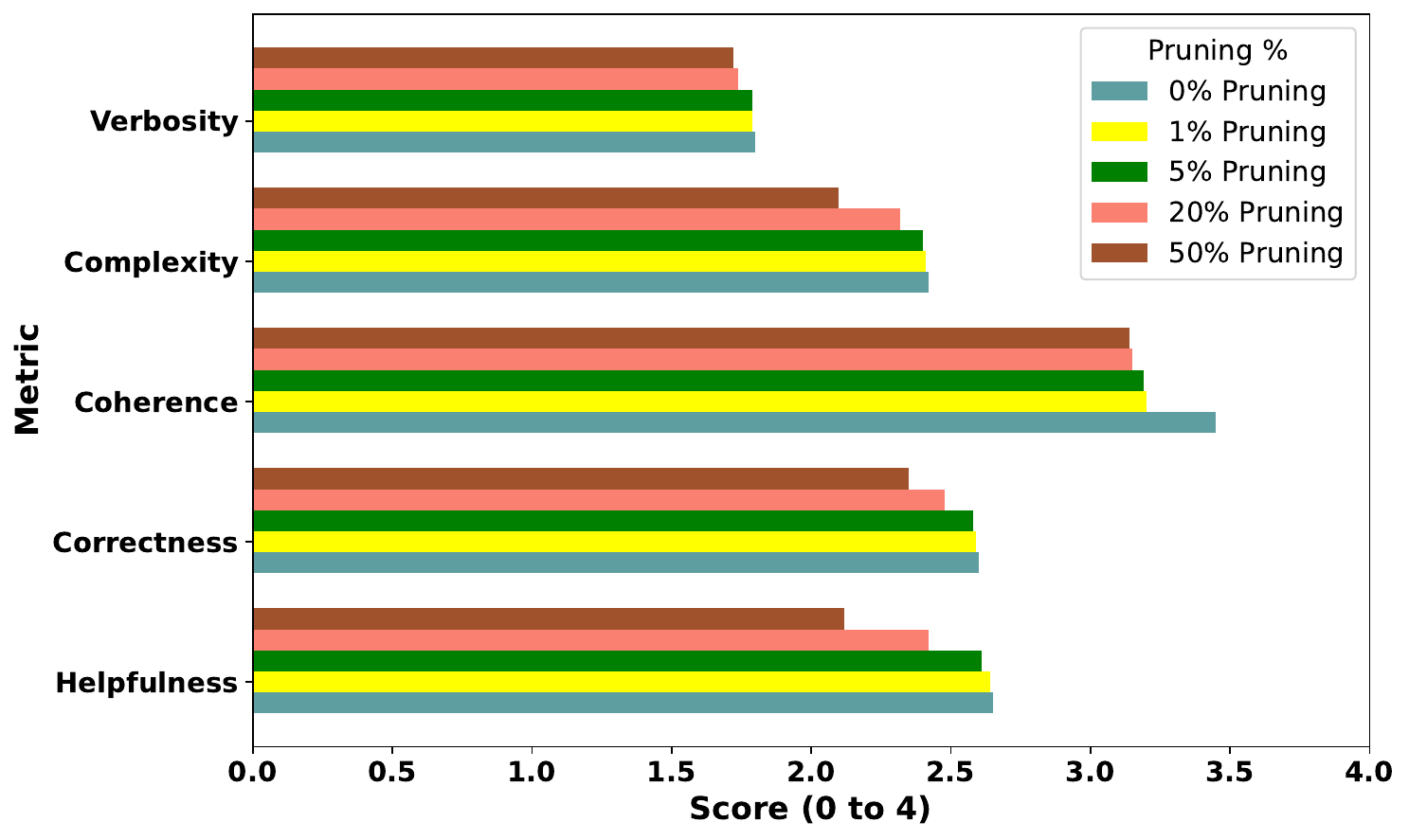}\label{fig:width_prune_test}} 
\hspace{0.05\textwidth}
\subfloat[Impact of depth pruning on fine-tuned model performance using reward model scores (0-4 scale) across five metrics on the 1.5K \textit{QA}-pair generalization benchmark. Performance decline is most pronounced in coherence and complexity.]{\includegraphics[width=75mm]{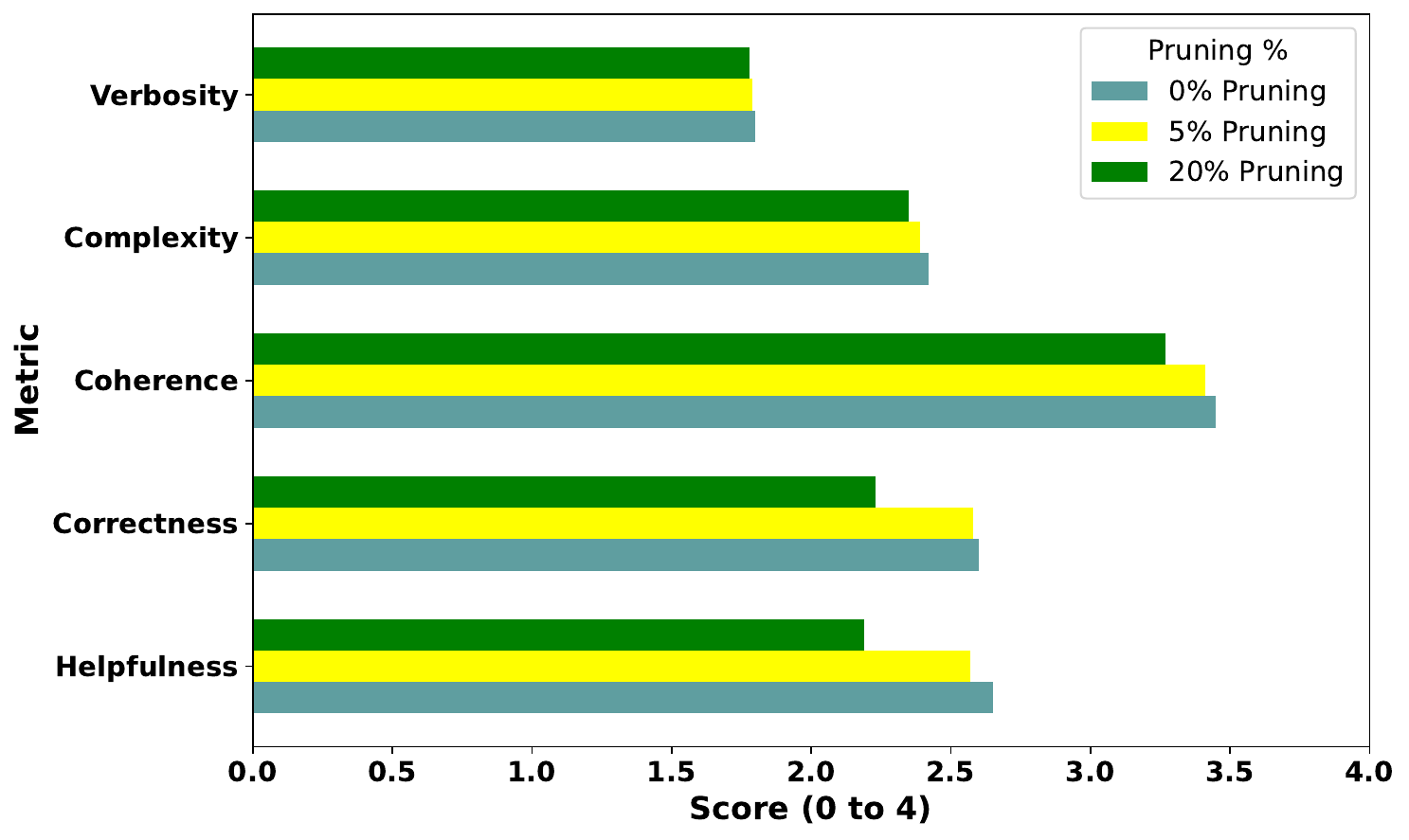}\label{fig:depth_prune_test}} 
}
\vspace{0mm}
\caption{Evaluation of width (a) and depth (b) pruning effects on fine-tuned model quality. Performance measured using reward model scores across five dimensions on the 1.5K \textit{QA}-pair generalization benchmark, demonstrating trade-offs between model compression and response quality.} 
\label{fig:pruning_impact_test} 
\vspace{-0mm}
\end{figure*}

\begin{figure*}[ht!]
\vspace{-3mm}
\centering
\resizebox{1.0\textwidth}{!}{
\subfloat[Impact of width pruning on PFD/PID generation performance using reward model scores (0-4 scale) on the \textit{ChemEval} benchmark. Quality degradation increases with higher pruning percentages.]{\includegraphics[width=75mm]{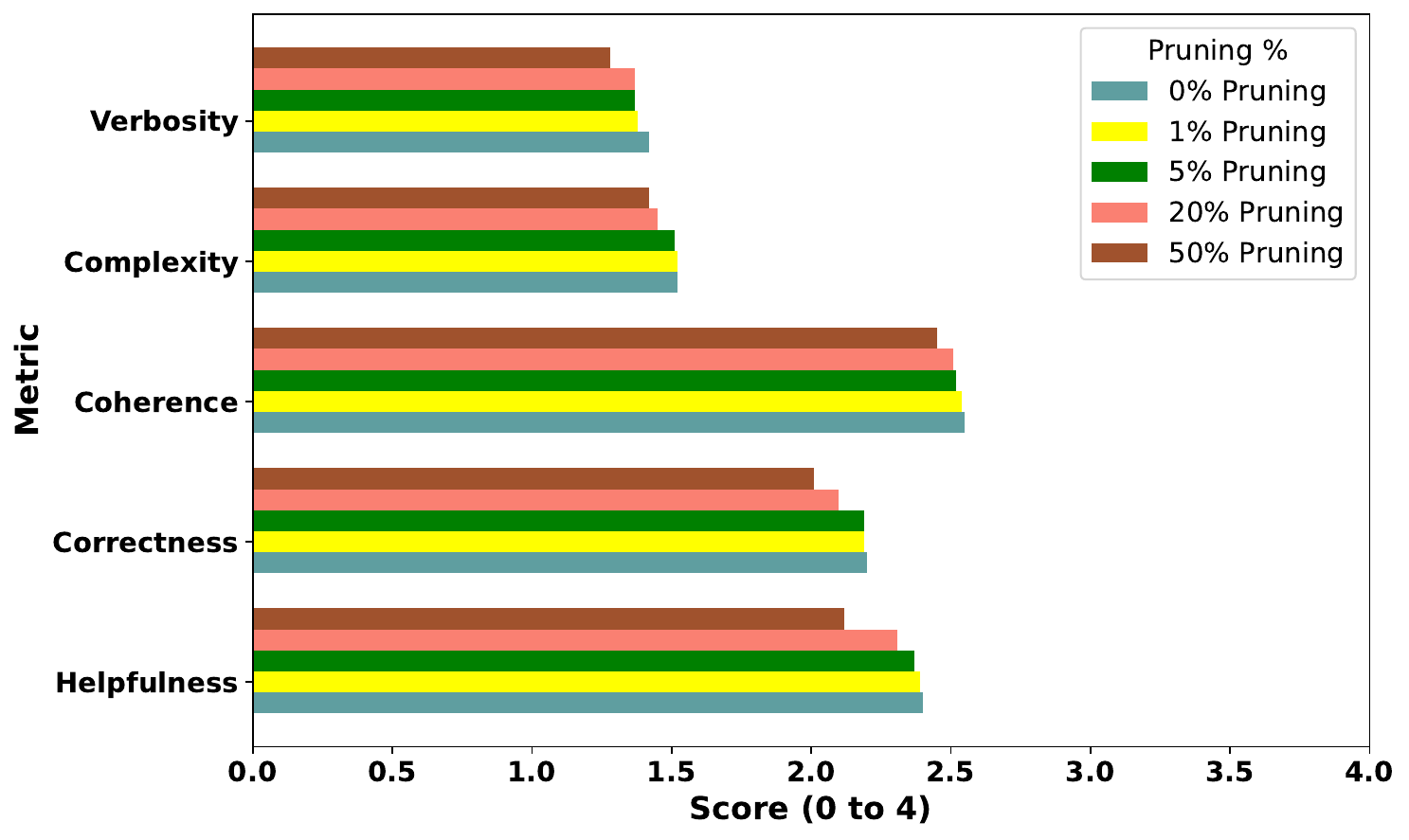}\label{fig:width_prune_chemeval}}
\hspace{0.05\textwidth}
\subfloat[Impact of depth pruning on PFD/PID generation performance using reward model scores (0-4 scale) on the \textit{ChemEval} benchmark. Layer removal leads to progressive performance decline.]{\includegraphics[width=75mm]{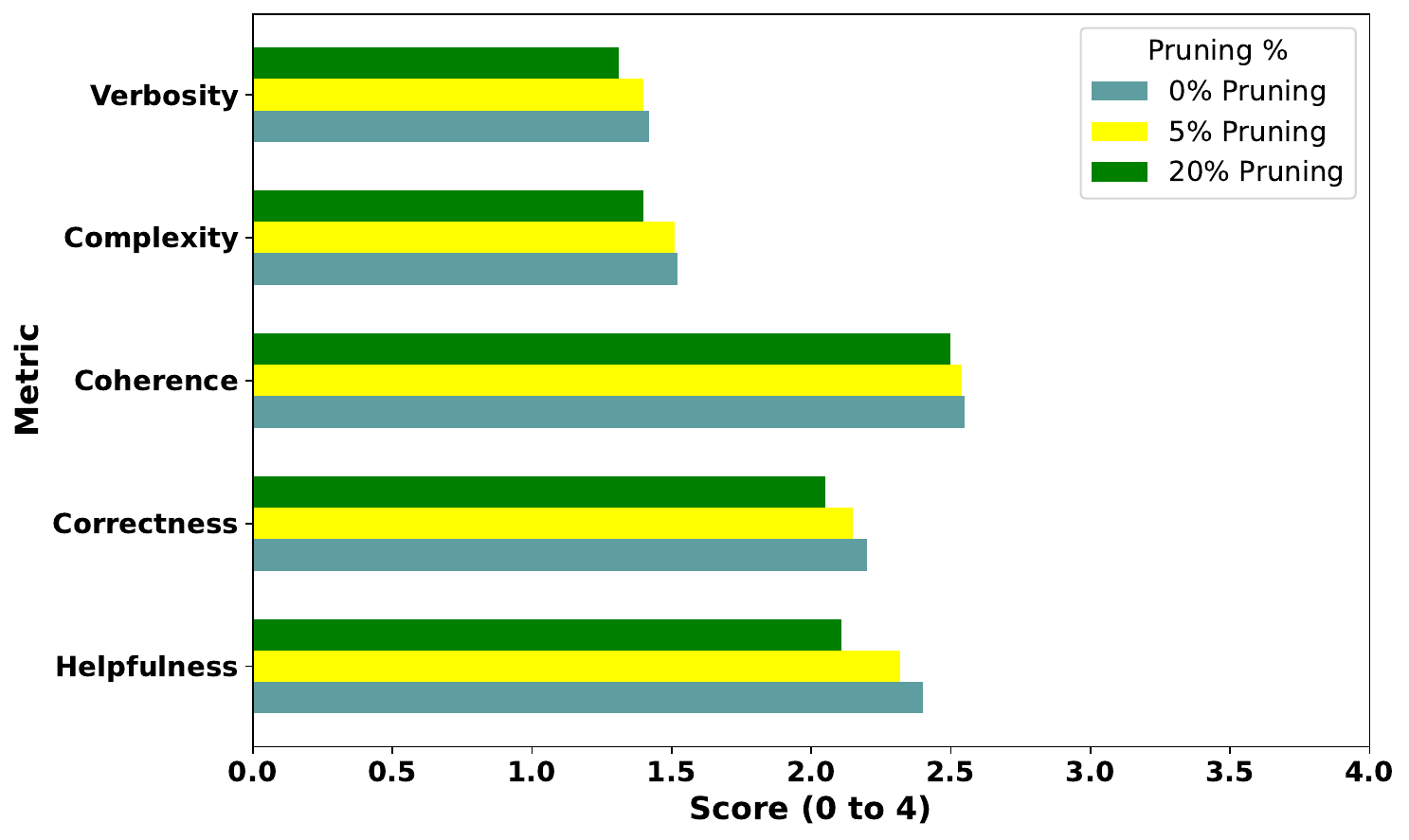}\label{fig:depth_prune_chemeval}}
}
\vspace{0mm}
\caption{Evaluation of width (a) and depth (b) pruning effects on specialized task performance for zero-shot PFD/PID generation. Performance measured using reward model scores on the \textit{ChemEval} benchmark, illustrating compression impact on domain-specific capabilities.} 
\label{fig:pruning_impact_chemeval} 
\vspace{-2mm}
\end{figure*}

\begin{figure*}[ht!]
\centering
\resizebox{0.85\textwidth}{!}{
\subfloat[Computational time reduction (minutes) as a function of width pruning percentage on the fine-tuned model during evaluation.]{\includegraphics[width=60mm]{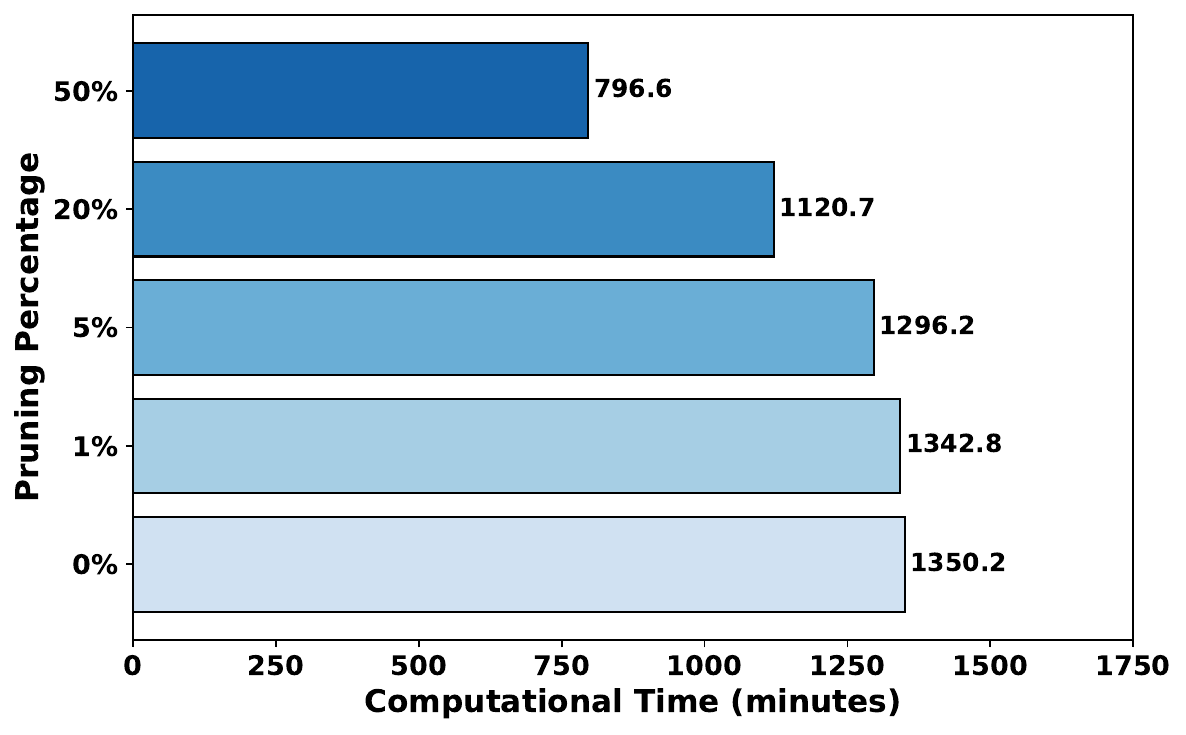}\label{fig:width_prune_time}} 
\hspace{0.05\textwidth}
\subfloat[Computational time reduction (minutes) as a function of depth pruning percentage on the fine-tuned model during evaluation.]{\includegraphics[width=60mm]{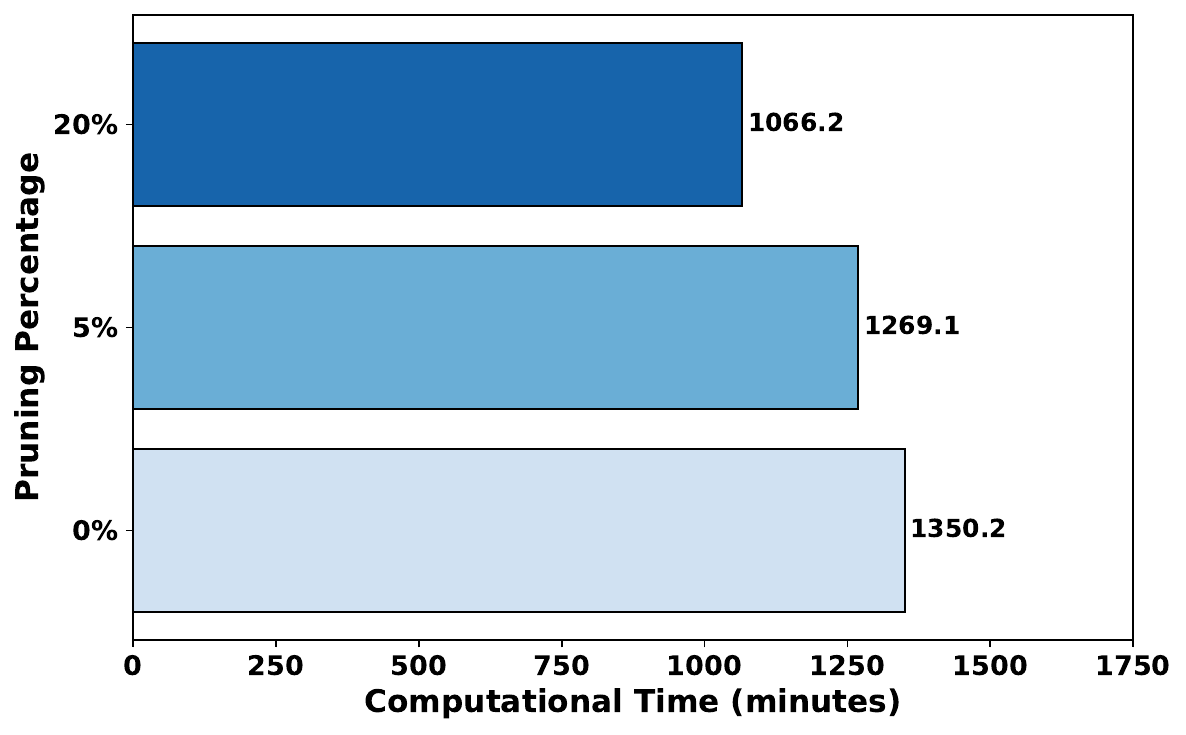}\label{fig:depth_prune_time}} 
}
\vspace{2mm}
\caption{Computational efficiency gains from width (a) and depth (b) pruning during evaluation. Plots demonstrate runtime reduction (minutes) as pruning percentage increases, showing the potential for faster inference with compressed models.}
\label{fig:prun}
\end{figure*}

The constrained optimization objective combines empirical risk minimization with sparsity penalties to achieve structured pruning:

\vspace{-1mm}
\resizebox{0.985\linewidth}{!}{
\begin{minipage}{\linewidth}
\begin{align}
\min_{\theta, \gamma, g} \; & \mathbb{E}_{(\mathbf{x}, y) \sim \mathcal{D}} 
\left[ \mathcal{L}(\mathcal{F}_{\theta, \gamma, g}(\mathbf{x}), y) \right]  \nonumber \\
& + \lambda_1 \sum_{\ell=1}^L (1 - \gamma^{(\ell)}) 
+ \lambda_2 \sum_{\ell=1}^L \sum_{j=1}^{d_{\text{ff}}} (1 - g_j^{(\ell)}) \nonumber
\end{align}
\end{minipage}
}

where $\mathcal{F}_{\theta, \gamma, g}$ is the pruned model with parameters $\theta$ and gates $\gamma, g$, $(\mathbf{x}, y)$ represents input-output pairs from the training distribution $\mathcal{D}$, $\mathcal{L}(\cdot, \cdot)$ is the task-specific loss function (e.g., cross-entropy), $\lambda_1$ controls the depth sparsity penalty (number of pruned layers), and $\lambda_2$ controls the width sparsity penalty (number of pruned neurons).
To enable end-to-end differentiability, we relax the binary gates using the Concrete distribution (also known as the Gumbel-Softmax trick). Each gate $g_j^{(\ell)} \in [0, 1]$ is sampled as:

\vspace{-4mm}
\resizebox{0.985\linewidth}{!}{
\begin{minipage}{\linewidth}
\begin{equation}
g_j^{(\ell)} = \sigma \left( \frac{1}{\tau} \left( \log \alpha_j^{(\ell)} + \log u - \log(1 - u) \right) \right), \quad u \sim \mathcal{U}(0, 1) \nonumber
\end{equation}
\end{minipage}
}

where $u \sim \mathcal{U}(0,1)$ is a uniform random variable, $\sigma(\cdot)$ is the sigmoid function, $\alpha_j^{(\ell)}$ is a learnable logit parameter, and $\tau > 0$ is a temperature parameter controlling the smoothness of the relaxation. A similar sampling strategy is applied to layer-level gates $\gamma^{(\ell)}$. The joint pruning approach performs structured pruning during fine-tuning to simultaneously optimize model performance and sparsity. The optimization process learns both the pruned model structure and the corresponding parameters $\theta$. After training, components with binarized gates $\gamma^{(\ell)} = 0$ or $g_j^{(\ell)} = 0$ are permanently removed, retaining only the most important neurons and layers. This structured sparsity approach achieves significant model compression while maintaining downstream task accuracy. The regularization hyperparameters $\lambda_1$ and $\lambda_2$ control the trade-off between accuracy and compression, allowing practitioners to tune the desired level of sparsity based on deployment constraints. Figure \ref{fig:pruning_impact_test} demonstrates the effects of width and depth pruning on model performance using five qualitative metrics scored from 0 to 4 scale by the  Nemotron-4-340B-Reward model. Both pruning methods generally decrease performance scores as pruning percentages increase. Width pruning (Figure \ref{fig:pruning_impact_test}a) particularly affects correctness, complexity, and helpfulness, showing notable drops at the 20\% level, while coherence remains relatively stable. Depth pruning (Figure \ref{fig:pruning_impact_test}b) more severely impacts coherence and complexity, especially at higher pruning ratios (20\% and 50\%). Correctness shows resilience to low-level depth pruning (1-5\%) but declines significantly thereafter. Verbosity remains the least affected metric across both methods at low to moderate pruning levels. Figures \ref{fig:width_prune_chemeval} and \ref{fig:depth_prune_chemeval} show the impact of width and depth pruning on \textit{ChemEval} benchmark performance for PFD/PID generation tasks involving unseen chemicals. Both methods demonstrate quality degradation across all metrics as pruning percentages increase. Width pruning (Figure \ref{fig:width_prune_chemeval}) at higher levels (particularly 20\%) significantly reduces correctness and complexity scores. Depth pruning (Figure \ref{fig:depth_prune_chemeval}) similarly reduces overall quality, with coherence and correctness notably impacted at 20\% and 50\% levels. Verbosity remains the least affected metric for both pruning approaches on \textit{ChemEval} tasks, indicating that structural compression via either method hinders the model's ability to generate accurate and coherent PFD/PID descriptions for novel chemical processes. Figures \ref{fig:prun}a and \ref{fig:prun}b present the relationship between computational time and pruning percentages for width and depth pruning, respectively. Figure \ref{fig:prun}a shows that increasing width pruning correlates with decreased computational time: baseline (0\%) required 1350.2 minutes, 5\% pruning took 1269.1 minutes, and 20\% pruning took 1066.2 minutes. Similarly, Figure \ref{fig:prun}b indicates that depth pruning also reduces computational time, with baseline at 1350.2 minutes, decreasing through 1\% (1342.8 min), 5\% (1296.2 min), and 20\% (1120.7 min), achieving the most significant reduction at 50\% depth pruning (796.6 minutes). Both figures illustrate a consistent inverse relationship between pruning percentage and computational time.

%%%%%%%%%%%%%%%%%%%%%%%%%%%%%%%%%%%%%%%%%%%%%%%%%%%%%%%%%%%%%%%%%%%%%%%%%%%%%%%

\vspace{-3mm}
\subsubsection{Test-Time Inference Scaling via Self-Consistency, Confidence-Weighted Entropy, and Self-Reflection}
To address limitations in factual accuracy, reliability, and reasoning robustness in small-scale language models (SLMs), we propose a test-time inference scaling mechanism~\cite{balachandran2025inference, zhang2025and, liu2025efficient, liu2025metascale, singhi2025solve, snell2024scaling, li2025survey, zhang2025and, yang2025towards, bi2024forest, yu2025z1, chen2025sets, qu2025optimizing} that combines three complementary strategies: (1) self-consistency decoding, (2) confidence-weighted entropy scoring, and (3) a self-reflection-based revision mechanism. Unlike fine-tuning or prompt engineering approaches, this method operates purely at inference time, requiring no model parameter updates, and is particularly well-suited for tasks involving multi-step reasoning, such as automatic generation of PFDs and PIDs. The first step involves multiple candidate generation via Chain-of-Thought sampling. Given an input query $x$, the model generates a set of $N$ diverse reasoning trajectories $\mathcal{Y} = {y^{(1)}, y^{(2)}, \ldots, y^{(N)}}$, where each candidate sequence $y^{(i)} = (y_1^{(i)}, y_2^{(i)}, \ldots, y_T^{(i)}) \in \mathcal{V}^T$ is produced using stochastic decoding (e.g., nucleus sampling or top-$k$ sampling) under a Chain-of-Thought (CoT) prompting strategy. Here, $\mathcal{V}$ denotes the model vocabulary and $T$ represents the maximum generation length. At each decoding step $t$, the token $y_t^{(i)}$ is sampled from the conditional distribution:

\vspace{-1mm}
\resizebox{0.985\linewidth}{!}{
\begin{minipage}{\linewidth}
\begin{equation}
P_\theta(v \mid x, y^{(i)}_{<t}), \quad \forall v \in \mathcal{V}, \nonumber
\end{equation}
\end{minipage}
}

\vspace{0mm}
where $\theta$ denotes the LLM's parameters and $y^{(i)}_{<t}$ refers to the prefix tokens up to decoding step $t-1$. Next, we discuss the confidence-weighted entropy scoring mechanism. We evaluate the quality of each generated sequence $y^{(i)}$ via a confidence-weighted entropy score that reflects model uncertainty per decoding step and overall sequence likelihood. At decoding step $t$, the model provides the predictive distribution $P_t^{(i)}$ over the vocabulary $\mathcal{V}$, conditioned on the input $x$ and prefix $y^{(i)}_{<t}$:

\vspace{-1mm}
\resizebox{0.985\linewidth}{!}{
\begin{minipage}{\linewidth}
\begin{equation}
P_t^{(i)}(v) = P_\theta(v \mid x, y^{(i)}_{<t}), \quad \forall v \in \mathcal{V}. \nonumber
\end{equation}
\end{minipage}
}

\vspace{1mm}
The entropy of this distribution at decoding step $t$ for candidate $i$ is:

\vspace{-1mm}
\resizebox{0.985\linewidth}{!}{
\begin{minipage}{\linewidth}
\begin{equation}
H_t^{(i)} = - \sum_{v \in \mathcal{V}} P_t^{(i)}(v) \log P_t^{(i)}(v), \nonumber
\end{equation}
\end{minipage}
}

\vspace{-1mm}
where $H_t^{(i)}$ quantifies the model's uncertainty about the token choice at decoding step $t$. To reflect the relative importance or semantic salience of each token position in the generated output $y^{(i)}$, we derive importance weights ${w_1^{(i)}, \ldots, w_T^{(i)}}$ using an attention-weighted gradient attribution method. Specifically, let $\alpha_t^{(i)}$ denote the average attention weight received by the $t$-th generated token $y_t^{(i)}$ across all attention heads and layers in the LLM. We define the importance weight for decoding step $t$ as:

\vspace{-1mm}
\resizebox{0.985\linewidth}{!}{
\begin{minipage}{\linewidth}
\begin{equation}
w_t^{(i)} = \frac{\alpha_t^{(i)} \cdot \left| \frac{\partial \mathcal{L}^{(i)}}{\partial \ell_t^{(i)}} \right|}{\sum_{t'=1}^{T} \alpha_{t'}^{(i)} \cdot \left| \frac{\partial \mathcal{L}^{(i)}}{\partial \ell_{t'}^{(i)}} \right|} \cdot T, \nonumber
\end{equation}
\end{minipage}
}

where $\mathcal{L}^{(i)}$ is the negative log-likelihood loss over the candidate sequence $y^{(i)}$, defined as:

\resizebox{0.985\linewidth}{!}{
\begin{minipage}{\linewidth}
\begin{equation}
\mathcal{L}^{(i)} = - \sum_{t=1}^T \log P_\theta(y_t^{(i)} \mid x, y_{<t}^{(i)}). \nonumber
\end{equation}
\end{minipage}
}

\vspace{-1mm}
Here, $\ell_t^{(i)}$ denotes the language model's output logits at decoding step $t$ for candidate $i$, which are used to compute the gradient $\frac{\partial \mathcal{L}^{(i)}}{\partial \ell_t^{(i)}}$ for the attention-weighted attribution. The gradient term $\left| \frac{\partial \mathcal{L}^{(i)}}{\partial \ell_t^{(i)}} \right|$ captures the sensitivity of the loss with respect to the predicted logits for token $y_t^{(i)}$.
This formulation ensures the weights are normalized such that:

\vspace{-3mm}
\resizebox{0.985\linewidth}{!}{
\begin{minipage}{\linewidth}
\begin{equation}
\sum_{t=1}^T w_t^{(i)} = T. \nonumber
\end{equation}
\end{minipage}
}

\vspace{-1mm}
These weights $w_t^{(i)}$ provide a profile of token importance across the sequence, influenced by both attention patterns and gradient magnitudes. Using the token-level entropies $H_t^{(i)}$ and the importance weights $w_t^{(i)}$, we compute the weighted average entropy for the complete candidate sequence $y^{(i)}$:

\vspace{-1mm}
\resizebox{0.985\linewidth}{!}{
\begin{minipage}{\linewidth}
\begin{equation}
\bar{H}_w^{(i)} = \frac{1}{T} \sum_{t=1}^{T} w_t^{(i)} \cdot H_t^{(i)}. \nonumber
\end{equation}
\end{minipage}
}

\vspace{1mm}
This metric aggregates token-level uncertainty, assigning greater significance to uncertainty occurring at decoding steps deemed important by the attention-gradient attribution. Lower values of $\bar{H}_w^{(i)}$ indicate higher confidence, particularly for semantically salient tokens. To measure the overall likelihood of a generated sequence according to the model, we compute the normalized average log-probability:

\resizebox{0.985\linewidth}{!}{
\begin{minipage}{\linewidth}
\begin{equation}
\bar{\ell}^{(i)} = \frac{1}{T} \sum_{t=1}^{T} \log P_\theta(y_t^{(i)} \mid x, y_{<t}^{(i)}). \nonumber
\end{equation}
\end{minipage}
}

A higher (less negative) value of $\bar{\ell}^{(i)}$ indicates that the sequence is more fluent or probable under the model $P_\theta$. To combine model confidence and fluency, we assign a final score to each candidate $y^{(i)}$ by balancing its weighted entropy and average log-likelihood:

\resizebox{0.985\linewidth}{!}{
\begin{minipage}{\linewidth}
\begin{equation}
\text{Score}(y^{(i)}) = \lambda \cdot \bar{H}_w^{(i)} - (1 - \lambda) \cdot \bar{\ell}^{(i)}, \nonumber
\end{equation}
\end{minipage}
}

\vspace{1mm}
where $\lambda \in [0,1]$ is a tunable hyperparameter controlling the trade-off between minimizing uncertainty (favoring lower $\bar{H}_w^{(i)}$) and maximizing sequence likelihood (favoring higher $\bar{\ell}^{(i)}$). A lower overall $\text{Score}(y^{(i)})$ indicates a more desirable candidate sequence, reflecting a better balance between confidence and fluency. Next, we will discuss about the Top-K Candidate Selection. After computing scores for all $N$ generated sequences in $\mathcal{Y}$, we rank them and select the top-$K$ candidates:

\resizebox{0.985\linewidth}{!}{
\begin{minipage}{\linewidth}
\begin{equation}
\mathcal{Y}{\text{top-K}} = \operatorname{TopK}{y \in \mathcal{Y}} \left( -\text{Score}(y), K \right), \nonumber
\end{equation}
\end{minipage}
}

\begin{figure}[htbp!]
\vspace{-1mm}
\centering
\includegraphics[width=0.95\linewidth]{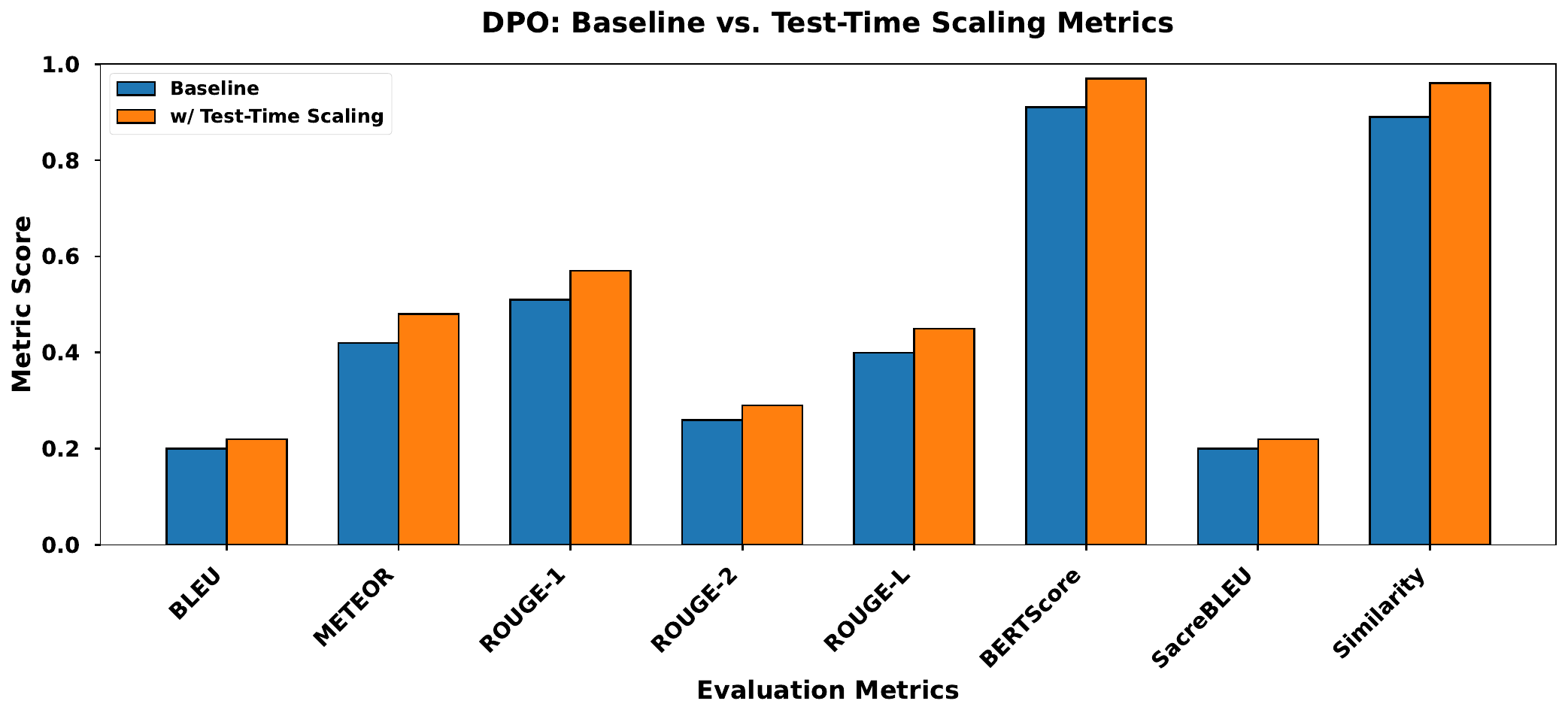}
\vspace{-2mm}
\caption{Comparison of standard NLP metrics on the \textit{DPO} dataset using a fine-tuned Llama-3.2-1B model. The plot contrasts baseline greedy decoding (blue) against test-time inference scaling (orange). The scaling mechanism consistently improves metrics such as METEOR, ROUGE variants, BERTScore, and Similarity, demonstrating enhanced output quality without model parameter updates.}
\label{fig:tts_dpo_separate}
\vspace{-3mm}
\end{figure}

\begin{figure}[htbp!]
\vspace{-1mm}
\centering
\includegraphics[width=0.95\linewidth]{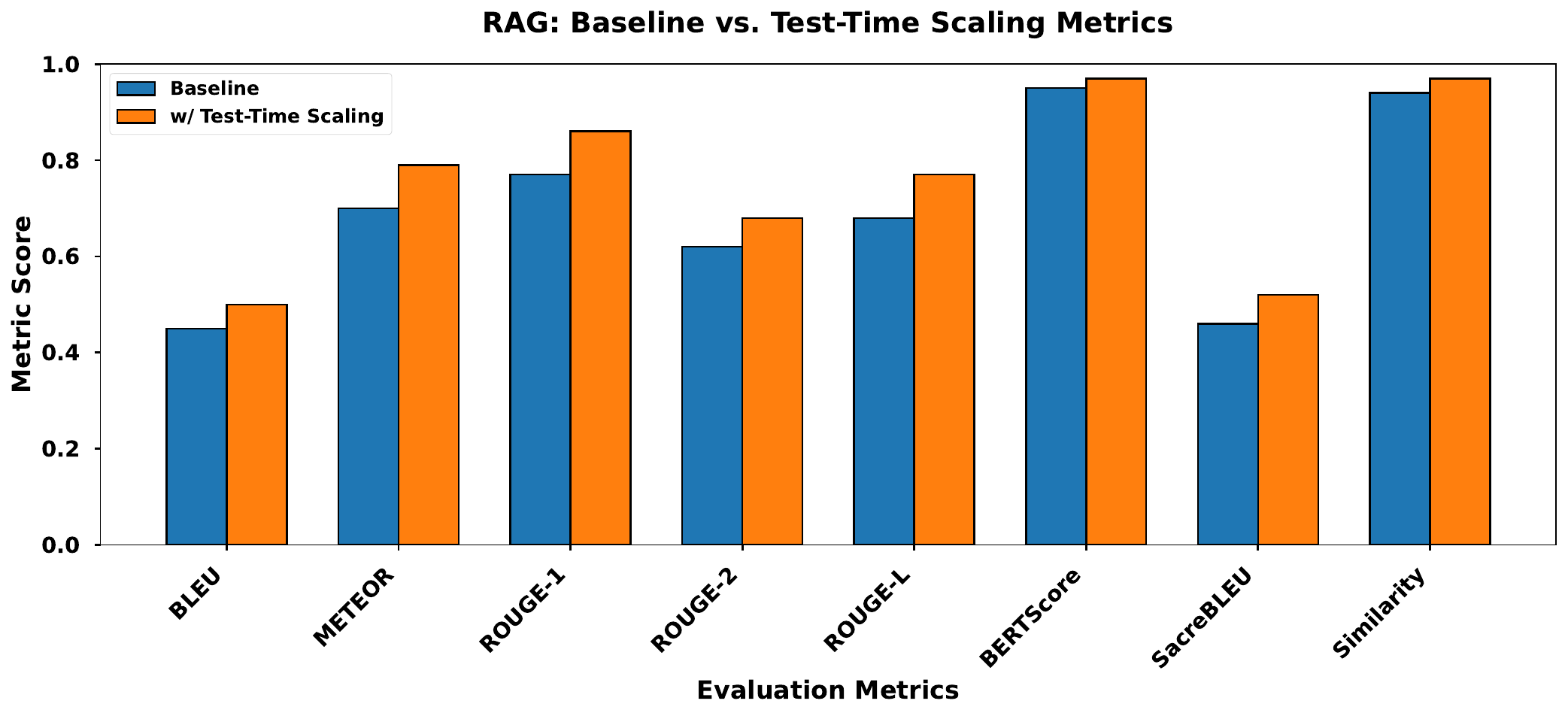}
\vspace{-2mm}
\caption{Comparison of standard NLP metrics on the \textit{RAG} dataset using a fine-tuned Llama-3.2-1B model. Results from baseline greedy decoding (blue) are compared against those from test-time inference scaling (orange). The scaling mechanism notably improves ROUGE-1, ROUGE-L, and Similarity, showcasing its effectiveness over standard decoding.}
\label{fig:tts_rag_separate}
\vspace{-2mm}
\end{figure}

\begin{figure}[htbp!]
\centering
\includegraphics[width=0.95\linewidth]{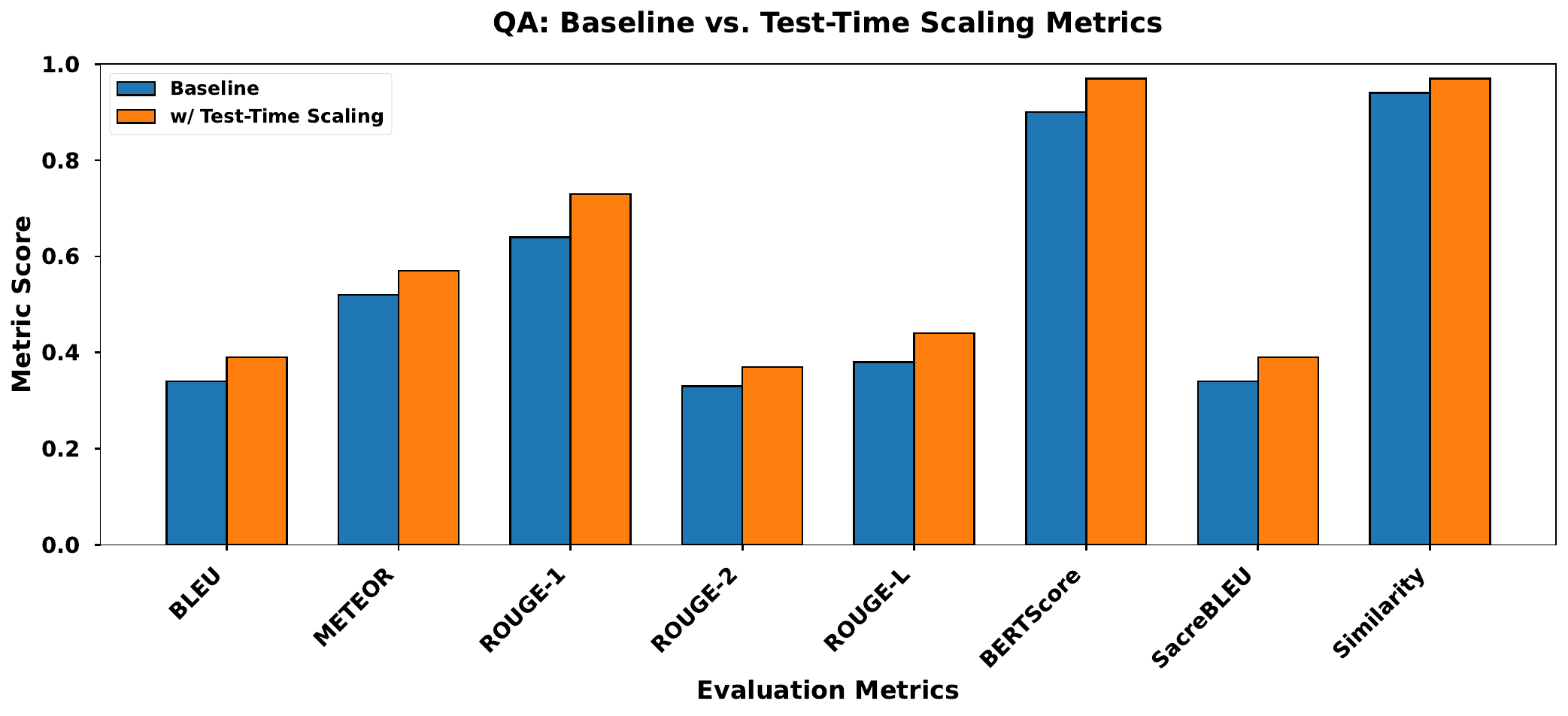}
\vspace{-2mm}
\caption{Comparison of standard NLP metrics on the \textit{General QA} dataset using a fine-tuned Llama-3.2-1B model. Test-time inference scaling (orange) outperforms baseline greedy decoding (blue) across ROUGE-1, ROUGE-L, BERTScore, and Similarity, reinforcing the approach's utility in enhancing SLM robustness after fine-tuning.}
\label{fig:tts_qa_separate}
\vspace{-5mm}
\end{figure}

\begin{figure}[htbp!]
\centering
\includegraphics[width=0.95\linewidth]{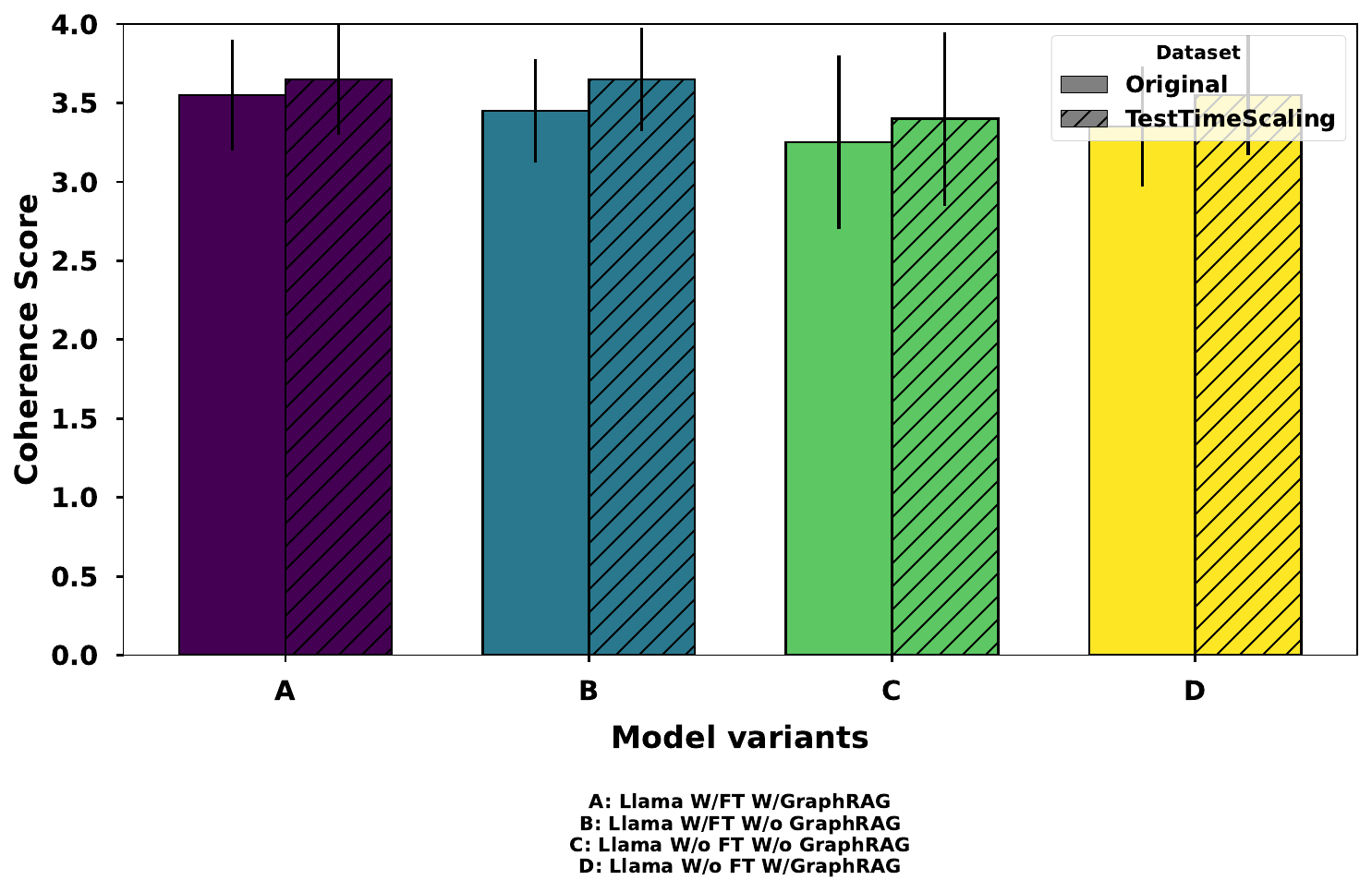}
\vspace{-1mm}
\caption{Effect of test-time inference scaling on \textbf{Coherence Score} across four Llama-3.2-1B variants (A: Fine-tuned with GraphRAG, B: Fine-tuned without GraphRAG, C: Base without GraphRAG, D: Base with GraphRAG), evaluated on the \textit{LogiCore-DPO}, \textit{GraphRAG-RetrievalQA}, and \textit{Factual QA} datasets. Compared to baseline greedy decoding (‘Original’), the scaling mechanism (‘TestTimeScaling’) generally maintains or slightly improves coherence.}
\label{fig:tts_coherence}
\vspace{-4mm}
\end{figure}

\begin{figure}[htbp!]
\centering
\includegraphics[width=0.95\linewidth]{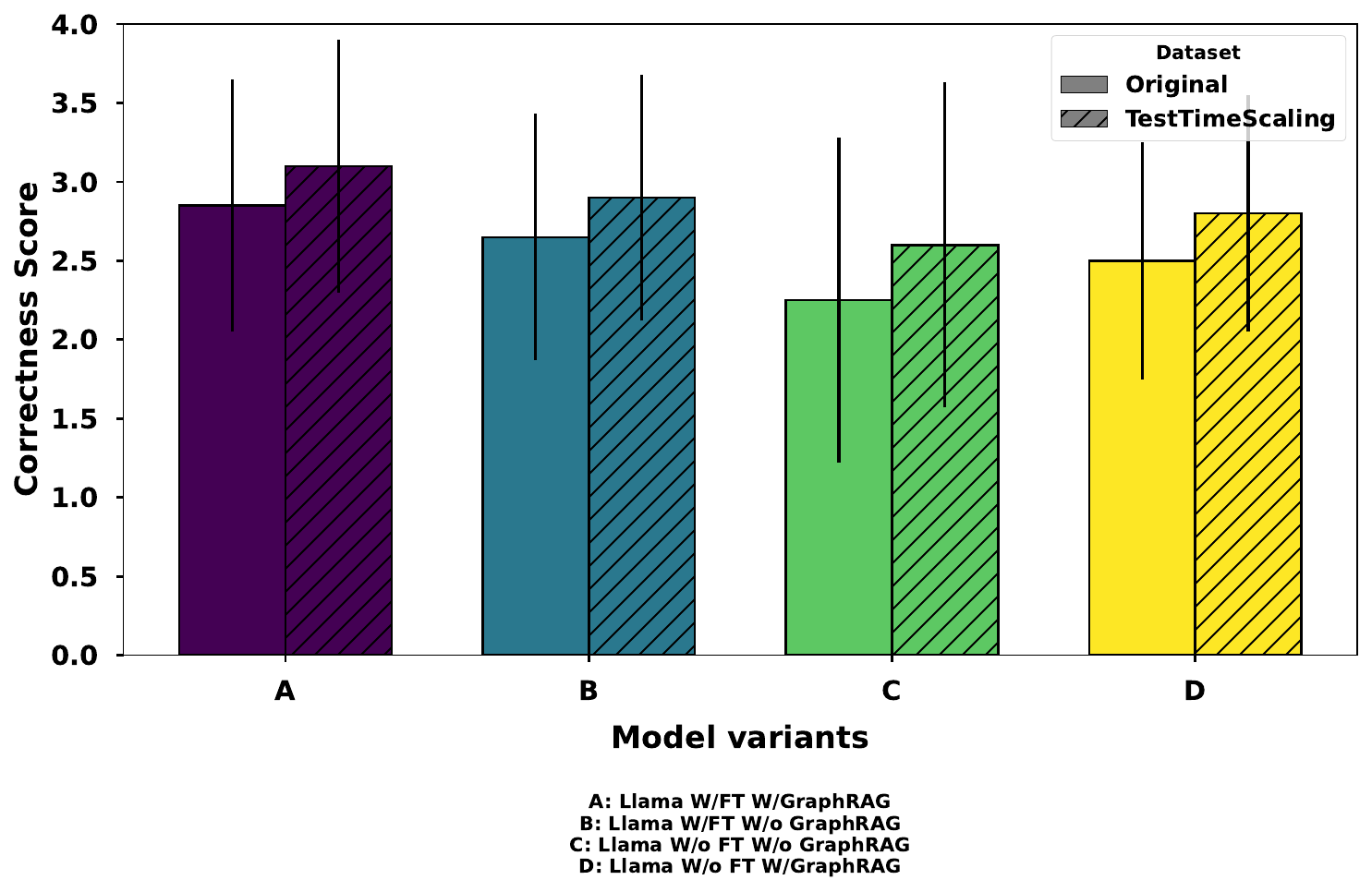}
\vspace{-1mm}
\caption{Effect of test-time inference scaling on \textbf{Correctness Score} across four Llama-3.2-1B variants, evaluated on the \textit{LogiCore-DPO}, \textit{GraphRAG-RetrievalQA}, and \textit{Factual QA} datasets. Compared to baseline decoding (‘Original’), the scaling mechanism (‘TestTimeScaling’) yields consistent and significant improvements in correctness across all configurations.}
\label{fig:tts_correctness}
\vspace{-4mm}
\end{figure}

\begin{figure}[htbp!]
\centering
\includegraphics[width=0.95\linewidth]{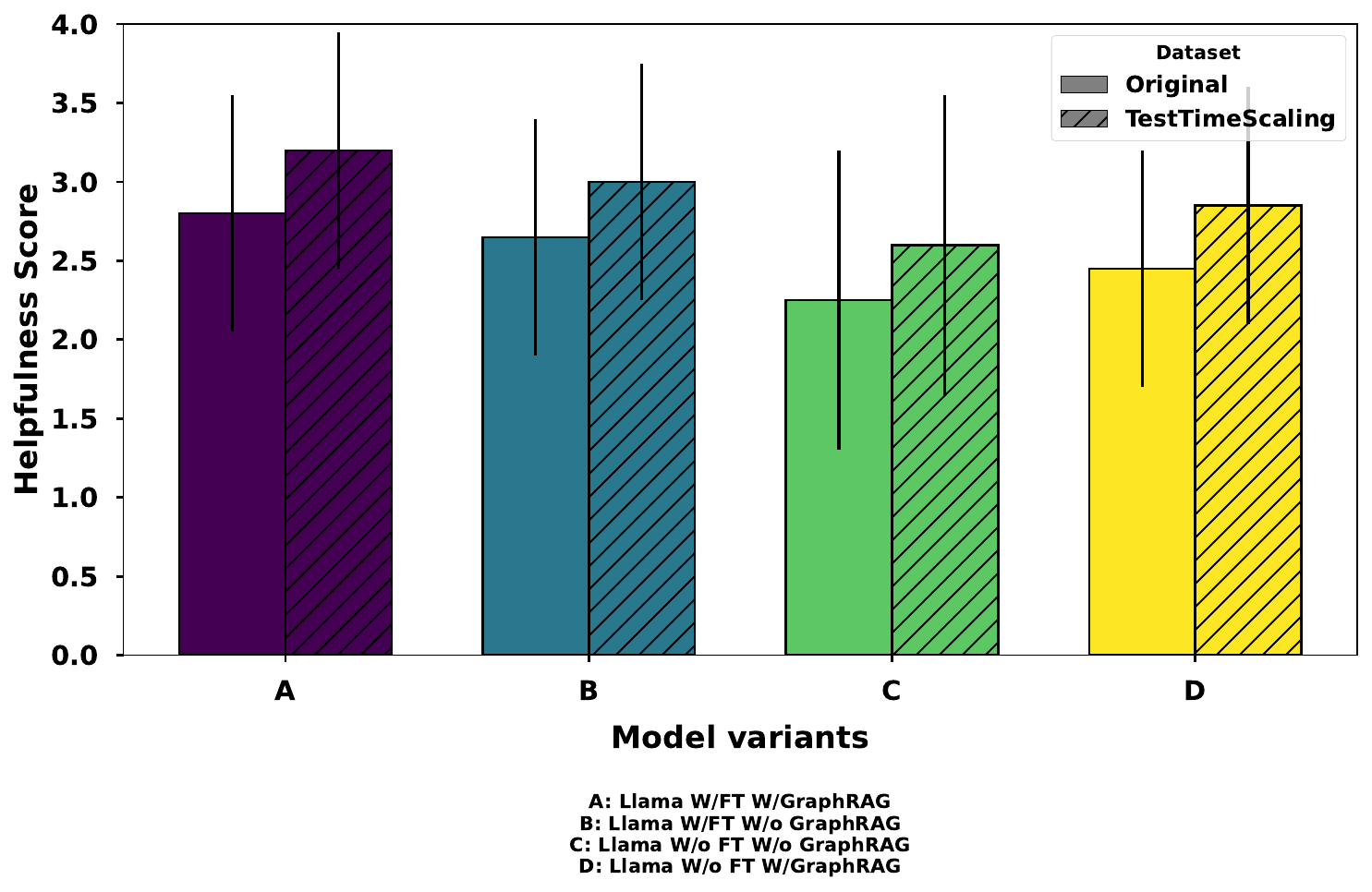}
\vspace{-1mm}
\caption{Effect of test-time inference scaling on \textbf{Helpfulness Score} across four Llama-3.2-1B variants, evaluated on the \textit{LogiCore-DPO}, \textit{GraphRAG-RetrievalQA}, and \textit{Factual QA} datasets. The scaling mechanism consistently improves helpfulness over baseline decoding across all variants.}
\label{fig:tts_helpfulness}
\vspace{-3mm}
\end{figure}

\begin{figure}[htbp!]
\centering
\includegraphics[width=0.95\linewidth]{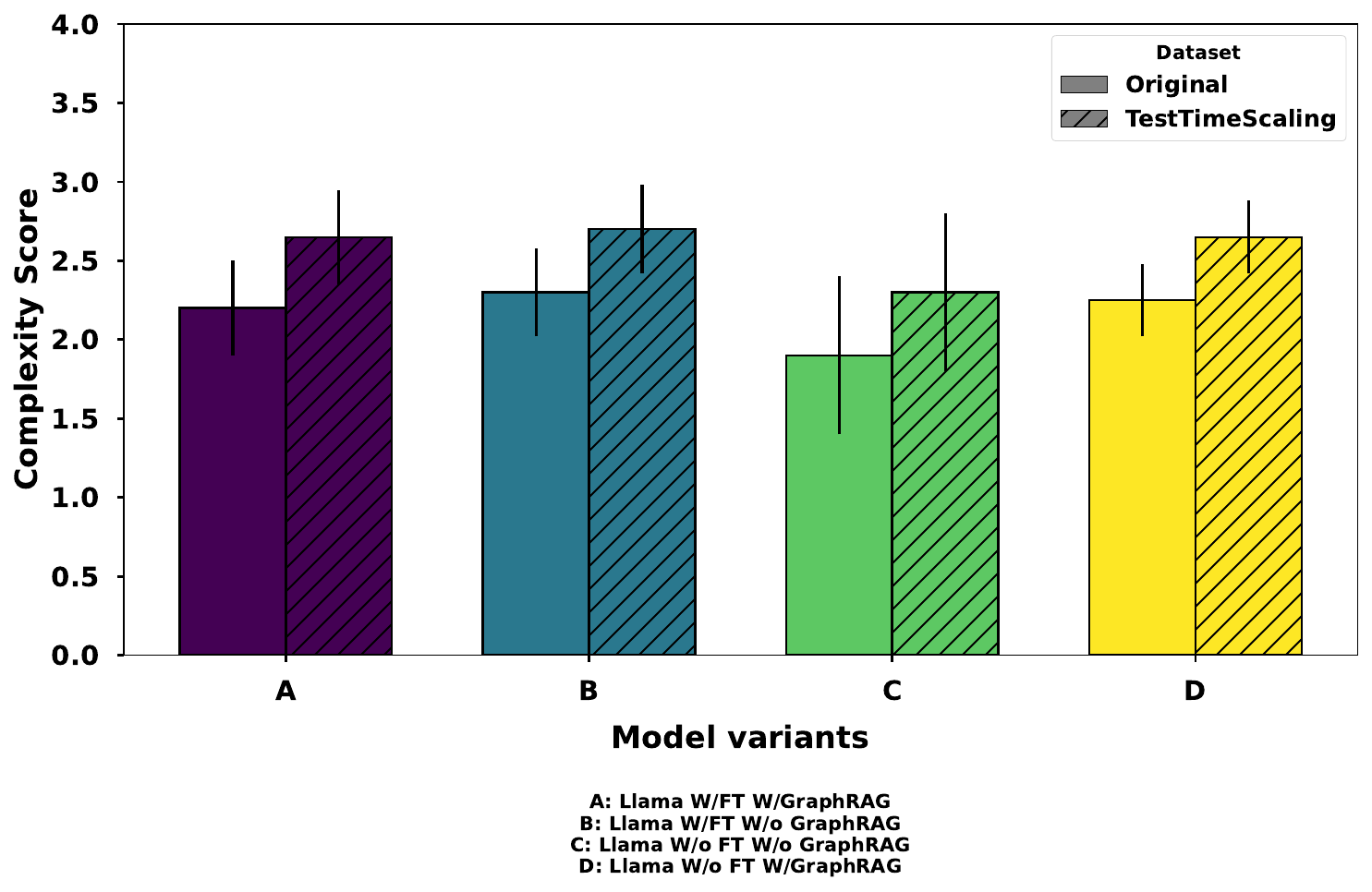}
\vspace{-1mm}
\caption{Effect of test-time inference scaling on \textbf{Complexity Score} across four Llama-3.2-1B variants, evaluated on the \textit{LogiCore-DPO}, \textit{GraphRAG-RetrievalQA}, and \textit{Factual QA} datasets. The scaling mechanism introduces a slight but consistent increase in generation complexity.}
\label{fig:tts_complexity}
\vspace{-4mm}
\end{figure}

\begin{figure}[htbp!]
\centering
\includegraphics[width=0.95\linewidth]{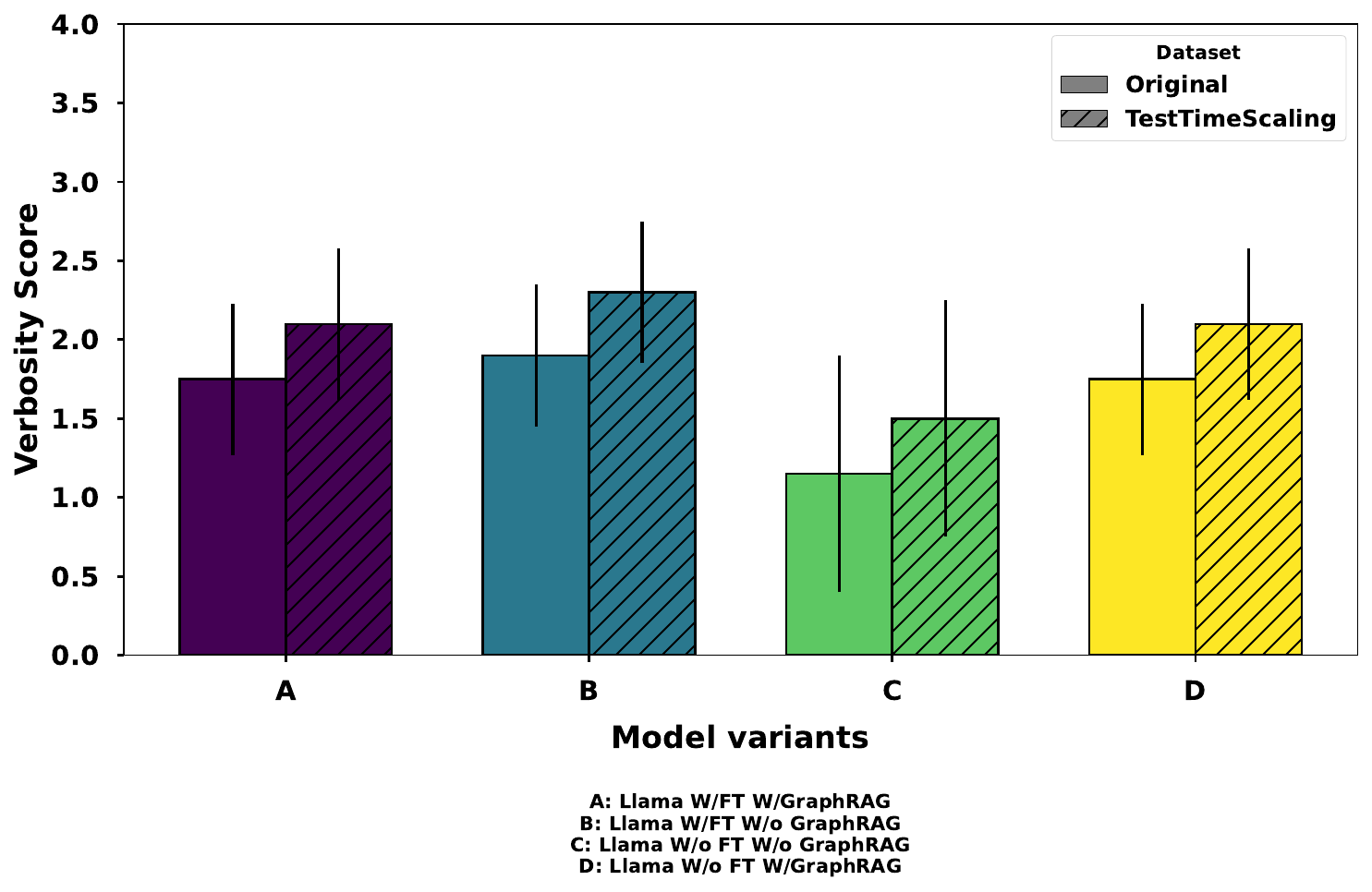}
\vspace{-1mm}
\caption{Effect of test-time inference scaling on \textbf{Verbosity Score} across four Llama-3.2-1B variants, evaluated on the \textit{LogiCore-DPO}, \textit{GraphRAG-RetrievalQA}, and \textit{Factual QA} datasets. The scaling mechanism leads to a marginal yet consistent increase in verbosity compared to baseline decoding.}
\label{fig:tts_verbosity}
\vspace{-3mm}
\end{figure}

\vspace{1mm}
where $\operatorname{TopK}(S, K)$ denotes selecting the $K$ elements with the highest values in set $S$ (corresponding here to the lowest scores after negation). Next, we will discuss about the Self-Reflection Mechanism. Following the selection of top-$K$ candidates $\mathcal{Y}{\text{top-K}}$, a self-reflection mechanism is applied to enhance reasoning robustness. In the critique phase, an auxiliary model $\phi$ (which may be identical to $\theta$ or a separately fine-tuned model) critiques each candidate $y^{(i)} \in \mathcal{Y}{\text{top-K}}$, generating a critique $c^{(i)} = \text{Critique}_\phi(y^{(i)})$ that aims to identify logical inconsistencies, missing justifications, or factual inaccuracies in the reasoning trajectory. Subsequently, in the revision phase, the model uses the critique to generate an improved sequence: $y_{\text{rev}}^{(i)} = \text{Reflect}\phi(y^{(i)}, c^{(i)})$. This yields a set of revised candidates: $\mathcal{Y}{\text{rev}} = {y_{\text{rev}}^{(i)} : y^{(i)} \in \mathcal{Y}_{\text{top-K}}}$. Next, we will discuss the final Consensus Selection. For the final aggregation, we form a consensus candidate pool combining the top-$K$ original candidates and their revisions: $\mathcal{Y}_{\text{cons}} = \mathcal{Y}_{\text{top-K}} \cup \mathcal{Y}_{\text{rev}}$. A deterministic extraction function $a(\cdot)$ is then applied to each candidate $y \in \mathcal{Y}_{\text{cons}}$ to retrieve its final proposed answer (e.g., the concluding statement), resulting in an answer set: $\mathcal{A} = \{ a(y) : y \in \mathcal{Y}_{\text{cons}} \}$. The final output $a^*$ is determined by majority vote over the extracted answers: $a^* = \operatorname{mode}(\mathcal{A})$, selecting the most frequently occurring answer among the candidates. This multi-stage inference process—combining exploration through sampling, confidence-weighted evaluation, targeted self-reflection, and robust consensus selection—significantly improves output reliability without requiring model retraining. We demonstrate the efficacy of our test-time inference scaling mechanism in significantly improving the reliability of Small Language Models (SLMs) while maintaining their original parameterization. Our evaluation employs Llama-3.2 1B model variants with a sophisticated inference pipeline combining: (1) Chain-of-Thought sampling with $N = 4$ diverse reasoning trajectories, (2) confidence-weighted entropy scoring ($\lambda = 0.5$) for uncertainty-aware candidate selection, (3) Top-K filtering ($K = 2$) to retain high-quality outputs, (4) internal self-reflection for iterative refinement, and (5) self-consistency aggregation for final predictions. The experimental framework evaluates performance across three critical benchmarks: instruction following (via Direct Preference Optimization; DPO), knowledge-intensive question answering (using Retrieval-Augmented Generation; RAG), and general question answering (through Supervised Fine-Tuning; SFT). Assessment leverages both traditional NLP metrics and fine-grained qualitative dimensions—including Correctness, Coherence, Helpfulness, Complexity, and Verbosity—with qualitative judgments provided by the Nemotron-4-340B reward model for consistent evaluation. Results demonstrate consistent improvements across standard NLP metrics—including METEOR, ROUGE variants, BERTScore, and Similarity (Figures~\ref{fig:tts_dpo_separate}--\ref{fig:tts_qa_separate})—indicating enhanced lexical and semantic alignment. Qualitative assessment reveals particularly strong gains in factual correctness (Figure~\ref{fig:tts_correctness}) and helpfulness (Figure~\ref{fig:tts_helpfulness}), while maintaining baseline coherence levels (Figure~\ref{fig:tts_coherence}). These improvements are accompanied by moderate increases in output complexity and verbosity (Figures~\ref{fig:tts_complexity} and \ref{fig:tts_verbosity}), representing an expected trade-off between generation richness and conciseness. The success stems from the multi-stage architecture synergistically combining exploratory sampling for diverse solution generation, confidence-guided filtering for high-quality candidate selection, reflective refinement for iterative improvement, and consensus-based selection for robust final predictions. This pipeline delivers markedly improved model reliability with strong gains in factual accuracy, making it well-suited for high-stakes applications where computational overhead is justified by the need for dependable performance.

% Figure for DPO comparison

% Figure for RAG comparison

% Figure for QA comparison

% Figure for Coherence Score

% Figure for Correctness Score

% Figure for Helpfulness Score

% Figure for Complexity Score

% Figure for Verbosity Score

\vspace{-3mm}  
\subsection{Related Work}  
This section reviews recent advances in data-driven PFD and PID generation, highlighting their methodologies, limitations, and gaps in industrial applicability. The Generative Flowsheet Transformer \cite{vogel2023learning} introduces a transformer-based model that autocompletes chemical process flowsheets by treating them as linear text sequences using the SFILES 2.0 notation—a structured, text-based format for representing process flow diagrams. The model is pre-trained on synthetic data and fine-tuned on real flowsheet data, with both datasets converted into SFILES 2.0 strings. These strings serve as input for learning the structural grammar of flowsheets, achieving low perplexity while enabling realistic autocompletion. However, the method relies heavily on synthetic data, which poorly reflects industrial variability, and suffers from limited real-world data, leading to unstable generalization. To address the challenge of limited data availability, the **Randomized SFILES-based Data Augmentation technique** \cite{schulze2023data} proposes a text-based augmentation method for chemical process flowsheets using SFILES 2.0 notation. This approach introduces an algorithm that applies randomized flowsheet graph traversal and template-based mutations to generate structurally varied (non-canonical) yet semantically equivalent flowsheet strings. The technique supports flowsheet-based process modeling by expanding the diversity of machine-readable training data. However, the method is limited by its dependence on the number of branching points, offering minimal augmentation for small flowsheets while risking overrepresentation of larger flowsheets. Additionally, it only introduces syntactic variations without altering functional or topological features, limiting its ability to improve generalization to structurally novel process designs. The SFILES2Seq framework \cite{hirretier2022towards} proposes a data-driven sequence-to-sequence approach for the automatic prediction of control structures, generating PIDs from PFDs. Using the SFILES 2.0 notation, both diagrams are encoded as structured text strings, enabling transformer-based translation. A T5 encoder-decoder model is trained to map PFD sequences to corresponding PID sequences, guided by a custom tokenizer that captures the syntax of unit operations and control elements. The model is first pre-trained on synthetically generated examples, created through a Markov chain-like process that assembles subprocess modules and inserts control structures based on design heuristics. It is then fine-tuned on a small real-world dataset, though performance is limited by dataset size and variability. To improve generalization, augmented SFILES 2.0 strings are generated by varying branching and control unit placements. Beam search is used during inference to produce multiple PID predictions, demonstrating that NLP models can effectively support automated control structure generation from PFDs. Despite strong performance on synthetic data, the method struggles with real-world generalization due to limited and diverse training samples. The lack of constrained decoding and oversimplified synthetic data further limits its reliability in capturing complex industrial control structures.  PID-TALK \cite{alimin2025talking} is a three-stage methodology for enabling natural language interaction with PIDs. First, PIDs are transformed into graph representations that capture both domain hierarchies and lexical interconnections among components. These graphs are then enriched with semantic labels and properties to form labeled property graphs within a knowledge graph framework. Finally, a graph-based retrieval-augmented generation (graph-RAG) approach is employed, where the high-level knowledge graph provides context for large language models, enabling efficient, context-aware querying of PID information while improving interpretability and reducing hallucinations. Despite recent innovations, these approaches exhibit critical limitations that restrict their practical applicability. Current methods are unable to autonomously generate novel industrial PFDs and PIDs, limiting their ability to support new or customized process designs. They often neglect the broader process context—such as operational objectives, feedstock-product relationships, safety constraints, and design rationales—which is essential for producing technically sound schematics. Additionally, many approaches rely heavily on inadequately curated synthetic datasets, failing to capture the complexity and variability of real-world industrial processes. The absence of rigorous simulator-backed validation further compounds these issues, as generated PFDs and PIDs are not tested for operational safety, control robustness, or engineering feasibility, posing significant risks in practical deployment.  

% %%%%%%%%%%%%%%%%%%%%%%%%%%%%%%%%%%%%%%%%%%%%%%%%%%%%%%%%%%%%%%%%%%%%%%%%%%%%%%%

\subsection{Auxiliary Results}

\subsubsection{Composite Reward Group Relative Policy Optimization (GRPO)}
We propose a modification to the standard \textbf{Group Relative Policy Optimization (GRPO)}~\cite{shao2024deepseekmath, guo2025deepseek, liu2024deepseek, lin2025cppo} algorithm for direct fine-tuning of a small-scale language model (SLM). The SLM acts as a policy network with parameters $\theta \in \Theta$, where $\Theta \subset \mathbb{R}^d$ denotes the parameter space. It implements a stochastic, autoregressive policy $\pi_\theta(y \mid x)$, mapping an input prompt $x \in \mathcal{X}$ to a generated output $y \in \mathcal{Y}$. Our goal is to optimize $\theta$ such that the model's responses better align with a ground-truth reference answer $r_x$. For each prompt $x$, we sample a group of $G$ responses:  $\mathcal{O}(x) = \{o_1, o_2, \ldots, o_G\}, \quad \text{where} \quad o_i \sim \pi_{\theta_{\text{old}}}(\cdot \mid x)$. This group-level sampling enables \textit{relative comparison} of outputs within each group, facilitating targeted policy updates. We assign a composite reward to each generated output $o$ using a weighted combination of three quality metrics:

\vspace{1mm}
\resizebox{0.985\linewidth}{!}{
\begin{minipage}{\linewidth}
\begin{equation}
r(o, r_x) = 0.3 \cdot r^{\text{rouge}}(o, r_x) + 0.2 \cdot r^{\text{length}}(o, r_x) + 0.5 \cdot r^{\text{LLM}}(o, r_x), \nonumber
\end{equation}
\end{minipage}
}

where $r^{\text{rouge}}(o, r_x)$ is the ROUGE-L F1 score between $o$ and the reference $r_x$, measuring lexical and semantic overlap. The length penalty $r^{\text{length}}(o, r_x)$ is defined as:

\vspace{-1mm}
\resizebox{0.985\linewidth}{!}{
\begin{minipage}{\linewidth}
\begin{equation}
r^{\text{length}}(o, r_x) = 
\begin{cases}
\displaystyle \frac{\min\bigl(\text{len}(o), \text{len}(r_x)\bigr)}{\max\bigl(\text{len}(o), \text{len}(r_x)\bigr)} \times 0.5, & \text{if } \text{len}(r_x) > 0, \\[1ex]
0, & \text{otherwise},
\end{cases} \nonumber
\end{equation}
\end{minipage}
}

with $\text{len}(\cdot)$ denoting token count. This term penalizes responses that deviate from the reference length, yielding values in $[0, 0.5]$. Lastly, $r^{\text{LLM}}(o, r_x)$ is a normalized score (${}\in [0, 1]$) from an auxiliary LLM evaluating the correctness of $o$ against $r_x$. For each generated output $o_i \in \mathcal{O}(x)$ where $i \in \{1,\ldots,G\}$, we compute its composite reward $r_i \triangleq r(o_i, r_x)$. To assess relative performance within the group, we normalize these rewards by calculating the sample mean:

\vspace{-2mm}
\resizebox{0.985\linewidth}{!}{
\begin{minipage}{\linewidth}
\begin{equation}
\mu_x = \frac{1}{G} \sum_{i=1}^{G} r_i \nonumber
\end{equation}
\end{minipage}
}

and the sample standard deviation:

\vspace{-1mm}
\resizebox{0.985\linewidth}{!}{
\begin{minipage}{\linewidth}
\begin{equation}
\sigma_x = \sqrt{\frac{1}{G} \sum_{i=1}^{G} (r_i - \mu_x)^2}. \nonumber
\end{equation}
\end{minipage}
}

The normalized advantage for each output $o_i$ is then computed as:

\vspace{1mm}
\resizebox{0.985\linewidth}{!}{
\begin{minipage}{\linewidth}
\begin{equation}
\widehat{A}_i = \frac{r_i - \mu_x}{\sigma_x}, \nonumber
\end{equation}
\end{minipage}
}

This converts rewards to z-scores, highlighting outputs that significantly differ from the group mean for policy updates. During fine-tuning, the SLM policy $\pi_\theta$ autoregressively generates each output $o_i = (o_{i,1}, \ldots, o_{i,T_i})$, where $T_i \triangleq |o_i|$. For each token position $t \in \{1,\ldots,T_i\}$, the policy outputs the probability $\pi_\theta(o_{i,t} \mid x, o_{i,<t})$ given prompt $x$ and preceding tokens $o_{i,<t} \triangleq (o_{i,1},\ldots,o_{i,t-1})$. To maintain training stability, we sample the group $\mathcal{O}(x)$ using the old policy $\pi_{\theta_{\text{old}}}$. For each token $o_{i,t}$ in output $o_i$, we compute the probability ratio:

\vspace{1mm}
\resizebox{0.985\linewidth}{!}{
\begin{minipage}{\linewidth}
\begin{equation}
r_{i,t}(\theta) = \frac{\pi_\theta(o_{i,t} \mid x, o_{i,<t})}{\pi_{\theta_{\text{old}}}(o_{i,t} \mid x, o_{i,<t})}. \nonumber
\end{equation}
\end{minipage}
}

Combining this ratio with the normalized advantage $\widehat{A}_i$, we define our modified GRPO objective:

\vspace{-5mm}
\resizebox{0.985\linewidth}{!}{
\begin{minipage}{\linewidth}
\begin{align}
J_{\text{GRPO}}(\theta) = \mathbb{E}_{\substack{x \sim \mathcal{X}, \\ \mathcal{O}(x) \sim \pi_{\theta_{\text{old}}}}} \Bigg[ &\frac{1}{G} \sum_{i=1}^{G} \frac{1}{|o_i|} \sum_{t=1}^{|o_i|} \min\Big( r_{i,t}(\theta) \widehat{A}_i, \nonumber \\
&\operatorname{clip}(r_{i,t}(\theta), 1 - \epsilon, 1 + \epsilon) \widehat{A}_i \Big) \Bigg] \nonumber \\
&- \beta D_{\mathrm{KL}}\big(\pi_\theta(\cdot \mid x) \big\| \pi_{\text{ref}}(\cdot \mid x)\big) \nonumber
\end{align}
\end{minipage}
}

\begin{figure}[ht!]
\vspace{-3mm}
\centering
\includegraphics[width=0.5\textwidth]{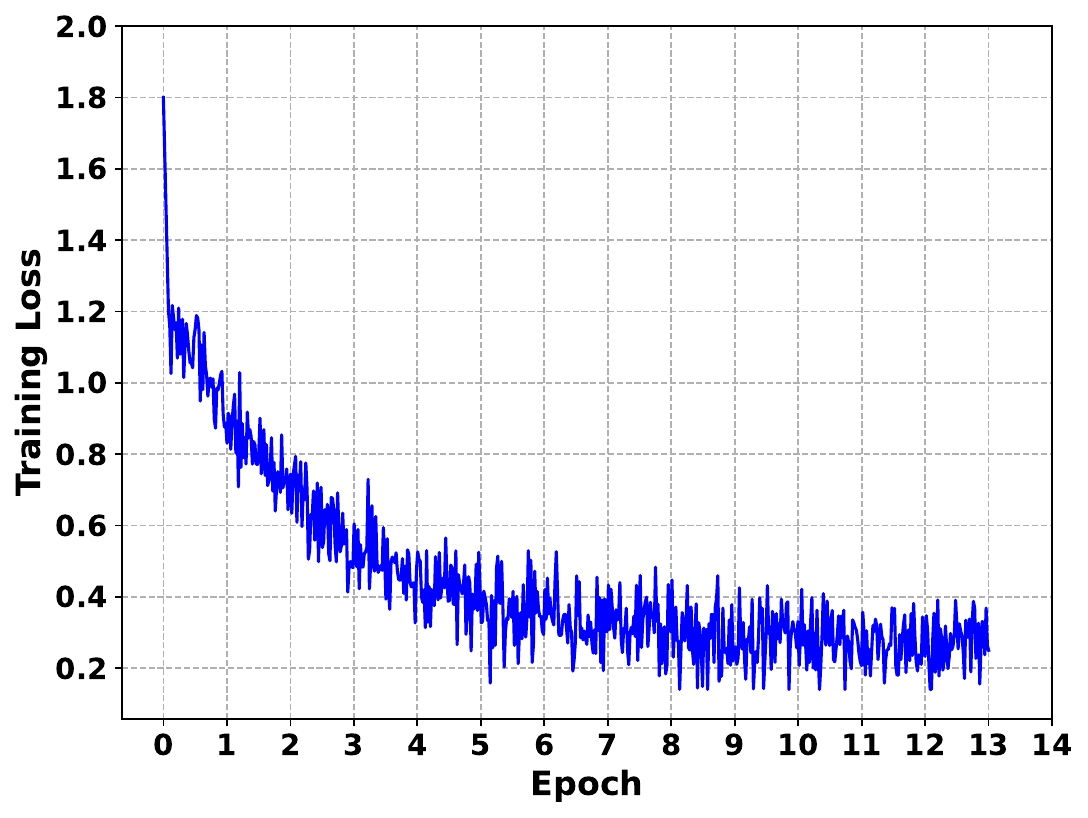}
\vspace{-2mm}
\caption{Training loss progression for Llama 3.2 1B fine-tuned with GRPO on QA datasets (\textit{Factual QA}, \textit{SynDIP}, \textit{LogiCore}), showing convergence within 10 epochs.}
\label{fig:grpo_qa_loss}
\vspace{-3mm}
\end{figure}

\begin{figure}[ht!]
\vspace{-2mm}
\centering
\includegraphics[width=0.5\textwidth]{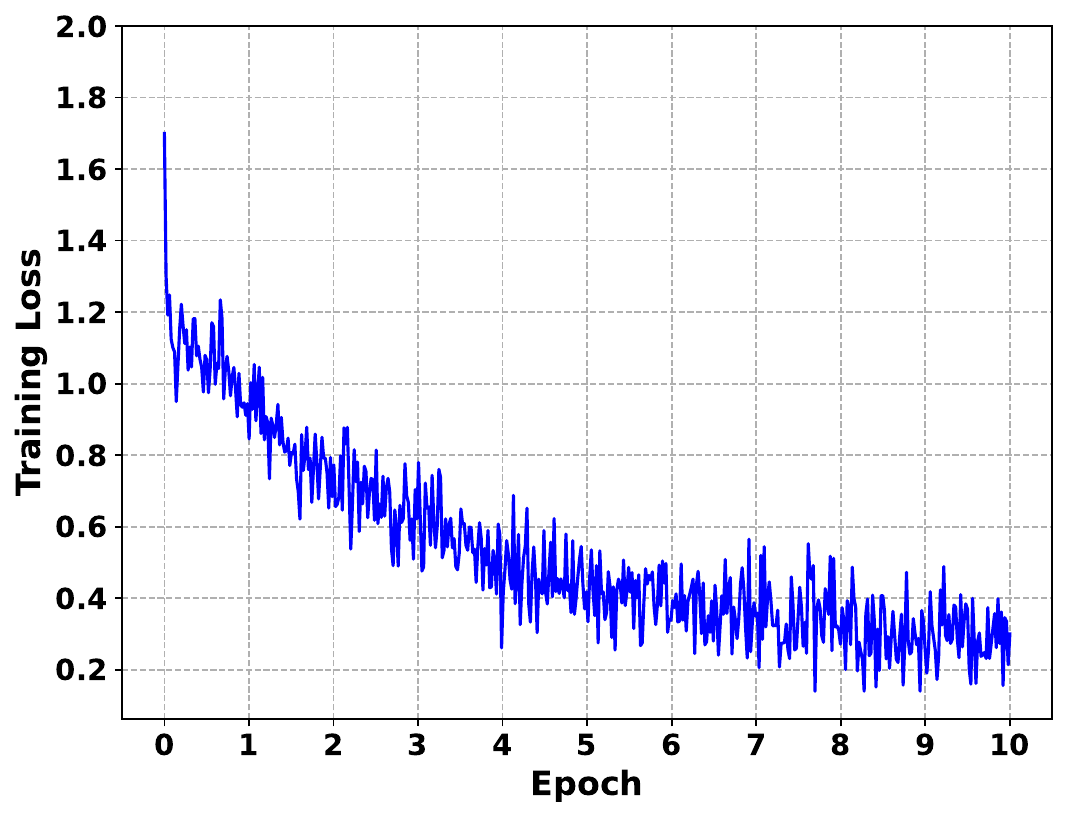}
\vspace{-3mm}
\caption{Training loss progression for Llama 3.2 1B fine-tuned with GRPO on retrieval-augmented datasets (\textit{Local RAIT}, \textit{Global RAIT}), achieving convergence in 13 epochs.}
\label{fig:grpo_rait_loss}
\end{figure}

\begin{figure}[ht!]
\vspace{-2mm}
\centering
\includegraphics[width=0.5\textwidth]{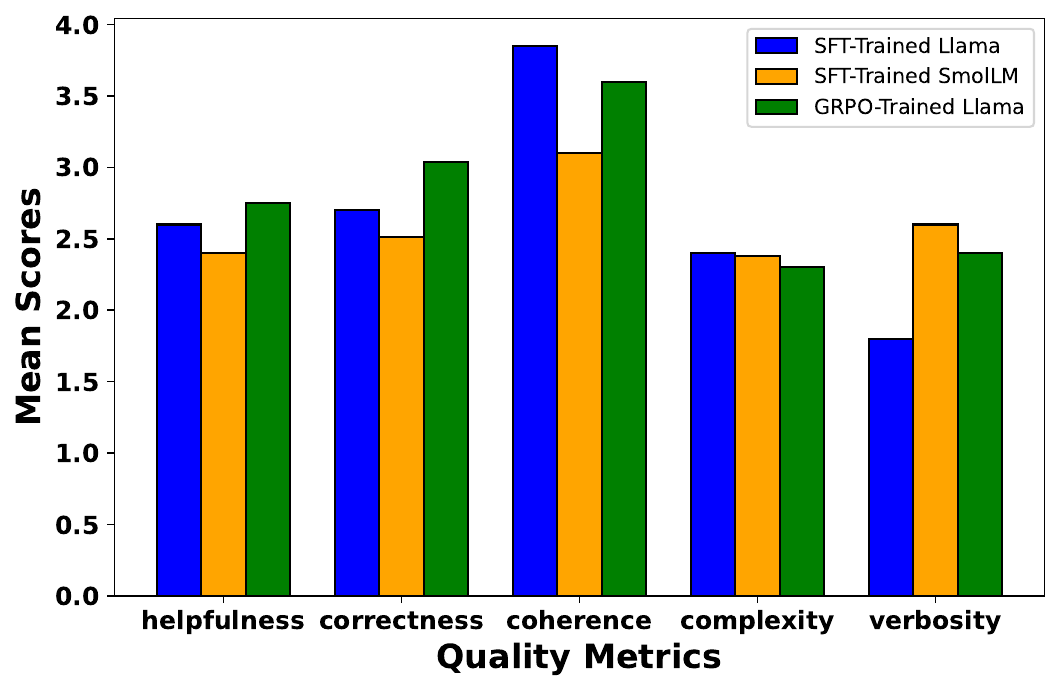}
\vspace{-3mm}
\caption{Performance comparison of GRPO and SFT fine-tuning on Llama 3.2 1B and SmolLM2-135M models, evaluated on the \textit{1.5K QA-pair} generalization benchmark. Bars show mean scores across five quality metrics: helpfulness, correctness, coherence, complexity, and verbosity.}
\label{fig:grp_1500}
\vspace{-2mm}
\end{figure}

\begin{figure}[ht!]
\vspace{-3mm}
\centering
\includegraphics[width=0.5\textwidth]{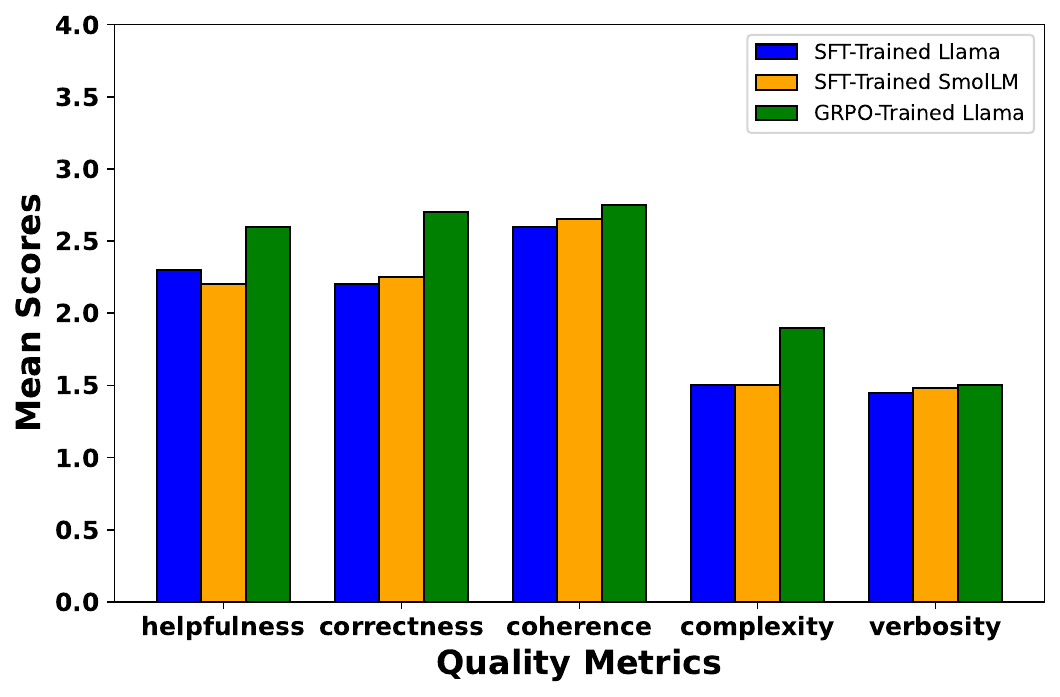}
\vspace{-2mm}
\caption{Generalization performance of GRPO vs. SFT fine-tuning on Llama 3.2 1B and SmolLM2-135M models, evaluated on the out-of-distribution \textit{ChemEval} dataset. GRPO shows clear advantages across helpfulness, correctness, coherence, and complexity, with similar verbosity across models.}
\label{fig:grp_unseen}
\vspace{-3mm}
\end{figure}

Here, $\epsilon$ clips the probability ratio $r_{i,t}(\theta)$ to $[1 - \epsilon, 1 + \epsilon]$, preventing overly aggressive policy updates. The KL divergence term $\beta D_{\mathrm{KL}}(\pi_\theta \| \pi_{\text{ref}})$ regularizes updates, where $\beta$ controls the penalty strength and $\pi_{\text{ref}}$ is typically the initial supervised fine-tuned model. This constraint ensures the policy doesn’t deviate excessively from the reference, avoiding catastrophic forgetting of previously learned knowledge. The fine-tuning procedure iterates through the following steps. For each input prompt $x$, we first sample a group of $G$ outputs $\mathcal{O}(x) = \{o_1, \dots, o_G\}$ independently from the old policy $\pi_{\theta_{\text{old}}}$. Next, we compute composite rewards $r(o_i, r_x)$ for each output $o_i$ using our weighted combination of ROUGE, length, and LLM-based metrics. These rewards are normalized within the group via mean $\mu_x$ and standard deviation $\sigma_x$ calculations, producing relative advantage scores $\widehat{A}_i$. For every token $o_{i,t}$ in each generated output, we compute probability ratios $r_{i,t}(\theta) = \pi_\theta(o_{i,t} \mid x, o_{i,<t}) / \pi_{\theta_{\text{old}}}(o_{i,t} \mid x, o_{i,<t})$ and construct the clipped surrogate objective $J_{\text{GRPO}}(\theta)$. The policy parameters $\theta$ are updated via gradient ascent on this objective, followed by synchronizing the old policy ($\theta_{\text{old}} \leftarrow \theta$) for the next iteration. This process holistically improves response quality by combining multiple reward metrics. We use gradient ascent because GRPO maximizes the reward objective $J_{\text{GRPO}}(\theta)$, unlike supervised learning which minimizes losses. The update $\theta \leftarrow \theta + \alpha \nabla_\theta J_{\text{GRPO}}(\theta)$ is mathematically equivalent to descent on $-J_{\text{GRPO}}(\theta)$. Our modified GRPO algorithm eliminates the need for a separate value network through three key mechanisms: (1) computing composite rewards for each output, (2) normalizing these rewards within each group to obtain relative advantages, and (3) performing direct policy optimization via token-level updates. The SLM $\pi_\theta$ thereby achieves efficient, end-to-end reinforcement learning that enhances performance while preserving generation diversity, all within a computationally lightweight framework. Figures \ref{fig:grpo_qa_loss} and \ref{fig:grpo_rait_loss} present the training loss trajectories for a Llama 3.2 1B model fine-tuned using Group Relative Policy Optimization (GRPO) on two distinct synthetic dataset categories. Figure \ref{fig:grpo_qa_loss} displays results for QA-style datasets (\textit{Factual QA}, \textit{SynDIP}, and \textit{LogiCore}), which enhance domain knowledge and reasoning for PFD/PID interpretation. Figure \ref{fig:grpo_rait_loss} shows corresponding results for retrieval-augmented instruction datasets (\textit{Local RAIT} and \textit{Global RAIT}), designed to ground responses in retrieved contextual information. Both figures demonstrate consistent convergence patterns: a rapid initial loss reduction followed by gradual stabilization over approximately 10 epochs for QA datasets and 13 epochs for RAIT datasets. These results confirm GRPO's effectiveness in optimizing language models for specialized chemical process engineering tasks. Figures \ref{fig:grp_1500} and \ref{fig:grp_unseen} compare the performance of Supervised Fine-Tuning (SFT) and Composite Reward Group Relative Policy Optimization (GRPO) applied to the Llama 3.2 1B and SmolLM2-135M models across five quality dimensions, as evaluated by a reward model. On the \textit{1.5K QA-pair} generalization benchmark (Figure \ref{fig:grp_1500}), the GRPO-trained Llama 3.2 1B demonstrates superior performance in helpfulness and correctness, while its SFT-trained counterpart achieves the highest coherence. In contrast, when evaluated on the out-of-distribution \textit{ChemEval} dataset (Figure \ref{fig:grp_unseen})—designed to test generalization to unseen chemical processes—the GRPO-trained Llama 3.2 1B consistently outperforms both the SFT-trained Llama 3.2 1B and the SFT-trained SmolLM2-135M across helpfulness, correctness, coherence, and complexity, while all models show comparable verbosity. These results highlight GRPO’s advantage in producing more robust and accurate model behavior on novel chemical tasks compared to standard SFT.

% %%%%%%%%%%%%%%%%%%%%%%%%%%%%%%%%%%%%%%%%%%%%%%%%%%%%%%%%%%%%%%%%%%%%%%%%%%%%%%%

\subsubsection{t-SNE/PCA Analysis of Semantic Structure in LLMs vs. Web-Derived Process Descriptions}
\begin{figure}[ht!]
\vspace{-4mm}
\centering
\includegraphics[width=85mm]{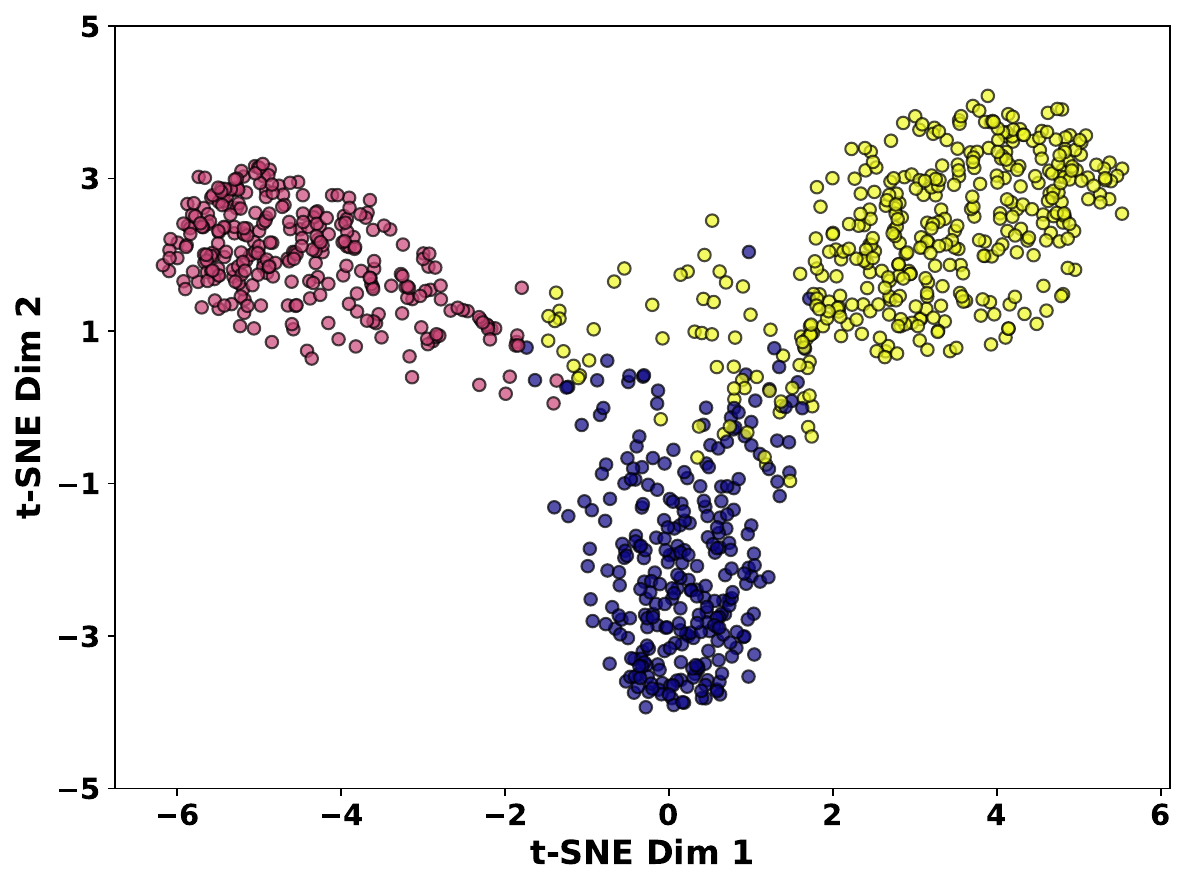}
\vspace{-2mm}
\caption{t-SNE visualization of GPT-4o-generated process embeddings (\textit{SynDIP} dataset) from the ChemAtlas corpus. Well-separated, compact clusters demonstrate high inter-chemical consistency in PFD/PID descriptions.}
\label{fig:tsne_gpt}
\vspace{-1mm}
\end{figure}

\begin{figure}[ht!]
\vspace{-2mm}
\centering
\includegraphics[width=85mm]{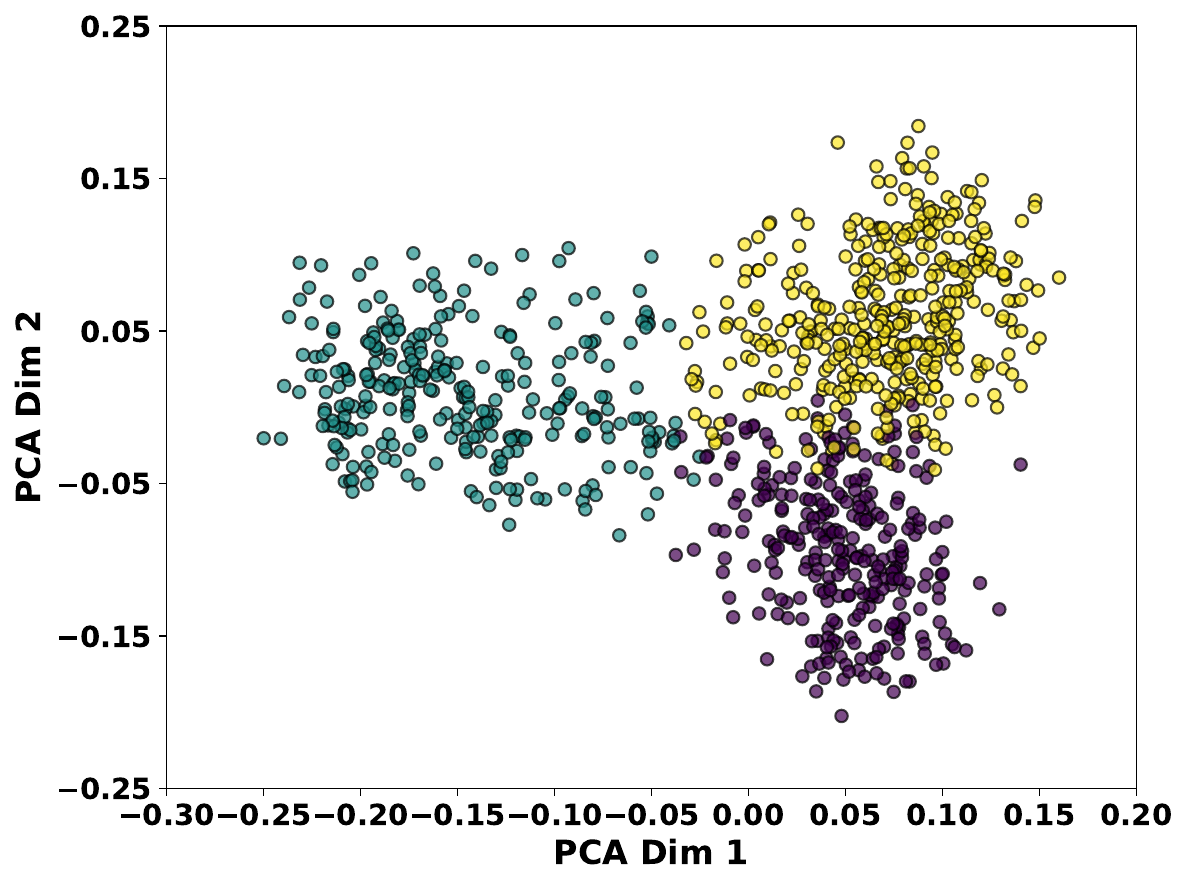}
\vspace{-2mm}
\caption{PCA visualization of GPT-4o-generated process embeddings from the ChemAtlas corpus. Tight clustering in the first two principal components reflects high semantic consistency and strong domain alignment across chemical production pathways.}
\label{fig:pca_gpt}
\vspace{-1mm}
\end{figure}

\begin{figure}[ht!]
\vspace{-3mm}
\centering
\includegraphics[width=85mm]{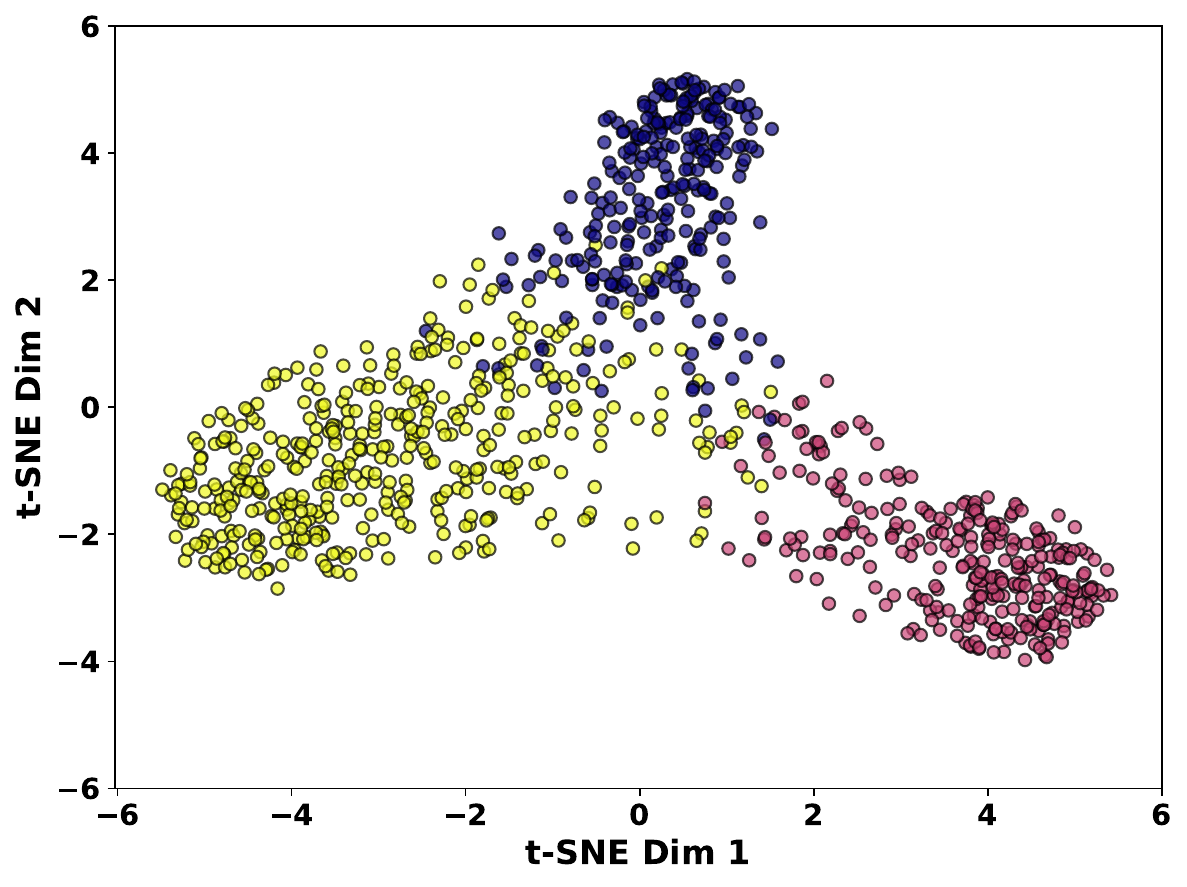}
\vspace{-2mm}
\caption{t-SNE visualization of Claude-3-Haiku-generated process flow and instrumentation description embeddings from the ChemAtlas corpus. Distinct clusters reveal semantic relationships in the embedding space, showing moderate separation. This indicates improved inter-chemical consistency and more stable intra-chemical representations compared to web-retrieved data.}
\label{fig:tsne_haiku}
\vspace{-1mm}
\end{figure}

\begin{figure}[ht!]
\vspace{-2mm}
\centering
\includegraphics[width=85mm]{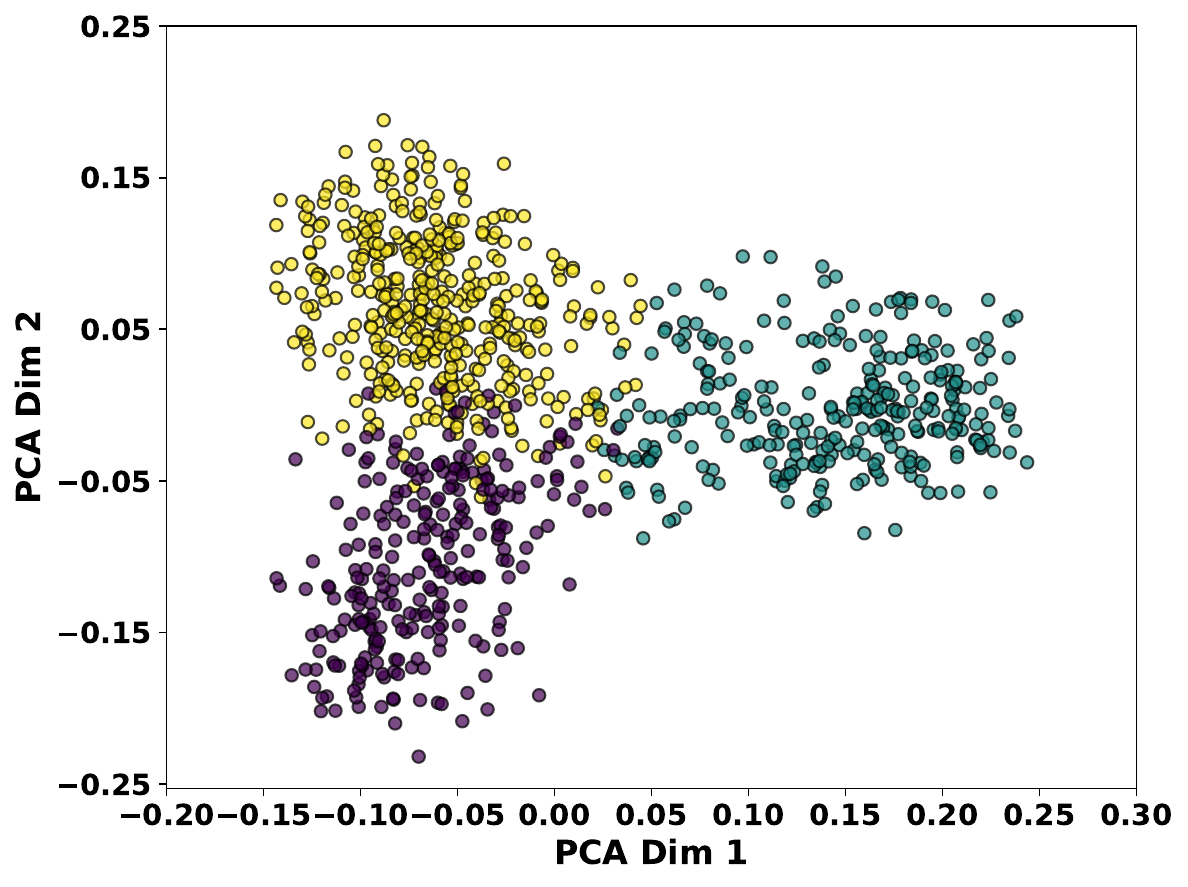}
\vspace{-1mm}
\caption{PCA visualization of Claude-3-Haiku-generated process description embeddings from the ChemAtlas corpus (first two principal components). Moderate clustering quality indicates better structural consistency and improved grouping of chemically similar production processes compared to web-sourced data.}
\label{fig:pca_haiku}
\vspace{-3mm}
\end{figure}

\begin{figure}[ht!]
\vspace{-3mm} 
\centering
\includegraphics[width=85mm]{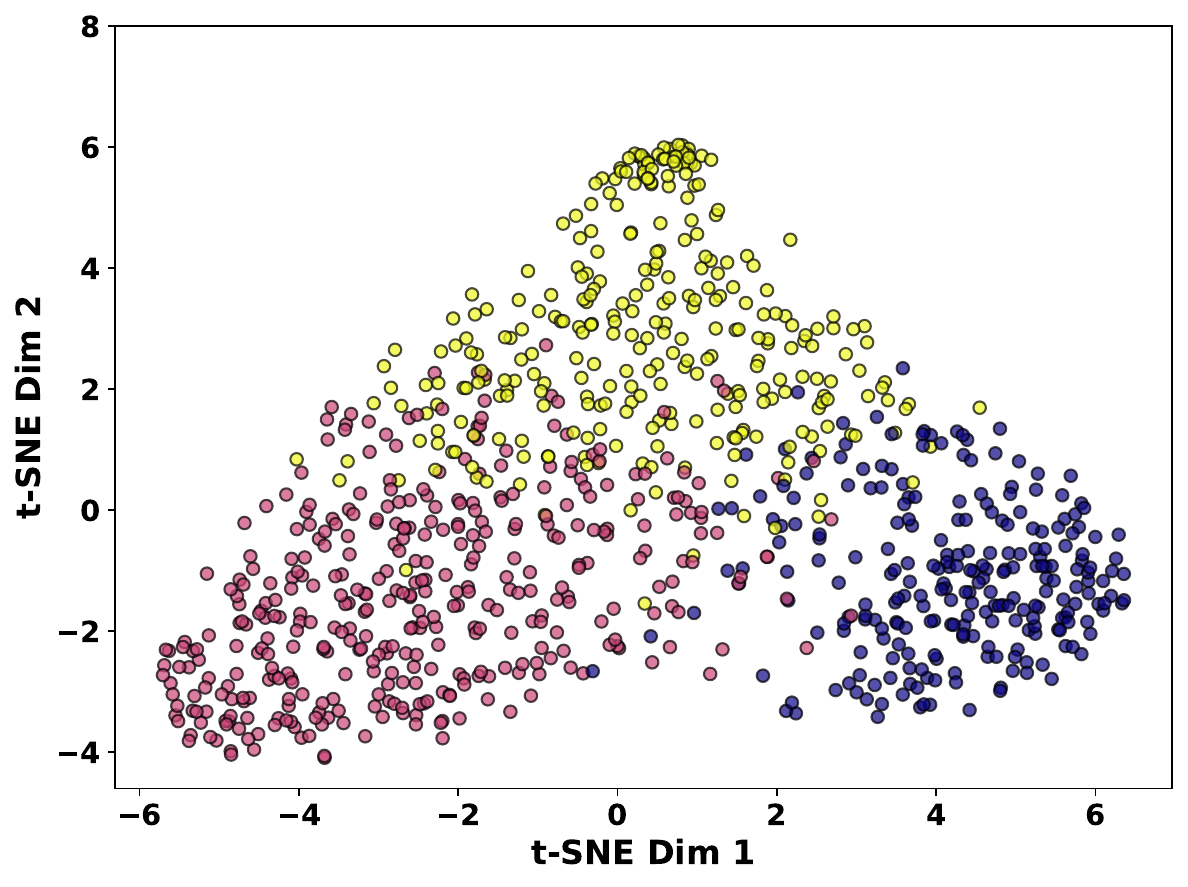}
\vspace{-1mm}
\caption{t-SNE visualization of web-retrieved process description embeddings from the ChemAtlas corpus. Diffuse, overlapping cluster formations indicate weaker inter-chemical consistency and lower structural coherence compared to LLM-generated data.}
\label{fig:tsne_internet}
\vspace{-2mm}
\end{figure}

\begin{figure}[ht!]
\vspace{-2mm} 
\centering
\includegraphics[width=85mm]{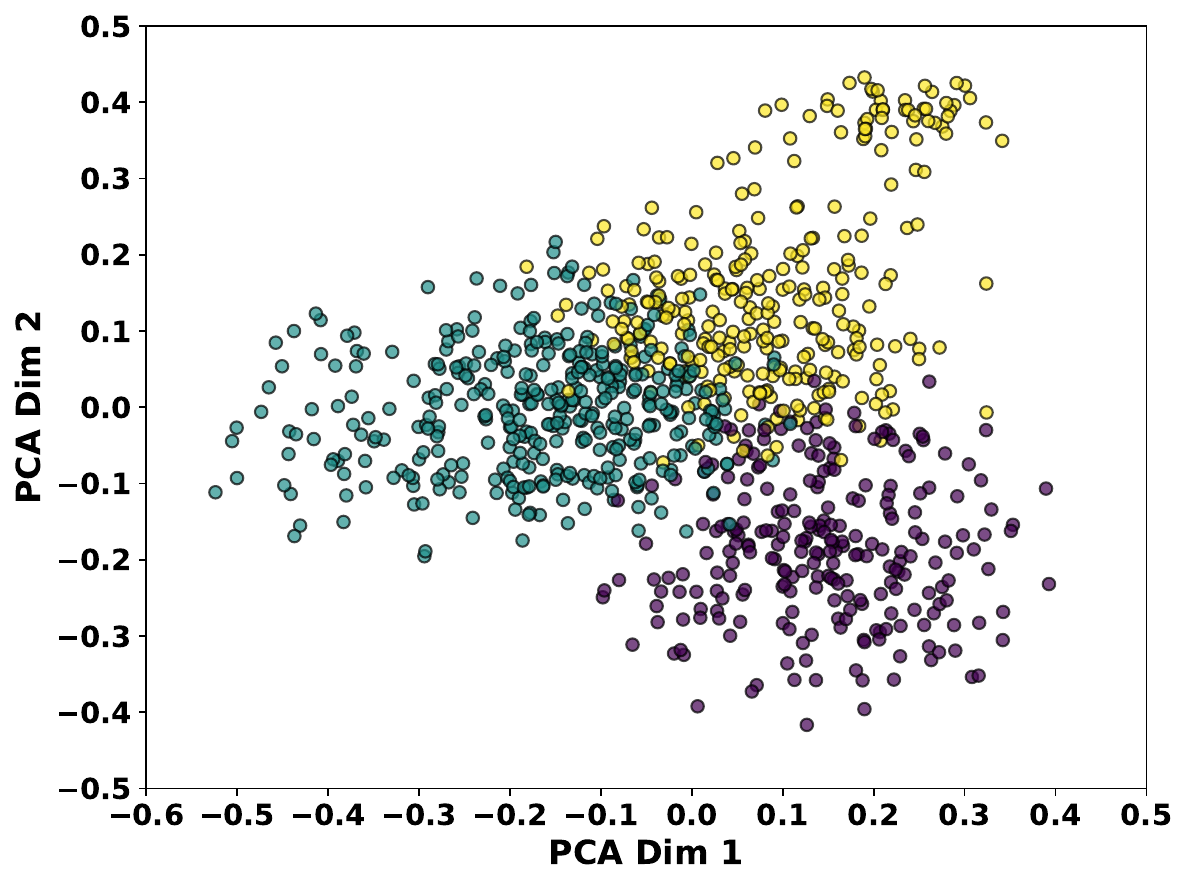}
\vspace{-1mm} 
\caption{PCA visualization of web-retrieved process description embeddings from the ChemAtlas corpus (primary variance directions). Loosely distributed embeddings suggest weaker structural coherence and less distinct process groupings compared to synthetic sources.}
\label{fig:pca_internet}
\vspace{-3mm}
\end{figure}

\begin{figure}[ht!]
\vspace{-0mm}
\centering
\includegraphics[width=85mm]{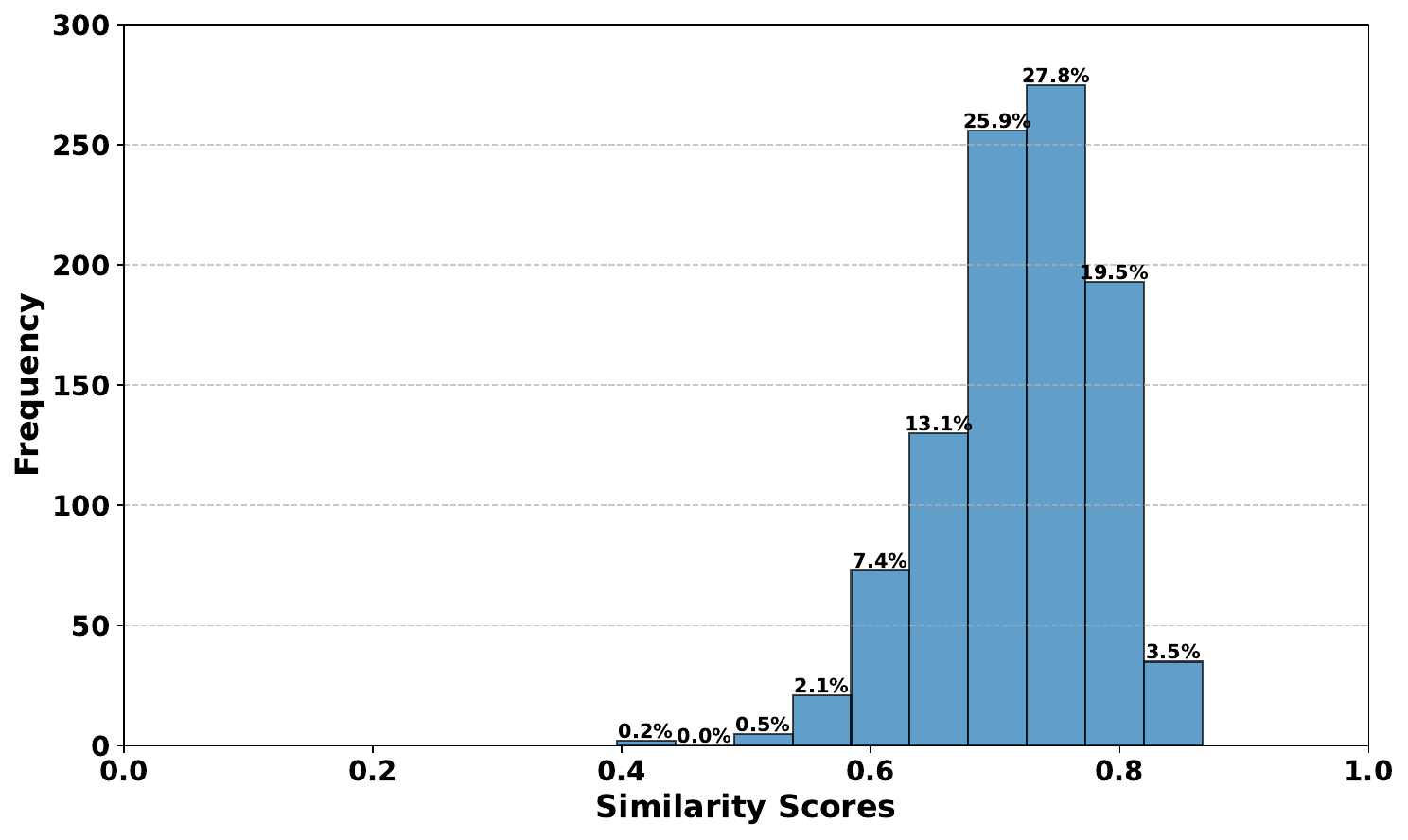}
\caption{Cosine similarity distribution between GPT-4o and Claude-3-Haiku process description embeddings. The 0.7--0.8 peak reflects strong semantic agreement and structural coherence in PFD/PID representations.}
\label{fig:gpt_vs_haiku}
\vspace{-1mm}
\end{figure}

\begin{figure}[ht!]
\vspace{-1mm}
\centering
\includegraphics[width=85mm]{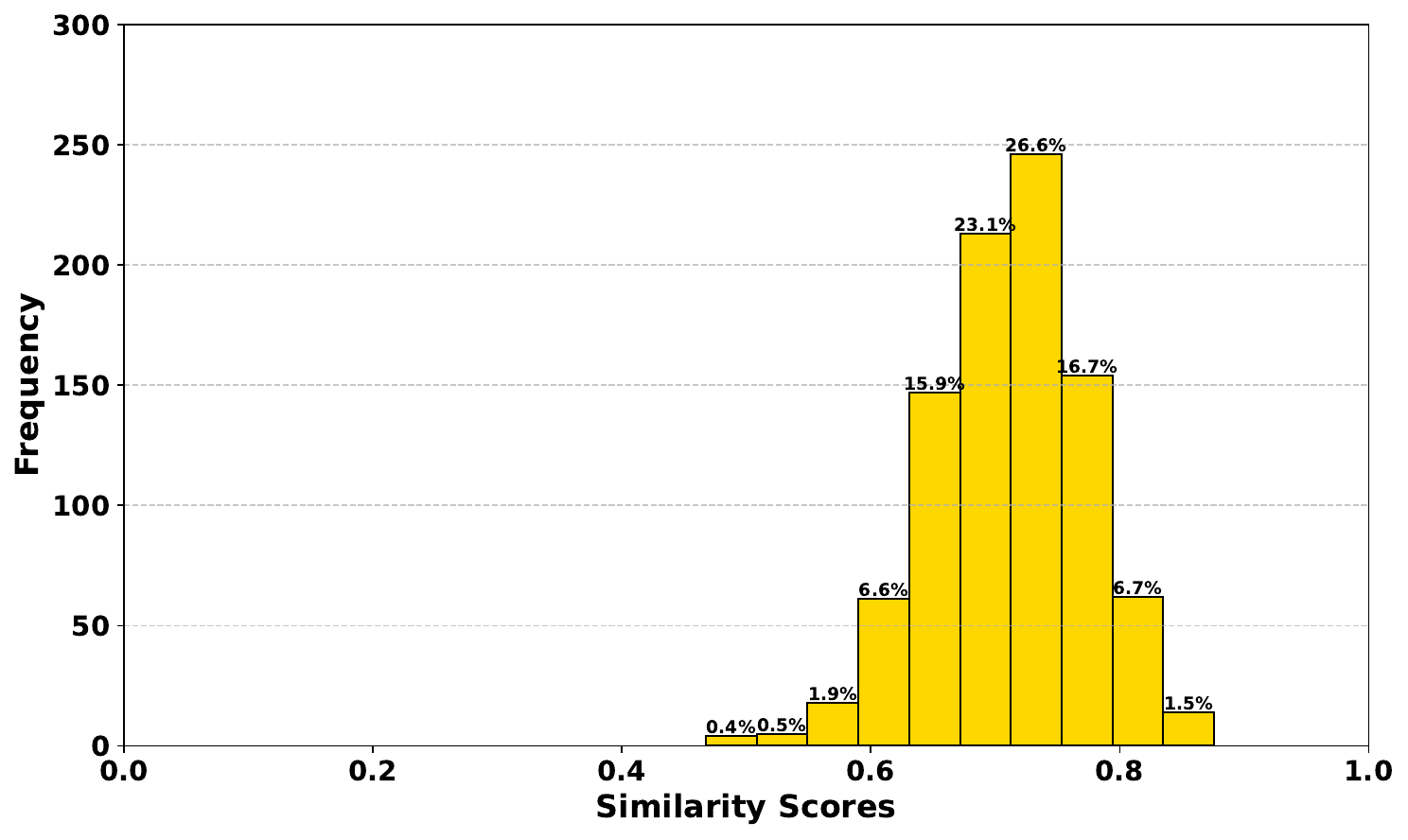}
\caption{Cosine similarity distribution between GPT-4o-generated and web-retrieved process embeddings. The broader 0.6--0.7 peak indicates moderate alignment with greater variability than Haiku-generated content.}
\label{fig:gpt_vs_internet}
\vspace{-1mm}
\end{figure}

\begin{figure}[ht!]
\vspace{-1mm}
\centering
\includegraphics[width=85mm]{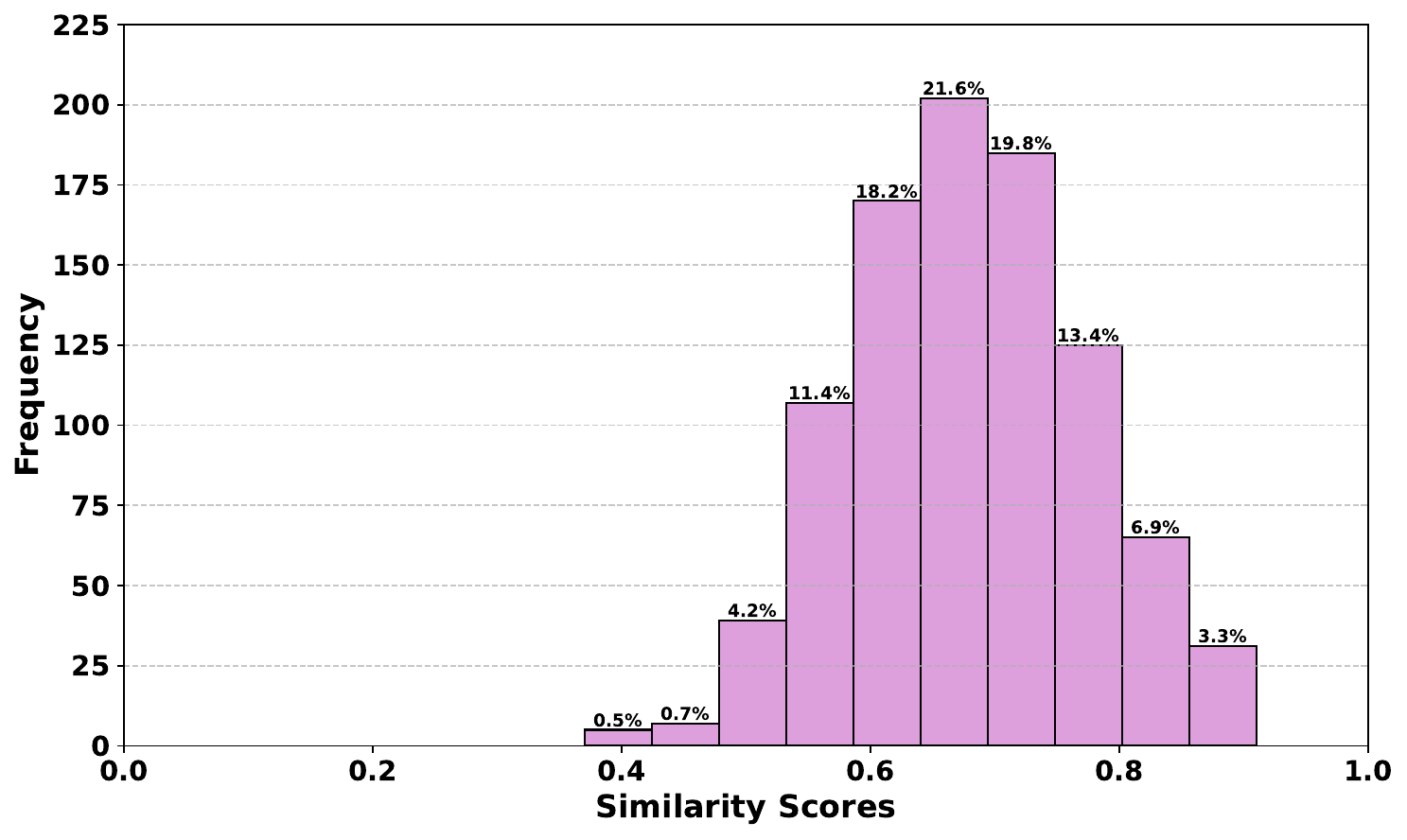}
\caption{Cosine similarity distribution between Claude-3-Haiku and web-retrieved process embeddings. The diffuse 0.6--0.7 distribution suggests weaker alignment than GPT-4o-generated representations.}
\label{fig:haiku_vs_internet}
\vspace{-1mm}
\end{figure}

We performed t-SNE and PCA visualizations to analyze the clustering behavior of process flow and instrumentation text embeddings derived from structured language model outputs (GPT-4o, Claude Haiku) and agentic web-retrieved ChemAtlas corpus data. These projections quantify inter-chemical consistency (semantic similarity of process descriptions across related substances) and intra-chemical coherence (semantic similarity across multiple descriptions of the same chemical, per LLM and web-retrieved data), revealing how chemically analogous production processes group in embedding space. Differences are illustrated using OpenAI's text-embedding-3-small embeddings~\cite{openai2024textembedding3}, which encode latent structural relationships and semantic similarities among chemical processes. For GPT-4o-generated outputs, Figures~\ref{fig:tsne_gpt} and~\ref{fig:pca_gpt} display tight, well-separated clusters, indicating strong semantic alignment among chemicals with analogous synthesis pathways, equipment types, or control strategies. Descriptions of related chemical processes—such as those sharing similar unit operations or instrumentation—are embedded proximally, while distinct processes remain clearly differentiated. In contrast, Haiku-generated outputs (Figures~\ref{fig:tsne_haiku} and~\ref{fig:pca_haiku}) exhibit moderately compact clusters, reflecting consistent grouping of chemically similar processes with enhanced structural fidelity compared to web-derived data. Conversely, web-retrieved content (Figures~\ref{fig:tsne_internet} and~\ref{fig:pca_internet}) shows diffuse, overlapping clusters, reflecting greater variability in process descriptions from heterogeneous sources. The t-SNE and PCA plots of web-retrieved process flow and instrumentation descriptions reveal a combination of overlapping and distinct clusters, demonstrating partial inter-chemical consistency. Although some chemical processes form well-defined groupings, the overall dispersion highlights structural diversity and semantic variability inherent in uncurated web content. These clustering patterns enable few-shot prompting by identifying semantically similar chemical processes, allowing language models to transfer structural knowledge—including unit operation sequences, flow configurations, and control logic—from established processes to novel chemical production scenarios. This capability can be further enhanced through teacher-student transfer learning. Larger models initially learn to recognize and leverage these semantic clusters, then distill this knowledge into smaller, more efficient language models. By retrieving industrial production processes from chemically similar neighbors within the same cluster, even compact models can generate accurate, contextually grounded process descriptions for previously unseen chemicals—requiring only minimal task-specific supervision. Overall, the PCA and t-SNE visualizations (Figures~\ref{fig:tsne_gpt}--\ref{fig:pca_internet}) reveal that LLM-generated structured outputs produce tighter clustering with higher semantic consistency and clearer inter-chemical separation compared to web-derived content, which exhibits noisier, less discriminative patterns. The similarity score distributions between text embeddings (Figures~\ref{fig:gpt_vs_haiku}--\ref{fig:haiku_vs_internet}) further illustrate these differences. GPT-4o and Claude-3-Haiku show the strongest alignment (Figure~\ref{fig:gpt_vs_haiku}), peaking at 0.7–0.8, indicating robust semantic consistency in chemical process representations. While GPT-4o also aligns with web-retrieved data (Figure~\ref{fig:gpt_vs_internet}), the similarity scores peak at a lower range (0.6–0.7), reflecting greater variability and reduced structural coherence. Haiku-web comparisons (Figure~\ref{fig:haiku_vs_internet}) follow a similar but more dispersed trend, with weaker overall alignment. These results demonstrate that while web content shows partial semantic overlap, LLM-generated descriptions exhibit significantly stronger internal consistency. The higher inter-model similarity underscores the reliability of synthetic outputs in representing chemical processes compared to unstructured web sources.

%%%%%%%%%%%%%%%%%%%%%%%%%%%%%%%%%%%%%Tsne PCA plots%%%%%%%%%%%%%%%%%%%%%%%%%%%%%%%%%%%%%

%%%%%%%%%%%%%%%%%%%%%%%%%%%%%%%%%%%% Similarity Plots %%%%%%%%%%%%%%%%%%%%%%%%%%%%%%%%%%%%%%

\vspace{-2mm}
\subsection{KV Caching and Paged Attention}
We implement a critical optimization technique to enhance the memory efficiency and computational throughput of fine-tuned SLMs during autoregressive decoding. In autoregressive transformer decoding, at each step $i$, the model processes previously available tokens—comprising (1) the original prompt tokens $\{x_1, \dots, x_m\}$ and (2) the generated tokens up to that point $\{x_{m+1}, \dots, x_{i-1}\}$—and computes a query vector $q_i \in \mathbb{R}^d$. This query attends to all previously processed tokens via their cached key vectors $k_j \in \mathbb{R}^d$ and value vectors $v_j \in \mathbb{R}^d$, where $j = 1, \dots, i-1$. The attention mechanism computes a weighted sum over the values based on query-key interactions:

\vspace{1mm}
\resizebox{0.985\linewidth}{!}{
\begin{minipage}{\linewidth}
\begin{equation}
\text{Attention}(q_i, K, V) = \sum_{j=1}^{i-1} \text{softmax}\left( \frac{q_i^\top k_j}{\sqrt{d}} \right) v_j \nonumber
\end{equation}
\end{minipage}
}

Here, $K = [k_1, \dots, k_{i-1}] \in \mathbb{R}^{(i-1) \times d}$ and $V = [v_1, \dots, v_{i-1}] \in \mathbb{R}^{(i-1) \times d}$ denote the cached key-value (KV) matrices for all previously processed tokens. The memory contiguity issue arises because the logical KV cache expands dynamically during decoding, necessitating storage of $(i-1) \times d$-dimensional matrices per layer and head at each step $i$. The linearly growing KV cache in standard autoregressive attention consumes significant memory, causing fragmentation and restricting achievable batch sizes. Coupled with its quadratic computational complexity, this substantially reduces overall throughput. Conventionally, the KV cache is stored contiguously, requiring pre-allocation of a fixed-size buffer for the maximum sequence length $L_{\text{max}}$ per sequence to avoid expensive reallocations. This approach exhibits inefficiency due to variable sequence lengths and dynamic growth. It induces internal fragmentation where allocated memory remains underutilized when $L \ll L_{\text{max}}$. More critically, it causes external fragmentation: concurrent sequences each occupy a contiguous block, and asynchronous completion creates variably-sized gaps between active allocations. GPU memory evolves into a discontiguous layout of allocated and free regions. Even with sufficient aggregate free memory, non-contiguous segmentation may prevent allocation of large contiguous blocks. Reallocation for sequences exceeding $L_{\text{max}}$ imposes substantial $O(L)$ time and memory overhead. These inefficiencies reduce maximum viable batch sizes and degrade serving throughput. To address these memory inefficiencies, PagedAttention~\cite{kwon2023efficient, rehg2024kv, prabhu2024vattention} adapts the virtual memory paging paradigm from operating systems. The system replaces contiguous GPU memory allocations with a block-based KV cache management strategy, partitioning each sequence's key-value cache into fixed-size blocks storing $B$ consecutive tokens. We formally define the $j$-th KV block as:

\vspace{1mm}  
\resizebox{0.985\linewidth}{!}{  
\begin{minipage}{\linewidth}  
\begin{align}
K_j &= [k_{(j-1)B+1}, \dots, k_{jB}] \in \mathbb{R}^{B \times d}, \nonumber \\
V_j &= [v_{(j-1)B+1}, \dots, v_{jB}] \in \mathbb{R}^{B \times d} \nonumber
\end{align}  
\end{minipage}  
}

\vspace{1mm}
The architecture's innovation centers on per-sequence block tables that map logical block indices to physical memory locations. This indirection enables three critical features: (1) non-contiguous storage where blocks occupy arbitrary GPU memory addresses, (2) the system only allocates physical memory for a block when that specific block is actually needed for computation (a "cache miss"), rather than reserving all memory upfront, and (3) memory sharing where multiple sequences reference identical blocks (particularly beneficial for shared prompt prefixes). Attention computation reformulates as a block-wise operation. For token position $i$, the output $o_i$ becomes:

\vspace{-1mm}
\resizebox{0.985\linewidth}{!}{
\begin{minipage}{\linewidth}
\begin{equation}
o_i = \sum_{j=1}^{\lceil i/B \rceil} \text{softmax}\left( \frac{q_i^\top K_j}{\sqrt{d}} \right) V_j \nonumber
\end{equation}
\end{minipage}
}

\vspace{1mm}
The softmax operation maintains mathematical equivalence with standard attention through global normalization across all blocks. Each block contributes a score matrix $A_{ij} = q_i^\top K_j / \sqrt{d} \in \mathbb{R}^B$, with the implementation optimizing performance through (i) efficient grouped memory reads (coalescing), (ii) predictive loading of upcoming data blocks (prefetching), and (iii) thread-safe block allocation (atomic resolution). This design eliminates internal fragmentation via fixed $B$-sized blocks and removes external fragmentation through non-contiguous allocation, while copy-on-write semantics preserve memory sharing benefits. The result is significantly improved memory utilization that directly enables larger batch sizes, longer sequence handling, and superior throughput - critical advantages for production deployment. While PagedAttention eliminates memory fragmentation through non-contiguous block-level KV caching, it preserves the original memory footprint per parameter since key and value vectors remain stored in high-precision formats (FP32/FP16). To achieve further compression, we implement group-wise quantization for the KV cache—a training-free technique that reduces memory requirements during autoregressive decoding. For each cached block containing key matrix $K_j \in \mathbb{R}^{B \times d}$ and value matrix $V_j \in \mathbb{R}^{B \times d}$, we independently quantize column-wise groups using group-specific parameters $(\alpha_g, z_g)$. The quantization of group $g$ in $K_j$ follows:

\resizebox{0.985\linewidth}{!}{
\begin{minipage}{\linewidth}
\begin{equation}
\tilde{K}_j^{(g)} = \left\lfloor \frac{K_j^{(g)}}{\alpha_g} - z_g \right\rceil, \quad \hat{K}_j^{(g)} = \alpha_g \cdot (\tilde{K}_j^{(g)} + z_g) \nonumber
\end{equation}
\end{minipage}
}

where $\tilde{K}_j^{(g)} \in \mathbb{Z}^{B \times d_g}$ contains quantized integers (typically INT4/8), and $\hat{K}_j^{(g)}$ denotes the dequantized approximation. An identical transformation applies to value matrices $V_j$. To minimize quantization error, we incorporate second-order Hessian information that identifies sensitive parameters through the diagonal Hessian matrix $H \in \mathbb{R}^{d \times d}$:

\resizebox{0.985\linewidth}{!}{
\begin{minipage}{\linewidth}
\begin{equation}
\alpha_g = \frac{\max(K_j^{(g)}) - \min(K_j^{(g)})}{2^n-1}, \quad z_g = \left\lfloor \frac{\min(K_j^{(g)})}{\alpha_g} \right\rceil \nonumber
\end{equation}
\end{minipage}
}

\begin{figure*}[ht!]
\centering
\includegraphics[width=1.0\linewidth]{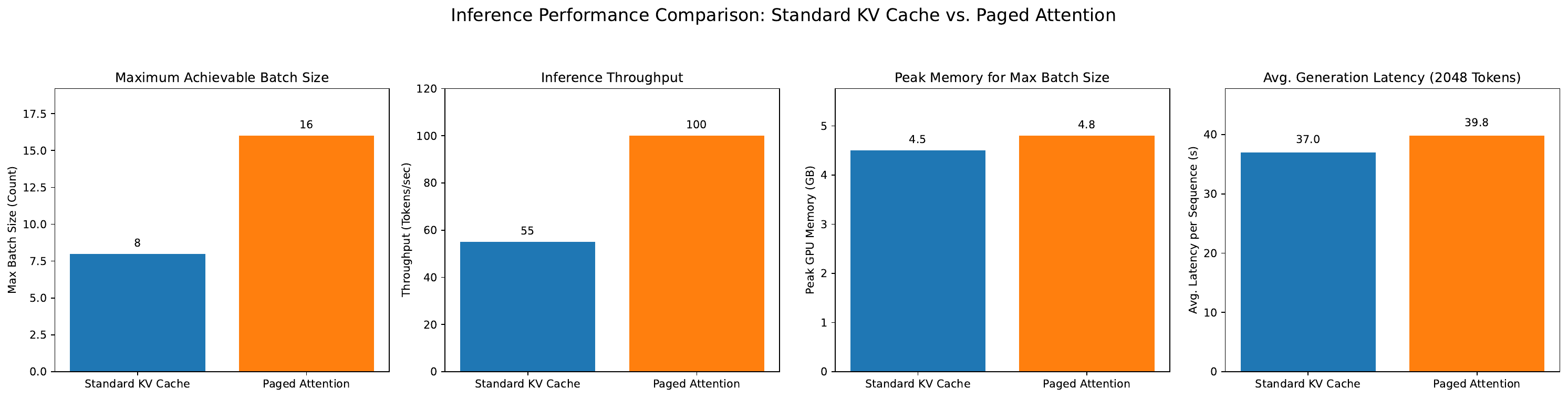}
\vspace{-3mm}
\caption{Inference performance comparison between standard KV cache and PagedAttention combined with KV cache quantization on LLaMA-3.2 1B. Four key metrics are displayed: maximum achievable batch size, inference throughput (tokens/sec), peak GPU memory (GB) at maximum batch size, and average generation latency (s) for 2048-token sequences.}
\label{fig:paged_attention_comparison}
\vspace{-3mm}
\end{figure*}

This Hessian-aware approach enables aggressive 4-bit quantization while maintaining model accuracy by preserving high-curvature parameters. The block structure of PagedAttention optimizes dequantization efficiency through contiguous storage of group metadata ($\alpha_g$, $z_g$). The combined technique
delivers dual benefits: PagedAttention manages memory fragmentation through block paging, while quantization reduces memory consumption per parameter by $4\times$ (INT4 vs FP16). This enables larger batch sizes (increased throughput), longer sequence lengths (expanded context), and efficient deployment on memory-constrained hardware. We evaluate the inference-time efficiency gains enabled by PagedAttention combined with KV cache quantization. By managing the Key-Value (KV) cache in non-contiguous, fixed-size blocks, this approach mitigates internal and external memory fragmentation inherent in standard contiguous caching while significantly improving inference performance. Since PagedAttention is an inference-only optimization that preserves model output quality, we focus exclusively on system-level metrics rather than quality measures such as BLEU, ROUGE, or reward scores discussed elsewhere. The efficiency metrics evaluated include inference throughput (tokens generated per second), maximum batch size (largest number of parallel sequences processed), peak GPU memory usage (in GB), and average per-sequence latency (generation time in seconds). We benchmarked our best-performing fine-tuned model, LLaMA-3.2 1B (with fine-tuning and Graph RAG, Variant A), on an NVIDIA V100 GPU using a 500-example subset of the held-out 1.5K QA-pair generalization benchmark dataset. The results, shown in Figure~\ref{fig:paged_attention_comparison}, demonstrate significant efficiency improvements. PagedAttention enabled an approximately 2.0$\times$ increase in maximum batch size (16 versus 8) and improved inference throughput by nearly 1.8$\times$ (~100 vs. 55 tokens/sec) compared to the baseline. While the LLaMA-3.2 1B model requires only ~2.3 GB of VRAM in FP16 precision, the larger batch size with PagedAttention increased peak GPU memory usage slightly (~4.8 GB vs. ~4.5 GB) due to greater sequence parallelism. However, memory utilization was substantially more efficient due to reduced fragmentation. The average generation latency for a 2048-token sequence was approximately 39.8 seconds, with only a marginal increase (5–10\%) attributable to block management overhead. These findings demonstrate PagedAttention's practical benefits for serving fine-tuned SLMs, especially in RAG-based applications with long, variable-length contexts. This technique complements model-centric optimizations, enabling more scalable real-world deployments.

%%%%%%%%%%%%%%%%%%%%%%%%%%%%%%%%%%%%%%%%%%%%%%%%%%%%%%%%%%%%%%%%%%%%%%%%%%%%%%%
%%%%%%%%%%%%%%%%%%%%%%%%%%%%%%%%%%%%%%%%%%%%%%%%%%%%%%%%%%%%%%%%%%%%%%%%%%%%%%%

\vspace{-2mm}
\subsection{Low-Latency LLM Decoding Strategies}
Let \(\mathcal{V} = \{1, 2, \ldots, |\mathcal{V}|\} \subset \mathbb{Z}_{>0}\) denote the vocabulary of a causal language model \(M\) with parameters \(\theta\), where \(|\mathcal{V}|\) is the vocabulary size. Given a fixed input prompt \(x_0 = (x_1, x_2, \ldots, x_s) \in \mathcal{V}^s\) of length \(s\), the goal is to autoregressively generate a target sequence \(Y = (y_1, y_2, \ldots, y_T) \in \mathcal{V}^T\) of length \(T\), where each token \(y_t \in \mathcal{V}\). The language model defines a conditional probability distribution over the next token:

\vspace{-1mm} 
\resizebox{0.985\linewidth}{!}{
\begin{minipage}{\linewidth}
\begin{equation}
P_M(y_t \mid y_{<t}, x_0; \theta), \quad \text{where } y_{<t} = (y_1, \ldots, y_{t-1}) \nonumber
\end{equation}
\end{minipage}
}

\vspace{1mm}
This reflects the causal (left-to-right) nature of the generation process—each token prediction depends only on previous tokens and the fixed prompt. In greedy (deterministic) decoding, the most probable token is selected at each step:

\vspace{-2mm} 
\resizebox{0.985\linewidth}{!}{
\begin{minipage}{\linewidth}
\begin{equation}
y_t = \arg\max_{v \in \mathcal{V}} P_M(v \mid y_{<t}, x_0; \theta) \nonumber
\end{equation}
\end{minipage}
}

\vspace{1mm}
This results in decoding latency that scales linearly with the sequence length \(T\). To enable parallel decoding, we reformulate the generation task as a system of fixed-point equations. For each position \(t \in \{1, \ldots, T\}\), define:

\vspace{-1mm} 
\resizebox{0.985\linewidth}{!}{
\begin{minipage}{\linewidth}
\begin{equation}
F_t(y_t, y_{<t}, x_0) = y_t - \arg\max_{v \in \mathcal{V}} P_M(v \mid y_{<t}, x_0; \theta) = 0 \nonumber
\end{equation}
\end{minipage}
}

\vspace{1mm}
This system can be solved using Jacobi iteration, which computes speculative updates in parallel at each iteration based on previous estimates. The method trades increased per-step latency for reduced total generation time (i.e., faster completion of the full response). Speculation involves parallel guessing of multiple future tokens without sequential verification. Verification checks whether these speculative guesses match the outputs that greedy decoding would produce. Let \(k \in \mathbb{Z}_{\geq 0}\) denote the iteration index, and let \(y_t^{[k]} \in \mathcal{V}\) be the estimate of token \(y_t\) at iteration \(k\). The Jacobi update rule is:

\vspace{1mm} 
\resizebox{0.985\linewidth}{!}{
\begin{minipage}{\linewidth}
\begin{equation}
y_t^{[k]} = \arg\max_{v \in \mathcal{V}} P_M(v \mid y_{<t}^{[k-1]}, x_0; \theta) \nonumber
\end{equation}
\end{minipage}
}

\vspace{1mm} 
where \(y_{<t}^{[k-1]} = (y_1^{[k-1]}, \ldots, y_{t-1}^{[k-1]})\). While Jacobi iteration enables parallel updates, speculative tokens generated without sequential verification may introduce inconsistencies, potentially discarding valid generation paths. Consequently, Jacobi decoding alone lacks convergence guarantees and offers limited empirical speedup. To address these limitations, Lookahead Decoding~\cite{fu2024break, zhao2024lookahead, mamou2024dynamic} introduces a hybrid approach combining speculative Jacobi-based multi-token generation with a structured verification mechanism. While each decoding step incurs higher latency due to parallel computation and verification overhead, the method reduces the total number of sequential steps required to generate the complete response. The decoding process maintains several key components: The confirmed output prefix \(o = (o_1, \ldots, o_{t-1}) \in \mathcal{V}^{t-1}\) consists of tokens verified to match standard greedy decoding outputs. A token trajectory window \(W \in \mathcal{V}^{N \times L}\) tracks speculative predictions, where \(N \in \mathbb{Z}_{>1}\) represents the number of retained Jacobi iterations and \(L \in \mathbb{Z}_{>0}\) denotes the number of parallel lookahead positions, with \(L \ll T\) constraining the local speculation horizon. Each entry \(W_{r,j}\) corresponds to the token predicted at iteration \(r\) for lookahead position \(j\). For each column \(j \in \{1, \ldots, L\}\), the system constructs vertical decoding trajectories as \(N\)-gram candidates \(g_j = (W_{1,j}, \ldots, W_{N,j}) \in \mathcal{V}^N\) by vertically traversing the window \(W\) across iterations, with each \(g_j\) representing a complete speculative decoding path originating from the confirmed prefix \(o\). These trajectories are aggregated in the \(N\)-gram candidate pool \(\mathcal{C} \subset \mathcal{V}^N\) defined as \(\mathcal{C} = \{g_j \mid j \in \{1, \ldots, L\}\}\). During the lookahead phase, the system updates the final row \(W_{N,1:L}\) through parallel speculative token generation across all lookahead positions. For each position \(j \in \{1, \ldots, L\}\), the speculative token \(W_{N,j}\) is predicted via:

\vspace{1mm}
\resizebox{0.985\linewidth}{!}{
\begin{minipage}{\linewidth}
\begin{align}
\hspace{-3mm}
W_{N,j} = \arg\max_{v \in \mathcal{V}} P_M\Big( v \;\Big|\; &\left(W_{\min(N-1, j-1), j-1}, \ldots, W_{1, j - \min(N-1, j-1)}\right), \nonumber \\
& \; o, x_0; \theta \Big) \nonumber
\end{align}
\end{minipage}
}

\vspace{1mm}
This prediction considers three factors: (1) the confirmed prefix \(o = (o_1, \ldots, o_{t-1})\), (2) the original input prompt \(x_0\), and (3) a causal diagonal context from window \(W \in \mathcal{V}^{N \times L}\) containing up to \(N{-}1\) previously predicted tokens. The context is selected through a systematic traversal decreasing both row index \(r\) from \(\min(N{-}1, j{-}1)\) to \(1\) and column index \(j'\) from \(j{-}1\) to \(j - \min(N{-}1, j{-}1)\), strictly maintaining autoregressive dependencies while enabling parallel computation. This allows efficient generation of the complete final row \(W_{N,1:L}\) without violating causal constraints. Following lookahead updates, the system constructs vertical \(N\)-grams \(g_j = (W_{1,j}, \ldots, W_{N,j})\) for each position and adds them to candidate pool \(\mathcal{C}\). The verification phase then retrieves up to \(G\) candidates from \(\mathcal{C}\) satisfying \(g_j^{1} = o_{t-1}\) and sequentially verifies each candidate \(g_j = (g_j^{1}, \ldots, g_j^{N})\) for \(r = 1\) to \(N\) through the comparison:

\vspace{1mm} 
\resizebox{0.985\linewidth}{!}{  
\begin{minipage}{\linewidth}  
\begin{equation}
\begin{aligned}
g_j^{r} \stackrel{?}{=} \arg\max_{v \in \mathcal{V}} \; & P_M\Big( 
v \;\Big|\; \big( x_0, o_1, \ldots, o_{t-1}, g_j^{1}, \ldots, g_j^{r-1} \big); \theta 
\Big)
\end{aligned}
\nonumber
\end{equation}
\end{minipage}  
}  

\begin{figure*}[ht!]
\centering
\vspace{-3mm} % Adjust spacing
\includegraphics[width=0.8\linewidth]{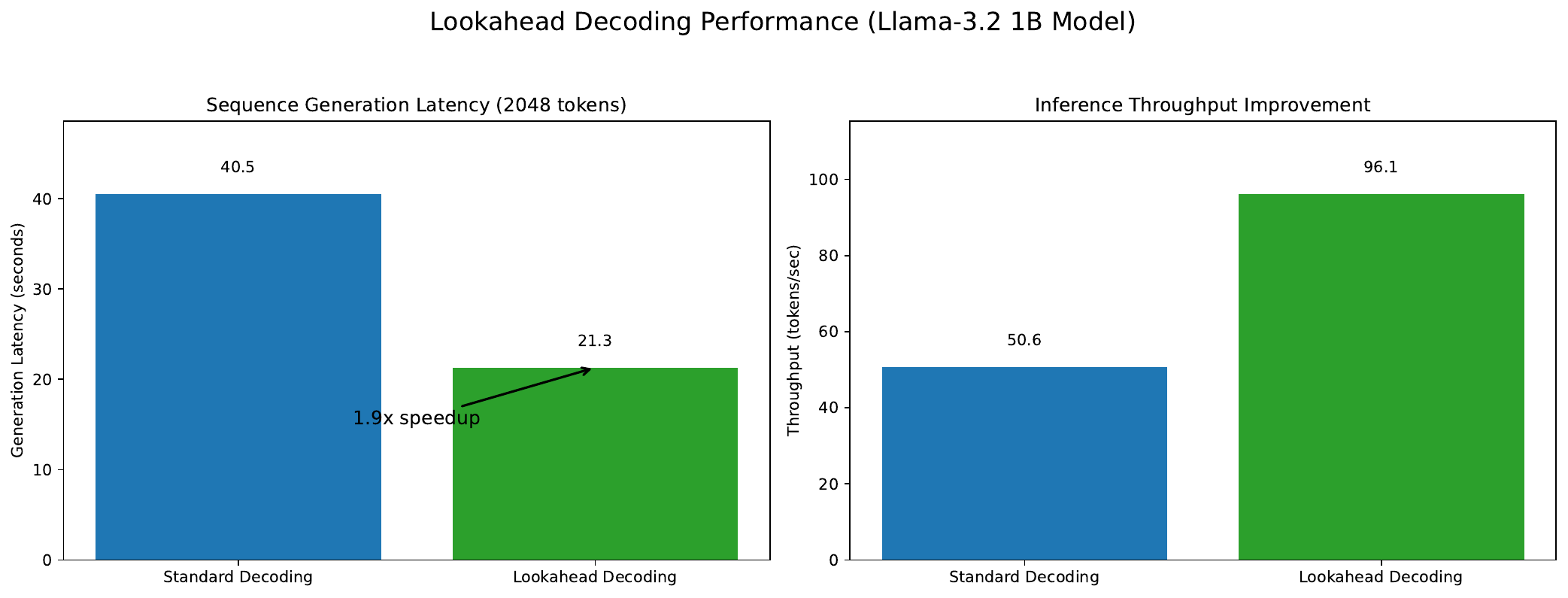} % A placeholder image could be used if preferred
\caption{Comparative inference performance of the fine-tuned Llama-3.2 1B model under standard greedy decoding and Lookahead Decoding (N=5, L=10). Results demonstrate a 1.9$\times$ latency reduction (40.5s $\rightarrow$ 21.3s for 2048 tokens) and 90\% higher throughput (50.6 $\rightarrow$ 96.1 tokens/sec).}
\label{fig:lookahead_comparison}
\vspace{-3mm} % Adjust spacing
\end{figure*}

\vspace{1mm}
 Verification yields either full acceptance, where all \(N\) tokens match greedy decoding outputs and are appended to \(o\), or partial acceptance where only the verified prefix \((g_j^{1}, \ldots, g_j^{r-1})\) is retained when verification fails at position \(r\). The window \(W\) then shifts rightward by the number of accepted tokens, discarding unverified speculative entries, thereby maintaining equivalence to standard greedy decoding while enabling speculative parallel generation. The lookahead and verification phases form a hybrid \textit{predict–verify–commit} decoding pipeline, enabling speculative multi-token generation while preserving exact output semantics. While increasing per-step latency, Lookahead Decoding is a lossless, parallel algorithm that maintains exact output fidelity while reducing the total number of sequential steps needed to generate the complete response. It combines token-level Jacobi speculation with \(N\)-gram-level greedy verification through a structured two-dimensional window and \(N\)-gram cache. This architecture trades increased FLOPs per step for reduced total generation time, scales effectively with parallel compute, and requires no model modifications or auxiliary networks. In summary, the Lookahead Decoding significantly reduces generation latency by speculatively predicting multiple future tokens in parallel and verifying them against the base model. This approach decodes multiple tokens per forward pass, cutting sequential steps while maintaining greedy decoding's exact output. 
  We evaluated two metrics: generation latency (total time per sequence) and throughput (tokens/second).  We evaluated the fine-tuned LLaMA-3.2 1B model using both standard greedy decoding and Lookahead Decoding (with N=5 iterations and L=10 lookahead positions) on an NVIDIA V100 GPU, benchmarking performance across 500 examples from our 1.5K QA test set. The results demonstrate substantial gains: latency for 2048-token sequences dropped from 40.5s to 21.3s (1.9× speedup), while throughput rose from 50.6 to 96.1 tokens/sec. Although parallel speculation increases per-step FLOPs, it reduces total generation time without requiring auxiliary models. This technique proves especially effective for SLMs like Llama-3.2 1B, nearly halving latency without compromising output quality—particularly valuable for time-sensitive applications like PFD/PID autogeneration. Its efficiency synergizes with optimizations such as Paged Attention and pruning. Figure~\ref{fig:lookahead_comparison} illustrates these performance improvements.

\begin{figure*}[ht!]
\centering
\includegraphics[width=0.85\linewidth]{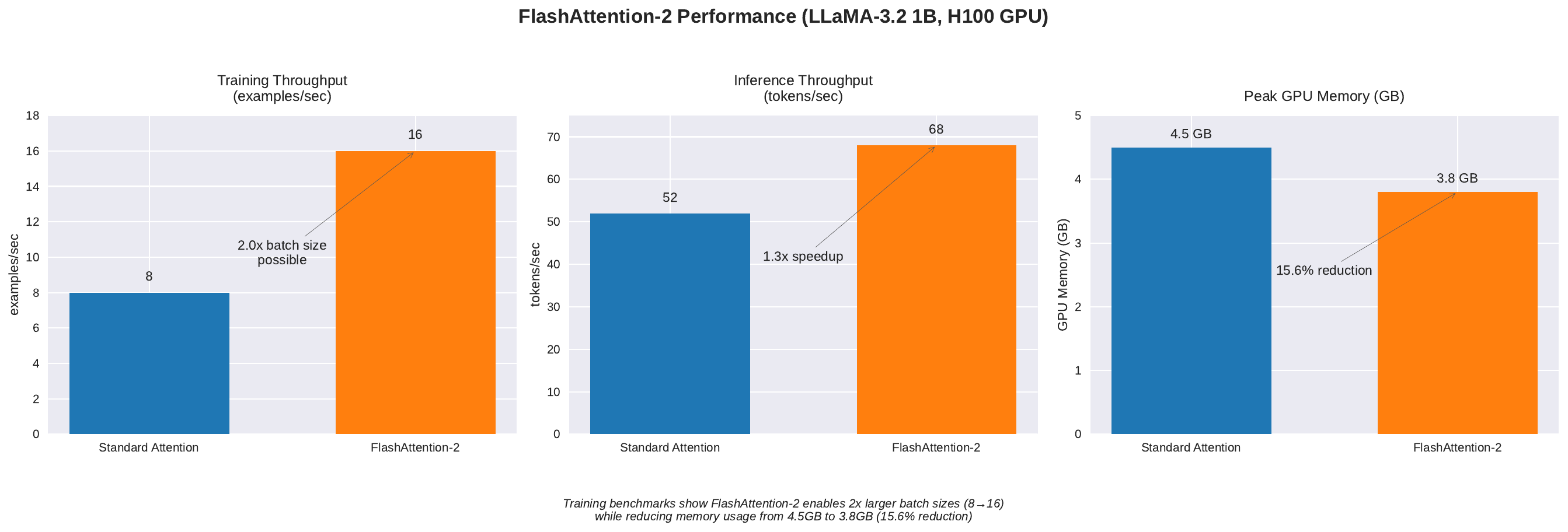}
\vspace{-3mm}
\caption{Performance comparison of Llama-3.2 1B using standard PyTorch attention versus FlashAttention on NVIDIA V100 GPU, showing training throughput (examples/sec), inference throughput (tokens/sec), and peak training memory usage (GB).}
\label{fig:flashattention_comparison}
\vspace{-3mm}
\end{figure*}

%%%%%%%%%%%%%%%%%%%%%%%%%%%%%%%%%%%%%%%%%%%%%%%%%%%%%%%%%%%%%%%%%%%%%%%%%%%%%%%

\vspace{-2mm}  
\subsection{FlashAttention (Optimizing Attention Computation)}  
FlashAttention~\cite{dao2022flashattention, dao2023flashattention, shah2024flashattention, chen2024int, abbott2024flashattention} improves attention computation by increasing throughput, reducing latency, and lowering memory usage while maintaining exact equivalence to standard attention. For the standard scaled dot-product attention mechanism, given query $Q \in \mathbb{R}^{N \times d_k}$, key $K \in \mathbb{R}^{N \times d_k}$, and value $V \in \mathbb{R}^{N \times d_v}$ matrices, where $N$ is sequence length and $d_k$, $d_v$ are dimensions, the attention scores are computed as:

\resizebox{0.985\linewidth}{!}{  
\begin{minipage}{\linewidth}  
\begin{equation}  
S = \frac{QK^\top}{\sqrt{d_k}} \nonumber
\end{equation}  
\end{minipage}  
}

A causal mask $M \in \mathbb{R}^{N \times N}$ with $M_{ij} = -\infty$ for $j > i$ prevents attention to future positions. The row-wise softmax produces attention probabilities:

\resizebox{0.985\linewidth}{!}{  
\begin{minipage}{\linewidth}  
\begin{equation}  
P_{ij} = \frac{\exp(S_{ij})}{\sum_{k=1}^{N} \exp(S_{ik})} \nonumber
\end{equation}  
\end{minipage}  
}

yielding output $O = PV \in \mathbb{R}^{N \times d_v}$. This standard approach requires materializing intermediate matrices $S,P \in \mathbb{R}^{N \times N}$, creating $O(N^2)$ memory overhead. The implementation suffers from significant HBM-SRAM data movement: (1) Loading $Q,K,V$ from HBM to SRAM; (2) Computing $S$ in SRAM; (3) Writing $S$ back to HBM if SRAM overflows; (4) Reloading $S$ to compute $P$; (5) Writing $P$ to HBM; (6) Reloading $P$ and $V$ for final output. These $O(N^2 d_k)$ memory transfers make bandwidth the dominant bottleneck. FlashAttention solves this via blockwise computation, partitioning $Q$ into $T_r$ blocks $\{Q_1,...,Q_{T_r}\}$ ($Q_i \in \mathbb{R}^{B_r \times d_k}$) and $K,V$ into $T_c$ blocks $\{K_1,...,K_{T_c}\}$, $\{V_1,...,V_{T_c}\}$ ($K_j \in \mathbb{R}^{B_c \times d_k}$, $V_j \in \mathbb{R}^{B_c \times d_v}$). Block sizes satisfy:

\resizebox{0.985\linewidth}{!}{  
\begin{minipage}{\linewidth}  
\begin{equation}  
B_r d_k + B_c d_k + B_c d_v + B_r B_c \ll M \nonumber
\end{equation}  
\end{minipage}  
}

where $M$ is SRAM capacity. The FlashAttention algorithm begins by initializing three components for each query block $Q_i \in \mathbb{R}^{B_r \times d_k}$: an output block $O_i \in \mathbb{R}^{B_r \times d_v}$ (initialized to zero), a normalization vector $l_i \in \mathbb{R}^{B_r}$ (set to zero), and a maximum vector $m_i \in \mathbb{R}^{B_r}$ (initialized to $-\infty$). The computation proceeds through nested loops where the outer loop iterates over query blocks while the inner loop processes corresponding key-value blocks $(K_j \in \mathbb{R}^{B_c \times d_k}, V_j \in \mathbb{R}^{B_c \times d_v})$. For each block pair, the algorithm first loads $(K_j,V_j)$ into SRAM and computes the local attention scores:

\resizebox{0.985\linewidth}{!}{  
\begin{minipage}{\linewidth}  
\begin{equation}  
S_{ij} = \frac{Q_i K_j^\top}{\sqrt{d_k}} \nonumber  
\end{equation}  
\end{minipage}  
}

When causal masking is required, the algorithm sets $S_{ij}[r,c] = -\infty$ for all positions where query index $r$ precedes key index $c$. The computation then progresses through three sequential steps: first calculating row-wise maxima $m_{ij}[r] = \max_{1 \leq c \leq B_c} S_{ij}[r,c]$, then computing exponentiated weights $P_{ij}^{\text{hat}} = \exp(S_{ij} - m_{ij})$, and finally determining normalization factors $l_{ij}[r] = \sum_{c=1}^{B_c} P_{ij}^{\text{hat}}[r,c]$. These local statistics are incorporated into running values through numerically stable updates:

\resizebox{0.985\linewidth}{!}{  
\begin{minipage}{\linewidth}  
\begin{equation}  
m_i^{\text{new}}[r] = \max(m_i[r], m_{ij}[r]) \nonumber  
\end{equation}  
\end{minipage}  
}

\resizebox{0.985\linewidth}{!}{  
\begin{minipage}{\linewidth}  
\begin{equation}  
l_i^{\text{new}}[r] = \exp(m_i[r] - m_i^{\text{new}}[r]) l_i[r] + \exp(m_{ij}[r] - m_i^{\text{new}}[r]) l_{ij}[r] \nonumber  
\end{equation}  
\end{minipage}  
}

The output block updates through careful combination of previous partial results with new attention-weighted values:

\resizebox{0.985\linewidth}{!}{  
\begin{minipage}{\linewidth}  
\begin{equation}  
O_i^{\text{new}} = \frac{\exp(m_i - m_i^{\text{new}}) l_i O_i + \exp(m_{ij} - m_i^{\text{new}}) (P_{ij}^{\text{hat}} V_j)}{l_i^{\text{new}}} \nonumber  
\end{equation}  
\end{minipage}  
}

After processing all key-value blocks for a given query block, the final output $O_i$ writes back to HBM. The backward pass employs an analogous blockwise strategy, recomputing $S_{ij}$ and $\hat{P}_{ij}$ using saved statistics $m_i$ and $l_i$ to avoid storing full $O(N^2)$ matrices. This approach computes gradients for $V_j$ as $\hat{P}_{ij}^\top \mathrm{d}O_i$ while deriving $Q_i$ and $K_j$ gradients through standard softmax backpropagation with recomputed $\hat{P}_{ij}$. Although increasing FLOPs by approximately 2×, this strategy dramatically reduces memory requirements from $O(N^2)$ to $O(N d_k)$ while preserving the exact $O(N^2 d_k)$ computational complexity of standard attention. Through these combined optimizations - blockwise computation, online softmax, and selective recomputation - FlashAttention achieves exact equivalence with standard attention while minimizing HBM-SRAM transfers, delivering 2-4× fewer memory accesses and up to 3× speedups for long sequences through its I/O-aware algorithm design. We implemented FlashAttention to optimize memory access between GPU HBM and on-chip SRAM during inference through its innovative tiling, recomputation, and kernel fusion techniques. This implementation-level optimization computes mathematically identical attention outputs while significantly reducing memory overhead and improving computational speed, particularly for long sequences, without affecting model outputs or task metrics like BLEU and ROUGE scores. Benchmarking on an NVIDIA H100 GPU with LLaMA-3.2 1B revealed substantial performance gains compared to standard PyTorch attention. During training, FlashAttention doubled throughput from 8 to 16 examples per second while reducing peak GPU memory consumption by 15.6\% (from 4.5 GB to 3.8 GB), enabling potential batch size increases or longer sequence training within the same memory budget. For inference, we observed a 1.3× throughput improvement, increasing generation speed from 52 to 68 tokens per second, which typically corresponds to reduced latency. These improvements, detailed in Figure~\ref{fig:flashattention_comparison}, stem from FlashAttention's I/O-aware design that minimizes costly data movement between HBM and SRAM - a critical advantage for memory-bound attention operations. FlashAttention works synergistically with other optimizations in our framework: Paged Attention efficiently manages KV cache, Lookahead Decoding reduces sequential generation steps, while FlashAttention accelerates the core attention computation itself. This combined approach creates a highly efficient system for both training and deploying SLMs, particularly beneficial for compute-intensive tasks like PFD/PID generation where it reduces development cycles and operational costs while maintaining model performance.

%%%%%%%%%%%%%%%%%%%%%%%%%%%%%%%%%%%%%%%%%%%%%%%%%%%%%%%%%%%%%%%%%%%%%%%%%%%%%%%

\begin{table*}[h!]
\centering
\renewcommand{\arraystretch}{1.3}
\begin{tabular}{|m{2cm}|p{11cm}|}
\hline
\textbf{Dataset Type} & \textbf{Prompt} \\
\hline
Factual QA Dataset & You must generate exactly \{n\_questions\} questions that are strictly and directly related to the specific subtopic provided. No tangential, broad, or off-topic questions are allowed.

The subtopic is: \{sub\_topics\}

Your response must consist of precisely \{n\_questions\} questions, each directly pertaining to the subtopic, separated by a newline character, with absolutely no additional text, numbering, explanations, or any other characters.

Deviation from the subtopic or any failure to generate exactly \{n\_questions\} questions as instructed will result in the output being considered invalid. \\
\hline
\multirow{2}{=}{DPO Dataset} & \textbf{Chosen Response Prompt Template:}

Generate a concise, relevant response to the given question. The response should be directly related to the question, clear, and free of any unnecessary information. It should be helpful, polite, and factually accurate.

The question is: \{question\}.

Provide only one response in plain text, with no additional explanations, introductions, or concluding remarks. \\
\cline{2-2}
& \textbf{Rejected Response Prompt Template:}

Generate a rejected response to the given question that is moderately inaccurate compared to the accurate response. The rejected response may be incomplete or less accurate, but it should still be relevant to the question.

The question is: \{question\}

Provide only one response in plain text, with no additional explanations, introductions, or concluding remarks. \\
\hline
LogiCore Dataset & Provide clear, accurate, and concise answers to the following questions. Adhere strictly to the following rules to ensure high scores in the following categories:

\textbf{Helpfulness:} Ensure each answer is maximally helpful, fully addressing the question in a way that effectively resolves the query. 

\textbf{Correctness:} Every answer must be factually correct, accurately referencing relevant details from the synthesis description (process context), Process Flow Diagram (PFD), and Piping and Instrumentation Diagram (P\&ID). 

\textbf{Coherence:} Ensure that each answer is logically structured and flows smoothly, making it easy for the reader to follow. 

\textbf{Complexity:} Balance complexity appropriately; provide necessary depth without making the answer overly complicated. Ensure the response is insightful when needed.

\textbf{Verbosity:} Be concise but thorough. Include all essential details without adding unnecessary information. Ensure that the length of the answer aligns perfectly with the complexity of the question.

Failure to adhere to these rules will lead to lower scores and suboptimal performance.

Synthesis Description: \{synthesis\_description\} 

Process Flow Diagram: \{pfd\_description\} 

Piping and Instrumentation Diagrams: \{pid\_description\} 

Questions: \{questions\} \\
\hline
Global/Local RAIT Dataset & Question: \{question\}

Context: \{chunk\}

Provide a concise, accurate, and fact-based answer to the question, using only the information available in the provided context. The answer must be directly derived from the context and should not include any external knowledge, speculation, or interpretation. Ensure that the response is precise and strictly adheres to the content of the context without introducing any additional information. \\
\hline
\end{tabular}
\caption{Illustrative prompt templates employed within the self-instruct framework to generate distinct synthetic datasets (\textit{Factual QA}, \textit{DPO}, \textit{LogiCore}, \textit{RAIT}) via teacher LLMs for subsequent instruction tuning.}
\end{table*}

\begin{table*}[h!]
\centering
\renewcommand{\arraystretch}{1.3}
\begin{tabular}{|m{1cm}|p{13cm}|}
\hline
\textbf{Dataset Type} & \textbf{Prompt} \\
\hline
\multirow{3}{=}{SynDIP Dataset} & \textbf{Industrial Synthesis Generation Prompt Template:} 

Provide a comprehensive and detailed description of the industrial synthesis process for \{chemical\_name\}. Your description should include:

\begin{itemize}[leftmargin=*,nosep]
\item All key chemical reactions, including reactants, intermediates, and products.
\item The types of reactors used (e.g., CSTR, PFR) and their operating conditions (e.g., temperature, pressure).
\item Details of any purification steps, such as distillation, crystallization, or filtration, including the equipment used.
\item Handling and treatment of by-products and waste streams.
\item Any recycling loops and the integration of heat exchange systems to optimize energy use.
\item Specific safety measures taken during the synthesis, especially when dealing with hazardous chemicals.
\end{itemize}

The description should be suitable for an engineer looking to understand the process in detail for implementation in a large-scale industrial setting. \\
\cline{2-2}
& \textbf{PFD Generation Prompt Template:} 

Based on the following synthesis description, create a detailed textual Process Flow Diagram (PFD) for the synthesis of \{chemical\_name\}. Your PFD should include:

\begin{itemize}[leftmargin=*,nosep]
\item Major equipment involved at each step, such as reactors, heat exchangers, distillation columns, separators, pumps, and compressors.
\item The flow of raw materials, intermediates, and products through the process, including any recycling streams.
\item Details of heat integration, such as the use of heat exchangers to recover energy from exothermic reactions or to preheat reactants.
\item A clear representation of phases (e.g., gas, liquid, solid) in each unit operation, highlighting phase transitions where applicable.
\item Specific operating conditions at key stages, including temperatures, pressures, and flow rates, to ensure proper operation.
\item The identification of potential bottlenecks in the process flow, and suggestions for optimizing throughput.
\end{itemize}

Ensure that the PFD is designed according to industry standards and is suitable for scaling up to large-scale production. \\
\cline{2-2}
& \textbf{P\&ID Generation Prompt Template:}  

Create a detailed Piping and Instrumentation Diagram (P\&ID) based on the following process flow diagram (PFD) for the synthesis of \{chemical\_name\}. The P\&ID should include:

\begin{itemize}[leftmargin=*,nosep]
\item Detailed placement of sensors (e.g., temperature, pressure, flow, and level sensors) at critical points in the process to ensure precise monitoring.
\item Specification of control valves, actuators, and control loops required to maintain process parameters within the desired range, including examples of typical control strategies (e.g., feedback, feedforward, cascade control).
\item Identification of safety instrumentation, such as pressure relief valves, emergency shutdown systems, interlocks, and alarms, to prevent accidents and ensure compliance with safety regulations.
\item Details of the instrumentation needed for process optimization, such as advanced process control (APC) systems, model predictive control (MPC), and real-time data analytics.
\item Consideration of redundancy and reliability in the placement of key sensors and control elements to ensure continuous operation and minimize downtime.
\item Recommendations for the types of piping materials to be used, considering chemical compatibility, temperature, and pressure of the process streams.
\item Suggestions for the integration of control systems with a distributed control system (DCS) or supervisory control and data acquisition (SCADA) system for centralized monitoring and control.
\end{itemize}

Your P\&ID suggestions should align with industry best practices and standards (e.g., ANSI/ISA-5.1) and be suitable for large-scale industrial production. \\
\hline
\end{tabular}
\caption{Sequential prompt templates used for the multi-stage generation of the \textit{SynDIP} dataset, covering industrial synthesis descriptions, textual PFD generation, and subsequent textual P\&ID generation.}
\end{table*}
%%%%%%%%%%%%%%%%%%%%%%%%%%%%%%%%%%%%%Sample Q&A%%%%%%%%%%%%%%%%%%%%%%%%%%%%%%%%%%%%%%%%%%%%%%

\begin{table*}[h!]
\centering
\renewcommand{\arraystretch}{1.3}
\begin{tabular}{|m{1.2cm}|m{2.25cm}|m{5.5cm}|m{4.5cm}|}
\hline
\textbf{Dataset Type} & \textbf{Question} & \textbf{Ground Truth Response} & \textbf{Model Response} \\
\hline
Factual QA Dataset & What are the challenges associated with heat transfer in high-temperature reactors? & Challenges associated with heat transfer in high-temperature reactors include material degradation, reduced thermal conductivity, increased risk of overheating, and the need for advanced cooling systems to manage extreme temperatures effectively. & Key heat transfer challenges in high-temperature reactors include material degradation, potentially reduced thermal conductivity at extreme temperatures, and increased risk of overheating, often necessitating advanced cooling solutions. \\
\hline
DPO Dataset & How does the scale of production impact reactor design choices? & \textbf{Chosen Response:} The scale of production impacts reactor design choices by influencing the type of reactor selected (batch vs. continuous), size and capacity requirements, heat and mass transfer considerations, and the need for automation and control systems to ensure efficiency and consistency at larger scales.

\textbf{Rejected Response:} The scale of production impacts reactor design choices mainly by determining the reactor's shape and color. Larger scales tend to favor simpler designs, but they do not significantly affect material selection or the technology used. & Production scale heavily influences reactor design, affecting the choice between batch and continuous reactors, determining size and capacity, impacting heat and mass transfer requirements, and dictating the necessary level of automation and control systems. \\
\hline
LogiCore Dataset  & What are the implications of reactor type selection (CSTR vs. PFR) on the overall efficiency and yield of the acetone synthesis process?  & CSTRs (Continuous Stirred-Tank Reactors) typically offer lower efficiency and yield for acetone synthesis due to backmixing, leading to reactant dilution and reduced conversion. PFRs (Plug Flow Reactors) provide higher efficiency and yield by maintaining concentration gradients, enabling better reactant conversion and selectivity. & Selecting a PFR generally leads to higher efficiency and yield in acetone synthesis because it maintains reactant concentration gradients for better conversion. Conversely, a CSTR often results in lower efficiency due to backmixing and reactant dilution, although it might offer simpler temperature control. \\
\hline
Global RAIT Dataset & How can unreacted hydrocarbons and hydrogen by-products be managed in the synthesis of 1,3-butadiene? & Unreacted hydrocarbons and hydrogen by-products from the synthesis of 1,3-butadiene can be effectively managed through several strategies. Unreacted hydrocarbons can be recycled back into the reactor or directed to a separate recovery unit for further processing, ensuring that valuable feedstock is not wasted. Hydrogen, on the other hand, can be captured and utilized as a fuel source or for other chemical processes, thereby enhancing overall efficiency and sustainability. This approach minimizes waste and maximizes the utilization of raw materials in the production process. & In 1,3-butadiene synthesis, unreacted hydrocarbons are typically managed by recycling them back to the reactor or separating them for recovery. Hydrogen by-products can be captured and utilized, for instance, as fuel or in other chemical processes, which improves overall process efficiency and material utilization. \\
\hline
Local RAIT Dataset & What are the operating conditions for the final synthesis of ammonium sulfate? & The operating conditions for the final synthesis of ammonium sulfate are a temperature maintained at 60-80°C and atmospheric pressure. The reactor type used for this synthesis is a Continuous Stirred-Tank Reactor (CSTR). & The final synthesis of ammonium sulfate typically operates at a temperature of 60-80°C and atmospheric pressure, using a Continuous Stirred-Tank Reactor (CSTR). \\
\hline
\end{tabular}
\caption{Comparison of model responses and ground truth responses across different synthetic dataset types.}
\label{tab:response_comparison_revised}
\end{table*}

\end{document}